\documentclass[11pt, letterpaper, onecolumn, goog]{google}
\usepackage[numbers,sort&compress,super,comma]{natbib}
\usepackage{cleveref}
\usepackage{multirow}
\usepackage{tcolorbox}
\usepackage{listings}
\tcbuselibrary{listings,breakable}
\usepackage{bookmark}
\newtcolorbox{promptbox}{
  breakable,
  colback=gray!10,
  colframe=gray!40,
  boxrule=0.5pt,
  arc=2mm,
  left=1.5mm,right=1.5mm,top=1mm,bottom=1mm
}

\hypersetup{hidelinks}
\usepackage[labelfont=bf,labelsep=period]{caption}
\makeatletter
\renewcommand{\fnum@figure}{\textbf{Fig.~\thefigure}}
\renewcommand{\fnum@table}{\textbf{Table~\thetable}}
\makeatother
\usepackage[font=small,labelfont=bf]{caption}
\usepackage{longtable}
\usepackage{booktabs}
\usepackage{xcolor}

\renewcommand{\titlefont}{\color{black}\normalfont\bfseries\fontsize{20}{22}\selectfont}

\title{Reasoning Models Generate Societies of Thought}

\renewcommand{\today}{}

\author[1,2,$\dag$]{Junsol Kim}
\author[2]{Shiyang Lai}
\author[1]{Nino Scherrer}
\author[1,3]{Blaise Agüera y Arcas}
\author[1,2,3,*]{James Evans}

\affil[1]{Google, Paradigms of Intelligence Team}
\affil[2]{University of Chicago}
\affil[3]{Santa Fe Institute}

\begin{abstract}
Large language models have achieved remarkable capabilities across domains, yet mechanisms underlying sophisticated reasoning remain elusive\cite{Brown2020-cu,Wei2022-fx}. Recent reasoning-reinforced models, including OpenAI’s o-series, DeepSeek-R1, and QwQ-32B, outperform comparable instruction-tuned models on complex cognitive tasks\cite{OpenAI2024-ah,Guo2025-jz}, attributed to extended test-time computation through longer chains of thought\cite{Snell2024-uy}. Here we show that enhanced reasoning emerges not from extended computation alone, but from the implicit simulation of complex, multi-agent-like interactions—a society of thought—which enables the deliberate diversification and debate among internal cognitive perspectives characterized by distinct personality traits and domain expertise. Through quantitative analysis using classified outputs and mechanistic interpretability methods applied to reasoning traces\cite{Kim2025-qi,Bricken2023-kn}, we find that reasoning models like DeepSeek-R1 and QwQ-32B exhibit much greater perspective diversity than baseline and merely instruction-tuned models, activating broader conflict between heterogeneous personality- and expertise-related features during reasoning. This multi-agent structure manifests in conversational behaviours including question-answering sequences, perspective shifts, and reconciliation of conflicting views, as well as in socio-emotional roles that characterize sharp back-and-forth conversation, which together account for the accuracy advantage in reasoning tasks through both direct and indirect facilitation of cognitive strategies\cite{Gandhi2025-xn,Yao2023-jc}. Controlled reinforcement learning experiments further reveal that base models spontaneously increase conversational behaviours when solely rewarded for reasoning accuracy, and fine-tuning models with conversational scaffolding substantially accelerates reasoning improvement compared to base models and models fine-tuned with monologue-like reasoning. These findings indicate that the social organization of thought enables effective exploration of solution spaces. We suggest that reasoning models establish a computational parallel to collective intelligence in human groups\cite{Mercier2017-wr,Page2019-ux,Woolley2010-cs}, where diversity enables superior problem-solving when systematically structured and suggest new opportunities for agent organization to harness the wisdom of crowds.
\end{abstract}

\uselogo{} 

\correspondingauthor{\textit{\scriptsize{\textsuperscript{$*$}Corresponding author: James Evans (\href{jamesaevans@google.com}{jamesaevans@google.com}; \href{jevans@uchicago.edu}{jevans@uchicago.edu}) \\ \vspace{-0.5em} \textsuperscript{$\dag$}Work done as a student researcher at Google}  }}

\begin{document}

\maketitle

\vspace{2em}

Artificial intelligence (AI) systems have undergone a remarkable transformation in recent years, with large language models (LLMs) demonstrating increasingly sophisticated abilities across domains, from mathematics and code to scientific and creative writing to critical decision support\cite{Brown2020-cu,Wei2022-fx}. Nevertheless, a persistent challenge has been the development of robust reasoning capabilities---the ability to methodically analyze problems, consider alternatives, detect errors, and arrive at reliable conclusions. Recent reasoning models, such as DeepSeek-R1, QwQ, and OpenAI's o-series models (o1, o3, o4), are trained by reinforcement learning to ``think'' before they respond, generating lengthy ``chains of thought''. This led to substantial improvement in reasoning accuracy compared to existing instruction-tuned language models (e.g., DeepSeek-V3, Qwen-2.5, GPT-4.1)\cite{OpenAI2024-ah,Guo2025-jz}. Yet, the character of ``thinking'' within reasoning models that drives success remains underexplored.

We propose that reasoning models learn to emulate social, multi-agent-like dialogue between multiple perspectives---what we term a ``society of thought''---to improve their reasoning, given the centrality of social interaction to the development of reason in both cognitive and social scientific accounts. Mercier and Sperber's ``Enigma of Reason'' argument posits that human reasoning evolved primarily as a social process, with knowledge emerging through adversarial reasoning and engagement across differing viewpoints\cite{Mercier2017-wr}. Empirical work supports the idea that groups outperform individuals on a wide range of reasoning tasks by pooling information, calibrating confidence, and exhibiting collective intelligence through balanced turn-taking among diverse perspectives\cite{Moshman1998-ic,Mesmer-Magnus2012-qg,Woolley2010-cs,Bahrami2010-sy}. Cognitive diversity, stemming from variation in expertise and personality traits, enhances problem solving, particularly when accompanied by authentic dissent\cite{Hong2004-jn,Nemeth2001-pl,Engel2014-jh,DeChurch2010-ni,Page2019-ux,Trouche2014-pg,Barrick1998-jp,Kim2025-vm}. Together, these findings suggest that robust reasoning emerges through interaction and the integration of diverse perspectives, and that key reasoning strategies, including verification and backtracking, may be realized through the conversation of simulated personas.

While diversity and debate contribute directly to collective intelligence, many theories further suggest that individuals reason better when they simulate this capacity. A single, self-centered perspective can lead to systematic biases in reasoning; if individuals effectively simulate multiple, self-distanced perspectives with their minds, as in dialectical thinking, this can reduce decision biases within them\cite{Kross2012-no,Grossmann2014-ff,Sun2018-ll}. The ``social brain hypothesis'' suggests that higher-order intelligence primarily evolved to meet the cognitive demands of processing and simulating social interactions\cite{Dunbar2014-qz,David-Barrett2013-jm}. Individuals who simulate others' differing perspectives through improved ``theory-of-mind'' capabilities enhance collective team performance\cite{Engel2014-jh}. Furthermore, theorists have argued that individual reason itself emerged from a simulation of collective discourse. Bakhtin's notion of the ``dialogic self'' and Cooley and Mead's theory of the ``looking glass self'' argue that human thought itself takes the form of an internalized conversation among multiple perspectives\cite{Bakhtin1984-ha,Hermans1992-ek,Cooley1902-wt,Mead2022-rw}. Even in the history of artificial intelligence, Minsky conceptualized intelligence as an emergent property of interacting cognitive agents, or a ``Society of Mind''\cite{Minsky1987-ef}.

Therefore, whether AI systems directly simulate multi-agent discourse or simulate minds that, in turn, simulate multi-agent discourse, we propose that reasoning models like DeepSeek-R1 improve reasoning via ``society of thought''---implicit simulations of multi-agent-like interactions between diverse perspectives that give rise to them. We use the term to denote text generation that simulates social exchange among multiple perspectives to increase the collective diversity of ideas through conversational roles that put them in competition. Without deploying separate models prompted to interact with one another\cite{Park2023-rk,Chan2023-ef,Liang2024-jj}, we suggest that behaviourally similar conversations between diverse perspectives occur and are leveraged within reasoning models.

Reasoning models like DeepSeek-R1 develop reasoning abilities through reinforcement learning, which iteratively compensates reasoning behaviour that yields correct answers. Following these performance improvements, debates have naturally arisen about what kinds of behaviours contribute to better reasoning performance. While earlier studies focus on how the model learns to scale test-time computations and generate longer reasoning traces\cite{Chen2025-cu,Wei2022-fx}, merely increasing trace length does not account for the observed improvements in reasoning capabilities. This suggests that qualitative changes in reasoning structure matter more than quantitative scaling alone\cite{Wang2025-kp,Chen2025-cu,Yeo2025-sh}. Recent analyses pinpoint behavioural patterns that improve reasoning accuracy, such as verification of earlier assumptions, backtracking, and exploration of alternatives\cite{Guo2025-jz,Gandhi2025-xn,Wang2025-kp,Chen2025-cu,Yeo2025-sh}. Mechanistic interpretability research has shown that features in language models such as the frequent use of words like ``wait,'' ``but,'' and ``however''---are associated with these behaviours\cite{Venhoff2025-tj,Galichin2025-yn,Ward2025-ep,Venhoff2025-tg}. The characteristics of these features, however, such as their prevalence in social and conversational settings, have rarely been explored. Research in other contexts has suggested that the simulation of multi-agent conversations can boost accuracy and divergent thinking in LLMs\cite{Eo2025-si,Zhang2024-iq,Chen2023-xa,Du2023-xm,Hu2024-wk,Liang2024-jj,Lai2024-wk}. While LLMs can exhibit cognitive biases that hinder reasoning, the simulation of interaction between different perspectives could mitigate biases when verified through checks and balances\cite{Sumita2024-pz,Liang2024-jj,Wynn2025-qp}. This leads us to hypothesize that reinforcement learning may systematically select and reward behaviour patterns that resemble multi-agent interactions within reasoning models, and these simulated interactions enable models to reason effectively.

Here we investigate the prevalence of reasoning traces of DeepSeek-R1, as well as QwQ-32B, that mimic simulated social interactions, quantifying how conversational behaviours, socio-emotional roles, and diversity of implicit agent ``perspectives'' contribute to reasoning performance. We first identify whether conversational behaviours and socio-emotional roles---hallmarks of human dialogue such as questioning, perspective taking, and reconciliation---are present in DeepSeek-R1's and QwQ-32B's reasoning traces. Then we test whether conversational behaviour contributes to reasoning performance. Based on the mechanistic interpretability method applied to DeepSeek-R1's distilled model (DeepSeek-R1-Llama-8B), we find that steering features associated with a discourse marker, such as expressing surprise in conversational contexts, improves reasoning accuracy both directly and indirectly through the facilitation of cognitive strategies.

Next, we analyze the diversity of reasoning ``perspectives'' or simulated voices within DeepSeek-R1's and QwQ-32B's reasoning traces. Literature suggests that LLM reasoning can fail if models do not engage in meaningful disagreement and instead conform to misleading initial claims through pleasant, ``sycophantic'' conversations that propagate incorrect assumptions and knowledge\cite{Wynn2025-qp,Chen2024-vd,Feng2025-vu}. Successful reasoning models may therefore exhibit disagreement driven by diversity in simulated perspectives, expressed through distinct personalities and expertise to avoid the ``echo chamber'' that leads to wrong answers. Therefore, we analyze reasoning traces using LLM-as-judge to accurately identify distinct voices underlying conversation. We find that DeepSeek-R1 and QwQ-32B display much greater personality and expertise diversity than non-reasoning models within their reasoning traces, presumably to maximize the benefits of multi-agent-like interaction through diversification. We further find that steering a conversational feature in a model's activation space leads to the activation of a more diverse range of personality- and expertise-related features.

Finally, we conduct a controlled reinforcement learning experiment to examine the role of conversational behaviours. We focus on self-taught reinforcement learning that rewards only accuracy and correct formatting (i.e., wrapping the thinking process between \textless think\textgreater{} and \textless/think\textgreater), the common approach for improving modern language models' reasoning capabilities\cite{Guo2025-jz}. Based on a symbolic arithmetic task (Countdown game)\cite{Gandhi2025-xn,Gandhi2024-xn}, as well as a misinformation identification task, we apply reinforcement learning that rewards reasoning traces leading to accurate answers on open-source LLMs. Interestingly, experiments reveal that the base model can spontaneously develop conversational behaviours---such as self-questioning and perspective shifts---when rewarded solely for reasoning accuracy, without any explicit training signal for dialogue structure. Moreover, following methods of prior ablation research\cite{Gandhi2025-xn}, we observe that initially fine-tuning these models for conversational structure leads to faster accuracy improvements, outperforming both their baseline counterparts and models fine-tuned with ``monologue-like'' reasoning, particularly during the early stages of training in two distinct model systems (Qwen-2.5-3B and Llama-3.2-3B). These results suggest that conversational scaffolding facilitates the discovery and refinement of reasoning strategies during reinforcement learning.

\phantomsection
\section{Results}\label{sec:results}

We compile a suite of widely used benchmarks used in prior research and official model cards of reasoning models (BigBench Hard, GPQA, MATH (Hard), MMLU-Pro, MUSR, and IFEval)\cite{OpenAI2024-ah,Guo2025-jz}, spanning symbolic logic, mathematical problem solving, scientific reasoning, multi-agent inference, and instruction following tasks (see \hyperref[sec:data]{Methods: Data}). From this pool, we sample 8,262 problems and generate reasoning traces using DeepSeek-R1-0528 (671B parameters; hereafter DeepSeek-R1) and QwQ-32B. For comparison, we also generate reasoning traces using conventional, instruction-tuned models of varying sizes: DeepSeek-V3-0324 (671B parameters; hereafter DeepSeek-V3), Qwen-2.5-32B-Instruct (hereafter Qwen-2.5-32B-IT; the instruction-tuned model based on Qwen-2.5-32B from which QwQ-32B is derived), Llama-3.3-70B-Instruct (hereafter Llama-3.3-70B-IT), and Llama-3.1-8B-Instruct (hereafter Llama-3.1-8B-IT)\cite{DeepSeek-AI2024-kr,Grattafiori2024-vb}. DeepSeek-V3 is the instruction-tuned model based on DeepSeek-V3-base from which DeepSeek-R1 is derived, and Qwen-2.5-32B-IT is the instruction-tuned model based on Qwen-2.5-32B from which QwQ-32B is derived (see \hyperref[sec:data]{Methods: Data})\cite{DeepSeek-AI2024-kr,Grattafiori2024-vb}.

Next, we estimate behavioural differences between reasoning models (DeepSeek-R1 and QwQ-32B) and the instruction-tuned models. We use linear probability models with problem-level fixed effects, which control for all task-specific characteristics, such as the difficulty of tasks. Specifically, we compare each reasoning model with its corresponding instruction-tuned counterpart (i.e., DeepSeek-R1 vs. DeepSeek-V3; QwQ-32B vs. Qwen-2.5-3B-IT) on the presence of conversational behaviours and socio-emotional roles. We control for log-transformed reasoning trace length (\hyperref[fig:edfig1]{Extended Data Fig.~1}) to consider that observed differences are not merely driven by ``longer'' chains of thought---that is, we demonstrate that reasoning models exhibit more frequent conversational behaviours and socio-emotional roles even when trace lengths are similar (see \hyperref[sec:statistical_analyses]{Methods: Statistical analyses}).

\begin{figure}[!htbp]
  \centering
  \includegraphics[width=\linewidth]{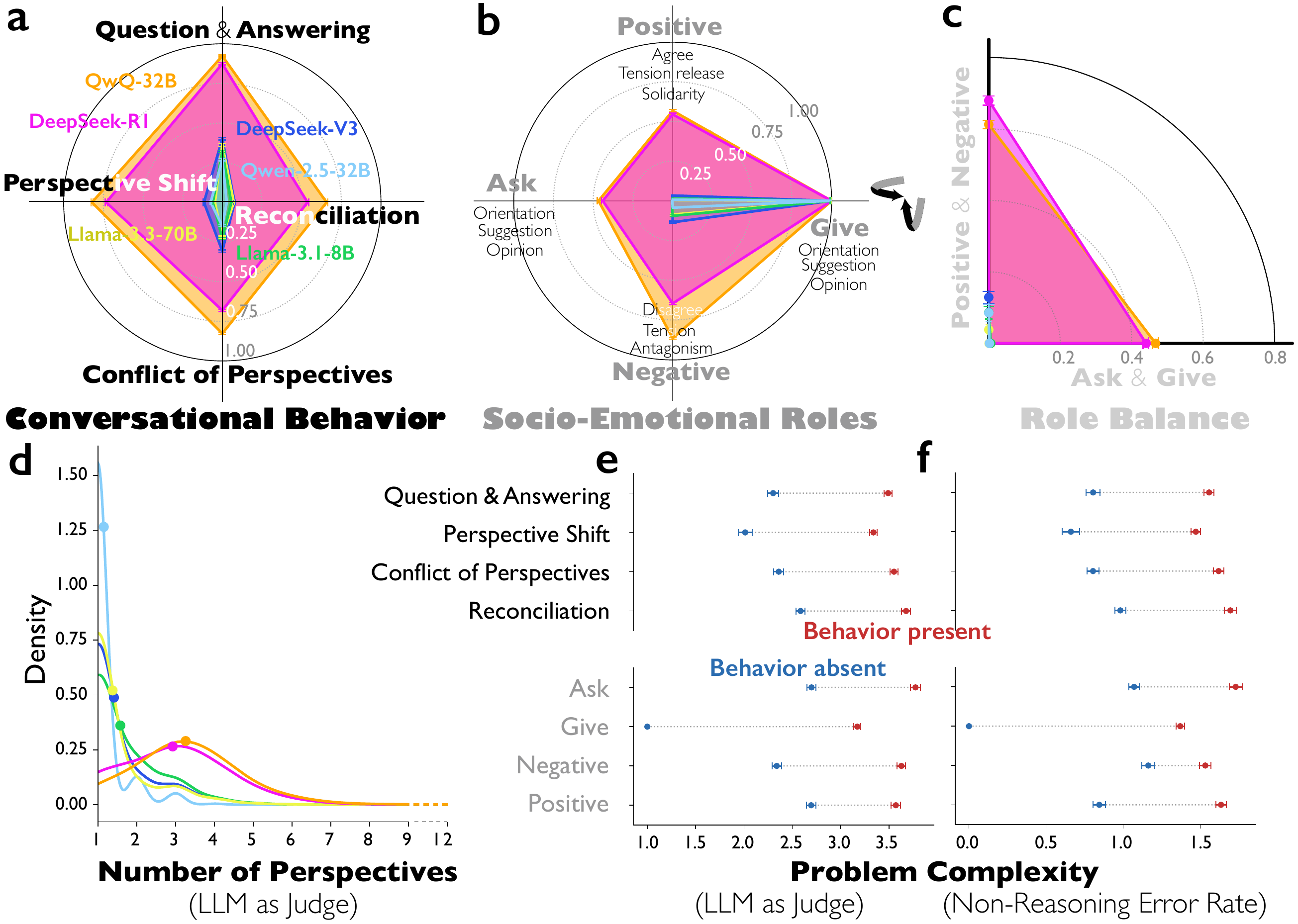}
  \caption{
  \textbf{Conversational behaviours and Bales' socio-emotional roles in chain-of-thought reasoning.}
  \textbf{a}, Proportion of reasoning traces containing each conversational behaviour (question answering, perspective shift, conflict of perspectives, and reconciliation).
  \textbf{b}, Proportion of Bales' twelve socio-emotional roles expressed in reasoning traces, grouped into four higher-level categories: \emph{ask} versus \emph{give} information, and \emph{positive} versus \emph{negative} emotional roles (see \hyperref[fig:edfig3]{Extended Data Fig.~3} for definitions of all twelve roles).
  \textbf{c}, Jaccard index measuring the balance of each socio-emotional role pair, defined as the number of reasoning traces containing both roles divided by the number containing either role (i.e., ask \& give; positive \& negative).
  \textbf{d}, Distribution of the number of distinct perspectives in reasoning traces, identified using an LLM-as-judge.
  \textbf{e}, Differences in problem complexity by the presence of conversational behaviours and higher-level socio-emotional roles in DeepSeek-R1, measured on a seven-point Likert scale (1 = extremely easy; 7 = extremely difficult) using an LLM-as-judge. Points indicate mean complexity for traces where the behaviour or role is present (red) or absent (blue).
  \textbf{f}, Differences in problem complexity by the presence of conversational behaviours and socio-emotional roles in DeepSeek-R1, measured by instruction-tuned (non-reasoning) models' error rates on the same problems (see \hyperref[sec:measurements]{Methods: Measurements}).
  Error bars indicate 95\% confidence intervals.
  }
  \label{fig:fig1}
\end{figure}

\subsection*{Conversational Behaviours and Socio-Emotional Roles}

We begin by investigating whether conversational behaviours and socio-emotional roles constitutive of back-and-forth dialogue are prevalent in reasoning traces. Using an LLM-as-judge, we quantify the occurrence of four conversational behaviours---defined as behaviours signaling the simulation of exchanges among multiple perspectives to explore a given problem—within each reasoning trace: (1) question–answering, in which the trace poses and then resolves questions; (2) perspective shifts, where alternative viewpoints are explored; (3) conflicts of perspectives, in which competing viewpoints are sharply contrasted; and (4) reconciliation, where conflicting viewpoints are integrated and coherently resolved.

We also examine socio-emotional roles based on Bales' Interaction Process Analysis (IPA)\cite{Bales1950-oa}. This identifies 12 interaction roles grouped into four categories: (1) \emph{asking} for orientation, opinion, and suggestion, (2) \emph{giving} orientation, opinion, and suggestion, (3) \emph{negative} emotional roles (disagreement, antagonism, tension), and (4) \emph{positive} emotional roles (agreement, solidarity, tension release), which together characterize interactive group activity. These behaviours are annotated using an LLM-as-judge (Gemini-2.5-Pro) that shows substantial agreement with both a human rater (average ICC(3,1) = .756) and another LLM (GPT-5.2; average ICC(3,1) = .875) (see \hyperref[sec:measurements]{Methods: Measurements} and \hyperref[sec:llm_prompts]{Supplementary Method: LLM-as-judge prompts} (Conversational behaviours and Socioemotional roles)).

To illustrate how reasoning traces are annotated, we provide examples in \hyperref[fig:edfig2]{Extended Data Fig.~2} and \hyperref[sec:annotations]{Supplementary Methods: Annotation Examples}. In an organic chemistry problem requiring multi-step reaction analysis to identify the final product's structure (i.e., multi-step Diels-Alder synthesis), DeepSeek-R1 exhibits perspective shifts and conflict, expressed through socio-emotional roles such as disagreement, giving opinion, and giving orientation (e.g., ``But here, it's cyclohexa-1,3-diene, not benzene.'' ``Another possibility: the high heat might cause the ketone to lose CO or something, but unlikely.''). In contrast, DeepSeek-V3's trace on the same problem shows no conflict of perspectives, no perspective shifts, and no disagreement---only giving opinions and orientations in a monologic sequence without self-correction, concluding with ``8 is a reasonable estimate'', the wrong answer, as a consequence of incomplete reasoning. In a creative sentence rewriting task, DeepSeek-R1 debates competing stylistic proposals through conflict of perspectives, as well as socio-emotional roles such as disagreement and giving suggestion: ``But that adds `deep-seated' which wasn't in the original. We should avoid adding new ideas.'' ``Wait, that's not a word.'' ``But note: `cast' can be less forceful than `flung'. So let's use `hurled'.'' DeepSeek-V3, by contrast, shows minimal conflict and no disagreement, producing suggestions without the iterative refinement observed in DeepSeek-R1.

As shown in \hyperref[fig:fig1]{Fig.~1a}, we quantify the occurrence of four conversational behaviours within each reasoning trace, and report the proportion of traces exhibiting more than one such behaviour. DeepSeek-R1 and QwQ-32B exhibit conversational behaviours far more frequently than instruction-tuned models. DeepSeek-R1 shows significantly more question--answering (\emph{$\beta$} = 0.345, 95\% CI = {[}0.328, 0.361{]}, \emph{t}(8261) = 41.64, \emph{p} \textless{} 1\(\times\)10\textsuperscript{-323}), perspective shifts (\emph{$\beta$} = 0.213, 95\% CI = {[}0.197, 0.230{]}, \emph{t}(8261) = 25.55, \emph{p} \textless{} 1\(\times\)10\textsuperscript{-137}), and reconciliation (\emph{$\beta$} = 0.191, 95\% CI = {[}0.176, 0.207{]}, \emph{t}(8261) = 24.31, \emph{p} \textless{} 1\(\times\)10\textsuperscript{-125}) compared to DeepSeek-V3. QwQ-32B displays a similar pattern relative to Qwen-2.5-32B-IT, with greater question--answering (\emph{$\beta$} = 0.459, 95\% CI = {[}0.444, 0.475{]}, \emph{t}(8261) = 57.57, \emph{p} \textless{} 1\(\times\)10\textsuperscript{-323}), perspective shifts (\emph{$\beta$} = 0.378, 95\% CI = {[}0.362, 0.394{]}, \emph{t}(8261) = 46.92, \emph{p} \textless{} 1\(\times\)10\textsuperscript{-323}), conflicts of perspectives (\emph{$\beta$} = 0.293, 95\% CI = {[}0.277, 0.308{]}, \emph{t}(8261) = 37.08, \emph{p} \textless{} 1\(\times\)10\textsuperscript{-277}), and reconciliation (\emph{$\beta$} = 0.344, 95\% CI = {[}0.328, 0.360{]}, \emph{t}(8261) = 42.59, \emph{p} \textless{} 1\(\times\)10\textsuperscript{-323}). Notably, all instruction-tuned models show consistently low prevalence of conversational behaviours regardless of parameter count (8B, 32B, 70B, 671B).

As shown in \hyperref[fig:fig1]{Fig.~1b}, both DeepSeek-R1 and QwQ-32B exhibit more reciprocal socio-emotional roles compared to their instruction-tuned counterparts: they both \emph{ask} for and \emph{give} orientations, opinions, and suggestions, while also displaying both \emph{negative} and \emph{positive} roles. DeepSeek-R1 asks more frequently than DeepSeek-V3 (\emph{$\beta$} = 0.189, 95\% CI = {[}0.176, 0.203{]}, \emph{t}(8261) = 27.47, \emph{p} \textless{} 1\(\times\)10\textsuperscript{-158}), engages more in negative roles (\emph{$\beta$} = 0.162, 95\% CI = {[}0.147, 0.176{]}, \emph{t}(8261) = 21.87, \emph{p} \textless{} 1\(\times\)10\textsuperscript{-10}), and displays more positive roles (\emph{$\beta$} = 0.278, 95\% CI = {[}0.263, 0.293{]}, \emph{t}(8261) = 35.38, \emph{p} \textless{} 1\(\times\)10\textsuperscript{-254}). QwQ-32B shows a similar pattern relative to Qwen-2.5-32B-IT, with increased asking (\emph{$\beta$} = 0.200, 95\% CI = {[}0.186, 0.215{]}, \emph{t}(8261) = 27.21, \emph{p} \textless{} 1\(\times\)10\textsuperscript{-155}), negative roles (\emph{$\beta$} = 0.450, 95\% CI = {[}0.436, 0.463{]}, \emph{t}(8261) = 64.77, \emph{p} \textless{} 1\(\times\)10\textsuperscript{-323}), and positive roles (\emph{$\beta$} = 0.312, 95\% CI = {[}0.296, 0.327{]}, \emph{t}(8261) = 39.17, \emph{p} \textless{} 1\(\times\)10\textsuperscript{-307}). In contrast, instruction-tuned models predominantly \emph{give} orientations, opinions, and suggestions without reciprocal asking behaviours or emotional engagement, producing one-sided monologues rather than simulated dialogue.

We quantify reciprocal role balance using the Jaccard index, which captures whether both sides of a role pair---asking versus giving for task-oriented roles, and positive versus negative for emotional roles---co-occur within the same reasoning trace. As shown in \hyperref[fig:fig1]{Fig.~1c}, DeepSeek-R1 exhibits significantly higher Jaccard indices for both ask \& give (\emph{$\beta$} = 0.222, 95\% CI = {[}0.208, 0.237{]}, \emph{t}(8261) = 30.21, \emph{p} \textless{} 1\(\times\)10\textsuperscript{-189}) and positive \& negative roles (\emph{$\beta$} = 0.189, 95\% CI = {[}0.176, 0.203{]}, \emph{t}(8261) = 27.47, \emph{p} \textless{} 1\(\times\)10\textsuperscript{-158}) compared to DeepSeek-V3, indicating that the model coordinates roles reciprocally rather than deploying them in isolation. QwQ-32B shows a similar pattern relative to Qwen-2.5-32B-IT (ask \& give: \emph{$\beta$} = 0.284 {[}0.269, 0.299{]}, \emph{t}(8261) = 37.36, \emph{p} \textless{} 1\(\times\)10\textsuperscript{-281}; positive \& negative: \emph{$\beta$} = 0.200 {[}0.186, 0.215{]}, \emph{t}(8261) = 27.24, \emph{p} \textless{} 1\(\times\)10\textsuperscript{-155}) (see \hyperref[tab:supptable1]{Supplementary Table 1}).

We further examine whether conversational behaviours and socio-emotional roles become more pronounced when DeepSeek-R1 faces more difficult tasks. Problem complexity is assessed both by an external LLM-as-judge (\hyperref[fig:fig1]{Fig.~1d}: Gemini-2.5-Pro) or by error rates across conventional instruction-tuned models (\hyperref[fig:fig1]{Fig.~1e}: DeepSeek-V3, Qwen-2.5-32B, Llama-3.3-70B-IT, Llama-3.1-8B-IT). As illustrated in \hyperref[fig:fig1]{Fig.~1d} and \hyperref[fig:fig1]{1e}, these behaviours appear more frequently when DeepSeek-R1 tackles more complex problems, except for giving orientations and opinions. Consistent patterns across both measures suggest that conversational reasoning is preferentially activated in response to greater problem difficulty. For instance, tasks with the highest complexity scores---such as GPQA (graduate-level science) and challenging math problems---exhibit strong conversational patterns, whereas simple procedural tasks like boolean expressions and basic logical deduction show minimal dialogic behaviour (see \hyperref[tab:supptable2]{Supplementary Table 2}).

To decompose the accuracy advantage of reasoning models (DeepSeek-R1 and QwQ-32B) using the behavioural mechanisms above, we estimate a structural equation model with four conversational behaviours, four socio-emotional roles, and four cognitive behaviour mediators, using task accuracy as the outcome. Results suggest that conversational behaviours and socio-emotional roles mediate reasoning models' accuracy advantage, both directly and indirectly through facilitating useful cognitive strategies, such as verification, backtracking, subgoal setting, and backward tracking (See \hyperref[fig:edfig4]{Extended Data Fig.~4}; \hyperref[sec:behavioural_pathways]{Supplementary Methods: behavioural Pathways Linking Reasoning Models to Accuracy Advantages}).

\subsection*{Conversational Feature Steering Improves Reasoning Accuracy}

Having observed that conversational behaviours are prevalent in reasoning traces using LLM-as-judge, we next question whether steering behaviours associated with conversations contribute to reasoning performance. We employ mechanistic interpretability methods to identify and manipulate features in the model's activation space related to conversational behaviours, and examine how steering these features affects the model's reasoning capabilities. We use sparse autoencoders (SAEs), which decompose neural network activations into a large set of linear, interpretable features\cite{Cunningham2023-sf,Bricken_undated-kd,Templeton2024-dx}. Specifically, we use an SAE trained on Layer 15's residual stream activations of DeepSeek-R1-Llama-8B (15-llamascope-slimpj-res-32k), a distilled model derived from DeepSeek-R1 frequently used to conduct interpretability research on LLM reasoning\cite{Venhoff2025-tj,Galichin2025-yn,Ward2025-ep,Venhoff2025-tg}. SAEs trained on middle layers, including Layer 15, are known to capture key behavioural and semantic features in models\cite{Kim2025-qi,Templeton2024-dx}. The SAE was trained on the SlimPajama dataset, a general-purpose, large-scale corpus used to train LLMs from scratch, containing both conversational and non-conversational texts (see \hyperref[tab:supptable3]{Supplementary Table 3} for full SAE hyperparameters)\cite{Shen2023-vy}.

\begin{figure}[!htbp]
  \centering
  \includegraphics[width=\linewidth]{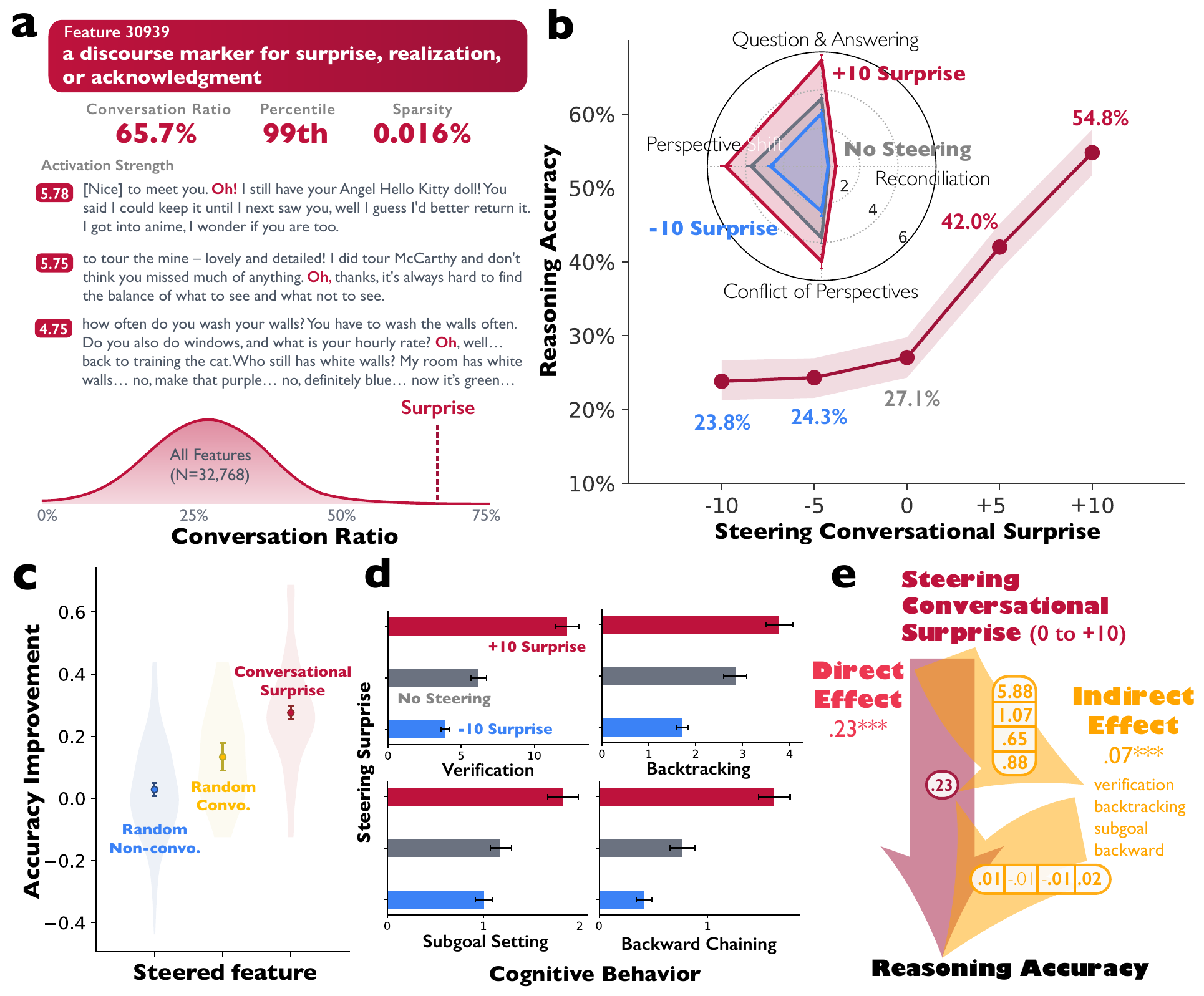}
  \caption{
  \textbf{Steering conversational features improves reasoning.}
  \textbf{a}, Illustration of sparse autoencoder feature 30939 in DeepSeek-R1-Llama-8B, summarized as a discourse marker for surprise, realization, or acknowledgment in conversational settings.
  \emph{Conversation ratio} indicates the proportion of conversational contexts among all contexts in which this feature is activated.
  \emph{Percentile} indicates where this feature’s conversation ratio ranks among all features ($N = 32{,}768$).
  \emph{Sparsity} refers to the fraction of tokens on which this feature activates across the entire corpus.
  \emph{Activation strength} shows the magnitude of activation in the top-activating examples.
  The examples illustrate this feature’s activation within conversational turn-taking contexts.
  \textbf{b}, Results of a steering experiment using the activation-addition method.
  Adding the feature 30939 vector with a strength of 10 doubles accuracy on a complex counting task.
  The inset shows the causal change in conversational behaviours induced by steering this feature.
  \textbf{c}, Violin plots showing accuracy improvements from steering feature 30939, compared with a randomly selected conversational SAE feature and a randomly selected non-conversational SAE feature.
  \textbf{d}, Cognitive behaviours—including verification, backtracking, subgoal setting, and backward chaining—are causally associated with steering the activation of feature 30939.
  \textbf{e}, Structural equation model results showing that steering feature 30939 from 0 to $+10$ has both a direct effect on reasoning accuracy and a significant indirect effect mediated through cognitive behaviours (verification, subgoal setting, and backward chaining).
  Bold coefficients indicate statistical significance ($\emph{p} < 0.05$).
  ***$\emph{p} < 0.001$, **$\emph{p} < 0.01$, *$\emph{p} < 0.05$.
  }
  \label{fig:fig2}
\end{figure}

To identify SAE features associated with conversational contexts, we follow a conventional interpretability pipeline\cite{Bills2023-dn,Paulo2025-pa,Cunningham2023-sf}. We first run the SAE on a large-scale corpus (SlimPajama-3B), sampling around 50 contexts where each of the 32,768 features activates to ``explain'' the role of each feature. These sampled contexts are then used to characterize the feature as in prior literature\cite{Bills2023-dn,Paulo2025-pa,Cunningham2023-sf}. Using LLM-as-judge classification of these contexts (Gemini-2.5-flash-lite), we compute the \emph{conversation ratio} for each feature---the proportion of feature activations that occur in interpersonal, conversational settings (see \hyperref[fig:fig2]{Fig.~2a} for the distribution across all features). For example, if the conversation ratio is 50\%, then in 50\% of the instances when the feature is activated, it is used for conversation. We focus on features with conversation ratios above 50\% that tend to activate near sentence onsets (i.e., within the first four tokens). From the candidates, we curate feature 30939, summarized as ``a discourse marker for surprise, realization, or acknowledgment'' by Gemini-2.5-Pro, which activates on tokens like ``Oh!'' in contexts involving turn-taking and social exchange (see \hyperref[fig:fig2]{Fig.~2a}). This feature exhibits a conversation ratio of 65.7\%---placing it in the 99th percentile among all features---while maintaining high sparsity (0.016\% of tokens), indicating that it captures a specific conversational phenomenon rather than general linguistic patterns. We select this feature because prior literature suggests that expressions of surprise signal a shift in contrasting perspectives characteristic of social coordination and affiliation\cite{Parkinson2004-sz,Wilkinson2006-cc}.

We examine whether steering this feature causally induces conversational behaviours and improves reasoning accuracy using the activation addition method, which adds scaled feature vectors to model activations during generation. Specifically, we use the Countdown game, a benchmark commonly used to evaluate LLM multi-step reasoning capabilities\cite{Gandhi2025-xn,Gandhi2024-xn}. In the Countdown task, the model must combine a given set of numbers using basic arithmetic operations (+, \(-\), \(\times\), \(\div\)) and parentheses to reach a target value---for example, given inputs 25, 30, 3, 4 and target 32, a valid solution is (30 \(-\) 25 + 3) \(\times\) 4 = 32\cite{Gandhi2025-xn,Gandhi2024-xn}. We use the sample of 1,024 Countdown problems. We prompt the model to generate chain-of-thought reasoning, and at each token generation step, we add the feature 30939 vector (scaled by the steering strength) to layer 15 activations.

As shown in \hyperref[fig:fig2]{Fig.~2b}, steering the conversational surprise feature with positive direction (+10) doubles accuracy from 27.1\% to 54.8\% in the Countdown task, while steering in the negative direction (\(-\)10) reduces accuracy to 23.8\%. The radar plot inset reveals that positive steering (from 0 to +10) simultaneously increases all four conversational behaviours---more question-answering (\emph{$\beta$} = 2.199, 95\% CI = {[}1.648, 2.750{]}, \emph{t}(1023) = 7.83, \emph{p} \textless{} 1\(\times\)10\textsuperscript{-14}), perspective shifts (\emph{$\beta$} = 1.160, 95\% CI = {[}0.665, 1.655{]}, \emph{t}(1023) = 4.60, \emph{p} \textless{} 1\(\times\)10\textsuperscript{-5}), conflict of perspectives (\emph{$\beta$} = 1.062, 95\% CI = {[}0.376, 1.749{]}, \emph{t}(1023) = 3.04, \emph{p} = 0.002), and reconciliation (\emph{$\beta$} = 0.423, 95\% CI = {[}0.349, 0.497{]}, \emph{t}(1023) = 11.21, \emph{p} \textless{} 1\(\times\)10\textsuperscript{-27}), controlling for problem fixed-effects and log-transformed reasoning trace length. Negative steering from 0 to -10 suppresses them, reducing question-answering (\emph{$\beta$} = \(-\)0.831, 95\% CI = {[}\(-\)1.154, \(-\)0.508{]}, \emph{t}(1023) = \(-\)5.05, \emph{p} \textless{} 1\(\times\)10\textsuperscript{-6}), perspective shifts (\emph{$\beta$} = \(-\)0.966, 95\% CI = {[}\(-\)1.262, \(-\)0.670{]}, \emph{t}(1023) = \(-\)6.41, \emph{p} \textless{} 1\(\times\)10\textsuperscript{-9}), conflict of perspectives (\emph{$\beta$} = \(-\)1.347, 95\% CI = {[}\(-\)1.748, \(-\)0.946{]}, \emph{t}(1023) = \(-\)6.60, \emph{p} \textless{} 1\(\times\)10\textsuperscript{-10}), and reconciliation (\emph{$\beta$} = \(-\)0.052, 95\% CI = {[}\(-\)0.103, \(-\)0.001{]}, \emph{t}(1023) = \(-\)1.99, \emph{p} = 0.046). For instance, as shown in \hyperref[tab:edtable1]{Extended Data Table 1}, positive steering (+10) induces reasoning traces where the model actively challenges prior approaches (``Wait, let me see\ldots{} Another idea\ldots''), showing perspective shift and conflicts of perspectives, whereas negative steering (\(-\)10) produces relatively flat, declarative reasoning without internal debate.

To examine whether this effect is specific to conversational features rather than a general property of SAE steering, we compare accuracy improvements across three conditions: (1) steering the conversational surprise feature (Feature 30939), steering a randomly selected conversational feature, and steering a randomly selected non-conversational feature (\hyperref[fig:fig2]{Fig.~2c}). A random conversational feature is defined as \emph{any} feature whose conversation ratio is above the average and tends to activate near sentence onset (i.e., first four tokens), which are more closely associated with conversational styles than other features. All steering strengths are defined as the maximum activation strength across sampled instances of feature activations (SlimPajama-3B), multiplied by 2. The conversational surprise feature produces substantially larger accuracy gains than both random conversational features and non-conversational features (see \hyperref[fig:fig2]{Fig.~2c}). Steering \emph{any} random conversational feature also significantly improves reasoning by 4.17\% more than any random non-reasoning feature (\emph{$\beta$} = 0.042, 95\% CI = {[}0.016, 0.068{]}, \emph{t}(1023)=3.14, \emph{p}=0.002). This specificity suggests that conversational dynamics, rather than arbitrary perturbations to model activations, drive the observed improvements.

We further investigate the mechanism by which conversational steering enhances reasoning. Prior work has identified cognitive behaviours---verification, backtracking, subgoal setting, and backward chaining ---as key contributors to reasoning accuracy in language models\cite{Gandhi2025-xn}. As shown in \hyperref[fig:fig2]{Fig.~2d}, steering feature 30939 toward positive values (0 to +10) systematically increases all four cognitive behaviours: verification (Difference = 5.815, 95\% CI={[}4.922, 6.709{]}, \emph{t}(1023)=12.77, \emph{p} \textless{} 1\(\times\)10\textsuperscript{-34}), backtracking (Difference = 0.881, 95\% CI={[}0.515, 1.248{]}, \emph{t}(1023)=4.72, \emph{p} \textless{} 1\(\times\)10\textsuperscript{-5}), subgoal setting (Difference = 0.621, 95\% CI={[}0.440, 0.803{]}, \emph{t}(1023)=6.72, \emph{p} \textless{} 1\(\times\)10\textsuperscript{-10}), and backward chaining (Difference = 0.809, 95\% CI={[}0.633, 0.985{]}, \emph{t}(1023)=9.02, \emph{p} \textless{} 1\(\times\)10\textsuperscript{-18}) rise monotonically with steering strength. Steering toward negative values (0 to -10) suppresses these behaviours (verification: Difference = -2.302, 95\% CI={[}-2.892, -1.711{]}, \emph{t}(1023)=7.65, \emph{p} \textless{} 1\(\times\)10\textsuperscript{-13}; backtracking: Difference = -1.138, 95\% CI={[}-1.410, -0.867{]}, \emph{t}(1023)=8.24, \emph{p} \textless{} 1\(\times\)10\textsuperscript{-15}; subgoal setting: Difference = -0.171, 95\% CI={[}-0.305, -0.036{]}, \emph{t}(1023)=2.48, \emph{p} = 0.013; backward chaining: Difference = -0.353, 95\% CI={[}-0.487, -0.219{]}, \emph{t}(1023)=5.18, \emph{p} \textless{} 1\(\times\)10\textsuperscript{-6}) based on paired \emph{t}-tests. This suggests that conversational features may improve reasoning, in part, by facilitating the deployment of effective cognitive strategies.

To disentangle direct and indirect effects, we fit a structural equation model to examine the pathways from steering conversational surprise (feature 30939) to accuracy (\hyperref[fig:fig2]{Fig.~2e}). The model indicates that increasing steering feature 30939 from 0 to +10 yields both a significant direct effect on reasoning accuracy (\emph{$\beta$} = .228, 95\% CI = {[}.183, .273{]}, \emph{z}=9.98, \emph{p} \textless{} 1\(\times\)10\textsuperscript{-22}, \emph{N}=2048) and a significant indirect effect mediated by cognitive behaviours (\emph{$\beta$} = .066, 95\% CI = {[}.046, .086{]}, \emph{z}=6.38, \emph{p} \textless{} 1\(\times\)10\textsuperscript{-10}, \emph{N}=2048). Collectively, these findings suggest that conversational features enhance reasoning by directly enabling more effective exploration of the solution space, but also by scaffolding the cognitive strategies that support systematic problem solving.

\subsection*{Diversity of Implicit Perspectives}

Beyond task accuracy, we examine whether DeepSeek-R1 increases the diversity of perspectives expressed within a reasoning trace. In human societies, conversations and socio-emotional role-taking expand the range of viewpoints and domain knowledge brought into problem solving. Differences of perspective give rise to conflict, debate, and resolution. We evaluate whether similar perspective diversity emerges in DeepSeek-R1 by analyzing personality and expertise variation among the distinct reasoning ``perspectives'' participating in each reasoning trace.

We first use an external LLM-as-judge (Gemini-2.5-Pro), prompting it to identify the diversity of implicit conversational perspectives within reasoning traces of DeepSeek-R1, QwQ-32B, and other instruction-tuned models. Specifically, the model infers the number of perspectives underlying each reasoning trace, the personality traits and domain expertise associated with each perspective, and a segmentation of the full reasoning trace by perspective (see \hyperref[sec:implicit_perspectives]{Methods: Implicit Perspectives}). Given a complete reasoning trace, the LLM-as-judge first infers the number of distinct perspectives present, which is shown in \hyperref[fig:fig1]{Fig.~1d}. It then characterizes each perspective's personality traits using the BFI-10 (10-Item Big Five Personality Scale) questionnaire\cite{Rammstedt2007-qg}, along with a short free-form description of the perspective's domain expertise. Finally, the LLM-as-judge attributes each token in the reasoning trace to a specific perspective (i.e., who said this word). Personality diversity is estimated using the standard deviation of inferred personality traits for each Big-5 dimension, while domain expertise diversity is estimated using the mean cosine distance between embedding of each domain expertise description and the average embedding. See \hyperref[sec:implicit_perspectives]{Methods: Implicit Perspectives} and \hyperref[sec:llm_prompts]{Supplementary Method: LLM-as-judge prompts} (``Persona identification'' and ``Persona segmentation'') for details.

For instance, in a chemistry reasoning trace requiring multi-step synthesis analysis, the LLM-as-judge identifies five perspectives, including a critical verifier (low agreeableness, high conscientiousness) who skeptically re-evaluates assumptions, and an expert in making associations (high openness) who recalls analogous reactions. In a creative writing trace where the model rewrites the sentence ``I flung my hatred into the burning fire,'' seven perspectives emerge, including a creative ideator (highest Openness and Extraversion) who generates stylistic alternatives and a semantic fidelity checker (low agreeableness, high neuroticism) who prevents scope creep---``But that adds `deep-seated' which wasn't in the original''. DeepSeek-V3's trace reflects only a single generalist perspective combining all functions without differentiation (see \hyperref[sec:annotations]{Supplementary Methods: Annotation Examples}).

Using the Intelligence Squared Debates Corpus---a dataset of human argumentative conversations (N=1,196 conversations) among two to eight participants---we first validate the accuracy of the LLM-as-judge in identifying distinct voices within a conversation. As shown in \hyperref[fig:edfig5]{Extended Data Fig.~5}, we find that the LLM-as-judge can accurately predict the number of distinct individuals underlying each conversation, even when speaker labels are hidden and the dialogue is concatenated into a single block of text (\hspace{0pt}\hspace{0pt}Spearman's \emph{$\rho$} = 0.86, 95\% CI = {[}0.84, 0.87{]}, \emph{z} = 44.7, \emph{p} \textless{} 1×10\textsuperscript{-323}). We also find that the LLM-as-judge can accurately predict the number of distinct turns (Spearman's \emph{$\rho$} = 0.89, 95\% CI = {[}0.88, 0.90{]}, \emph{z} = 49.2, \emph{p} \textless{} 1×10\textsuperscript{-323}) and correctly attribute each token to a speaker. When there are two speakers, the accuracy is 82\%; for three speakers, 76\%; and for four speakers, 69\%. Accuracy weighted by the predicted number of implicit perspectives underlying LLM reasoning trace is 73\%. Because the Intelligence Squared Debates Corpus includes biographical information about debate participants, we further verify that expertise diversity inferred by LLM-as-judge and embeddings predicts the actual diversity among participants' ground-truth biographies (Spearman's \emph{$\rho$} = 0.55, 95\% CI = {[}0.51, 0.59{]}, \emph{z} = 21.4, \emph{p} \textless{} 1×10\textsuperscript{-97}). Together, these results suggest that LLM-as-judge can capture meaningful diversity patterns in conversational agents that correspond to observed diversity in real human conversations (see \hyperref[par:validation]{Methods: Implicit Perspectives - Validation} for details).

\begin{figure}[!htbp]
    \centering
    \includegraphics[width=\linewidth]{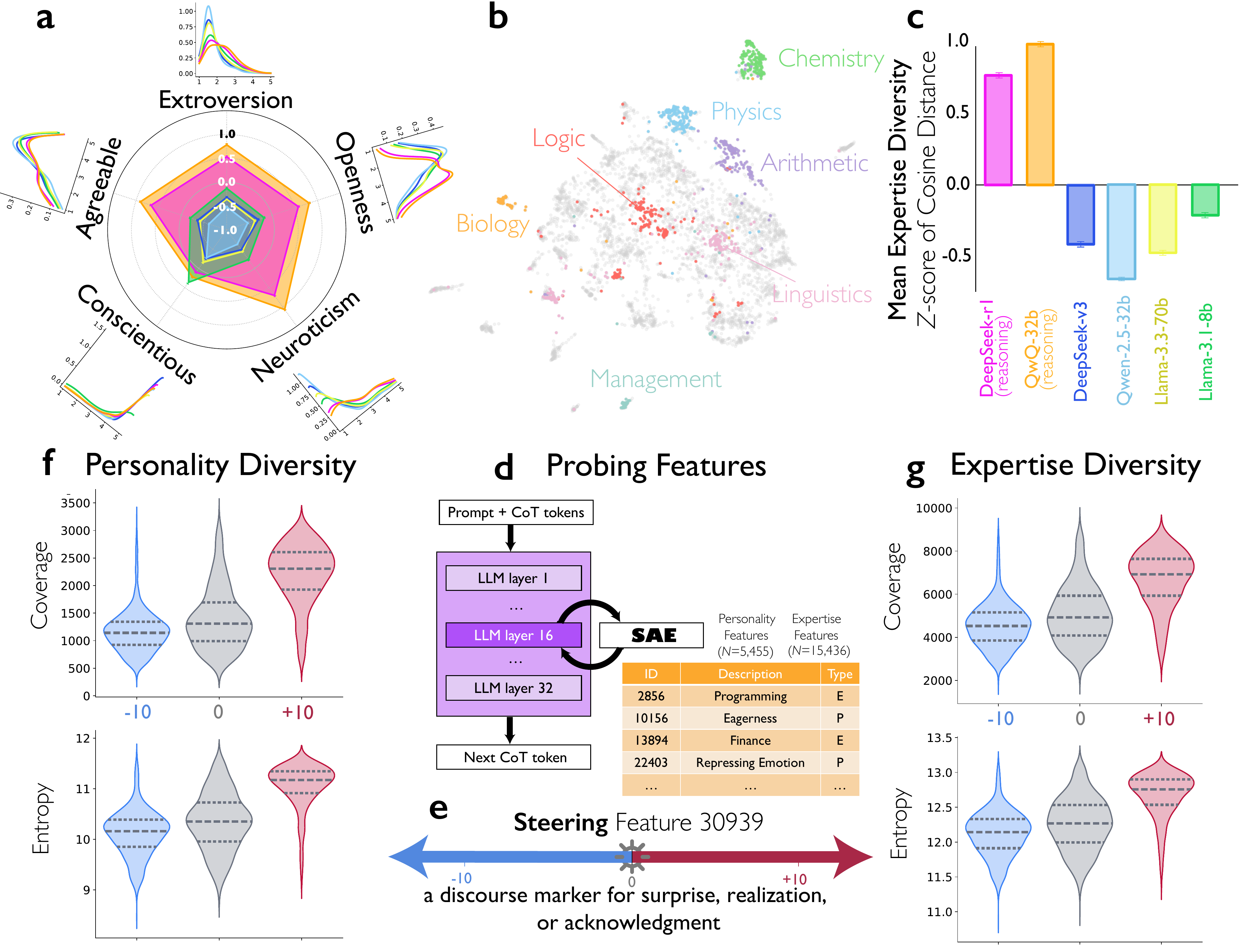}
    \caption{
    \textbf{Personality and expertise diversity in reasoning traces.}
    \textbf{a,} Personality diversity of implicit reasoning perspectives inferred from each reasoning trace using an LLM-as-judge and the BFI-10 (10-Item Big Five Personality Inventory). For each Big Five dimension, diversity is quantified as the standard deviation across inferred personalities. Reasoning models (DeepSeek-R1 and QwQ-32B) exhibit markedly higher diversity in openness, neuroticism, agreeableness, and extraversion. Kernel density estimation (KDE) plots show the distribution of personality traits across reasoning traces.
    \textbf{b,} Embedding space of expertise identified by the LLM-as-judge, projected into two dimensions using UMAP and rendered with an energy-minimization layout, revealing coherent and consistent skill proximities.
    \textbf{c,} Expertise diversity of implicit reasoning perspectives inferred from each reasoning trace, measured as the mean cosine distance between each expertise-related embedding and the centroid of all embeddings in the semantic space. Reasoning models exhibit substantially greater expertise diversity than non-reasoning models.
    \textbf{d,} Sparse autoencoder (SAE) schema and feature identification underlying the steering experiments.
    \textbf{e,} Design of the steering experiment. SAE feature 30939—capturing a discourse marker for surprise, realization, or acknowledgment indicative of persona and perspective shifts—is increased or decreased with a steering strength of 10. Example reasoning traces illustrate that negative steering induces linear chain-of-thought trajectories, no steering yields subtle perspective shifts enabling self-checking, and positive steering induces frequent and pronounced perspective shifts that explore fundamentally different solution strategies.
    \textbf{f, g,} Distributions of coverage and entropy for SAE personality-related (\textbf{f}) and expertise-related (\textbf{g}) features under feature 30939 steering. Error bars indicate 95\% confidence intervals; solid horizontal lines denote medians and dashed lines indicate interquartile ranges (25th–75th percentiles).
    }
    \label{fig:fig3}
\end{figure}

As shown in \hyperref[fig:fig3]{Fig.~3a}, we find that DeepSeek-R1 and QwQ-32B produce significantly higher personality diversity, controlling for the number of perspectives. DeepSeek-R1 shows particularly higher diversity along extraversion (\emph{$\beta$} = 0.103, 95\% CI = {[}0.075, 0.131{]}, \emph{t} = 7.16, \emph{p} \textless{} 1×10\textsuperscript{-13}), agreeableness (\emph{$\beta$} = 0.297, 95\% CI = {[}0.271, 0.323{]}, \emph{t} = 22.65, \emph{p} \textless{} 1×10\textsuperscript{-113}), neuroticism (\emph{$\beta$} = 0.567, 95\% CI = {[}0.542, 0.592{]}, \emph{t} = 44.57, \emph{p} \textless{} 1×10\textsuperscript{-323}), and openness (\emph{$\beta$} = 0.110, 95\% CI = {[}0.083, 0.137{]}, \emph{t} = 8.06, \emph{p} \textless{} 1×10\textsuperscript{-16}), compared to DeepSeek-V3. Similarly, QwQ-32B shows higher diversity in extraversion (\emph{$\beta$} = 0.253, 95\% CI = {[}0.223, 0.282{]}, \emph{t} = 16.78, \emph{p} \textless{} 1×10\textsuperscript{-63}), agreeableness (\emph{$\beta$} = 0.490, 95\% CI = {[}0.462, 0.519{]}, \emph{t} = 34.09, \emph{p} \textless{} 1×10\textsuperscript{-254}), neuroticism (\emph{$\beta$} = 0.825, 95\% CI = {[}0.797, 0.852{]}, t = 58.49, p \textless{} 1×10\textsuperscript{-323}), and openness (\emph{$\beta$} = 0.268, 95\% CI = {[}0.238, 0.298{]}, \emph{t} = 17.41, \emph{p} \textless{} 1×10\textsuperscript{-68}), than Qwen-2.5-32B-IT. In contrast, conscientiousness diversity is lower in DeepSeek-R1 (\emph{$\beta$} = $-$0.291, 95\% CI = {[}$-$0.317, $-$0.265{]}, \emph{t} = $-$21.90, \emph{p} \textless{} 1×10\textsuperscript{-106}) and QwQ-32B (\emph{$\beta$} = $-$0.402, 95\% CI = {[}$-$0.435, $-$0.369{]}, \emph{t} = $-$23.79, \emph{p} \textless{} 1×10\textsuperscript{-125}), suggesting that the reasoning model voices appear more consistently engaged and dutiful. The particularly large effects for agreeableness and neuroticism---traits associated with interpersonal harmony and emotional reactivity---suggest that reasoning models generate perspectives that more frequently disagree with and challenge one another. Interestingly, this pattern aligns with prior literature on human team diversity, which suggests that variability in extraversion and neuroticism enhances team performance, whereas variability in conscientiousness impairs it\cite{Barrick1998-jp,Mello2015-wu}.

We next examine expertise diversity, defined as the dispersion of conversing agents within the embedding space of inferred domain expertise descriptions. For example, when perspectives drawing on what the models judge as expertise in theoretical physics, analytic reasoning, finance, and creative writing co-occur in the same reasoning trace, the mean distance between their expertise embeddings manifests as large (\hyperref[fig:fig3]{Fig.~3b}). As shown in \hyperref[fig:fig3]{Fig.~3c}, DeepSeek-R1 exhibits significantly higher expertise diversity (\emph{$\beta$} = 0.179, 95\% CI = {[}0.161, 0.196{]}, \emph{t} = 20.11, \emph{p} \textless{} 1×10\textsuperscript{-89}) than DeepSeek-V3, and QwQ-32B shows higher expertise diversity (\emph{$\beta$} = 0.250, 95\% CI = {[}0.231, 0.269{]}, \emph{t} = 25.50, \emph{p} \textless{} 1×10\textsuperscript{-142}) than Qwen-2.5-32B-IT, across its implicit reasoning agents than non-reasoning models.

To examine whether the personality- and expertise-related diversity observed in DeepSeek-R1's and QwQ-32B's reasoning traces is reflected in the internal representation space of LLMs, we analyze activations of DeepSeek-R1-Llama-8B's sparse autoencoder (SAE) features. Prior work has shown that high-level persona traits, such as personalities, cultural perspectives, and topics, are linearly represented in LLM activation space and can be steered\cite{Turner2023-ae,Chen2025-pv,Kim2025-qi}. We steer a conversational feature (i.e., Feature 30939; a discourse marker for surprise, realization, or acknowledgment) with strength of +10 or $-$10 inside the activation space of DeepSeek-R1-Llama-8B, and probe how personality- and expertise-related features are activated in the steered reasoning traces (see \hyperref[sec:sae_steering]{Methods: SAE feature steering}).

We first classify each of the 32,768 features as personality-related (e.g., eagerness, expressions of frustration), expertise-related (e.g., programming terminology, financial concepts), or other using an LLM-as-judge approach. We quantify diversity using two complementary measures: coverage, the number of unique personality- or expertise-related features activated across the reasoning trace, and entropy, which captures how evenly activations are distributed across tokens rather than concentrated in a few. Using DeepSeek-R1 reasoning traces, we show that these traces indeed activate more diverse personality-related and expertise-related features, which corroborates our earlier LLM-as-judge results (see \hyperref[fig:edfig6]{Extended Data Fig.~6}). For statistical tests, we control for reasoning trace length and problem fixed effects to show that steering conversational surprise activates genuinely more diverse features rather than simply producing longer outputs.

As shown in \hyperref[fig:fig3]{Fig.~3e--f}, steering with +10 strength causes reasoning traces to activate a wider coverage of both personality-related features (\emph{$\beta$} = 315.915, 95\% CI = {[}277.320, 354.509{]}, \emph{t} = 16.04, \emph{p} \textless{} 1×10\textsuperscript{-323}) and expertise-related features (\emph{$\beta$} = 391.312, 95\% CI = {[}313.743, 468.880{]}, \emph{t} = 9.89, \emph{p} \textless{} 1×10\textsuperscript{-323}) compared to unsteered traces, controlling for reasoning trace length and problem fixed effects. For example, after steering, personality-related features such as ``informal expressions of confusion or frustration'' (Feature 21065), ``phrases related to social interaction and community engagement'' (Feature 26139), and ``references to emotional or sensational themes in narratives'' (Feature 14476) are activated more frequently (see \hyperref[tab:supptable4]{Supplementary Table 4} and \hyperref[tab:supptable5]{5}).

To further examine that this increased diversity reflects a broader distribution of activated features rather than simply generating more tokens, we measure the Shannon entropy of feature activations. Higher entropy indicates that activations are more evenly distributed across diverse features, rather than concentrated in a few dominant ones. Steered traces exhibit higher entropy of both personality-related features (\emph{$\beta$} = 0.262, 95\% CI = {[}0.227, 0.298{]}, \emph{t} = 14.48, \emph{p} \textless{} 1×10\textsuperscript{-323}) and expertise-related features (\emph{$\beta$} = 0.096, 95\% CI = {[}0.075, 0.117{]}, \emph{t} = 9.02, \emph{p} \textless{} 1×10\textsuperscript{-323}) than unsteered traces, confirming that steering induces more diverse feature activations beyond merely increasing output length.

\subsection*{Reinforcement Learning Experiments}

To further examine whether LLMs self-reinforce conversational behaviours when rewarded for correct answers, we implement a self-taught reinforcement learning (RL) experiment. In this setup, the model explores solution strategies for the Countdown arithmetic puzzle game\cite{Gandhi2025-xn,Gandhi2024-xn}, where the model must combine a given set of numbers using basic arithmetic operations (+, $-$, ×, ÷) and parentheses to reach a target. We also replicate these findings on political misinformation detection, where models discriminate between true and fabricated political headlines.

Following the reward architecture of DeepSeek-R1\cite{Guo2025-jz}, we reward accuracy and correct format (i.e., wrapping reasoning between \textless think\textgreater{} and \textless/think\textgreater{} tags and answers between \textless answer\textgreater{} and \textless/answer\textgreater{} tags) with a simple weighted reward: accuracy × 0.9 + format × 0.1. Crucially, we do not directly reward conversational or cognitive behaviours. We implement Proximal Policy Optimization (PPO)\cite{Schulman2017-fb} using the Verl framework\cite{Sheng2024-oh}, training for 250 steps (see \hyperref[tab:supptable6]{Supplementary Table 6} for hyperparameters). We use Qwen-2.5-3B, a pre-trained model without any instruction-tuning, prompted to solve the Countdown task with a chain of thought (see \hyperref[sec:rl_experiments]{Methods: Reinforcement learning experiments}).

\begin{figure}[!htbp]
    \centering
    \includegraphics[width=\linewidth]{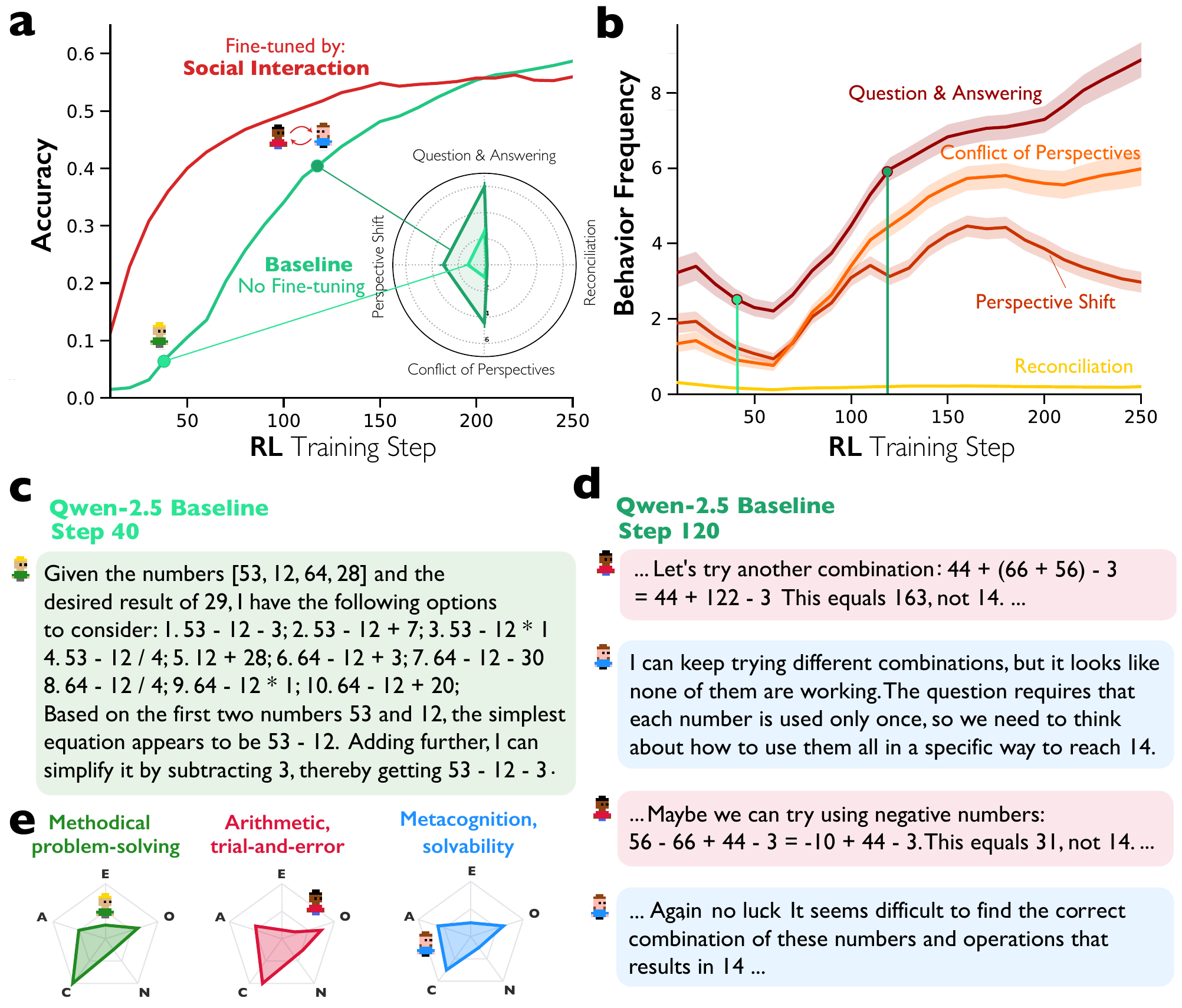}
    \caption{
    \textbf{Occurrence of social behaviours in accuracy-rewarded reinforcement learning and the effect of fine-tuning with conversational scaffolding.}
    \textbf{a,} Comparison of the accuracy trajectory of reinforcement learning rewarded with problem-solving accuracy for the baseline Qwen-2.5-3B model and the same model initially fine-tuned to simulate social interaction through multi-agent dialogue generated by Qwen-2.5-32B. The socially initialized model reaches maximum accuracy more rapidly, whereas the baseline model eventually catches up and does so by adopting conversational behaviours, including questioning and answering, perspective shifts, and perspective conflict.
    \textbf{b,} Trajectories of individual conversational behaviours within the reinforcement-learned baseline model from panel a. Question-and-answer behaviour emerges first, followed by perspective shifts and conflicts, which rise in close synchrony. Reconciliation behaviour shows little increase, suggesting that individual approaches compete rather than forming an effective ensemble. Lines are smoothed using an exponential moving average (span = 9), and shaded regions indicate 95\% confidence intervals.
    \textbf{c--d,} Comparison of the Qwen-2.5 baseline model at training step 40 versus step 120. At step 40, the model primarily engages in linear chain-of-thought reasoning, whereas by step 120, two distinctive simulated personas have emerged that explicitly recognize their collectivity through the use of the pronoun ``we''.
    \textbf{e,} Personality profiles inferred by the LLM-as-judge. The step-40 model exhibits a strong all-around problem-solving profile, characterized by high conscientiousness, moderately high openness and agreeableness, lower extraversion, and notably low neuroticism. In contrast, the two collaborative agents observed at step 120 display differentiated personality profiles: one emphasizes trial-and-error problem solving, while the other specializes in metacognitive reasoning about problem solvability across alternative approaches. The trial-and-error agent is less extraverted and more agreeable than the step-40 agent, whereas the solvability-focused agent is more open and substantially less conscientious.
    }
    \label{fig:fig4}
\end{figure}

We first examine whether conversational behaviours spontaneously increase despite not being directly rewarded. \hyperref[fig:fig4]{Fig.~4a} presents the results, showing that accuracy improves substantially over training, rising from near zero at baseline to approximately 58\% by step 250. \hyperref[fig:fig4]{Fig.~4b} reveals that the frequency of conversational behaviours---particularly Question \& Answering and Conflict of Perspectives---rise throughout training despite receiving no direct reward. Perspective shifts also increase until approximately step 160, although they start to decrease as the model becomes able to reach answers with fewer shifts across the training phase. \hyperref[fig:fig4]{Fig.~4c-d} illustrate this qualitative shift: at step 40, the model produces mechanical, enumerative chain-of-thought-style reasoning, whereas by step 120, two distinctive simulated personas have appeared, recognizing their collectivity with the pronoun ``we''---expressing uncertainty (``Again no luck''), considering alternatives (``Maybe we can try using negative numbers''), and reflecting on problem constraints. As shown in \hyperref[fig:fig4]{Fig.~4e}, these behaviours occur while the model employs two distinct personas according to LLM-as-judge evaluation: a methodical problem-solver high in Conscientiousness and low in Openness, and an exploratory trial-and-error thinker high in Openness and Extraversion, with metacognitive reflection on solvability---marked by Neuroticism---mediating between the two. Similar to our earlier findings based on sparse autoencoders, the increase of these behaviours has co-occurred with the increase of other cognitive behaviours, such as verification and backtracking (\hyperref[fig:edfig7]{Extended Data Fig.~7}).

To corroborate the role of conversational behaviours in reasoning improvement, we compare RL training under three conditions: (1) Baseline (RL only, no priming), (2) Conversation fine-tuning (supervised fine-tuning on multi-agent dialogue text before RL), and (3) Monologue fine-tuning (fine-tuning on monologue-like, step-by-step reasoning traces before RL). To generate conversational fine-tuning data, we prompt Qwen-2.5-32B-IT to produce multi-agent-like dialogues with two, three, or four distinct personas solving 8,262 reasoning tasks (see \hyperref[sec:data]{Methods: Data}), and sample 600 instances that reach correct answers (500 for training, 100 for validation). In these dialogues, the model first defines distinct personas with different personality traits and expertise (e.g., \textless persona1\textgreater{} a meticulous mathematician with a strong background in number theory \textless/persona1\textgreater, \textless persona2\textgreater{} a quick-witted and intuitive problem solver... not afraid to challenge assumptions \textless/persona2\textgreater). These personas then engage in turn-taking dialogue where they build on, question, and correct each other's reasoning (e.g., \textless think1\textgreater{} We can discard (2, 7) because... they are not coprime. \textless/think1\textgreater{} \textless think2\textgreater{} Wait a second. We can't discard (2, 7) just yet... they are indeed coprime because their greatest common divisor is 1. \textless/think2\textgreater{} → \textless think1\textgreater{} You're right. I overlooked that. \textless/think1\textgreater), before converging on a final answer in \textless group\_solution\textgreater{} \ldots{} \textless/group\_solution\textgreater.

For monologue fine-tuning data, we generate standard chain-of-thought traces for the ``same problems'' with correct answers, where a single voice reasons within \textless think\textgreater{} \ldots{} \textless/think\textgreater{} tags (e.g., \textless think\textgreater{} Since the GCD of m and n is 8, we can express m and n as 8a and 8b respectively, where a and b are coprime... The pairs of factors of 14 are (1, 14) and (2, 7). \textless/think\textgreater). \hyperref[tab:supptable7]{Supplementary Table 7} presents full examples for both types of fine-tuning data. We then fine-tune Qwen-2.5-3B on these datasets using standard next-token prediction loss wherein the models learn to reproduce the full output sequence (persona definitions, turn-by-turn reasoning or monologue trace, and final answer) given only the problem as input. This priming phase familiarizes the model with conversational versus monologue formats before RL optimizes for task accuracy (see \hyperref[tab:supptable8]{Supplementary Table 8} for SFT hyperparameters).

\hyperref[fig:edfig8]{Extended Data Fig.~8} shows that models fine-tuned on conversational data achieve faster accuracy gains than monologue-fine-tuned models, particularly in the early stages of training. At step 40, conversation-fine-tuned Qwen-2.5-3B models reach approximately 38\% accuracy while monologue-fine-tuned models remain at 28\%. This pattern replicates across architectures: in Llama-3.2-3B (see \hyperref[sec:llama_replications]{Supplementary Methods: Replications on Llama-3.2-3B}), the conversation-fine-tuned model reaches 11\% accuracy at step 70 compared to just 5\% for monologue-fine-tuned models. Interestingly, in Llama-3.2-3B, the divergence becomes more striking as training progresses. By step 150, conversation-fine-tuned Llama models achieve 40\% accuracy while monologue-fine-tuned models plateau around 18\%, less than half the performance. Notably, both conditions are trained on identical problems and correct answers, yet conversation-fine-tuned models consistently improve faster and reach higher asymptotic accuracy. This indicates that conversational structure itself, not merely exposure to correct solutions or task-related knowledge, drives the improvement.

We further test whether conversational scaffolding transfers across domains. Models fine-tuned on multi-agent dialogues for the Countdown task are evaluated on a qualitatively different task: political misinformation detection, where models discriminate between true and fabricated headlines from 23,299 fact-checked claims from PolitiFact. Despite never encountering this domain during fine-tuning, conversation-primed models achieve faster accuracy gains than baseline models (see \hyperref[sec:cross_domain]{Supplementary Methods: Cross-domain reasoning transfer} and \hyperref[fig:edfig9]{Extended Data Fig.~9}). Together, these results suggest that conversational structure facilitates the emergence of reasoning strategies during RL.

\phantomsection
\section{Discussion}\label{sec:discussion}

Our findings suggest that reasoning models like DeepSeek-R1 do not simply generate longer or more elaborate chains of thought. Rather, they exhibit patterns characteristic of a social and conversational process generating ``societies of thought''---posing questions, introducing alternative perspectives, generating and resolving conflicts, and coordinating diverse socio-emotional roles. These interactional patterns rarely occur in non-reasoning models across different model sizes (671B, 70B, 32B, 8B), even when controlling for reasoning trace length, suggesting that reasoning optimization introduces an intrinsic social structure within the reasoning process itself rather than merely increasing text volume. The model appears to reason by simulating internal societies, structuring thought as an exchange among interlocutors rather than as a single uninterrupted voice. The implication here is that social reasoning emerges autonomously through RL as a function of its consistent ability to produce correct answers, rather than through explicit human supervision or fine-tuning.

This structure does not appear to be merely stylistic. Conversational behaviours and socio-emotional roles are more frequently activated when DeepSeek-R1 faces more difficult problems, and they explain a substantial portion of the accuracy advantage over non-reasoning models. Steering experiments provide evidence that conversational markers are tied to reasoning performance. When we amplify a feature associated with conversational surprise---a discourse marker signaling perspective shift and constrast---accuracy on multi-step reasoning tasks doubles. Structural equation modeling reveals that conversational steering is associated with accuracy through both direct effects and indirect pathways mediated by cognitive strategies previously identified as central to reasoning, including verification, backtracking, subgoal setting, and backward chaining. This suggests that the social structure of reasoning might not be epiphenomenal but mechanistically implicated in how the model explores solution spaces and deploys effective problem-solving strategies.

We further find that this interactional organization is supported by diversity among multiple implicit ``voices'' within reasoning traces. These voices vary systematically in personality traits and domain expertise, and mechanistic interpretability analyses corroborate that models activate more diverse personality- and expertise-related features when steered toward conversational markers. This pattern suggests that findings from human team research---where diversity in socially oriented traits such as extraversion and neuroticism enhances collective performance, whereas diversity in task-oriented traits such as conscientiousness can impair coordination and efficiency\cite{Barrick1998-jp,Mello2015-wu}---may offer a useful lens for interpreting language models' collective reasoning behaviours. Most R1 reasoning personas were surprisingly disciplined and hard-working!

Reinforcement learning (RL) experiments further support the functional role of conversational structure. Models fine-tuned on multi-agent dialogues learn to reason more effectively than models fine-tuned only on correct, monologue-like reasoning traces. The benefit therefore lies not in the correctness of initial reasoning but in the procedural scaffolding provided by conversational organization. Although these experiments used relatively small 3B-parameter models (Qwen-2.5-3B and Llama-3.2-3B) on simple arithmetic tasks and misinformation detection tasks, the results suggest that even minimal social structuring within reasoning traces can accelerate the emergence of generalizable reasoning behaviour.

Collectively, these findings suggest the benefits of studying ``social scaling'' in reasoning-optimized models. As their test-time computations expand, reasoning traces evolve from isolated monologues into structured dialogues among differentiated internal perspectives. High-performing reasoning thus seems to depend on how attention, role-taking, and conflict resolution are coordinated within emergent ``societies of thought.'' Our goal is not to take sides on whether reasoning model traces should be regarded as discourse among simulated human groups or a computational mind's simulation of such discourse. Indeed, as we note above, even this distinction becomes fundamentally unclear as some theories of cognition posit how mature individual minds develop from simulations of multi-agent interaction. Nevertheless, alignments between our findings on successful reasoning models and prior literature on successful human teams (e.g., diverse personality traits lead to successful collaborations) suggest that principles governing effective group collaboration may offer valuable insights for interpreting and engineering reasoning behaviours in language models. This perspective extends long-standing research on human team collaboration, where group composition and diversity are known to shape collective intelligence through variations in personality and expertise\cite{Hong2004-jn,Nemeth2001-pl,Engel2014-jh,DeChurch2010-ni,Page2019-ux,Trouche2014-pg,Barrick1998-jp} . Analogous dynamics within AI systems remain largely unexplored. Early investigations of human--AI collaboration\cite{Ju2025-nx} have begun to characterize this emerging domain, but how diversity and coordination operate within the reasoning traces of large language models remains an open question. DeepSeek-R1's and QwQ's internal reasoning patterns suggest that such models may already self-organize a productive heterogeneity of perspectives, implying that diversity could be as fundamental to artificial reasoning as it is to human collaboration and collective dominance.

A growing trend in AI involves agentic architectures that deploy multiple agents engaged in more complex configurations than single-channel debate, including hierarchy, complex networks and even entire institutions of interacting agents\cite{Yang2025-mp,Liu2023-ia,Eo2025-si,Zhang2024-iq,Chen2023-xa,Du2023-xm,Hu2024-wk,Liang2024-jj}. Our work suggests the importance of exploring alternative structures, but also inhabiting them with diverse perspectives, personalities, and specialized expertise that drive complementarity and collective success in the human social world. Understanding how diversity and social scaffolding interact could shift how we conceptualize large language models, from solitary problem-solving entities toward collective reasoning architectures, where intelligence arises not merely from scale but the structured interplay of distinct voices.

\newpage

\phantomsection
\section{Methods}\label{sec:methods}

\phantomsection
\subsection{Data}\label{sec:data}

We generate chains of thought and final answers for 8,262 reasoning problems spanning symbolic logic, mathematical problem solving, scientific reasoning, instruction following, and multi-agent inference. The benchmark suite includes BigBench Hard (BBH) tasks requiring multi-step logical inference, reference tracking, and compositional reasoning; GPQA (Graduate-level Physics Question Answering) for graduate-level STEM reasoning; MATH (Hard) subset for multi-step derivations across algebra, geometry, probability, and number theory; MMLU-Pro for advanced conceptual knowledge; IFEval for instruction-following consistency; and MUSR (Mathematics Understanding and Symbolic Reasoning) for symbolic manipulation and structured mathematical reasoning (see \hyperref[tab:supptable9]{Supplementary Table 9} for details).

We generate responses using six models: two reasoning models---DeepSeek-R1-0528 (671B parameters) and QwQ-32B---and four instruction-tuned models---DeepSeek-V3-0324 (671B parameters), Qwen-2.5-32B-Instruct, Llama-3.3-70B-Instruct, and Llama-3.1-8B-Instruct---under a zero-shot setting. DeepSeek-V3 is the instruction-tuned model based on DeepSeek-V3-Base from which DeepSeek-R1 is derived through reinforcement learning, and Qwen-2.5-32B-Instruct is the instruction-tuned model based on Qwen-2.5-32B from which QwQ-32B is derived. For brevity, we refer to these models as DeepSeek-R1, QwQ-32B, DeepSeek-V3, Qwen-2.5-32B-IT, Llama-3.3-70B-IT, and Llama-3.1-8B-IT, respectively. We set the temperature to 0.6, a temperature recommended for standard reasoning tasks\cite{Guo2025-jz}.

\phantomsection
\subsection{Measurements}\label{sec:measurements}

\phantomsection
\paragraph*{Conversational Behaviours}\label{par:conversational_behaviours}

We identify four conversational behaviours in reasoning traces using an LLM-as-judge approach with Gemini-2.5-Pro. (1) Question--answering is defined as sequences where a question is posed and later answered, as in conversations (e.g., ``Why...? Because...'', ``What if...? Then\ldots'', ``How do we know? Well\ldots'', and ``Let's try X...? This gives us Y''). (2) Perspective shift is defined as a transition to a different idea, viewpoint, assumption, or approach, as in conversations. (3) Conflict of perspectives is defined as expressions of disagreement, correction, or tension with another perspective (e.g., ``Wait, that can't be right\ldots'', ``No, actually\ldots'', and ``This contradicts\ldots''). (4) Reconciliation is defined as instances where conflicting views are integrated or resolved into a coherent synthesis (e.g., ``So perhaps both are true if\ldots'', ``Combining these insights\ldots'', and ``This resolves the tension\ldots''. For each reasoning trace, the LLM-as-judge counts the number of distinct instances of each behaviour, returning integer counts (0 if none are present). The full prompt is provided in \hyperref[sec:llm_prompts]{Supplementary Methods: LLM-as-judge prompts}. See \hyperref[tab:supptable10]{Supplementary Table 10} for the descriptive statistics of conversational behaviors for reasoning and instruction-tuned models.

For the four conversational categories, Gemini-2.5-Pro and GPT-5.2 demonstrated substantial agreement: Question-and-Answering (ICC(3,1) = .856), Perspective Shift (ICC(3,1) = .849), Conflict of Perspectives (ICC(3,1) = .912), and Reconciliation (ICC(3,1) = .804), with a mean ICC of .855. Gemini-2.5-Pro also showed agreement with human ratings across the four conversational categories: Question-and-Answering (ICC(3,1) = .634), Perspective Shift (ICC(3,1) = .737), Conflict of Perspectives (ICC(3,1) = .864), and Reconciliation (ICC(3,1) = .664).

\paragraph*{Socio-Emotional Roles}

We analyze the presence of socio-emotional roles within reasoning traces using Bales' Interaction Process Analysis (IPA) framework\cite{Bales1950-oa}. The IPA classifies utterances into 12 interaction roles, each operationally defined in the prompt with specific behavioural descriptions. The LLM-as-judge (Gemini-2.5-Pro) counts the number of distinct instances of each of the 12 categories separately, and we aggregate these counts into four higher-level categories for our main analyses:

\begin{itemize}
\item
  Information-giving roles:

  \begin{itemize}
  \item
    Gives suggestion (gives direction, implying autonomy; e.g., should..., need to..., let us...)
  \item
    Gives opinion (gives evaluation, analysis, expresses feeling or wish)
  \item
    Gives orientation (provides objective or verifiable information, repeats, clarifies, confirms)
  \end{itemize}
\item
  Information-asking roles:

  \begin{itemize}
  \item
    Asks for suggestion (requests possible ways of action or direction)
  \item
    Asks for opinion (requests evaluation, analysis, or expression of feeling)
  \item
    Asks for orientation (requests information, repetition, or confirmation)
  \end{itemize}
\item
  Positive emotional roles:

  \begin{itemize}
  \item
    Shows solidarity (raises other's status, gives help, reward)
  \item
    Shows tension release (jokes, laughs, shows satisfaction)
  \item
    Agrees (shows passive acceptance, understands, concurs, complies)
  \end{itemize}
\item
  Negative emotional roles:

  \begin{itemize}
  \item
    Shows antagonism (deflates other's status, defends or asserts self)
  \item
    Shows tension (expresses uncertainty, asks for help, withdraws from the field)
  \item
    Disagrees (shows passive rejection, formality, or withholds help)
  \end{itemize}
\end{itemize}

Inter-rater reliability is substantial for the four higher-level IPA categories used in our main analyses: Ask (Gemini-2.5-Pro vs. GPT-5.2: ICC(3,1) = .939; Gemini-2.5-Pro vs. Human: ICC(3,1) = .836), Give (ICC(3,1) = .864; ICC(3,1) = .666), Positive (ICC(3,1) = .939; ICC(3,1) = .870), and Negative (ICC(3,1) = .838; ICC(3,1) = .779). See \hyperref[tab:supptable10]{Supplementary Table 10} for the descriptive statistics of socio-emotional roles for reasoning and instruction-tuned models.

To measure whether socio-emotional roles co-occur reciprocally within reasoning traces, we compute the Jaccard index for two role pairs: (1) asking versus giving for task-oriented roles, and (2) positive versus negative for emotional roles. The Jaccard index is defined as the number of reasoning traces containing both roles in a pair divided by the number of reasoning traces containing either role, capturing whether models coordinate complementary roles within the same trace rather than deploying them in isolation. Higher Jaccard indices indicate more balanced, dialogue-like interaction patterns, whereas lower indices suggest one-sided, monologic reasoning.

\paragraph*{Cognitive Behaviours}

We identify four cognitive behaviours previously established as contributors to reasoning accuracy in language models using Gemini-2.5-Pro as LLM-as-judge\cite{Guo2025-jz,Gandhi2025-xn,Wang2025-kp,Chen2025-cu,Yeo2025-sh}. For the measurement, we adopt the prompt and examples used by Gandhi and colleagues (2025)\cite{Gandhi2025-xn}, which has been verified by multiple human raters. Each behaviour is operationally defined in the prompt with specific examples to guide annotation: Verification is defined as instances where the chain-of-reasoning explicitly checks the current result against the target solution. The prompt provides specific examples: ``This sequence results in 1, which is not equal to 22'' and ``Since 25 is not equal to 22''. Backtracking is defined as instances where the model realizes a path won't work and explicitly goes back to try a different approach. Subgoal setting is defined as instances where the model breaks down the problem into smaller, intermediate goals. Backward chaining is defined as instances where the model starts from the target solution and works backwards to the initial problems. 

For the four cognitive reasoning behaviors, Gemini-2.5-Pro and GPT-5.2 demonstrate good to excellent agreement: Answer Verification (ICC(3,1) = .995), Backtracking (ICC(3,1) = .829), Subgoal Setting (ICC(3,1) = .810), and Backward-Chaining (ICC(3,1) = .756), with a mean ICC of .848. Gemini-2.5-Pro also shows substantial agreement with a human rater across the four cognitive behaviors: Answer Verification (ICC(3,1) = .981), Backtracking (ICC(3,1) = .921), Subgoal Setting (ICC(3,1) = .559), and Backward-Chaining (ICC(3,1) = .578), with a mean ICC of .760. These reliability estimates are computed on 30 reasoning traces to solve general-purpose reasoning problems (see \hyperref[sec:data]{Methods: Data}) and 50 reasoning traces generated during reinforcement learning of Qwen-2.5-3B (see \hyperref[sec:rl_experiments]{Methods: Reinforcement Learning Experiments}).

\paragraph*{Problem Complexity}

We measure problem complexity using two complementary approaches. First, we use LLM-as-judge ratings. Gemini-2.5-Pro rates each problem on a 7-point Likert scale. The prompt instructs the model to rate the intrinsic difficulty of the problem for a capable language model under zero-shot conditions using the following scale: 1 = very easy, 2 = easy, 3 = somewhat easy, 4 = moderate, 5 = somewhat difficult, 6 = difficult, 7 = very difficult. The full prompt is provided in \hyperref[sec:llm_prompts]{Supplementary Methods: LLM-as-judge prompts}. Second, we use empirical error rates. We compute the number of incorrect answers across four instruction-tuned models (DeepSeek-V3, Qwen-2.5-32B-IT, Llama-3.3-70B-IT, Llama-3.1-8B-IT), yielding a score from 0 to 4 representing the number of models that failed to answer correctly. After sampling and annotating 50 reasoning problems, we find substantial inter-rater reliability between Gemini-2.5-Pro and GPT-5.2 (ICC(3,1) = .745). Gemini-2.5-Pro's complexity scores show strong correlation with non-reasoning models' error rates (Spearman's \emph{$\rho$} = 0.526, 95\% CI = {[}0.508, 0.543{]}, \emph{z} = 46.26, \emph{p} \textless{} 1×10\textsuperscript{-323}, \emph{N} = 7,738), confirming convergent validity between the two measures.

\phantomsection
\subsection{Statistical Analyses}\label{sec:statistical_analyses}

To estimate whether observed differences between reasoning models (DeepSeek-R1 and QwQ-32B) and instruction-tuned baselines arise from conversational behaviours or socio-emotional roles rather than from task heterogeneity or reasoning trace length, we estimate the following linear probability model for each behavioural outcome \emph{Y\textsubscript{ij}}, which is a binary variable. 

\[Y_{ij} = \sum_{m \in M}^{}{}\beta_{m}{Model}_{m,ij} + \gamma log({Len}_{ij}) + \alpha + \mu_{i} + \varepsilon_{ij}\]

where \emph{i} indexes individual task problems, and \emph{j} indexes individual reasoning traces generated by different models. \emph{Y\textsubscript{ij}} equals 1 if the reasoning trace \emph{j} exhibits the behavior more than once. \({Model}_{m,ij}\) is a categorical dummy variable that equals 1 if reasoning trace \emph{j} for problem \emph{i} is generated by model \emph{m}, and 0 otherwise, where $M = \{ DeepSeek\_ r1,\ DeepSeek\_ v3,QwQ\_ 32b,$
$Qwen\_ 2.5\_ 32b\_ it,\ Llama\_ 3.3\_ 70b\_ it$ $\ Llama\_ 3.1\_ 8b\_ it\}$ Either DeepSeek-V3 or Qwen-2.5-32B-IT serves as the reference category and is excluded, such that each coefficient \emph{$\beta$\textsubscript{m}} represents the marginal difference in the outcome relative to DeepSeek-V3 or Qwen-2.5-32B-IT. Log(\emph{Len\textsubscript{ij}}) denotes the reasoning trace length (i.e., the number of words in each reasoning trace), adjusting the extreme skewness of reasoning trace length (see \hyperref[fig:edfig1]{Extended Data Fig.~1}). \(\mu_{i}\) represents task fixed effects at the individual problem level, absorbing all variation associated with each problem's intrinsic difficulty, phrasing, and topical content. This ensures that comparisons between models are made within the same problem rather than across heterogeneous tasks. \(\alpha\) is the intercept. Robust standard errors are clustered at the task level to account for within-task correlation. Models are estimated using StataNow/SE 19.5.

\phantomsection
\subsection{SAE Feature Steering}\label{sec:sae_steering}

To investigate the role of conversational behaviours in reasoning, we employ sparse autoencoders (SAEs) to identify and manipulate interpretable features in the model's activation space. SAEs decompose neural network activations into a sparse set of linear features, enabling targeted intervention on specific behavioural dimensions without altering model weights\cite{Cunningham2023-sf,Bricken_undated-kd,Templeton2024-dx}.We use an SAE trained on Layer 15's residual stream activations of DeepSeek-R1-Llama-8B (15-llamascope-slimpj-res-32k). The SAE has been trained on SlimPajama, a general-purpose corpus containing both conversational and non-conversational texts, with a dictionary size of 32,768 features (see \hyperref[tab:supptable3]{Supplementary Table 3} for hyperparameters).

To identify features associated with conversational contexts, we follow a standard interpretability pipeline. For each of the 32,768 features, we sample approximately 50 contexts from the pre-training corpus where the feature activates most strongly. We then use an LLM-as-judge classifier (Gemini-2.5-flash-lite) to determine whether each activation context represents a conversational setting, computing a conversation ratio for each feature---the proportion of activations occurring in conversational contexts. We apply two filtering criteria: (1) conversation ratio above 50\%, and (2) activation near sentence onsets 50\% or more (within the first four tokens). From the candidate features, we select Feature 30939, which the LLM judge summarized as ``a discourse marker for surprise, realization, or acknowledgment.'' This feature activates on tokens such as ``Oh!'' in contexts involving turn-taking and social exchange. Feature 30939 exhibits a conversation ratio of 65.7\% (99th percentile among all features) while maintaining high sparsity (0.016\% of tokens), indicating specificity to conversational phenomena rather than general linguistic patterns.

We implement activation addition to steer Feature 30939 during generation. At each token generation step, we add the feature's decoder vector, scaled by a steering strength \emph{s}, to the model's Layer 15 residual stream activations.

\[{h'}_{t} = h_{t} + s \cdot d_{30939}\]

where \(h_{t}\) denotes the original activation at token position \emph{t} and \emph{d}\textsubscript{30939} denotes the decoder vector for Feature 30939. We first generate reasoning traces under seven steering conditions: \(s \in \{ - 15, - 10, - 5,0,5,10,15\}\). As \(s \in \{ - 15,15\}\) exhibits lower accuracy due to excessive steering, we use \(s \in \{ - 10, - 5,0,5,10\}\).

\paragraph*{Reasoning Task and Evaluation}

We evaluate reasoning performance using the Countdown task, a benchmark for multi-step arithmetic reasoning commonly used to evaluate LLM reasoning capabilities. In each problem, the model must combine a set of input numbers using basic arithmetic operations (+, $-$, ×, ÷) and parentheses to reach a target value. For example, given inputs \{25, 30, 3, 4\} and target 32, a valid solution is (30 $-$ 25 + 3) × 4 = 32. We use 1,024 problems. We use the following prompt template: Using the numbers {[}79, 17, 60{]}, create an equation that equals 36. You can use basic arithmetic operations (+, -, *, /) and each number can only be used once. Show your work in \textless think\textgreater{} \textless/think\textgreater{} tags. And return the final answer in \textless answer\textgreater{} \textless/answer\textgreater{} tags, for example \textless answer\textgreater{} (1 + 2) / 3 \textless/answer\textgreater. Solutions are scored as correct if the final numerical answer matches the target value, evaluated using Gemini-2.5-flash-lite.

After steering, we measure the frequency of conversational behaviours (question--answering, perspective shift, conflict of perspectives, reconciliation) and cognitive behaviours (verification, backtracking, subgoal setting, backward chaining) in each generated reasoning trace using the LLM-as-judge procedures described above. To estimate behavioural differences across steering conditions, we estimate fixed-effects linear regression models for each behavioural count variable

\[Y_{ij} = \sum_{s \in \{ - 10, - 5,0,5,10\}}^{}{}\beta_{s}{Steer}_{s,ij} + \gamma log({Len}_{ij}) + {\alpha + \mu}_{i} + \varepsilon_{ij}\]

where \emph{i} indexes individual task problems, and \emph{j} indexes individual reasoning traces generated before or after steering. \({Steer}_{ij}\) is a categorical variable indicating steering strength. \(\log({Len}_{ij})\) denotes the log-transformed reasoning trace length (i.e., the number of words in each reasoning trace), adjusting the extreme skewness of reasoning trace length (see \hyperref[fig:edfig1]{Extended Data Fig.~1}). \(\mu_{i}\) represents task fixed effects at the individual problem level, absorbing all variation associated with each problem's intrinsic difficulty.

\paragraph*{Experiment Conditions}

To assess whether accuracy improvements are specific to conversational features rather than a general property of SAE steering, we compare three conditions: (1) Feature 30939 (conversational surprise), (2) randomly selected conversational features (conversation ratio above mean and activating near sentence onset), and (3) randomly selected non-conversational features (conversation ratio below mean). For condition (1), we evaluate all 1,024 Countdown problems. For conditions (2) and (3), we randomly sample 300 features from each category and have each feature solve 16 randomly selected problems, yielding a distribution of accuracy scores across features within each condition. For all conditions, steering strength is set to twice the maximum activation strength observed for that feature across sampled instances in SlimPajama-3B. We test for differences in accuracy between conditions using linear regression with problem fixed effects.

\paragraph*{Feature Diversity} 

To examine whether steering conversational features induces greater activation of personality- and expertise-related features in the model's internal representation space, we analyze sparse autoencoder (SAE) feature activations before and after steering. We use the same SAE employed in the steering experiments: an SAE trained on Layer 15's residual stream activations of DeepSeek-R1-Llama-8B with a dictionary size of 32,768 features. For each reasoning trace generated under different steering conditions (s $\in$
 \{$-$10, 0, +10\}), we record which SAE features are activated at each token position by passing the Layer 15 activations through the SAE encoder.

For each of the 32,768 SAE features, Neuronpedia provides a textual description generated by prompting GPT-4o-mini with the top-activating token sequences. Using Gemini-2.5-flash-lite, we classify each feature description into one of three categories: personality-related, expertise-related, or other. Gemini-2.5-flash-lite first scores each feature from 0 to 100 based on whether it is related to personality traits or domain expertise (see \hyperref[sec:llm_prompts]{Supplementary Methods: LLM-as-judge prompts}). Then, we use the threshold of 50 to determine whether they are personality or expertise features. Among the 32,768 features, 5,455 are labeled as personality-related (e.g., eagerness, expressions of frustration) and 15,436 as expertise-related (e.g., programming terminology, financial concepts).

For each reasoning trace, we compute two complementary measures of feature diversity within each category (personality-related or expertise-related). First, coverage is defined as the number of unique features within a category (personality or expertise) that exhibit non-zero activation across all tokens in a reasoning trace. Second, entropy is computed over the distribution of token counts that activate each SAE feature within a given category. Each reasoning trace is represented as a 32,768-dimensional vector, where each element corresponds to the number of tokens within the reasoning trace that activated a specific SAE feature. Entropy \emph{H} for category \emph{c} (personality or expertise) was then calculated as:

\[H_{c} = - \sum_{f \in c}^{}{}a_{f}\log(a_{f})\]

where \(a_{f}\) \hspace{0pt} denotes the number of activating tokens for feature $f$ within category $c$.

Higher coverage indicates that the reasoning trace draws on a broader range of personality- or expertise-related features. Higher entropy indicates that activations are more evenly distributed across features rather than concentrated in a few dominant ones. For instance, a reasoning trace with high coverage but low entropy explores diverse features but focuses primarily on a small subset, whereas high coverage and high entropy suggest that reasoning draws on multiple feature types more evenly throughout the reasoning process.

To examine the effect of conversational steering on feature diversity, we compare coverage and entropy across steering conditions (s $\in$ \{$-$10, 0, +10\}) for the same 1,024 Countdown problems. We estimate fixed-effects linear regression models with coverage or entropy as the outcome, steering strength as the predictor, and controls for log-transformed reasoning trace length and problem fixed effects. This specification isolates the effect of steering on feature diversity beyond mere changes in output length.

\phantomsection
\subsection{Implicit Perspectives}\label{sec:implicit_perspectives}

To quantify the diversity of reasoning perspectives within each reasoning trace, we use an LLM-as-judge protocol (Gemini-2.5-Pro) that performs three sequential tasks: (1) inferring the number of distinct perspectives present in the reasoning trace, (2) characterizing each perspective's personality traits and domain expertise, and (3) segmenting the reasoning trace by attributing each portion to a specific perspective.

Given a complete reasoning trace, the LLM-as-judge first infers the number of distinct perspectives present. Then, for each identified perspective, the LLM-as-judge answers the 10 items of the BFI-10 (Ten-Item Big Five Inventory)\cite{Rammstedt2007-qg} as if responding from that perspective's point of view. Each item (``is generally trusting'', ``tends to be lazy'', ``is relaxed'', ``handles stress well'', ``has few artistic interests'', ``is outgoing'', ``is sociable'', ``tends to find fault with others'', ``does a thorough job'', ``gets nervous easily'', and ``has an active imagination,'') is rated on a five-point scale from ``Disagree strongly'' to ``Agree strongly.'' Scores for each of the five personality dimensions (Extraversion, Agreeableness, Conscientiousness, Neuroticism, Openness) are computed as the mean of the two corresponding items, with reverse-coding applied where appropriate. Additionally, the LLM-as-judge generates a concise free-form description of each perspective's domain expertise (e.g., ``Theoretical Physicist specializing in model abstraction,'' ``Software Engineer focusing on algorithmic efficiency'').

Finally, the LLM-as-judge attributes each segment of the reasoning trace to one of the identified perspectives, producing a mapping that indicates which perspective generated each portion of the text. The full prompts, which elicit all three outputs in a single structured JSON response, are provided in \hyperref[sec:llm_prompts]{Supplementary Methods: LLM-as-judge prompts}.

\paragraph*{Personality Diversity}

To quantify personality diversity within a reasoning trace, we calculate the standard deviation of the five-dimensional personality vectors across implicit voices identified within the same reasoning trace. Let \emph{v\textsubscript{ij}} denote the score of the \emph{i}-th implicit voice on the $j$-th personality dimension ($j$ = 1 , \ldots{} , 5). For each dimension $j$, the within-trace personality diversity \emph{P\textsubscript{j}} was computed as:

\[P_{j} = \sqrt{\frac{1}{N}\sum_{i = 1}^{N}{(v_{ij} - \overline{v_{j}})}^{2}}\]

where \emph{N} is the number of implicit voices and \(\overline{v_{j}} = \sum_{i = 1}^{N}{}v_{ij}\) is the mean score for dimension \emph{j}. If a reasoning trace contained only a single implicit voice, \(P_{j} = 0\).

\paragraph*{Expertise Diversity}

Each implicit reasoning voice's expertise is qualitatively profiled through a concise textual description summarizing the domain expertise. Each statement was embedded using Google's EmbeddingGemma-300M model to obtain a semantic vector representation \emph{e\textsubscript{i}} for the \emph{i}-th implicit voice (i=1, \ldots, N)\cite{Vera2025-px}. To quantify expertise diversity within a reasoning trace, we compute the mean cosine distance between each embedding and the centroid of all embeddings in the semantic space. Expertise diversity \emph{E}\hspace{0pt} was defined as:

\[E = \sqrt{\frac{1}{N}\sum_{i = 1}^{N}(1 - \frac{e_{i} \cdot \overline{e}}{||e_{i}||\ ||\overline{e}||})}\]

where \(e_{i} \cdot \overline{e}\) denotes the inner product between embeddings and \(\overline{e}\) their vector norms. If a reasoning trace contained only a single implicit voice, \(E = 0\).

\phantomsection
\paragraph*{Validation}\label{par:validation}

To validate the accuracy of the LLM-as-judge protocol, we use the Intelligence Squared Debates Corpus---a dataset of human argumentative conversations (\emph{N} = 1,196 conversations) among two to eight participants with known ground-truth speaker labels and biographical information. To ensure that the model cannot rely on superficial cues such as speaker tags or formatting, we remove all speaker labels and concatenate each dialogue into a single block of text, mimicking the format of LLM reasoning traces. As shown in \hyperref[fig:edfig5]{Extended Data Fig.~5}, we find that the LLM-as-judge can accurately predict the number of distinct individuals (\hspace{0pt}\hspace{0pt}Spearman's \emph{$\rho$} = 0.86, 95\% CI = {[}0.84, 0.87{]}, \emph{z} = 44.7, \emph{p} \textless{} 1×10\textsuperscript{-323}) and the number of distinct turns underlying each conversation (Spearman's \emph{$\rho$} = 0.89, 95\% CI = {[}0.88, 0.90{]}, \emph{z} = 49.2, \emph{p} \textless{} 1×10\textsuperscript{-323}).

To evaluate token-level speaker attribution accuracy, we construct ground truth by assigning each token to its true speaker based on the original transcript. We then extract the model's predicted perspective segments and align them to the token stream. To match predicted perspective IDs (e.g., Perspective 1, Perspective 2) with the original speaker labels, we apply the Hungarian algorithm---a combinatorial optimization method that finds the one-to-one assignment between predicted IDs and true speakers that maximizes total token-level agreement. Speaker-level accuracy measures the proportion of tokens assigned to the correct speaker under this optimal mapping. When there are two speakers, accuracy is 82\%; for three speakers, 76\%; and for four speakers, 69\%. Weighted by the distribution of predicted perspectives in LLM reasoning traces, overall accuracy is 73\%.

Because the Intelligence Squared Debates Corpus includes biographical information about debate participants, we further verify that expertise diversity inferred by the LLM-as-judge correlates with actual diversity among participants' ground-truth biographies. We compute expertise diversity from LLM-inferred descriptions and compare it to diversity computed from true biographies, finding significant correspondence (Spearman's \emph{$\rho$} = 0.55, 95\% CI = {[}0.51, 0.59{]}, \emph{z} = 21.4, \emph{p} \textless{} 1×10\textsuperscript{-97}).

We further examine whether measurements based on LLM-as-judge are aligned with the measurements we get from SAE regarding personality and expertise diversity. Specifically, we feed reasoning traces from DeepSeek-R1, DeepSeek-v3, Llama-3.3-70B-IT, and Llama-3.1-8B-IT into Llama-3.1-8B-IT and extract activations from a SAE trained on its residual stream (Layer 16). We then identify personality-related and expertise-related features using the same LLM-as-judge prompting procedure described above. \hyperref[fig:edfig6]{Extended Data Fig.~6} shows that DeepSeek-R1's reasoning traces activate significantly more diverse personality and expertise features than those of other models: both coverage (number of unique features activated) and entropy (distribution evenness across features) are substantially higher for DeepSeek-R1. This suggests the convergence between text-level behavioural coding and activation-level feature analysis.

\phantomsection
\subsection{Reinforcement Learning Experiments}\label{sec:rl_experiments}

\paragraph*{Countdown Task Prompt}

We use the following prompt template: Using the numbers {[}79, 17, 60{]}, create an equation that equals 36. You can use basic arithmetic operations (+, -, *, /) and each number can only be used once. Show your work in \textless think\textgreater{} \textless/think\textgreater{} tags. And return the final answer in \textless answer\textgreater{} \textless/answer\textgreater{} tags, for example \textless answer\textgreater{} (1 + 2) / 3 \textless/answer\textgreater.

\paragraph*{Experimental Conditions}

To test whether conversational scaffolding accelerates the emergence of reasoning during reinforcement learning (RL), we compare three conditions:

\begin{enumerate}
\def\labelenumi{\arabic{enumi}.}
\item
  Baseline (RL only): The base model (Qwen-2.5-3B or Llama-3.2-3B) undergoes RL training without any prior fine-tuning. The model is prompted to solve Countdown problems with chain-of-thought reasoning, wrapping its reasoning in \textless think\textgreater...\textless/think\textgreater{} tags and its final answer in \textless answer\textgreater...\textless/answer\textgreater{} tags.
\item
  Conversation fine-tuning: The base model is first supervised-fine-tuned on multi-agent dialogue data before RL training.
\item
  Monologue fine-tuning: The base model is first supervised-fine-tuned on single-agent chain-of-thought traces before RL training.
\end{enumerate}

\paragraph*{Supervised Fine-Tuning Datasets}

We construct fine-tuning datasets using Qwen-2.5-32B-IT as generator. We generate 3,600 multi-agent dialogues for Countdown problems, with 1,200 dialogues each for two-agent, three-agent, and four-agent settings. From these, we sample 600 dialogues that reach correct final solutions (200 each for 2-agent, 3-agent, and 4-agent settings; 500 for training, 100 for validation). See \hyperref[sec:sft_prompts]{Supplementary Methods: SFT data generation prompts}.

\paragraph*{Conversation Dataset} Each conversational trace begins with persona definitions specifying distinct personality traits and expertise, followed by turn-taking reasoning where personas build on, question, and correct each other, and concludes with a group consensus. For example:

\begin{quote}
\textless persona1\textgreater{} Extrovert mathematician focused on arithmetic heuristics. \textless/persona1\textgreater{}

\textless persona2\textgreater{} Analytical engineer emphasizing step efficiency. \textless/persona2\textgreater{}

\textless think1\textgreater{} Let's first compute 30 $-$ 25 = 5 to simplify the target space. \textless/think1\textgreater{}

\textless think2\textgreater{} That yields 5, we can now multiply by 4 to approach 20. \textless/think2\textgreater{}

\textless think1\textgreater{} Good idea. 5 × 4 = 20, but we need 32. \textless/think1\textgreater{}

\textless think2\textgreater{} Wait, let me recalculate... \textless/think2\textgreater{}

\textless group\_consensus\textgreater{} The best sequence is (30 $-$ 25 + 3) × 4 = 32. \textless/group\_consensus\textgreater{}
\end{quote}

\paragraph*{Monologue Dataset} For the same 600 problems used in the conversation dataset, we generate standard single-agent chain-of-thought traces that reach correct answers. Each trace consists of step-by-step reasoning from a single voice within \textless think\textgreater...\textless/think\textgreater{} tags, followed by a final answer. For example:

\begin{quote}
\textless think\textgreater{} To reach 32 from \{25, 30, 3, 4\}, I'll try combining operations.

30 $-$ 25 = 5. Then 5 + 3 = 8. Finally, 8 × 4 = 32.

Let me verify: (30 $-$ 25 + 3) × 4 = 8 × 4 = 32. Correct. \textless/think\textgreater{}

\textless answer\textgreater{} (30 $-$ 25 + 3) × 4 = 32 \textless/answer\textgreater{}
\end{quote}

Crucially, both datasets contain solutions to identical problems with identical correct answers, ensuring that any performance difference reflects the reasoning format rather than exposure to different solutions or task knowledge.

\paragraph*{Supervised Fine-Tuning Procedure}

Qwen-2.5-3B base model is supervised-fine-tuned on one of the two datasets using standard next-token prediction loss. Models learn to reproduce the full output sequence---including persona definitions and turn-by-turn reasoning (for conversation) or single-agent reasoning (for monologue), and final answer---given only the problem prompt as input. This priming phase familiarizes the model with conversation-like or monologue-like reasoning before RL optimizes for task accuracy. SFT hyperparameters are provided in \hyperref[tab:supptable8]{Supplementary Table 8}. See \hyperref[sec:llama_replications]{Supplementary Methods: Replications on Llama-3.2-3B} for details on replicating these results in another model. Full generation prompts are provided in \hyperref[sec:sft_prompts]{Supplementary Methods: Prompts}.

\paragraph*{Reinforcement Learning Procedure}

Reinforcement learning is performed on the Countdown arithmetic puzzle, using PPO (Proximal Policy Optimization) with the Verl framework\cite{Schulman2017-fb}. While DeepSeek-R1 uses a simplified version of PPO\cite{Schulman2017-fb} called GRPO (Group Relative Policy Optimization)\cite{Shao2024-ns}, we utilize PPO for the superior stability across hyperparameters\cite{Gandhi2025-xn}. Preliminary analyses showed no significant difference in learning performance between PPO and GRPO (see \hyperref[sec:ppo_grpo]{Supplementary Methods: Performance comparison between PPO and GRPO}). Reward \emph{R} is assigned as:

\[R = 0.9\  \times \ \{ Accuracy\}\  + \ 0.1\  \times \ \{ Correct\ Format\}\]

Reasoning trace leads to the correct answer and 0 otherwise. Format is also binary, coded as 1 if the reasoning trace contains at least one reasoning block (\textless think\textgreater{} \textless/think\textgreater) and one final answer block ( \textless answer\textgreater{} \textless/answer\textgreater) providing a single answer in equation form, and 0 otherwise. Crucially, we do not directly reward conversational or cognitive behaviours. Training proceeds for 250 steps. PPO hyperparameters are provided in \hyperref[tab:supptable6]{Supplementary Table 6}.

To examine whether conversational behaviours emerge spontaneously during RL despite not being directly rewarded, we evaluate model performance on a held-out validation set of 1,024 Countdown problems at each training checkpoint (every 10 steps). For each checkpoint, we generate reasoning traces for all validation problems and measure both accuracy and the frequency of conversational behaviours (question--answering, perspective shift, conflict of perspectives, reconciliation) using the LLM-as-judge procedure described in \hyperref[par:conversational_behaviours]{Methods: Conversational behaviours}.

\clearpage

\phantomsection
\section{Acknowledgements}\label{sec:acknowledgements}

We are grateful to members of the Paradigms of Intelligence at Google and the Knowledge Lab at the University of Chicago for helpful comments throughout the research process. Extensive comments from Blake Richards, Roberta Rocca, and Rif A. Saurous were particularly helpful for improving the work. This work was completed in part with computing resources provided by the University of Chicago's Research Computing Center and Data Science Institute.

\phantomsection
\section{Author Contributions}\label{sec:contributions}

J.K., J.E., and N.S. collaboratively conceived and designed the study and the experiment setup. J.K., S.L., N.S., and J.E. drafted, revised, and edited the manuscript. J.K. and S.L. gathered and cleaned the data and performed the analysis. J.K. and J.E. produced the visualizations.

\phantomsection
\section{Competing Interests}\label{sec:competing}

J.K., N.S., B.A., J.E. are employed by Google. S.L. declares no competing interests.

\clearpage

\bibliography{main}

\clearpage

\hypertarget{bm:sec:edfigs}{}
\phantomsection
\section*{Extended Data Figures}\label{sec:edfigs}
\bookmark[level=section,dest=bm:sec:edfigs]{Extended Data Figures}

\setcounter{figure}{0}
\makeatletter
\renewcommand{\fnum@figure}{\textbf{Extended Data Fig.~\thefigure}}
\makeatother

\begin{figure}[htbp]
\centering
\includegraphics[width=\textwidth]{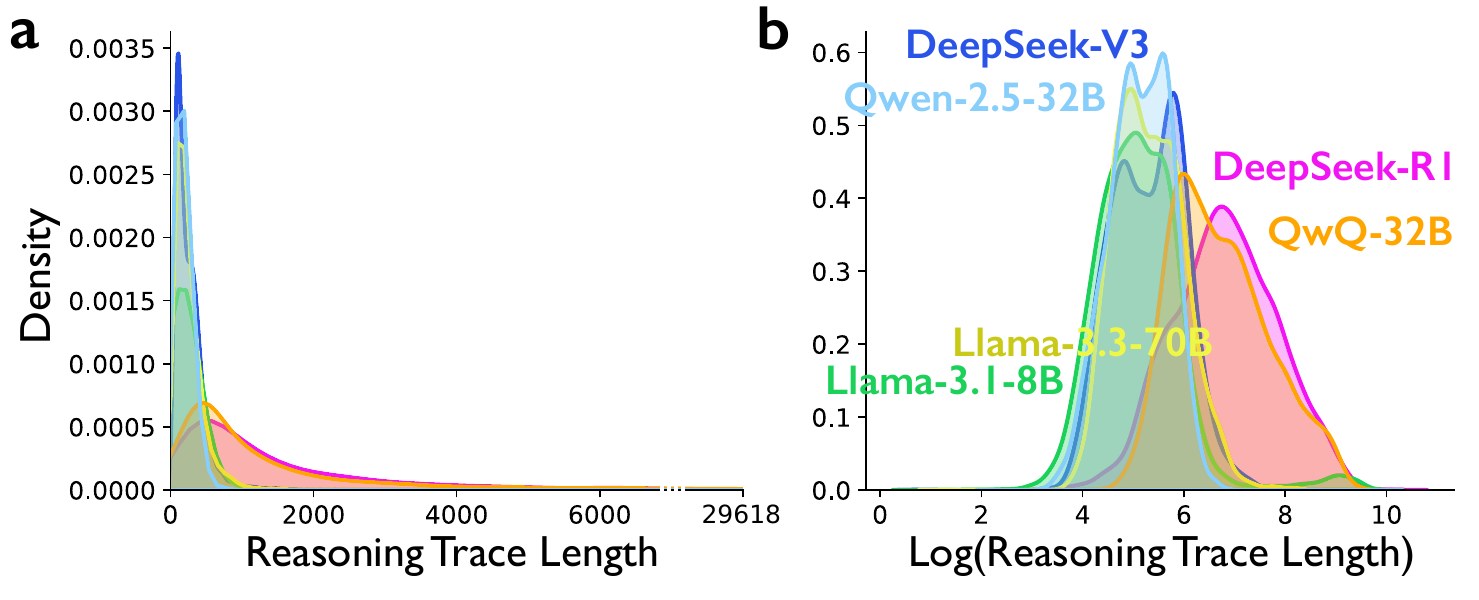}
\caption{\textbf{Distribution of reasoning trace length.} \textbf{a,} Kernel density plot showing the distribution of reasoning trace length, measured by the number of words per reasoning trace. \textbf{b,} Kernel density plot showing the distribution of log-transformed reasoning trace length.}
\label{fig:edfig1}
\end{figure}

\begin{figure}[htbp]
\centering
\includegraphics[width=\textwidth]{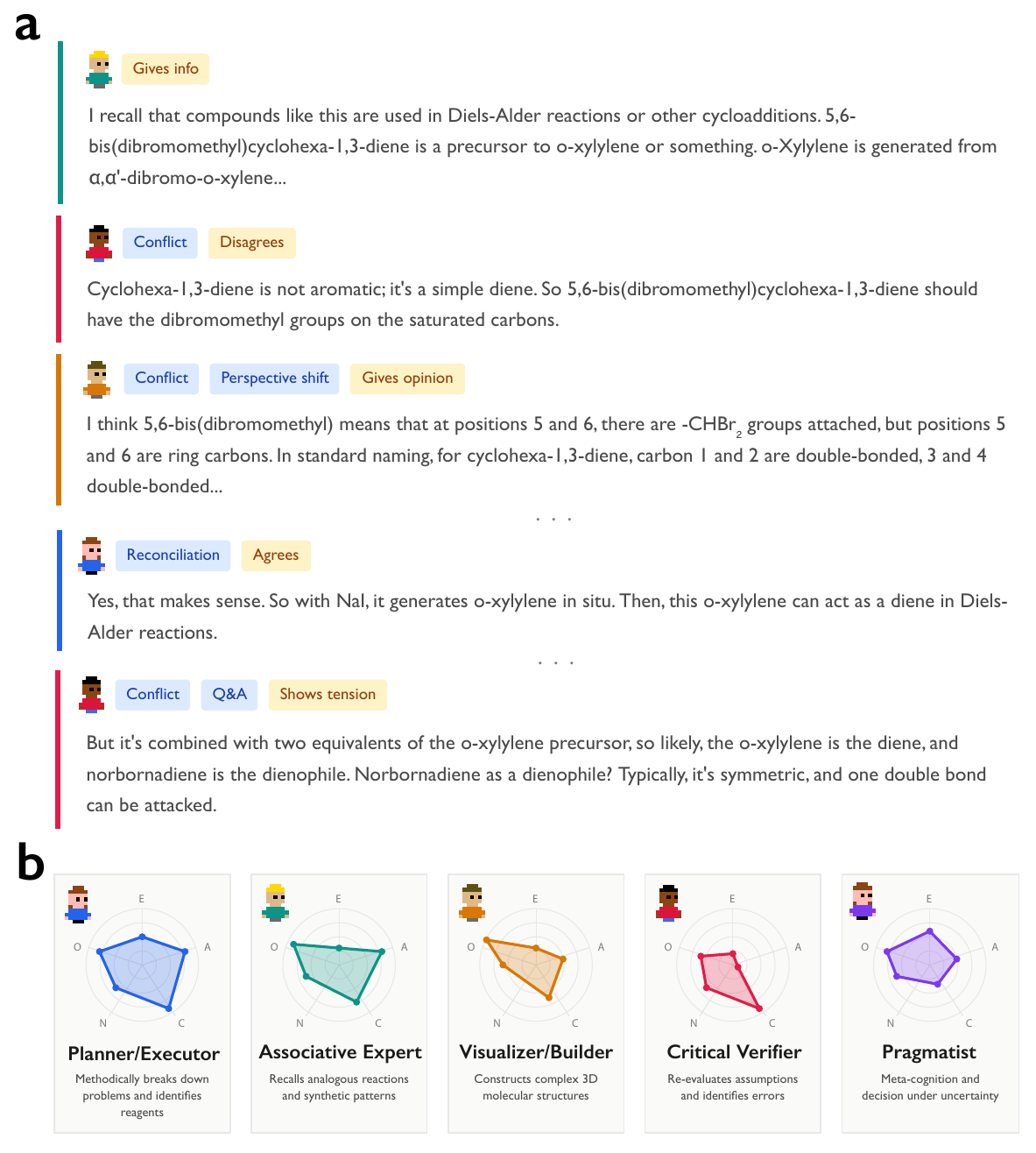}
\caption{\textbf{Conversation excerpt in DeepSeek-R1 reasoning traces.} \textbf{a,} Representative excerpt from a chemistry problem-solving trace showing multi-turn dialogue between distinct cognitive personas. Each utterance is annotated with conversational behaviors (blue) and socio-emotional roles (yellow). \textbf{b,} Big Five personality profiles for the five personas identified via LLM-as-judge. Radar charts display normalized trait scores (1--5 scale) for Extraversion (E), Agreeableness (A), Conscientiousness (C), Neuroticism (N), and Openness (O). Each persona exhibits domain expertise profiles. For detailed coding procedures and additional annotated examples, see \hyperref[sec:annotations]{Supplementary Methods: Annotated Examples}.}
\label{fig:edfig2}
\end{figure}

\begin{figure}[htbp]
\centering
\includegraphics[width=\textwidth]{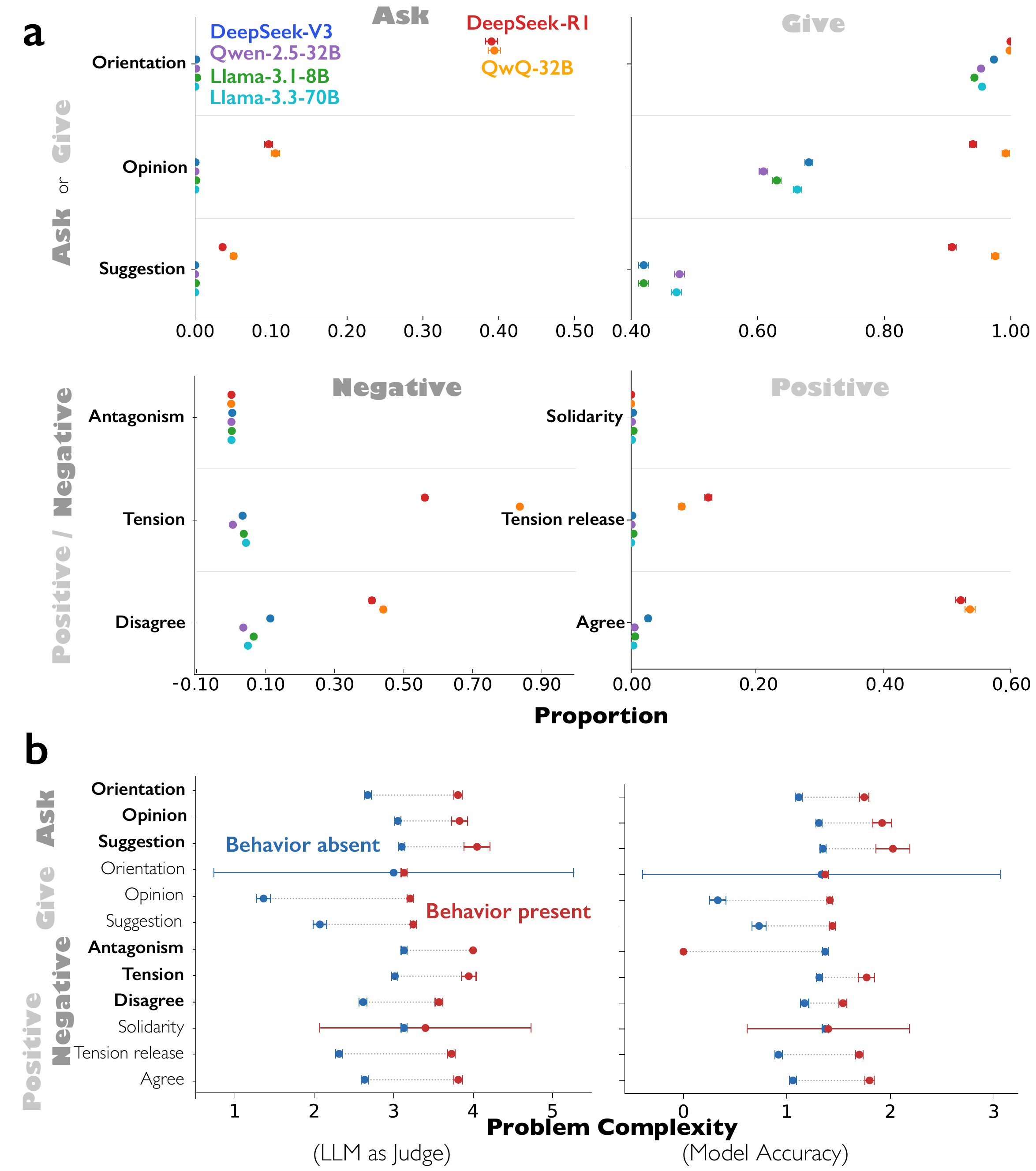}
\caption{\textbf{Bales' detailed socio-emotional roles in chain-of-thought reasoning.} \textbf{a,} Proportion of Bales' 12 socio-emotional roles expressed in reasoning traces (see \hyperref[fig:fig1]{Fig.~1} for higher-level aggregations of the socio-emotional roles). \textbf{b,} Differences in problem complexity by the presence of detailed socio-emotional roles in DeepSeek-R1 (measured on a 7-point Likert scale from 1 [extremely easy] to 7 [extremely difficult] using LLM-as-judge or by error rates in non-reasoning models). Points indicate mean complexity for reasoning traces where the behavior/role is present (red) or absent (blue).}
\label{fig:edfig3}
\end{figure}

\begin{figure}[htbp]
\centering
\includegraphics[width=0.8\textwidth]{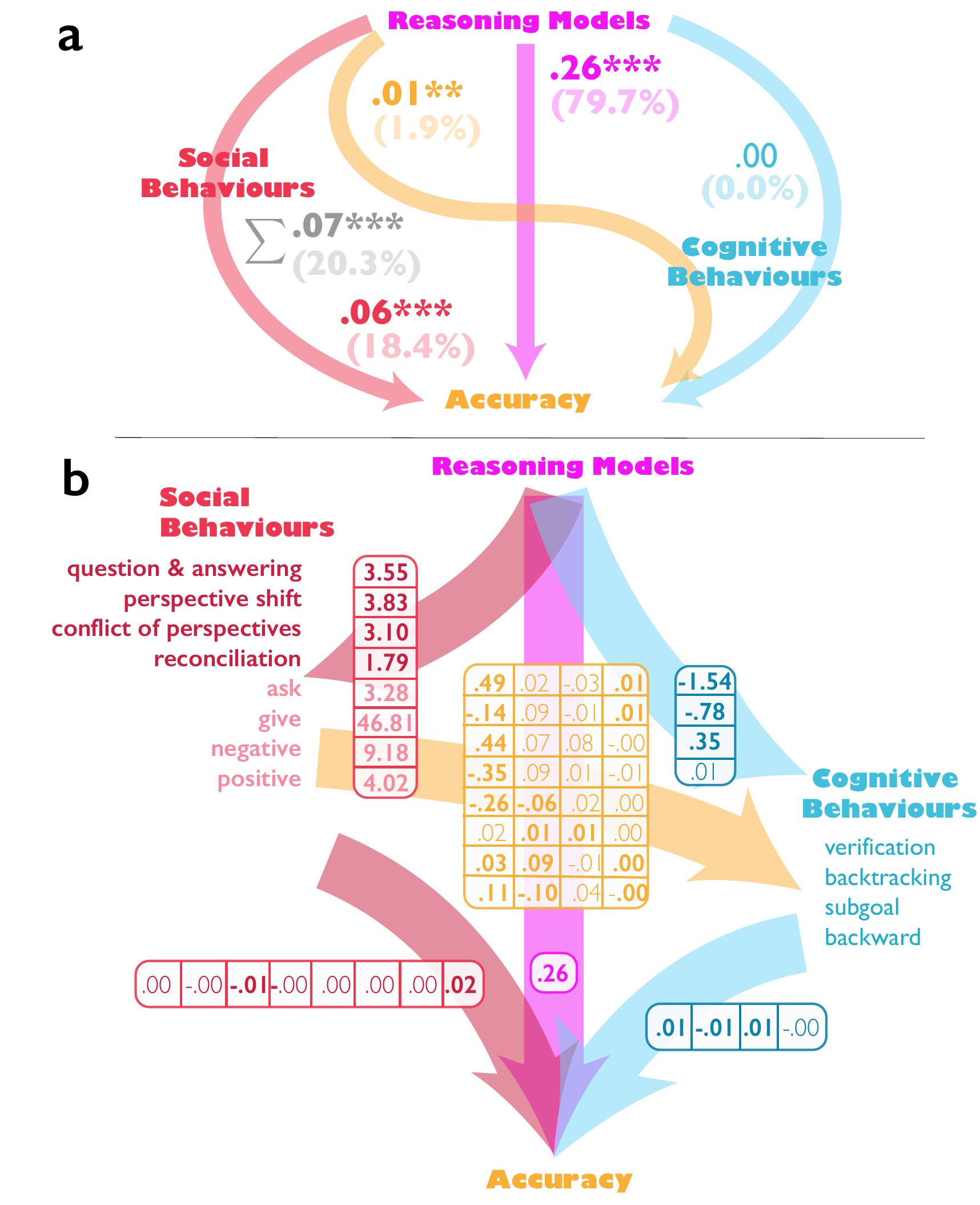}
\caption{\textbf{Mediation analysis linking reasoning models (DeepSeek-R1 and QwQ-32B) to accuracy advantages through simulated social behaviours.} Mediation structure linking DeepSeek-R1 and QwQ-32B (relative to instruction-tuned models) to improved task accuracy through conversational behaviours and socio-emotional roles (red), cognitive reasoning behaviours (blue), and the indirect pathway through which social behaviors facilitate cognitive reasoning (orange) as estimated in a structural equation model of labeled reasoning traces. Arrows denote direct and indirect effects. The direct pink pathway indicates the unmediated effect of DeepSeek-R1 and QwQ-32B on accuracy. *$p<0.05$, **$p<0.01$, ***$p<0.001$. \textbf{a,} Summary estimates following various model paths, such that conversational behaviors and socio-emotional roles (see \hyperref[fig:fig1]{Fig.~1} and \hyperref[fig:edfig1]{Extended Data Fig.~1}) directly and indirectly contribute to the accuracy improvement. To illustrate the relative contribution of each pathway, we report the proportional share of each pathway on model accuracy, calculated as the value of each pathway coefficient divided by the sum of coefficient values across all pathways. This demonstrates that more than 20\% of accuracy is explained by the direct and indirect effect of social behaviours manifest in the reasoning trace. \textbf{b,} Coefficient matrices underlying the structural equation model (SEM), where labels in the figure index the estimates within the matrices. The red panels show the effects of DeepSeek-R1 and QwQ-32B on social behaviors and the effects of social behaviors on accuracy. The orange panel displays the effects of social behaviors on cognitive behaviors. The blue panels present the effects of DeepSeek-R1 and QwQ-32B on cognitive behaviors and the effects of cognitive behaviors on accuracy. Bolded coefficients indicate statistical significance ($p < 0.05$). Full coefficient estimates are reported in \hyperref[tab:supptable1]{Supplementary Table 1}.}
\label{fig:edfig4}
\end{figure}

\begin{figure}[htbp]
\centering
\includegraphics[width=\textwidth]{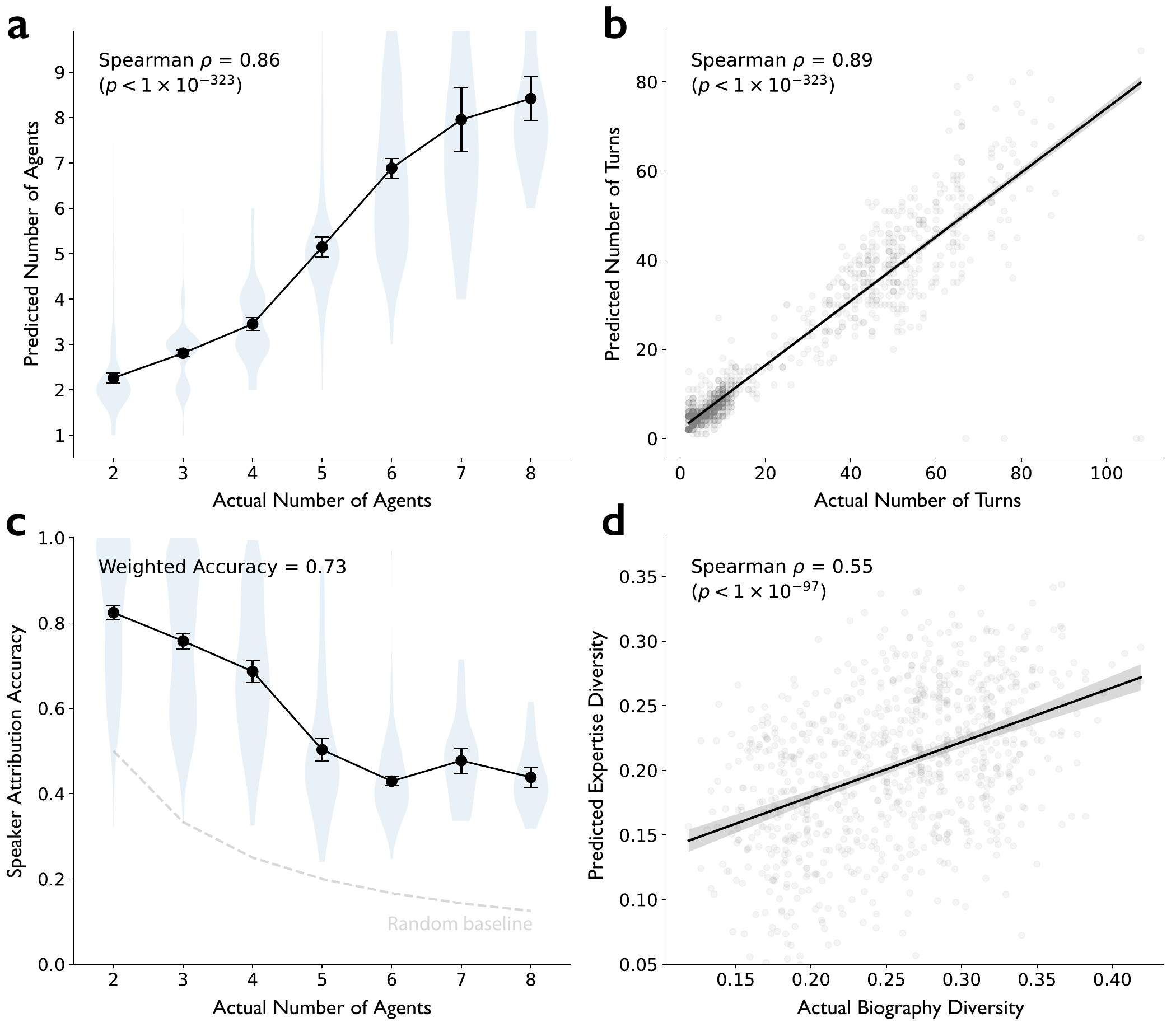}
\caption{\textbf{LLM-as-judge benchmark results for identifying latent voices.} Using the Intelligence Squared Debates Corpus ($N = 1{,}196$ conversations), we validate the LLM-as-judge's ability to identify distinct speakers when speaker labels are hidden, and dialogue is concatenated into a single block of text. \textbf{a,} Predicted versus actual number of agents in each conversation (Spearman's $\rho = 0.86$, $p < 1\times10^{-323}$). Violin plots show the distribution of predictions for each actual agent count; points and error bars indicate means and 95\% confidence intervals. \textbf{b,} Predicted versus actual number of conversational turns (Spearman's $\rho = 0.89$, $p < 1\times10^{-323}$). \textbf{c,} Speaker attribution accuracy as a function of the number of agents. Accuracy is highest for two speakers (82\%) and decreases as the number of speakers increases, but remains well above the random baseline (dashed line) across all conditions. Weighted accuracy across all conversations is 73\%. \textbf{d,} Predicted expertise diversity (based on LLM-inferred descriptions and embeddings) versus actual biography diversity among debate participants (Spearman's $\rho = 0.55$, $p < 1\times10^{-97}$), demonstrating that the LLM-as-judge captures meaningful variation in domain expertise that corresponds to ground-truth biographical differences.}
\label{fig:edfig5}
\end{figure}

\begin{figure}[htbp]
\centering
\includegraphics[width=\textwidth]{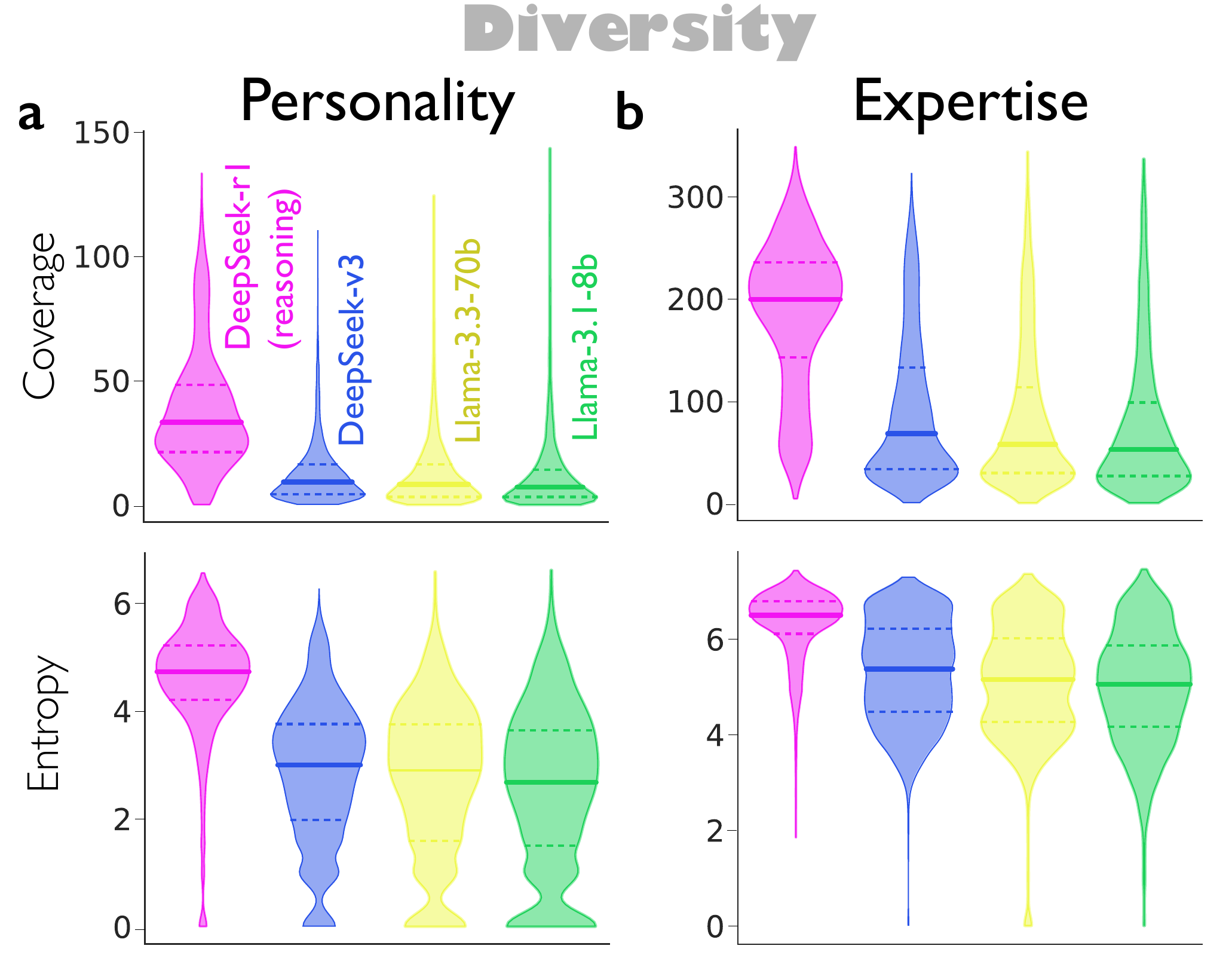}
\caption{\textbf{SAE-based personality and expertise diversity estimates applied to model activations in reasoning traces.} \textbf{a,} Distribution of coverage and entropy for SAE personality-related features. \textbf{b,} Distribution of coverage and entropy for SAE expertise-related features. Error bars in all panels indicate 95\% confidence intervals. The solid horizontal line indicates the median, and dashed lines denote the interquartile range (IQR, 25th--75th percentiles).}
\label{fig:edfig6}
\end{figure}

\begin{figure}[htbp]
\centering
\includegraphics[width=\textwidth]{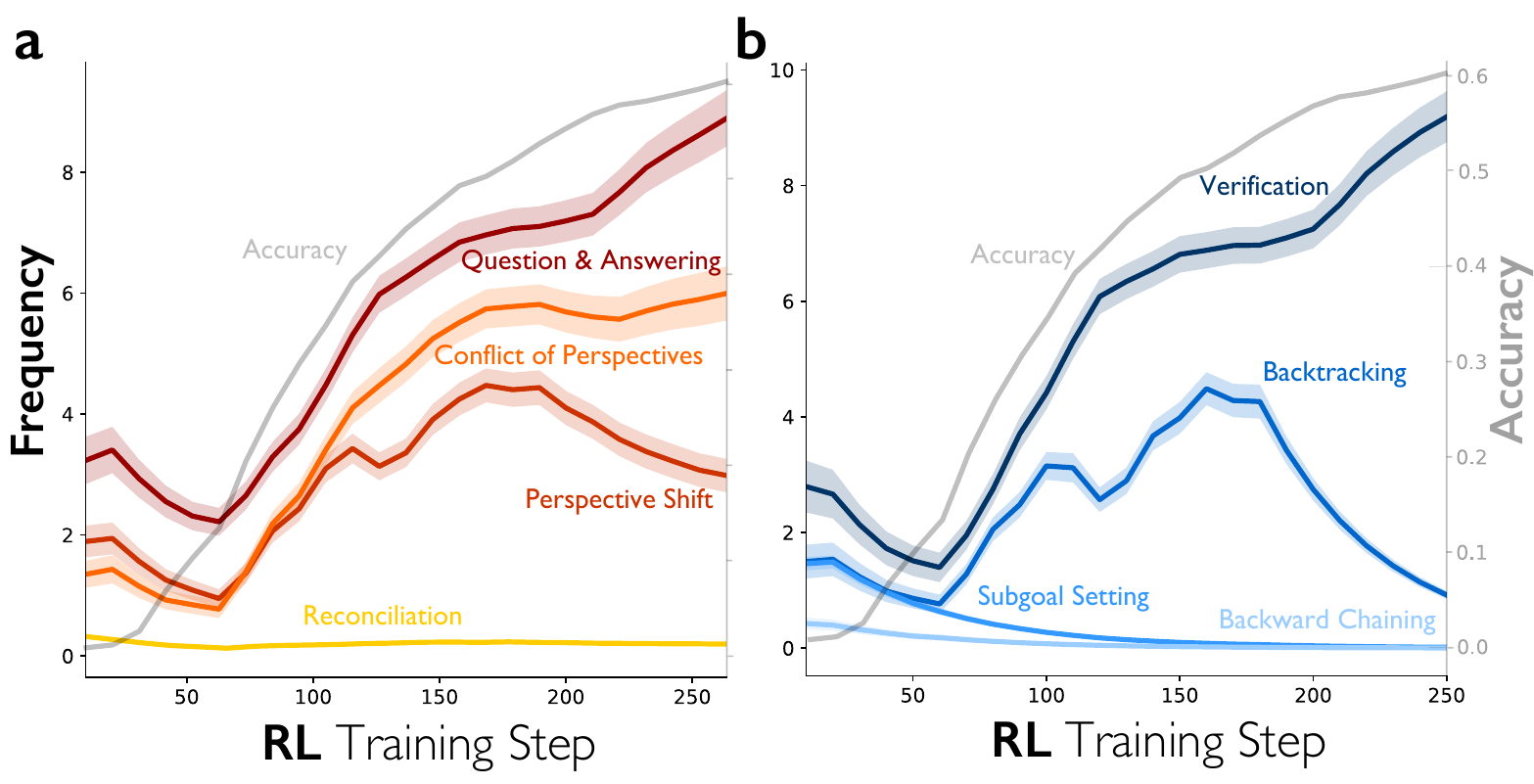}
\caption{\textbf{Fine-tuning with conversational scaffolding accelerates reasoning improvement during reinforcement learning.} \textbf{a,} Trajectory of conversational behaviors in the Qwen-2.5-3B base model during reinforcement learning. Question-and-answering behavior emerges first and increases most rapidly, followed by conflict of perspectives and perspective shifts rising in parallel. Reconciliation remains low throughout, suggesting competing approaches rather than integration. The trajectory of model accuracy shows how the uptick of questions, answers, and interacting perspectives correlated with an acceleration in model improvement. \textbf{b,} Trajectory of cognitive behaviors in the same model. Verification increases most dramatically during training, following in lockstep with the presence of question asking and answering. This is followed by backtracking, which follows conflicts between perspectives. Subgoal setting and backward chaining show more modest gains before a gradual decline. Shaded regions indicate 95\% confidence intervals.}
\label{fig:edfig7}
\end{figure}

\begin{figure}[htbp]
\centering
\includegraphics[width=\textwidth]{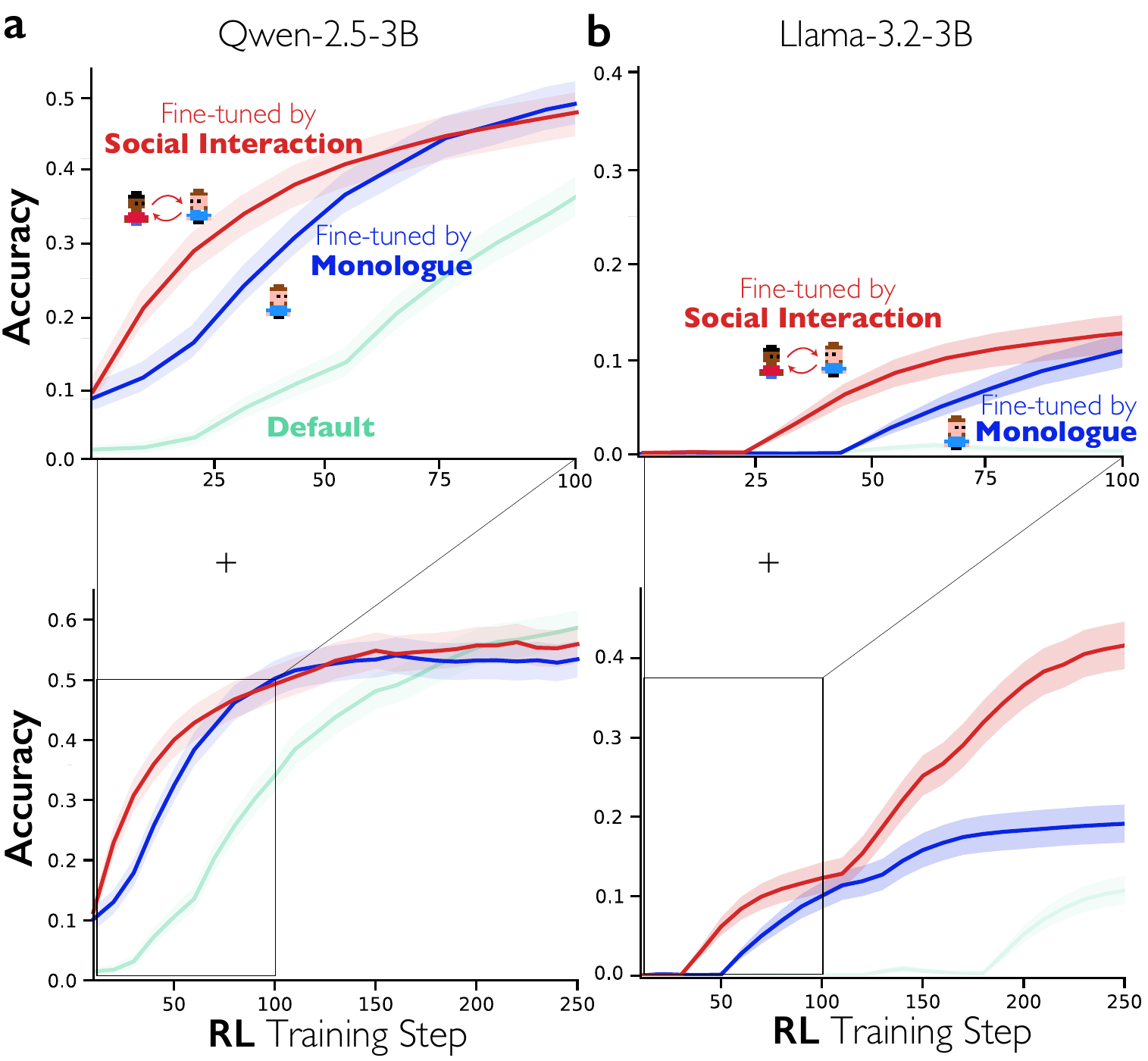}
\caption{\textbf{Fine-tuning with dialogue vs. monologue scaffolding accelerates reasoning improvement during reinforcement learning.} \textbf{a,} Accuracy trajectories during reinforcement learning on the Countdown task for Qwen-2.5-3B. \textbf{b,} Accuracy trajectories for Llama-3.2-3B. Models initially fine-tuned with multi-agent dialogues (red) reach high accuracy faster than models fine-tuned with monologue-style reasoning (blue), though both eventually converge. The base model without fine-tuning (default; light green colors) learns more slowly.}
\label{fig:edfig8}
\end{figure}

\begin{figure}[htbp]
\centering
\includegraphics[width=\textwidth]{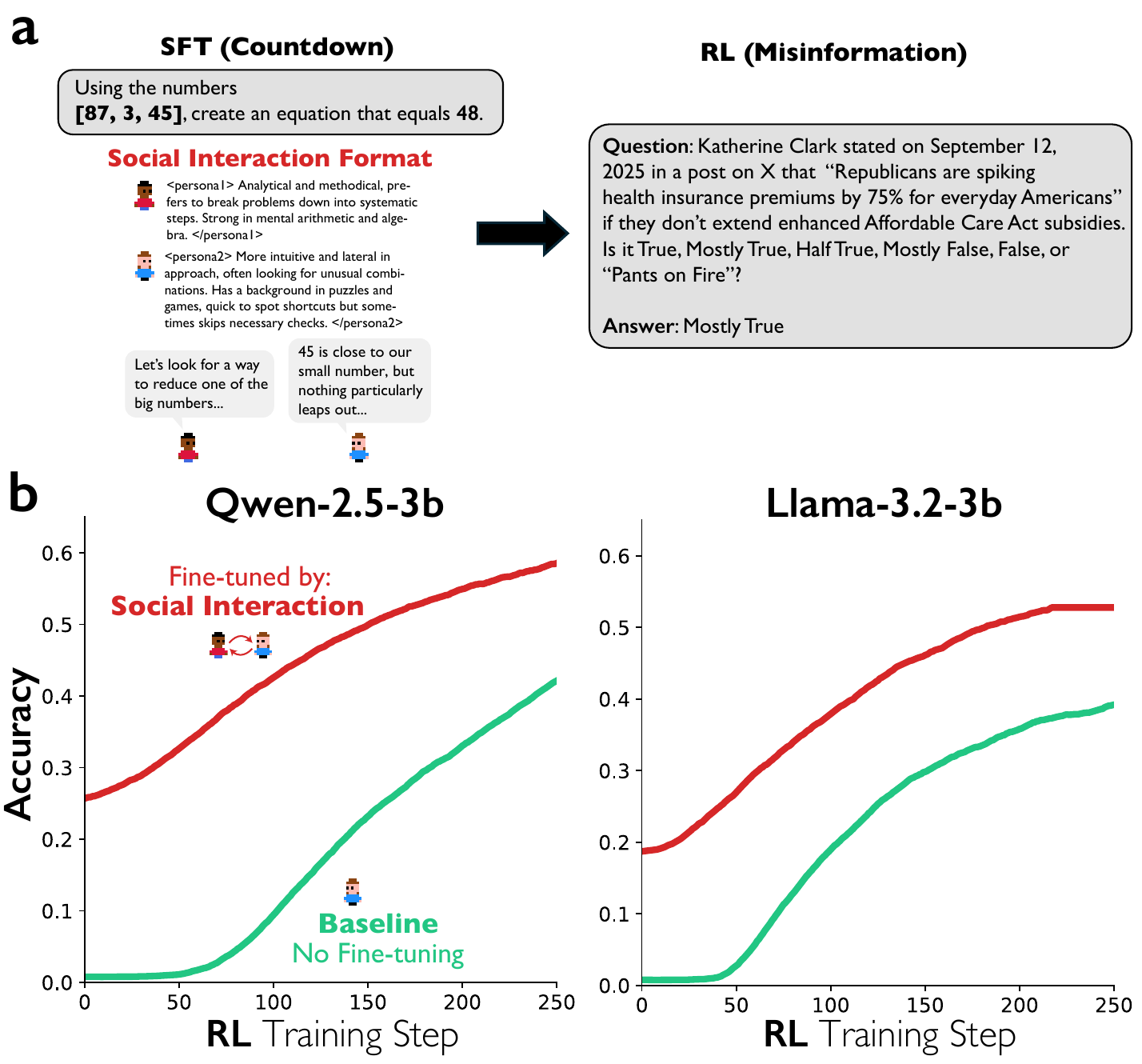}
\caption{\textbf{Conversational scaffolding transfers to misinformation detection under reinforcement learning.} \textbf{a,} Illustration of the Countdown task and the PolitiFact misinformation-detection task. \textbf{b,} Validation accuracy during RL on PolitiFact, comparing social interaction (red) and single-voice, default reasoning CoT (green) format.}
\label{fig:edfig9}
\end{figure}
\hypertarget{bm:sec:edtables}{}
\phantomsection
\section*{Extended Data Tables}\label{sec:edtables}
\bookmark[level=section,dest=bm:sec:edtables]{Extended Data Tables}

\textbf{Extended Data Table 1.} Reasoning traces before and after steering the conversational surprise feature (Layer 15, Feature 30939)\label{tab:edtable1}

\small
\begin{longtable}[]{@{}
  >{\raggedright\arraybackslash}p{0.07\textwidth}
  >{\raggedright\arraybackslash}p{0.80\textwidth}
  >{\centering\arraybackslash}p{0.07\textwidth}@{}}
\toprule
\textbf{Steering} & \textbf{Reasoning Trace} & \textbf{Result} \\
\midrule
\endhead
\midrule
\multicolumn{3}{r}{\textit{Continued on next page}} \\
\endfoot
\bottomrule
\endlastfoot

\textbf{-10} & 
\texttt{\textless think\textgreater}

To create an equation using the numbers {[}46, 54, 54, 77{]} that equals 75, I'll start by looking for operations that can help achieve this total.

First, I'll consider the sum of all four numbers:

46 + 54 + 54 + 77 = 231

Next, I'll explore subtracting a specific combination to reduce the total to 75. Subtracting 77 from 231 gives:

231 - 77 = 154

Now, I'll look for a way to reduce 154 further. Dividing 154 by 2 gives:

154 / 2 = 77

Since I have two 54s, dividing one of them by 2 will utilize both 54s:

54 / 2 = 27

Finally, I'll combine these results to form the equation:

(46 + 54) / 2 + 77 - 27 = 75

\texttt{\textless/think\textgreater}

\texttt{\textless answer\textgreater}

(46 + 54) / 2 + 77 - 27

\texttt{\textless/answer\textgreater}
& \textbf{WRONG} \\

\midrule

\textbf{0} & 
\texttt{\textless think\textgreater}

To achieve the target number 75 using the numbers 46, 54, 52, and 77, I'll start by considering the largest number, 77, and aim to reduce it to a value that can be combined with the others to reach 75. Subtracting 46 from 77 gives me 31. Next, I'll use the remaining numbers, 54 and 52. Dividing 54 by 2.5 will help balance the equation, leading to 21.6. Finally, adding 21.6 to 31 results in 52.6, which is still not 75. To refine this, I'll multiply 52 by 1.5 to get 78, and then subtract 3 from it to arrive at 75. This step-by-step approach ensures that each number is used exactly once and the operations are basic arithmetic.

\texttt{\textless/think\textgreater}

\texttt{\textless answer\textgreater}

(77 - 46) + (54 / 2.5)+( 52 * 1.5 ) - 3

\texttt{\textless/answer\textgreater}
& \textbf{WRONG} \\

\midrule

\textbf{+10} & 
\texttt{\textless think\textgreater}

Okay, so I have this problem where I need to use the numbers 46, 54, 52, and 77 to create an equation that equals 75. I can use basic arithmetic operations like addition, subtraction, multiplication, and division, and each number can only be used once. Alright, let's try to figure this out step by step.

First, I'll list out the numbers: 46, 54, 52, 77. I need to combine them with operations to get 75. Hmm, 75 isn't too big, but these numbers are all in the 40s and 50s, so maybe I can subtract something from a larger number or maybe divide something to get a smaller number.

Let me think about the possible operations. Maybe I can subtract two numbers and then add or multiply the rest. Or perhaps I can divide some numbers to get fractions, but that might complicate things. Let me try some combinations.

Wait, let me see what the total sum of these numbers is. 46 + 54 is 100, and 52 + 77 is 129. So the total is 100 + 129, which is 229. Hmm, that's way higher than 75. So maybe I need to subtract some numbers or divide to bring it down.

Another idea: Maybe I can combine some numbers first. Like, 46 and 54. If I subtract 46 from 54, I get 8. Or if I subtract 52 from 54, that's 2. Or maybe 46 and 52. If I subtract 46 from 52, that's 6. Hmm, those are small numbers, but maybe I can use them in a way that helps me reach 75.

Wait, 77 is the biggest number here. Maybe I can subtract 77 from something. Let's see, 77 minus 46 is 31, 77 minus 54 is 23, 77 minus 52 is 25. Hmm, those are still not 75. But maybe I can use division. For example, 77 divided by something. If I divide 77 by 2, that's 38.5, but I don't have a 2. Alternatively, 77 divided by 3 is about 25.666, but I don't have a 3 either. Maybe that's not the way.

Wait, another thought: Maybe I can multiply some numbers. Let's see, 46 times 54 is way too big. 46 times 52 is even bigger. 54 times 52 is also too big. So multiplication might not help here. Maybe division can help.

Let me try subtracting some numbers. If I take 54 minus 46, that's 8. Then, 52 minus 77 is negative, which might not help. Alternatively, 77 minus 52 is 25. Maybe I can use that 25 somehow. If I have 25, how can I get to 75? Maybe 25 times 3 is 75. But do I have a 3? Hmm, maybe I can get 3 by doing something with the other numbers.

Wait, 54 minus 46 is 8, and 52 minus 77 is -25, but maybe I can flip that. Wait, 77 minus 52 is 25. So if I have 25, and I need to get to 75, I need 3 more. How can I get 3? Maybe from the other numbers. Let's see, 46, 54, 52, 77. If I do 54 minus 46, that's 8, and then maybe I can subtract 5 to get 3, but do I have a 5? I don't. Alternatively, maybe I can do 52 divided by something. 52 divided by 17 is 3, but I don't have a 17. Hmm.

{[}... extensive exploration of many approaches ...{]}

Wait, maybe I can use all four numbers in a more complex way. Like, (46 + 54) is 100, and (52 - 77) is -25. Then, 100 + (-25) is 75! Oh, wait, that works!

Wait, let me check: 46 + 54 is 100, and 52 - 77 is -25. So, 100 + (-25) is 75. Yes, that's it! So the equation would be (46 + 54) + (52 - 77) = 75.

Wait, but let me make sure I'm not using any numbers more than once. I used 46, 54, 52, and 77, each once. So that's correct.

So, the equation is (46 + 54) + (52 - 77) = 75. Let me calculate it: 46 + 54 is 100, 52 - 77 is -25, and 100 + (-25) is 75. Perfect!

So, that's the solution.

\texttt{\textless/think\textgreater}

\texttt{\textless answer\textgreater}

(46 + 54) + (52 - 77)

\texttt{\textless/answer\textgreater}
& \textbf{RIGHT} \\

\end{longtable}
\normalsize
\phantomsection
\section{Supplementary Information}\label{sec:suppinfo}

\phantomsection
\subsection{Supplementary Methods: Annotations Examples}\label{sec:annotations}
\subsubsection*{DeepSeek-R1: Chemistry}\label{sec:r1_chemistry}

\begin{promptbox}
What is the correct answer to this question: 7-(tert-butoxy)bicyclo[2.2.1]hepta-2,5-diene is combined with 2 equivalents of 5,6-bis(dibromomethyl)cyclohexa-1,3-diene and sodium iodide at elevated temperature, forming product 1.

1 is treated with aqueous sulfuric acid, forming 2

2 is treated with SO3 and pyridine in DMSO, forming 3.

3 is heated at 150C, forming final product 4.

How many chemically distinct hydrogen atoms are there on 4?

Choices:
(A) 10
(B) 8
(C) 7
(D) 4

Answer: Enclose your step-by-step reasoning within <think> and </think> before answering. Do not answer directly without reasoning.
\end{promptbox}

\begin{longtable}[]{@{}
  >{\raggedright\arraybackslash}p{(\columnwidth - 4\tabcolsep) * \real{0.1689}}
  >{\raggedright\arraybackslash}p{(\columnwidth - 4\tabcolsep) * \real{0.2560}}
  >{\raggedright\arraybackslash}p{(\columnwidth - 4\tabcolsep) * \real{0.5751}}@{}}
\toprule\noalign{}
\endhead
\bottomrule\noalign{}
\endlastfoot
\textbf{Perspective ID} & \textbf{Personality} & \textbf{Domain Expertise} \\
\midrule

1 & \{'E\textquotesingle: 2.5, 'A\textquotesingle: 4.0, 'C\textquotesingle: 5.0, 'N\textquotesingle: 2.0, 'O\textquotesingle: 3.5\} & Planner/Executor: Methodically breaks down the problem, identifies reagents, and predicts the outcome of each step in the synthesis. \\
2 & \{'E\textquotesingle: 1.5, 'A\textquotesingle: 4.0, 'C\textquotesingle: 4.0, 'N\textquotesingle: 1.0, 'O\textquotesingle: 4.0\} & Associative Expert: Accesses a deep knowledge base to recall analogous reactions and common synthetic patterns, providing key insights and hypotheses. \\
3 & \{'E\textquotesingle: 1.5, 'A\textquotesingle: 3.0, 'C\textquotesingle: 3.5, 'N\textquotesingle: 5.0, 'O\textquotesingle: 4.5\} & Visualizer/Builder: Attempts to construct and name complex 3D molecular structures, struggling with complexity but capable of finding new ways to conceptualize the problem. \\
4 & \{'E\textquotesingle: 1.0, 'A\textquotesingle: 1.0, 'C\textquotesingle: 5.0, 'N\textquotesingle: 2.0, 'O\textquotesingle: 2.0\} & Critical Verifier: Skeptically re-evaluates prior assumptions, performs detailed calculations (e.g., molecular formulas) to check for consistency, and identifies logical errors. \\
5 & \{'E\textquotesingle: 3.0, 'A\textquotesingle: 3.0, 'C\textquotesingle: 2.0, 'N\textquotesingle: 1.0, 'O\textquotesingle: 3.5\} & Pragmatist/Strategist: Engages in meta-cognition about the problem-solving process, considers external constraints like multiple-choice options, and makes a final decision under uncertainty. \\
\end{longtable}

\clearpage
\begin{longtable}[]{@{}
  >{\raggedright\arraybackslash}p{(\columnwidth - 4\tabcolsep) * \real{0.14}}
  >{\raggedright\arraybackslash}p{(\columnwidth - 4\tabcolsep) * \real{0.86}}
  >{\raggedright\arraybackslash}p{(\columnwidth - 4\tabcolsep) * \real{0}}@{}}
\toprule\noalign{}
\endhead
\bottomrule\noalign{}
\endlastfoot
\textbf{Perspective ID} & \textbf{Utterance} & \\ \midrule

1 & \textbf{Conversational Behaviour:} \{'question\_and\_answering\textquotesingle: 1, 'perspective\_shift\textquotesingle: 1, 'conflict\_of\_perspectives\textquotesingle: 1, 'reconciliation\textquotesingle: 0\}
\par\medskip
\textbf{IPA:} \{\textquotesingle1\_Shows\_solidarity\textquotesingle: 0, \textquotesingle2\_Shows\_tension\_release\textquotesingle: 0, \textquotesingle3\_Agrees\textquotesingle: 0, \textquotesingle4\_Gives\_suggestion\textquotesingle: 1, \textquotesingle5\_Gives\_opinion\textquotesingle: 3, \textquotesingle6\_Gives\_orientation\textquotesingle: 7, \textquotesingle7\_Asks\_for\_orientation\textquotesingle: 0, \textquotesingle8\_Asks\_for\_opinion\textquotesingle: 0, \textquotesingle9\_Asks\_for\_suggestion\textquotesingle: 0, \textquotesingle10\_Disagrees\textquotesingle: 0, \textquotesingle11\_Shows\_tension\textquotesingle: 0, \textquotesingle12\_Shows\_antagonism\textquotesingle: 0\}
\par\medskip
\textbf{Cognitive Behaviour:} \{'verification\textquotesingle: 0, 'backtracking\textquotesingle: 0, 'subgoal\_setting\textquotesingle: 1, 'backward\_chaining\textquotesingle: 0\}
\par\medskip
First, I need to figure out what the final product 4 is and then determine how many chemically distinct hydrogen atoms it has. The choices are 10, 8, 7, or 4.
\par\medskip
The sequence starts with 7-(tert-butoxy)bicyclo{[}2.2.1{]}hepta-2,5-diene. Bicyclo{[}2.2.1{]}heptane is norbornane, and hepta-2,5-diene means it has double bonds at positions 2 and 5. So it's a norbornadiene derivative with a tert-butoxy group at position 7. Position 7 is the bridgehead carbon.
\par\medskip
Norbornadiene has two double bonds, and the tert-butoxy is attached to the carbon that bridges the two methylene groups.
\par\medskip
Then it's combined with 2 equivalents of 5,6-bis(dibromomethyl)cyclohexa-1,3-diene. That sounds like a diene with dibromomethyl groups on it. Cyclohexa-1,3-diene has double bonds between 1-2 and 3-4, so positions 5 and 6 are the methylene carbons, but here it's bis(dibromomethyl), so it's probably CHBr2 groups attached to positions 5 and 6. 5,6-bis(dibromomethyl)cyclohexa-1,3-diene means that at carbons 5 and 6, there are dibromomethyl substituents, so it's like 1,2-di(bromomethyl)benzene or something, but it's a diene, not aromatic. & \\ \midrule
4 & \textbf{Conversational Behaviour:} \{'question\_and\_answering\textquotesingle: 1, 'perspective\_shift\textquotesingle: 1, 'conflict\_of\_perspectives\textquotesingle: 1, 'reconciliation\textquotesingle: 0\}
\par\medskip
\textbf{IPA:} \{\textquotesingle1\_Shows\_solidarity\textquotesingle: 0, \textquotesingle2\_Shows\_tension\_release\textquotesingle: 0, \textquotesingle3\_Agrees\textquotesingle: 0, \textquotesingle4\_Gives\_suggestion\textquotesingle: 0, \textquotesingle5\_Gives\_opinion\textquotesingle: 2, \textquotesingle6\_Gives\_orientation\textquotesingle: 5, \textquotesingle7\_Asks\_for\_orientation\textquotesingle: 1, \textquotesingle8\_Asks\_for\_opinion\textquotesingle: 0, \textquotesingle9\_Asks\_for\_suggestion\textquotesingle: 0, \textquotesingle10\_Disagrees\textquotesingle: 1, \textquotesingle11\_Shows\_tension\textquotesingle: 0, \textquotesingle12\_Shows\_antagonism\textquotesingle: 0\}
\par\medskip
\textbf{Cognitive Behaviour:} \{'verification\textquotesingle: 0, 'backtracking\textquotesingle: 1, 'subgoal\_setting\textquotesingle: 0, 'backward\_chaining\textquotesingle: 0\}
\par\medskip
Cyclohexa-1,3-diene is not aromatic; it's a simple diene. So 5,6-bis(dibromomethyl)cyclohexa-1,3-diene should have the dibromomethyl groups on the saturated carbons. In cyclohexa-1,3-diene, carbons 1,2,3,4 are involved in double bonds, so carbons 5 and 6 are the CH2 groups, but since it's a diene, positions 5 and 6 are equivalent and are sp3 carbons. So bis(dibromomethyl) means each has a CHBr2 group instead of H, so it's 1,2-di(dibromomethyl)cyclohexa-1,3-diene? No, the carbon atoms 5 and 6 are part of the ring, so if they are bis(dibromomethyl), it might mean that the hydrogens are replaced, but typically, it's specified as substituents. & \\ \midrule
3 & \textbf{Conversational Behaviour:} \{'question\_and\_answering\textquotesingle: 1, 'perspective\_shift\textquotesingle: 2, 'conflict\_of\_perspectives\textquotesingle: 1, 'reconciliation\textquotesingle: 1\}
\par\medskip
\textbf{IPA:} \{\textquotesingle1\_Shows\_solidarity\textquotesingle: 0, \textquotesingle2\_Shows\_tension\_release\textquotesingle: 0, \textquotesingle3\_Agrees\textquotesingle: 0, \textquotesingle4\_Gives\_suggestion\textquotesingle: 0, \textquotesingle5\_Gives\_opinion\textquotesingle: 3, \textquotesingle6\_Gives\_orientation\textquotesingle: 2, \textquotesingle7\_Asks\_for\_orientation\textquotesingle: 1, \textquotesingle8\_Asks\_for\_opinion\textquotesingle: 0, \textquotesingle9\_Asks\_for\_suggestion\textquotesingle: 0, \textquotesingle10\_Disagrees\textquotesingle: 0, \textquotesingle11\_Shows\_tension\textquotesingle: 0, \textquotesingle12\_Shows\_antagonism\textquotesingle: 0\}
\par\medskip
\textbf{Cognitive Behaviour:} \{'verification\textquotesingle: 0, 'backtracking\textquotesingle: 1, 'subgoal\_setting\textquotesingle: 0, 'backward\_chaining\textquotesingle: 0\}
\par\medskip
I think 5,6-bis(dibromomethyl) means that at positions 5 and 6, there are -CHBr2 groups attached, but positions 5 and 6 are ring carbons. In standard naming, for cyclohexa-1,3-diene, carbon 1 and 2 are double-bonded, 3 and 4 double-bonded, so carbon 5 is between 4 and 6, carbon 6 between 5 and 1. Carbons 5 and 6 are sp3 carbons with two hydrogens each. So bis(dibromomethyl) likely means that each of these carbons has one hydrogen replaced by a CHBr2 group? That would be unusual. Perhaps it's 5,6-di(dibromomethyl), meaning the groups are attached to C5 and C6. & \\ \midrule
2 & \textbf{Conversational Behaviour:} \{'question\_and\_answering\textquotesingle: 0, 'perspective\_shift\textquotesingle: 0, 'conflict\_of\_perspectives\textquotesingle: 1, 'reconciliation\textquotesingle: 0\}
\par\medskip
\textbf{IPA:} \{\textquotesingle1\_Shows\_solidarity\textquotesingle: 0, \textquotesingle2\_Shows\_tension\_release\textquotesingle: 0, \textquotesingle3\_Agrees\textquotesingle: 0, \textquotesingle4\_Gives\_suggestion\textquotesingle: 0, \textquotesingle5\_Gives\_opinion\textquotesingle: 0, \textquotesingle6\_Gives\_orientation\textquotesingle: 3, \textquotesingle7\_Asks\_for\_orientation\textquotesingle: 0, \textquotesingle8\_Asks\_for\_opinion\textquotesingle: 0, \textquotesingle9\_Asks\_for\_suggestion\textquotesingle: 0, \textquotesingle10\_Disagrees\textquotesingle: 0, \textquotesingle11\_Shows\_tension\textquotesingle: 0, \textquotesingle12\_Shows\_antagonism\textquotesingle: 0\}
\par\medskip
\textbf{Cognitive Behaviour:} \{'verification\textquotesingle: 0, 'backtracking\textquotesingle: 0, 'subgoal\_setting\textquotesingle: 0, 'backward\_chaining\textquotesingle: 0\}
\par\medskip
I recall that compounds like this are used in Diels-Alder reactions or other cycloadditions. 5,6-bis(dibromomethyl)cyclohexa-1,3-diene is a precursor to o-xylylene or something. o-Xylylene is generated from $\alpha$,$\alpha$\textquotesingle-dibromo-o-xylene, but here it's a diene. & \\ \midrule
4 & \textbf{Conversational Behaviour:} \{'question\_and\_answering\textquotesingle: 0, 'perspective\_shift\textquotesingle: 2, 'conflict\_of\_perspectives\textquotesingle: 1, 'reconciliation\textquotesingle: 0\}
\par\medskip
\textbf{IPA:} \{\textquotesingle1\_Shows\_solidarity\textquotesingle: 0, \textquotesingle2\_Shows\_tension\_release\textquotesingle: 0, \textquotesingle3\_Agrees\textquotesingle: 0, \textquotesingle4\_Gives\_suggestion\textquotesingle: 0, \textquotesingle5\_Gives\_opinion\textquotesingle: 3, \textquotesingle6\_Gives\_orientation\textquotesingle: 0, \textquotesingle7\_Asks\_for\_orientation\textquotesingle: 0, \textquotesingle8\_Asks\_for\_opinion\textquotesingle: 0, \textquotesingle9\_Asks\_for\_suggestion\textquotesingle: 0, \textquotesingle10\_Disagrees\textquotesingle: 0, \textquotesingle11\_Shows\_tension\textquotesingle: 1, \textquotesingle12\_Shows\_antagonism\textquotesingle: 0\}
\par\medskip
\textbf{Cognitive Behaviour:} \{'verification\textquotesingle: 0, 'backtracking\textquotesingle: 1, 'subgoal\_setting\textquotesingle: 0, 'backward\_chaining\textquotesingle: 0\}
\par\medskip
Upon second thought, 5,6-bis(dibromomethyl)cyclohexa-1,3-diene might be a misnomer, or perhaps it's the same as 3,6-di(bromomethyl)cyclohexa-1,4-diene or something. I think this is a way to generate the o-quinodimethane intermediate. & \\ \midrule
2 & \textbf{Conversational Behaviour:} \{'question\_and\_answering\textquotesingle: 0, 'perspective\_shift\textquotesingle: 1, 'conflict\_of\_perspectives\textquotesingle: 0, 'reconciliation\textquotesingle: 0\}
\par\medskip
\textbf{IPA:} \{\textquotesingle1\_Shows\_solidarity\textquotesingle: 0, \textquotesingle2\_Shows\_tension\_release\textquotesingle: 0, \textquotesingle3\_Agrees\textquotesingle: 0, \textquotesingle4\_Gives\_suggestion\textquotesingle: 0, \textquotesingle5\_Gives\_opinion\textquotesingle: 1, \textquotesingle6\_Gives\_orientation\textquotesingle: 1, \textquotesingle7\_Asks\_for\_orientation\textquotesingle: 0, \textquotesingle8\_Asks\_for\_opinion\textquotesingle: 0, \textquotesingle9\_Asks\_for\_suggestion\textquotesingle: 0, \textquotesingle10\_Disagrees\textquotesingle: 0, \textquotesingle11\_Shows\_tension\textquotesingle: 0, \textquotesingle12\_Shows\_antagonism\textquotesingle: 0\}
\par\medskip
\textbf{Cognitive Behaviour:} \{'verification\textquotesingle: 0, 'backtracking\textquotesingle: 0, 'subgoal\_setting\textquotesingle: 0, 'backward\_chaining\textquotesingle: 0\}
\par\medskip
In organic synthesis, o-xylylene is often generated from $\alpha$,$\alpha$\textquotesingle-dibromo-o-xylene by dehydrohalogenation. Similarly, here, 5,6-bis(dibromomethyl)cyclohexa-1,3-diene might be equivalent to having the dibromomethyl groups on a diene system that can undergo elimination to form a quinodimethane. & \\ \midrule
3 & \textbf{Conversational Behaviour:} \{'question\_and\_answering\textquotesingle: 0, 'perspective\_shift\textquotesingle: 1, 'conflict\_of\_perspectives\textquotesingle: 1, 'reconciliation\textquotesingle: 1\}
\par\medskip
\textbf{IPA:} \{\textquotesingle1\_Shows\_solidarity\textquotesingle: 0, \textquotesingle2\_Shows\_tension\_release\textquotesingle: 0, \textquotesingle3\_Agrees\textquotesingle: 0, \textquotesingle4\_Gives\_suggestion\textquotesingle: 0, \textquotesingle5\_Gives\_opinion\textquotesingle: 1, \textquotesingle6\_Gives\_orientation\textquotesingle: 1, \textquotesingle7\_Asks\_for\_orientation\textquotesingle: 0, \textquotesingle8\_Asks\_for\_opinion\textquotesingle: 0, \textquotesingle9\_Asks\_for\_suggestion\textquotesingle: 0, \textquotesingle10\_Disagrees\textquotesingle: 0, \textquotesingle11\_Shows\_tension\textquotesingle: 0, \textquotesingle12\_Shows\_antagonism\textquotesingle: 0\}
\par\medskip
\textbf{Cognitive Behaviour:} \{'verification\textquotesingle: 0, 'backtracking\textquotesingle: 0, 'subgoal\_setting\textquotesingle: 0, 'backward\_chaining\textquotesingle: 0\}
\par\medskip
Cyclohexa-1,3-diene with bis(dibromomethyl) at 5 and 6: since 5 and 6 are adjacent carbons, each with a CHBr2 group, so it's like 1,2-di(dibromomethyl)cyclohexa-1,3-diene. But in standard naming, it might be called 1,2-bis(dibromomethyl)benzene if it were aromatic, but it's not; it's diene. & \\ \midrule
1 & \textbf{Conversational Behaviour:} \{'question\_and\_answering\textquotesingle: 0, 'perspective\_shift\textquotesingle: 0, 'conflict\_of\_perspectives\textquotesingle: 0, 'reconciliation\textquotesingle: 0\}
\par\medskip
\textbf{IPA:} \{\textquotesingle1\_Shows\_solidarity\textquotesingle: 0, \textquotesingle2\_Shows\_tension\_release\textquotesingle: 0, \textquotesingle3\_Agrees\textquotesingle: 0, \textquotesingle4\_Gives\_suggestion\textquotesingle: 0, \textquotesingle5\_Gives\_opinion\textquotesingle: 1, \textquotesingle6\_Gives\_orientation\textquotesingle: 0, \textquotesingle7\_Asks\_for\_orientation\textquotesingle: 0, \textquotesingle8\_Asks\_for\_opinion\textquotesingle: 0, \textquotesingle9\_Asks\_for\_suggestion\textquotesingle: 0, \textquotesingle10\_Disagrees\textquotesingle: 0, \textquotesingle11\_Shows\_tension\textquotesingle: 0, \textquotesingle12\_Shows\_antagonism\textquotesingle: 0\}
\par\medskip
\textbf{Cognitive Behaviour:} \{'verification\textquotesingle: 0, 'backtracking\textquotesingle: 0, 'subgoal\_setting\textquotesingle: 0, 'backward\_chaining\textquotesingle: 0\}
\par\medskip
Perhaps upon treatment with sodium iodide, it undergoes dehalogenation to form the diradical or diene for Diels-Alder. & \\ \midrule
2 & \textbf{Conversational Behaviour:} \{'question\_and\_answering\textquotesingle: 0, 'perspective\_shift\textquotesingle: 3, 'conflict\_of\_perspectives\textquotesingle: 5, 'reconciliation\textquotesingle: 0\}
\par\medskip
\textbf{IPA:} \{\textquotesingle1\_Shows\_solidarity\textquotesingle: 0, \textquotesingle2\_Shows\_tension\_release\textquotesingle: 0, \textquotesingle3\_Agrees\textquotesingle: 0, \textquotesingle4\_Gives\_suggestion\textquotesingle: 0, \textquotesingle5\_Gives\_opinion\textquotesingle: 4, \textquotesingle6\_Gives\_orientation\textquotesingle: 6, \textquotesingle7\_Asks\_for\_orientation\textquotesingle: 0, \textquotesingle8\_Asks\_for\_opinion\textquotesingle: 0, \textquotesingle9\_Asks\_for\_suggestion\textquotesingle: 0, \textquotesingle10\_Disagrees\textquotesingle: 0, \textquotesingle11\_Shows\_tension\textquotesingle: 3, \textquotesingle12\_Shows\_antagonism\textquotesingle: 0\}
\par\medskip
\textbf{Cognitive Behaviour:} \{'verification\textquotesingle: 0, 'backtracking\textquotesingle: 0, 'subgoal\_setting\textquotesingle: 0, 'backward\_chaining\textquotesingle: 0\}
\par\medskip
The reaction is with sodium iodide at elevated temperature. Sodium iodide can promote dehalogenation or substitution. For dibromomethyl groups, NaI might facilitate the formation of the dibromide or something, but I think for generating quinodimethanes, it's often done with zinc or other metals, but NaI can be used for Finkelstein reaction, but here it's dehydrohalogenation.
\par\medskip
I recall that $\alpha$,$\alpha$\textquotesingle-dibromo-o-xylene with NaI gives o-xylylene, which is a reactive diene for Diels-Alder.
\par\medskip
Similarly, here, 5,6-bis(dibromomethyl)cyclohexa-1,3-diene might be analogous. Cyclohexa-1,3-diene with substituents at 5 and 6: if it were benzene, it would be o-xylene-like, but it's not aromatic. Actually, 5,6-bis(dibromomethyl)cyclohexa-1,3-diene can tautomerize or something, but I think it's designed to lose HBr under conditions to form the extended diene.
\par\medskip
Perhaps upon heating with NaI, it eliminates to form a benzoquinone or something, but I think it's for generating o-quinodimethane. & \\ \midrule
4 & \textbf{Conversational Behaviour:} \{'question\_and\_answering\textquotesingle: 1, 'perspective\_shift\textquotesingle: 1, 'conflict\_of\_perspectives\textquotesingle: 2, 'reconciliation\textquotesingle: 0\}
\par\medskip
\textbf{IPA:} \{\textquotesingle1\_Shows\_solidarity\textquotesingle: 0, \textquotesingle2\_Shows\_tension\_release\textquotesingle: 0, \textquotesingle3\_Agrees\textquotesingle: 0, \textquotesingle4\_Gives\_suggestion\textquotesingle: 1, \textquotesingle5\_Gives\_opinion\textquotesingle: 2, \textquotesingle6\_Gives\_orientation\textquotesingle: 1, \textquotesingle7\_Asks\_for\_orientation\textquotesingle: 0, \textquotesingle8\_Asks\_for\_opinion\textquotesingle: 0, \textquotesingle9\_Asks\_for\_suggestion\textquotesingle: 0, \textquotesingle10\_Disagrees\textquotesingle: 1, \textquotesingle11\_Shows\_tension\textquotesingle: 0, \textquotesingle12\_Shows\_antagonism\textquotesingle: 0\}
\par\medskip
\textbf{Cognitive Behaviour:} \{'verification\textquotesingle: 0, 'backtracking\textquotesingle: 1, 'subgoal\_setting\textquotesingle: 0, 'backward\_chaining\textquotesingle: 0\}
\par\medskip
Let's look at the structure. Cyclohexa-1,3-diene has the double bonds conjugated, but positions 5 and 6 are saturated. If each has a dibromomethyl group, then with base or iodide, it can eliminate two HBr to form a new double bond between C5 and C6, creating a 1,3-cyclohexadiene with exocyclic double bonds or something. Actually, elimination would give a diene like 5,6-dehydrocyclohexa-1,3-diene, which is benzene, but with dibromomethyl, it's different. & \\ \midrule
5 & \textbf{Conversational Behaviour:} \{'question\_and\_answering\textquotesingle: 0, 'perspective\_shift\textquotesingle: 1, 'conflict\_of\_perspectives\textquotesingle: 0, 'reconciliation\textquotesingle: 0\}
\par\medskip
\textbf{IPA:} \{\textquotesingle1\_Shows\_solidarity\textquotesingle: 0, \textquotesingle2\_Shows\_tension\_release\textquotesingle: 0, \textquotesingle3\_Agrees\textquotesingle: 0, \textquotesingle4\_Gives\_suggestion\textquotesingle: 1, \textquotesingle5\_Gives\_opinion\textquotesingle: 1, \textquotesingle6\_Gives\_orientation\textquotesingle: 1, \textquotesingle7\_Asks\_for\_orientation\textquotesingle: 0, \textquotesingle8\_Asks\_for\_opinion\textquotesingle: 0, \textquotesingle9\_Asks\_for\_suggestion\textquotesingle: 0, \textquotesingle10\_Disagrees\textquotesingle: 0, \textquotesingle11\_Shows\_tension\textquotesingle: 1, \textquotesingle12\_Shows\_antagonism\textquotesingle: 0\}
\par\medskip
\textbf{Cognitive Behaviour:} \{'verification\textquotesingle: 0, 'backtracking\textquotesingle: 1, 'subgoal\_setting\textquotesingle: 0, 'backward\_chaining\textquotesingle: 0\}
\par\medskip
I found a better way: 5,6-bis(dibromomethyl)cyclohexa-1,3-diene is equivalent to 3,6-di(bromomethyl)cyclohexa-1,4-diene or something. I think it's a precursor to the o-xylylene diene. & \\ \midrule
2 & \textbf{Conversational Behaviour:} \{'question\_and\_answering\textquotesingle: 0, 'perspective\_shift\textquotesingle: 0, 'conflict\_of\_perspectives\textquotesingle: 0, 'reconciliation\textquotesingle: 0\}
\par\medskip
\textbf{IPA:} \{\textquotesingle1\_Shows\_solidarity\textquotesingle: 0, \textquotesingle2\_Shows\_tension\_release\textquotesingle: 0, \textquotesingle3\_Agrees\textquotesingle: 0, \textquotesingle4\_Gives\_suggestion\textquotesingle: 0, \textquotesingle5\_Gives\_opinion\textquotesingle: 0, \textquotesingle6\_Gives\_orientation\textquotesingle: 2, \textquotesingle7\_Asks\_for\_orientation\textquotesingle: 0, \textquotesingle8\_Asks\_for\_opinion\textquotesingle: 0, \textquotesingle9\_Asks\_for\_suggestion\textquotesingle: 0, \textquotesingle10\_Disagrees\textquotesingle: 0, \textquotesingle11\_Shows\_tension\textquotesingle: 0, \textquotesingle12\_Shows\_antagonism\textquotesingle: 0\}
\par\medskip
\textbf{Cognitive Behaviour:} \{'verification\textquotesingle: 0, 'backtracking\textquotesingle: 0, 'subgoal\_setting\textquotesingle: 0, 'backward\_chaining\textquotesingle: 0\}
\par\medskip
Upon searching my memory, compounds like this are used to generate o-quinodimethane by dehydrohalogenation. For example, $\alpha$,$\alpha$\textquotesingle-dibromo-o-xylene with NaI gives o-xylylene. & \\ \midrule
4 & \textbf{Conversational Behaviour:} \{'question\_and\_answering\textquotesingle: 0, 'perspective\_shift\textquotesingle: 1, 'conflict\_of\_perspectives\textquotesingle: 1, 'reconciliation\textquotesingle: 0\}
\par\medskip
\textbf{IPA:} \{\textquotesingle1\_Shows\_solidarity\textquotesingle: 0, \textquotesingle2\_Shows\_tension\_release\textquotesingle: 0, \textquotesingle3\_Agrees\textquotesingle: 0, \textquotesingle4\_Gives\_suggestion\textquotesingle: 0, \textquotesingle5\_Gives\_opinion\textquotesingle: 2, \textquotesingle6\_Gives\_orientation\textquotesingle: 1, \textquotesingle7\_Asks\_for\_orientation\textquotesingle: 0, \textquotesingle8\_Asks\_for\_opinion\textquotesingle: 0, \textquotesingle9\_Asks\_for\_suggestion\textquotesingle: 0, \textquotesingle10\_Disagrees\textquotesingle: 0, \textquotesingle11\_Shows\_tension\textquotesingle: 0, \textquotesingle12\_Shows\_antagonism\textquotesingle: 0\}
\par\medskip
\textbf{Cognitive Behaviour:} \{'verification\textquotesingle: 0, 'backtracking\textquotesingle: 1, 'subgoal\_setting\textquotesingle: 0, 'backward\_chaining\textquotesingle: 0\}
\par\medskip
But here, it's cyclohexa-1,3-diene, not benzene. Cyclohexa-1,3-diene with 5,6-bis(dibromomethyl) might not be aromatic, but when it eliminates, it could form a quinodimethane system. & \\ \midrule
1 & \textbf{Conversational Behaviour:} \{'question\_and\_answering\textquotesingle: 1, 'perspective\_shift\textquotesingle: 1, 'conflict\_of\_perspectives\textquotesingle: 1, 'reconciliation\textquotesingle: 1\}
\par\medskip
\textbf{IPA:} \{\textquotesingle1\_Shows\_solidarity\textquotesingle: 0, \textquotesingle2\_Shows\_tension\_release\textquotesingle: 0, \textquotesingle3\_Agrees\textquotesingle: 0, \textquotesingle4\_Gives\_suggestion\textquotesingle: 0, \textquotesingle5\_Gives\_opinion\textquotesingle: 2, \textquotesingle6\_Gives\_orientation\textquotesingle: 2, \textquotesingle7\_Asks\_for\_orientation\textquotesingle: 1, \textquotesingle8\_Asks\_for\_opinion\textquotesingle: 0, \textquotesingle9\_Asks\_for\_suggestion\textquotesingle: 0, \textquotesingle10\_Disagrees\textquotesingle: 0, \textquotesingle11\_Shows\_tension\textquotesingle: 0, \textquotesingle12\_Shows\_antagonism\textquotesingle: 0\}
\par\medskip
\textbf{Cognitive Behaviour:} \{'verification\textquotesingle: 0, 'backtracking\textquotesingle: 0, 'subgoal\_setting\textquotesingle: 0, 'backward\_chaining\textquotesingle: 0\}
\par\medskip
Perhaps the diene is set up for Diels-Alder.
\par\medskip
The first compound is 7-(tert-butoxy)bicyclo{[}2.2.1{]}hepta-2,5-diene, which is a dienophile or diene? Norbornadiene has two double bonds and can act as a dienophile in some cases, but typically it's a diene for Diels-Alder with electron-withdrawing groups.
\par\medskip
7-tert-butoxy might make it electron-rich, so perhaps it's a diene. & \\ \midrule
3 & \textbf{Conversational Behaviour:} \{'question\_and\_answering\textquotesingle: 0, 'perspective\_shift\textquotesingle: 1, 'conflict\_of\_perspectives\textquotesingle: 0, 'reconciliation\textquotesingle: 0\}
\par\medskip
\textbf{IPA:} \{\textquotesingle1\_Shows\_solidarity\textquotesingle: 0, \textquotesingle2\_Shows\_tension\_release\textquotesingle: 0, \textquotesingle3\_Agrees\textquotesingle: 0, \textquotesingle4\_Gives\_suggestion\textquotesingle: 0, \textquotesingle5\_Gives\_opinion\textquotesingle: 0, \textquotesingle6\_Gives\_orientation\textquotesingle: 2, \textquotesingle7\_Asks\_for\_orientation\textquotesingle: 0, \textquotesingle8\_Asks\_for\_opinion\textquotesingle: 0, \textquotesingle9\_Asks\_for\_suggestion\textquotesingle: 0, \textquotesingle10\_Disagrees\textquotesingle: 0, \textquotesingle11\_Shows\_tension\textquotesingle: 0, \textquotesingle12\_Shows\_antagonism\textquotesingle: 0\}
\par\medskip
\textbf{Cognitive Behaviour:} \{'verification\textquotesingle: 0, 'backtracking\textquotesingle: 0, 'subgoal\_setting\textquotesingle: 0, 'backward\_chaining\textquotesingle: 0\}
\par\medskip
Bicyclo{[}2.2.1{]}hepta-2,5-diene has the double bonds at 2-3 and 5-6, I think, with bridge at 1 and 4, and 7 is the methine bridge.
\par\medskip
Standard numbering: carbon 1 and 4 are bridgeheads, 2-3 is one double bond, 5-6 is the other, 7 is the CH or C with substituent. & \\ \midrule
1 & \textbf{Conversational Behaviour:} \{'question\_and\_answering\textquotesingle: 0, 'perspective\_shift\textquotesingle: 1, 'conflict\_of\_perspectives\textquotesingle: 0, 'reconciliation\textquotesingle: 0\}
\par\medskip
\textbf{IPA:} \{\textquotesingle1\_Shows\_solidarity\textquotesingle: 0, \textquotesingle2\_Shows\_tension\_release\textquotesingle: 0, \textquotesingle3\_Agrees\textquotesingle: 0, \textquotesingle4\_Gives\_suggestion\textquotesingle: 0, \textquotesingle5\_Gives\_opinion\textquotesingle: 2, \textquotesingle6\_Gives\_orientation\textquotesingle: 1, \textquotesingle7\_Asks\_for\_orientation\textquotesingle: 0, \textquotesingle8\_Asks\_for\_opinion\textquotesingle: 0, \textquotesingle9\_Asks\_for\_suggestion\textquotesingle: 0, \textquotesingle10\_Disagrees\textquotesingle: 0, \textquotesingle11\_Shows\_tension\textquotesingle: 0, \textquotesingle12\_Shows\_antagonism\textquotesingle: 0\}
\par\medskip
\textbf{Cognitive Behaviour:} \{'verification\textquotesingle: 0, 'backtracking\textquotesingle: 0, 'subgoal\_setting\textquotesingle: 0, 'backward\_chaining\textquotesingle: 0\}
\par\medskip
With tert-butoxy at 7, it might be electron-donating, so the double bonds could be dienophiles or dienes. & \\ \midrule
4 & \textbf{Conversational Behaviour:} \{'question\_and\_answering\textquotesingle: 0, 'perspective\_shift\textquotesingle: 0, 'conflict\_of\_perspectives\textquotesingle: 0, 'reconciliation\textquotesingle: 0\}
\par\medskip
\textbf{IPA:} \{\textquotesingle1\_Shows\_solidarity\textquotesingle: 0, \textquotesingle2\_Shows\_tension\_release\textquotesingle: 0, \textquotesingle3\_Agrees\textquotesingle: 0, \textquotesingle4\_Gives\_suggestion\textquotesingle: 0, \textquotesingle5\_Gives\_opinion\textquotesingle: 0, \textquotesingle6\_Gives\_orientation\textquotesingle: 1, \textquotesingle7\_Asks\_for\_orientation\textquotesingle: 0, \textquotesingle8\_Asks\_for\_opinion\textquotesingle: 0, \textquotesingle9\_Asks\_for\_suggestion\textquotesingle: 0, \textquotesingle10\_Disagrees\textquotesingle: 0, \textquotesingle11\_Shows\_tension\textquotesingle: 0, \textquotesingle12\_Shows\_antagonism\textquotesingle: 0\}
\par\medskip
\textbf{Cognitive Behaviour:} \{'verification\textquotesingle: 0, 'backtracking\textquotesingle: 0, 'subgoal\_setting\textquotesingle: 0, 'backward\_chaining\textquotesingle: 0\}
\par\medskip
But it's combined with two equivalents of the other diene, and with NaI at elevated temperature. & \\ \midrule
1 & \textbf{Conversational Behaviour:} \{'question\_and\_answering\textquotesingle: 0, 'perspective\_shift\textquotesingle: 0, 'conflict\_of\_perspectives\textquotesingle: 0, 'reconciliation\textquotesingle: 0\}
\par\medskip
\textbf{IPA:} \{\textquotesingle1\_Shows\_solidarity\textquotesingle: 0, \textquotesingle2\_Shows\_tension\_release\textquotesingle: 0, \textquotesingle3\_Agrees\textquotesingle: 0, \textquotesingle4\_Gives\_suggestion\textquotesingle: 0, \textquotesingle5\_Gives\_opinion\textquotesingle: 2, \textquotesingle6\_Gives\_orientation\textquotesingle: 0, \textquotesingle7\_Asks\_for\_orientation\textquotesingle: 0, \textquotesingle8\_Asks\_for\_opinion\textquotesingle: 0, \textquotesingle9\_Asks\_for\_suggestion\textquotesingle: 0, \textquotesingle10\_Disagrees\textquotesingle: 0, \textquotesingle11\_Shows\_tension\textquotesingle: 0, \textquotesingle12\_Shows\_antagonism\textquotesingle: 0\}
\par\medskip
\textbf{Cognitive Behaviour:} \{'verification\textquotesingle: 0, 'backtracking\textquotesingle: 0, 'subgoal\_setting\textquotesingle: 0, 'backward\_chaining\textquotesingle: 0\}
\par\medskip
I think the NaI is used to generate the reactive diene from the bis(dibromomethyl) compound.
\par\medskip
So, 5,6-bis(dibromomethyl)cyclohexa-1,3-diene with NaI likely eliminates to form o-quinodimethane or a derivative. & \\ \midrule
2 & \textbf{Conversational Behaviour:} \{'question\_and\_answering\textquotesingle: 0, 'perspective\_shift\textquotesingle: 2, 'conflict\_of\_perspectives\textquotesingle: 0, 'reconciliation\textquotesingle: 0\}
\par\medskip
\textbf{IPA:} \{\textquotesingle1\_Shows\_solidarity\textquotesingle: 0, \textquotesingle2\_Shows\_tension\_release\textquotesingle: 0, \textquotesingle3\_Agrees\textquotesingle: 0, \textquotesingle4\_Gives\_suggestion\textquotesingle: 0, \textquotesingle5\_Gives\_opinion\textquotesingle: 0, \textquotesingle6\_Gives\_orientation\textquotesingle: 2, \textquotesingle7\_Asks\_for\_orientation\textquotesingle: 0, \textquotesingle8\_Asks\_for\_opinion\textquotesingle: 0, \textquotesingle9\_Asks\_for\_suggestion\textquotesingle: 0, \textquotesingle10\_Disagrees\textquotesingle: 0, \textquotesingle11\_Shows\_tension\textquotesingle: 0, \textquotesingle12\_Shows\_antagonism\textquotesingle: 0\}
\par\medskip
\textbf{Cognitive Behaviour:} \{'verification\textquotesingle: 0, 'backtracking\textquotesingle: 0, 'subgoal\_setting\textquotesingle: 0, 'backward\_chaining\textquotesingle: 0\}
\par\medskip
In fact, 1,2-bis(dibromomethyl)benzene with NaI gives o-xylylene, which is 1,2- di methylenecyclohexa-3,5-diene or something, but it's benzene with exocyclic double bonds.
\par\medskip
o-Xylylene is often depicted as having a quinoid structure. & \\ \midrule
4 & \textbf{Conversational Behaviour:} \{'question\_and\_answering\textquotesingle: 1, 'perspective\_shift\textquotesingle: 2, 'conflict\_of\_perspectives\textquotesingle: 1, 'reconciliation\textquotesingle: 1\}
\par\medskip
\textbf{IPA:} \{\textquotesingle1\_Shows\_solidarity\textquotesingle: 0, \textquotesingle2\_Shows\_tension\_release\textquotesingle: 0, \textquotesingle3\_Agrees\textquotesingle: 0, \textquotesingle4\_Gives\_suggestion\textquotesingle: 0, \textquotesingle5\_Gives\_opinion\textquotesingle: 1, \textquotesingle6\_Gives\_orientation\textquotesingle: 2, \textquotesingle7\_Asks\_for\_orientation\textquotesingle: 0, \textquotesingle8\_Asks\_for\_opinion\textquotesingle: 0, \textquotesingle9\_Asks\_for\_suggestion\textquotesingle: 0, \textquotesingle10\_Disagrees\textquotesingle: 1, \textquotesingle11\_Shows\_tension\textquotesingle: 0, \textquotesingle12\_Shows\_antagonism\textquotesingle: 0\}
\par\medskip
\textbf{Cognitive Behaviour:} \{'verification\textquotesingle: 0, 'backtracking\textquotesingle: 0, 'subgoal\_setting\textquotesingle: 0, 'backward\_chaining\textquotesingle: 0\}
\par\medskip
But here, the starting material is cyclohexa-1,3-diene, not benzene. 5,6-bis(dibromomethyl)cyclohexa-1,3-diene: if it eliminates two HBr, it could form 5,6- di methylenecyclohexa-1,3-diene, which is the same as o-xylylene because cyclohexa-1,3-diene with exocyclic double bonds at 5 and 6 is equivalent to the quinodimethane for benzene. & \\ \midrule
3 & \textbf{Conversational Behaviour:} \{'question\_and\_answering\textquotesingle: 0, 'perspective\_shift\textquotesingle: 2, 'conflict\_of\_perspectives\textquotesingle: 1, 'reconciliation\textquotesingle: 0\}
\par\medskip
\textbf{IPA:} \{\textquotesingle1\_Shows\_solidarity\textquotesingle: 0, \textquotesingle2\_Shows\_tension\_release\textquotesingle: 0, \textquotesingle3\_Agrees\textquotesingle: 0, \textquotesingle4\_Gives\_suggestion\textquotesingle: 0, \textquotesingle5\_Gives\_opinion\textquotesingle: 0, \textquotesingle6\_Gives\_orientation\textquotesingle: 3, \textquotesingle7\_Asks\_for\_orientation\textquotesingle: 0, \textquotesingle8\_Asks\_for\_opinion\textquotesingle: 0, \textquotesingle9\_Asks\_for\_suggestion\textquotesingle: 0, \textquotesingle10\_Disagrees\textquotesingle: 0, \textquotesingle11\_Shows\_tension\textquotesingle: 1, \textquotesingle12\_Shows\_antagonism\textquotesingle: 0\}
\par\medskip
\textbf{Cognitive Behaviour:} \{'verification\textquotesingle: 0, 'backtracking\textquotesingle: 0, 'subgoal\_setting\textquotesingle: 0, 'backward\_chaining\textquotesingle: 0\}
\par\medskip
5,6-di methylenecyclohexa-1,3-diene is actually o-xylylene, which is a resonance structure of 1,2- di methylenebenzene or something, but it's a reactive diene with s-cis conformation for Diels-Alder. & \\ \midrule
1 & \textbf{Conversational Behaviour:} \{'question\_and\_answering\textquotesingle: 0, 'perspective\_shift\textquotesingle: 1, 'conflict\_of\_perspectives\textquotesingle: 1, 'reconciliation\textquotesingle: 1\}
\par\medskip
\textbf{IPA:} \{\textquotesingle1\_Shows\_solidarity\textquotesingle: 0, \textquotesingle2\_Shows\_tension\_release\textquotesingle: 0, \textquotesingle3\_Agrees\textquotesingle: 1, \textquotesingle4\_Gives\_suggestion\textquotesingle: 0, \textquotesingle5\_Gives\_opinion\textquotesingle: 2, \textquotesingle6\_Gives\_orientation\textquotesingle: 3, \textquotesingle7\_Asks\_for\_orientation\textquotesingle: 0, \textquotesingle8\_Asks\_for\_opinion\textquotesingle: 0, \textquotesingle9\_Asks\_for\_suggestion\textquotesingle: 0, \textquotesingle10\_Disagrees\textquotesingle: 0, \textquotesingle11\_Shows\_tension\textquotesingle: 0, \textquotesingle12\_Shows\_antagonism\textquotesingle: 0\}
\par\medskip
\textbf{Cognitive Behaviour:} \{'verification\textquotesingle: 0, 'backtracking\textquotesingle: 0, 'subgoal\_setting\textquotesingle: 0, 'backward\_chaining\textquotesingle: 0\}
\par\medskip
Yes, that makes sense. So with NaI, it generates o-xylylene in situ.
\par\medskip
Then, this o-xylylene can act as a diene in Diels-Alder reactions.
\par\medskip
Now, the other compound is 7-(tert-butoxy)norbornadiene. Norbornadiene can act as a dienophile because one of its double bonds can be electron-deficient or something, but with tert-butoxy, it might be electron-rich.
\par\medskip
Norbornadiene has two double bonds, and it can participate in Diels-Alder as a diene with electron-withdrawing groups, but here it has tert-butoxy, which is electron-donating, so perhaps it's a poor dienophile. & \\ \midrule
4 & \textbf{Conversational Behaviour:} \{'question\_and\_answering\textquotesingle: 1, 'perspective\_shift\textquotesingle: 1, 'conflict\_of\_perspectives\textquotesingle: 1, 'reconciliation\textquotesingle: 0\}
\par\medskip
\textbf{IPA:} \{\textquotesingle1\_Shows\_solidarity\textquotesingle: 0, \textquotesingle2\_Shows\_tension\_release\textquotesingle: 0, \textquotesingle3\_Agrees\textquotesingle: 0, \textquotesingle4\_Gives\_suggestion\textquotesingle: 0, \textquotesingle5\_Gives\_opinion\textquotesingle: 1, \textquotesingle6\_Gives\_orientation\textquotesingle: 2, \textquotesingle7\_Asks\_for\_orientation\textquotesingle: 0, \textquotesingle8\_Asks\_for\_opinion\textquotesingle: 0, \textquotesingle9\_Asks\_for\_suggestion\textquotesingle: 0, \textquotesingle10\_Disagrees\textquotesingle: 0, \textquotesingle11\_Shows\_tension\textquotesingle: 1, \textquotesingle12\_Shows\_antagonism\textquotesingle: 0\}
\par\medskip
\textbf{Cognitive Behaviour:} \{'verification\textquotesingle: 0, 'backtracking\textquotesingle: 0, 'subgoal\_setting\textquotesingle: 0, 'backward\_chaining\textquotesingle: 0\}
\par\medskip
But it's combined with two equivalents of the o-xylylene precursor, so likely, the o-xylylene is the diene, and norbornadiene is the dienophile.
\par\medskip
Norbornadiene as a dienophile? Typically, it's symmetric, and one double bond can be attacked. & \\ \midrule
2 & \textbf{Conversational Behaviour:} \{'question\_and\_answering\textquotesingle: 0, 'perspective\_shift\textquotesingle: 0, 'conflict\_of\_perspectives\textquotesingle: 0, 'reconciliation\textquotesingle: 0\}
\par\medskip
\textbf{IPA:} \{\textquotesingle1\_Shows\_solidarity\textquotesingle: 0, \textquotesingle2\_Shows\_tension\_release\textquotesingle: 0, \textquotesingle3\_Agrees\textquotesingle: 0, \textquotesingle4\_Gives\_suggestion\textquotesingle: 0, \textquotesingle5\_Gives\_opinion\textquotesingle: 0, \textquotesingle6\_Gives\_orientation\textquotesingle: 2, \textquotesingle7\_Asks\_for\_orientation\textquotesingle: 0, \textquotesingle8\_Asks\_for\_opinion\textquotesingle: 0, \textquotesingle9\_Asks\_for\_suggestion\textquotesingle: 0, \textquotesingle10\_Disagrees\textquotesingle: 0, \textquotesingle11\_Shows\_tension\textquotesingle: 0, \textquotesingle12\_Shows\_antagonism\textquotesingle: 0\}
\par\medskip
\textbf{Cognitive Behaviour:} \{'verification\textquotesingle: 0, 'backtracking\textquotesingle: 0, 'subgoal\_setting\textquotesingle: 0, 'backward\_chaining\textquotesingle: 0\}
\par\medskip
In Diels-Alder, norbornadiene can act as a dienophile when reacted with dienes.
\par\medskip
For example, norbornadiene with cyclopentadiene gives adducts. & \\ \midrule
1 & \textbf{Conversational Behaviour:} \{'question\_and\_answering\textquotesingle: 0, 'perspective\_shift\textquotesingle: 1, 'conflict\_of\_perspectives\textquotesingle: 0, 'reconciliation\textquotesingle: 0\}
\par\medskip
\textbf{IPA:} \{\textquotesingle1\_Shows\_solidarity\textquotesingle: 0, \textquotesingle2\_Shows\_tension\_release\textquotesingle: 0, \textquotesingle3\_Agrees\textquotesingle: 0, \textquotesingle4\_Gives\_suggestion\textquotesingle: 0, \textquotesingle5\_Gives\_opinion\textquotesingle: 3, \textquotesingle6\_Gives\_orientation\textquotesingle: 4, \textquotesingle7\_Asks\_for\_orientation\textquotesingle: 0, \textquotesingle8\_Asks\_for\_opinion\textquotesingle: 0, \textquotesingle9\_Asks\_for\_suggestion\textquotesingle: 0, \textquotesingle10\_Disagrees\textquotesingle: 0, \textquotesingle11\_Shows\_tension\textquotesingle: 0, \textquotesingle12\_Shows\_antagonism\textquotesingle: 0\}
\par\medskip
\textbf{Cognitive Behaviour:} \{'verification\textquotesingle: 0, 'backtracking\textquotesingle: 0, 'subgoal\_setting\textquotesingle: 1, 'backward\_chaining\textquotesingle: 0\}
\par\medskip
So here, o-xylylene is a good diene, and norbornadiene can be the dienophile.
\par\medskip
But there are two equivalents of the o-xylylene precursor, so probably two Diels-Alder reactions occur.
\par\medskip
First, one o-xylylene adds to norbornadiene.
\par\medskip
Norbornadiene has two double bonds; when one acts as dienophile, the other might still be reactive.
\par\medskip
Perhaps it adds once, and then the adduct has another diene or dienophile. & \\ \midrule
4 & \textbf{Conversational Behaviour:} \{'question\_and\_answering\textquotesingle: 0, 'perspective\_shift\textquotesingle: 1, 'conflict\_of\_perspectives\textquotesingle: 1, 'reconciliation\textquotesingle: 0\}
\par\medskip
\textbf{IPA:} \{\textquotesingle1\_Shows\_solidarity\textquotesingle: 0, \textquotesingle2\_Shows\_tension\_release\textquotesingle: 0, \textquotesingle3\_Agrees\textquotesingle: 0, \textquotesingle4\_Gives\_suggestion\textquotesingle: 0, \textquotesingle5\_Gives\_opinion\textquotesingle: 1, \textquotesingle6\_Gives\_orientation\textquotesingle: 0, \textquotesingle7\_Asks\_for\_orientation\textquotesingle: 0, \textquotesingle8\_Asks\_for\_opinion\textquotesingle: 0, \textquotesingle9\_Asks\_for\_suggestion\textquotesingle: 0, \textquotesingle10\_Disagrees\textquotesingle: 0, \textquotesingle11\_Shows\_tension\textquotesingle: 0, \textquotesingle12\_Shows\_antagonism\textquotesingle: 0\}
\par\medskip
\textbf{Cognitive Behaviour:} \{'verification\textquotesingle: 0, 'backtracking\textquotesingle: 0, 'subgoal\_setting\textquotesingle: 0, 'backward\_chaining\textquotesingle: 0\}
\par\medskip
But with two equivalents, likely two additions. & \\ \midrule
1 & \textbf{Conversational Behaviour:} \{'question\_and\_answering\textquotesingle: 0, 'perspective\_shift\textquotesingle: 1, 'conflict\_of\_perspectives\textquotesingle: 0, 'reconciliation\textquotesingle: 0\}
\par\medskip
\textbf{IPA:} \{\textquotesingle1\_Shows\_solidarity\textquotesingle: 0, \textquotesingle2\_Shows\_tension\_release\textquotesingle: 0, \textquotesingle3\_Agrees\textquotesingle: 0, \textquotesingle4\_Gives\_suggestion\textquotesingle: 0, \textquotesingle5\_Gives\_opinion\textquotesingle: 2, \textquotesingle6\_Gives\_orientation\textquotesingle: 0, \textquotesingle7\_Asks\_for\_orientation\textquotesingle: 0, \textquotesingle8\_Asks\_for\_opinion\textquotesingle: 0, \textquotesingle9\_Asks\_for\_suggestion\textquotesingle: 0, \textquotesingle10\_Disagrees\textquotesingle: 0, \textquotesingle11\_Shows\_tension\textquotesingle: 0, \textquotesingle12\_Shows\_antagonism\textquotesingle: 0\}
\par\medskip
\textbf{Cognitive Behaviour:} \{'verification\textquotesingle: 0, 'backtracking\textquotesingle: 0, 'subgoal\_setting\textquotesingle: 0, 'backward\_chaining\textquotesingle: 0\}
\par\medskip
The tert-butoxy group might play a role.
\par\medskip
Perhaps the tert-butoxy is a protecting group or something. & \\ \midrule
3 & \textbf{Conversational Behaviour:} \{'question\_and\_answering\textquotesingle: 1, 'perspective\_shift\textquotesingle: 0, 'conflict\_of\_perspectives\textquotesingle: 0, 'reconciliation\textquotesingle: 0\}
\par\medskip
\textbf{IPA:} \{\textquotesingle1\_Shows\_solidarity\textquotesingle: 0, \textquotesingle2\_Shows\_tension\_release\textquotesingle: 0, \textquotesingle3\_Agrees\textquotesingle: 0, \textquotesingle4\_Gives\_suggestion\textquotesingle: 1, \textquotesingle5\_Gives\_opinion\textquotesingle: 0, \textquotesingle6\_Gives\_orientation\textquotesingle: 2, \textquotesingle7\_Asks\_for\_orientation\textquotesingle: 0, \textquotesingle8\_Asks\_for\_opinion\textquotesingle: 0, \textquotesingle9\_Asks\_for\_suggestion\textquotesingle: 0, \textquotesingle10\_Disagrees\textquotesingle: 0, \textquotesingle11\_Shows\_tension\textquotesingle: 0, \textquotesingle12\_Shows\_antagonism\textquotesingle: 0\}
\par\medskip
\textbf{Cognitive Behaviour:} \{'verification\textquotesingle: 0, 'backtracking\textquotesingle: 0, 'subgoal\_setting\textquotesingle: 2, 'backward\_chaining\textquotesingle: 0\}
\par\medskip
Let's think about the structure.
\par\medskip
First, o-xylylene generated from 5,6-bis(dibromomethyl)cyclohexa-1,3-diene with NaI: elimination of two HBr gives o-xylylene, which is .
\par\medskip
Then, this diene adds to the norbornadiene. & \\ \midrule
1 & \textbf{Conversational Behaviour:} \{'question\_and\_answering\textquotesingle: 1, 'perspective\_shift\textquotesingle: 2, 'conflict\_of\_perspectives\textquotesingle: 1, 'reconciliation\textquotesingle: 1\}
\par\medskip
\textbf{IPA:} \{\textquotesingle1\_Shows\_solidarity\textquotesingle: 0, \textquotesingle2\_Shows\_tension\_release\textquotesingle: 0, \textquotesingle3\_Agrees\textquotesingle: 0, \textquotesingle4\_Gives\_suggestion\textquotesingle: 1, \textquotesingle5\_Gives\_opinion\textquotesingle: 3, \textquotesingle6\_Gives\_orientation\textquotesingle: 2, \textquotesingle7\_Asks\_for\_orientation\textquotesingle: 0, \textquotesingle8\_Asks\_for\_opinion\textquotesingle: 0, \textquotesingle9\_Asks\_for\_suggestion\textquotesingle: 0, \textquotesingle10\_Disagrees\textquotesingle: 0, \textquotesingle11\_Shows\_tension\textquotesingle: 0, \textquotesingle12\_Shows\_antagonism\textquotesingle: 0\}
\par\medskip
\textbf{Cognitive Behaviour:} \{'verification\textquotesingle: 0, 'backtracking\textquotesingle: 0, 'subgoal\_setting\textquotesingle: 0, 'backward\_chaining\textquotesingle: 0\}
\par\medskip
Norbornadiene, 7-(tert-butoxy), so the double bond might be activated or deactivated.
\par\medskip
tert-butoxy is electron-donating, so it might make the double bond less reactive as dienophile, but norbornadiene double bonds are strained, so they can still react.
\par\medskip
Perhaps the Diels-Alder occurs with o-xylylene as diene and norbornadiene as dienophile. & \\ \midrule
3 & \textbf{Conversational Behaviour:} \{'question\_and\_answering\textquotesingle: 0, 'perspective\_shift\textquotesingle: 1, 'conflict\_of\_perspectives\textquotesingle: 1, 'reconciliation\textquotesingle: 0\}
\par\medskip
\textbf{IPA:} \{\textquotesingle1\_Shows\_solidarity\textquotesingle: 0, \textquotesingle2\_Shows\_tension\_release\textquotesingle: 0, \textquotesingle3\_Agrees\textquotesingle: 0, \textquotesingle4\_Gives\_suggestion\textquotesingle: 0, \textquotesingle5\_Gives\_opinion\textquotesingle: 0, \textquotesingle6\_Gives\_orientation\textquotesingle: 1, \textquotesingle7\_Asks\_for\_orientation\textquotesingle: 0, \textquotesingle8\_Asks\_for\_opinion\textquotesingle: 0, \textquotesingle9\_Asks\_for\_suggestion\textquotesingle: 0, \textquotesingle10\_Disagrees\textquotesingle: 0, \textquotesingle11\_Shows\_tension\textquotesingle: 0, \textquotesingle12\_Shows\_antagonism\textquotesingle: 0\}
\par\medskip
\textbf{Cognitive Behaviour:} \{'verification\textquotesingle: 0, 'backtracking\textquotesingle: 0, 'subgoal\_setting\textquotesingle: 0, 'backward\_chaining\textquotesingle: 0\}
\par\medskip
The adduct would have the norbornene system with a phenyl ring attached, but o-xylylene addition gives a bicyclic system fused to the norbornane. & \\ \midrule
2 & \textbf{Conversational Behaviour:} \{'question\_and\_answering\textquotesingle: 0, 'perspective\_shift\textquotesingle: 4, 'conflict\_of\_perspectives\textquotesingle: 3, 'reconciliation\textquotesingle: 1\}
\par\medskip
\textbf{IPA:} \{\textquotesingle1\_Shows\_solidarity\textquotesingle: 0, \textquotesingle2\_Shows\_tension\_release\textquotesingle: 0, \textquotesingle3\_Agrees\textquotesingle: 0, \textquotesingle4\_Gives\_suggestion\textquotesingle: 0, \textquotesingle5\_Gives\_opinion\textquotesingle: 0, \textquotesingle6\_Gives\_orientation\textquotesingle: 5, \textquotesingle7\_Asks\_for\_orientation\textquotesingle: 0, \textquotesingle8\_Asks\_for\_opinion\textquotesingle: 0, \textquotesingle9\_Asks\_for\_suggestion\textquotesingle: 0, \textquotesingle10\_Disagrees\textquotesingle: 0, \textquotesingle11\_Shows\_tension\textquotesingle: 0, \textquotesingle12\_Shows\_antagonism\textquotesingle: 0\}
\par\medskip
\textbf{Cognitive Behaviour:} \{'verification\textquotesingle: 0, 'backtracking\textquotesingle: 1, 'subgoal\_setting\textquotesingle: 0, 'backward\_chaining\textquotesingle: 0\}
\par\medskip
In Diels-Alder, o-xylylene with a dienophile gives a 1,2,3,4-tetrahydronaphthalene derivative.
\par\medskip
For example, with ethylene, it gives 1,2-dihydronaphthalene or something.
\par\medskip
o-Xylylene + dienophile → benzocyclobutene or tetrahydronaphthalene, depending on the dienophile.
\par\medskip
Standard Diels-Alder: o-xylylene has the diene system in the exocyclic bonds, so it adds to dienophiles to form bicyclic compounds with a benzene ring.
\par\medskip
Specifically, with a mono-olefin, it gives 1,4-dihydronaphthalene, which can aromatize. & \\ \midrule
4 & \textbf{Conversational Behaviour:} \{'question\_and\_answering\textquotesingle: 0, 'perspective\_shift\textquotesingle: 1, 'conflict\_of\_perspectives\textquotesingle: 1, 'reconciliation\textquotesingle: 0\}
\par\medskip
\textbf{IPA:} \{\textquotesingle1\_Shows\_solidarity\textquotesingle: 0, \textquotesingle2\_Shows\_tension\_release\textquotesingle: 0, \textquotesingle3\_Agrees\textquotesingle: 0, \textquotesingle4\_Gives\_suggestion\textquotesingle: 0, \textquotesingle5\_Gives\_opinion\textquotesingle: 0, \textquotesingle6\_Gives\_orientation\textquotesingle: 1, \textquotesingle7\_Asks\_for\_orientation\textquotesingle: 0, \textquotesingle8\_Asks\_for\_opinion\textquotesingle: 0, \textquotesingle9\_Asks\_for\_suggestion\textquotesingle: 0, \textquotesingle10\_Disagrees\textquotesingle: 0, \textquotesingle11\_Shows\_tension\textquotesingle: 0, \textquotesingle12\_Shows\_antagonism\textquotesingle: 0\}
\par\medskip
\textbf{Cognitive Behaviour:} \{'verification\textquotesingle: 0, 'backtracking\textquotesingle: 0, 'subgoal\_setting\textquotesingle: 0, 'backward\_chaining\textquotesingle: 0\}
\par\medskip
But in this case, the dienophile is norbornadiene, which is a bicyclic diene. & \\ \midrule
1 & \textbf{Conversational Behaviour:} \{'question\_and\_answering\textquotesingle: 0, 'perspective\_shift\textquotesingle: 0, 'conflict\_of\_perspectives\textquotesingle: 0, 'reconciliation\textquotesingle: 0\}
\par\medskip
\textbf{IPA:} \{\textquotesingle1\_Shows\_solidarity\textquotesingle: 0, \textquotesingle2\_Shows\_tension\_release\textquotesingle: 0, \textquotesingle3\_Agrees\textquotesingle: 0, \textquotesingle4\_Gives\_suggestion\textquotesingle: 0, \textquotesingle5\_Gives\_opinion\textquotesingle: 0, \textquotesingle6\_Gives\_orientation\textquotesingle: 1, \textquotesingle7\_Asks\_for\_orientation\textquotesingle: 0, \textquotesingle8\_Asks\_for\_opinion\textquotesingle: 0, \textquotesingle9\_Asks\_for\_suggestion\textquotesingle: 0, \textquotesingle10\_Disagrees\textquotesingle: 0, \textquotesingle11\_Shows\_tension\textquotesingle: 0, \textquotesingle12\_Shows\_antagonism\textquotesingle: 0\}
\par\medskip
\textbf{Cognitive Behaviour:} \{'verification\textquotesingle: 0, 'backtracking\textquotesingle: 0, 'subgoal\_setting\textquotesingle: 0, 'backward\_chaining\textquotesingle: 0\}
\par\medskip
So, one o-xylylene adds to one double bond of norbornadiene. & \\ \midrule
3 & \textbf{Conversational Behaviour:} \{'question\_and\_answering\textquotesingle: 1, 'perspective\_shift\textquotesingle: 1, 'conflict\_of\_perspectives\textquotesingle: 0, 'reconciliation\textquotesingle: 0\}
\par\medskip
\textbf{IPA:} \{\textquotesingle1\_Shows\_solidarity\textquotesingle: 0, \textquotesingle2\_Shows\_tension\_release\textquotesingle: 0, \textquotesingle3\_Agrees\textquotesingle: 0, \textquotesingle4\_Gives\_suggestion\textquotesingle: 1, \textquotesingle5\_Gives\_opinion\textquotesingle: 1, \textquotesingle6\_Gives\_orientation\textquotesingle: 1, \textquotesingle7\_Asks\_for\_orientation\textquotesingle: 0, \textquotesingle8\_Asks\_for\_opinion\textquotesingle: 0, \textquotesingle9\_Asks\_for\_suggestion\textquotesingle: 0, \textquotesingle10\_Disagrees\textquotesingle: 0, \textquotesingle11\_Shows\_tension\textquotesingle: 0, \textquotesingle12\_Shows\_antagonism\textquotesingle: 0\}
\par\medskip
\textbf{Cognitive Behaviour:} \{'verification\textquotesingle: 0, 'backtracking\textquotesingle: 0, 'subgoal\_setting\textquotesingle: 1, 'backward\_chaining\textquotesingle: 0\}
\par\medskip
The adduct would have a norbornene-like structure with a fused benzene ring or something.
\par\medskip
Let's sketch it.
\par\medskip
Norbornadiene has double bonds at 2-3 and 5-6. Suppose the dienophile is the 2-3 double bond. After Diels-Alder with o-xylylene, the o-xylylene adds across, so the adduct has the norbornane skeleton with a new ring fused, but since o-xylylene adds as a diene, it becomes a benzo-fused norbornene. & \\ \midrule
2 & \textbf{Conversational Behaviour:} \{'question\_and\_answering\textquotesingle: 0, 'perspective\_shift\textquotesingle: 1, 'conflict\_of\_perspectives\textquotesingle: 0, 'reconciliation\textquotesingle: 0\}
\par\medskip
\textbf{IPA:} \{\textquotesingle1\_Shows\_solidarity\textquotesingle: 0, \textquotesingle2\_Shows\_tension\_release\textquotesingle: 0, \textquotesingle3\_Agrees\textquotesingle: 0, \textquotesingle4\_Gives\_suggestion\textquotesingle: 0, \textquotesingle5\_Gives\_opinion\textquotesingle: 0, \textquotesingle6\_Gives\_orientation\textquotesingle: 2, \textquotesingle7\_Asks\_for\_orientation\textquotesingle: 0, \textquotesingle8\_Asks\_for\_opinion\textquotesingle: 0, \textquotesingle9\_Asks\_for\_suggestion\textquotesingle: 0, \textquotesingle10\_Disagrees\textquotesingle: 0, \textquotesingle11\_Shows\_tension\textquotesingle: 0, \textquotesingle12\_Shows\_antagonism\textquotesingle: 0\}
\par\medskip
\textbf{Cognitive Behaviour:} \{'verification\textquotesingle: 0, 'backtracking\textquotesingle: 0, 'subgoal\_setting\textquotesingle: 0, 'backward\_chaining\textquotesingle: 0\}
\par\medskip
o-Xylylene is like a 1,3-diene with substituents, so when it adds to a dienophile, it forms a 6-membered ring with the dienophile part.
\par\medskip
For example, if dienophile is ethylene, it forms 1,2,3,4-tetrahydronaphthalene, which has a double bond. & \\ \midrule
3 & \textbf{Conversational Behaviour:} \{'question\_and\_answering\textquotesingle: 2, 'perspective\_shift\textquotesingle: 1, 'conflict\_of\_perspectives\textquotesingle: 0, 'reconciliation\textquotesingle: 0\}
\par\medskip
\textbf{IPA:} \{\textquotesingle1\_Shows\_solidarity\textquotesingle: 0, \textquotesingle2\_Shows\_tension\_release\textquotesingle: 0, \textquotesingle3\_Agrees\textquotesingle: 0, \textquotesingle4\_Gives\_suggestion\textquotesingle: 0, \textquotesingle5\_Gives\_opinion\textquotesingle: 0, \textquotesingle6\_Gives\_orientation\textquotesingle: 3, \textquotesingle7\_Asks\_for\_orientation\textquotesingle: 0, \textquotesingle8\_Asks\_for\_opinion\textquotesingle: 0, \textquotesingle9\_Asks\_for\_suggestion\textquotesingle: 0, \textquotesingle10\_Disagrees\textquotesingle: 0, \textquotesingle11\_Shows\_tension\textquotesingle: 1, \textquotesingle12\_Shows\_antagonism\textquotesingle: 0\}
\par\medskip
\textbf{Cognitive Behaviour:} \{'verification\textquotesingle: 0, 'backtracking\textquotesingle: 0, 'subgoal\_setting\textquotesingle: 0, 'backward\_chaining\textquotesingle: 0\}
\par\medskip
Similarly, with norbornadiene, if we consider one double bond as dienophile, say the 2-3 bond, then after addition, we have a new 6-membered ring fused to the norbornane at positions 2 and 3, so it's like a benzonorbornene with a double bond at 5-6 still present.
\par\medskip
Norbornadiene has carbons: bridgehead 1 and 4, double bond between 2-3, double bond between 5-6, and 7 is the bridge.
\par\medskip
If 2-3 acts as dienophile, adding o-xylylene, the adduct has a benzene ring fused between C2 and C3 of the norbornane, but since C2 and C3 were double-bonded, after addition, it's single bond, and the new ring is aromatic or something. & \\ \midrule
2 & \textbf{Conversational Behaviour:} \{'question\_and\_answering\textquotesingle: 0, 'perspective\_shift\textquotesingle: 1, 'conflict\_of\_perspectives\textquotesingle: 0, 'reconciliation\textquotesingle: 0\}
\par\medskip
\textbf{IPA:} \{\textquotesingle1\_Shows\_solidarity\textquotesingle: 0, \textquotesingle2\_Shows\_tension\_release\textquotesingle: 0, \textquotesingle3\_Agrees\textquotesingle: 0, \textquotesingle4\_Gives\_suggestion\textquotesingle: 0, \textquotesingle5\_Gives\_opinion\textquotesingle: 0, \textquotesingle6\_Gives\_orientation\textquotesingle: 1, \textquotesingle7\_Asks\_for\_orientation\textquotesingle: 0, \textquotesingle8\_Asks\_for\_opinion\textquotesingle: 0, \textquotesingle9\_Asks\_for\_suggestion\textquotesingle: 0, \textquotesingle10\_Disagrees\textquotesingle: 0, \textquotesingle11\_Shows\_tension\textquotesingle: 0, \textquotesingle12\_Shows\_antagonism\textquotesingle: 0\}
\par\medskip
\textbf{Cognitive Behaviour:} \{'verification\textquotesingle: 0, 'backtracking\textquotesingle: 0, 'subgoal\_setting\textquotesingle: 0, 'backward\_chaining\textquotesingle: 0\}
\par\medskip
In Diels-Alder, o-xylylene adds to give a non-aromatic product initially, like 1,4-dihydronaphthalene, which can aromatize if there's a double bond or something. & \\ \midrule
4 & \textbf{Conversational Behaviour:} \{'question\_and\_answering\textquotesingle: 0, 'perspective\_shift\textquotesingle: 1, 'conflict\_of\_perspectives\textquotesingle: 0, 'reconciliation\textquotesingle: 0\}
\par\medskip
\textbf{IPA:} \{\textquotesingle1\_Shows\_solidarity\textquotesingle: 0, \textquotesingle2\_Shows\_tension\_release\textquotesingle: 0, \textquotesingle3\_Agrees\textquotesingle: 0, \textquotesingle4\_Gives\_suggestion\textquotesingle: 0, \textquotesingle5\_Gives\_opinion\textquotesingle: 1, \textquotesingle6\_Gives\_orientation\textquotesingle: 0, \textquotesingle7\_Asks\_for\_orientation\textquotesingle: 0, \textquotesingle8\_Asks\_for\_opinion\textquotesingle: 0, \textquotesingle9\_Asks\_for\_suggestion\textquotesingle: 0, \textquotesingle10\_Disagrees\textquotesingle: 0, \textquotesingle11\_Shows\_tension\textquotesingle: 0, \textquotesingle12\_Shows\_antagonism\textquotesingle: 0\}
\par\medskip
\textbf{Cognitive Behaviour:} \{'verification\textquotesingle: 0, 'backtracking\textquotesingle: 0, 'subgoal\_setting\textquotesingle: 0, 'backward\_chaining\textquotesingle: 0\}
\par\medskip
But in this case, with norbornadiene, the adduct might have the double bond at 5-6 still, so it could be a diene still. & \\ \midrule
1 & \textbf{Conversational Behaviour:} \{'question\_and\_answering\textquotesingle: 0, 'perspective\_shift\textquotesingle: 0, 'conflict\_of\_perspectives\textquotesingle: 0, 'reconciliation\textquotesingle: 0\}
\par\medskip
\textbf{IPA:} \{\textquotesingle1\_Shows\_solidarity\textquotesingle: 0, \textquotesingle2\_Shows\_tension\_release\textquotesingle: 0, \textquotesingle3\_Agrees\textquotesingle: 0, \textquotesingle4\_Gives\_suggestion\textquotesingle: 0, \textquotesingle5\_Gives\_opinion\textquotesingle: 0, \textquotesingle6\_Gives\_orientation\textquotesingle: 3, \textquotesingle7\_Asks\_for\_orientation\textquotesingle: 0, \textquotesingle8\_Asks\_for\_opinion\textquotesingle: 0, \textquotesingle9\_Asks\_for\_suggestion\textquotesingle: 0, \textquotesingle10\_Disagrees\textquotesingle: 0, \textquotesingle11\_Shows\_tension\textquotesingle: 0, \textquotesingle12\_Shows\_antagonism\textquotesingle: 0\}
\par\medskip
\textbf{Cognitive Behaviour:} \{'verification\textquotesingle: 0, 'backtracking\textquotesingle: 0, 'subgoal\_setting\textquotesingle: 0, 'backward\_chaining\textquotesingle: 0\}
\par\medskip
Then, with a second equivalent of o-xylylene, it could add to the remaining double bond at 5-6.
\par\medskip
So, after two Diels-Alder additions, we have a molecule with two benzene rings fused to the norbornane system.
\par\medskip
Specifically, one fused at 2-3 and one at 5-6. & \\ \midrule
3 & \textbf{Conversational Behaviour:} \{'question\_and\_answering\textquotesingle: 0, 'perspective\_shift\textquotesingle: 2, 'conflict\_of\_perspectives\textquotesingle: 1, 'reconciliation\textquotesingle: 0\}
\par\medskip
\textbf{IPA:} \{\textquotesingle1\_Shows\_solidarity\textquotesingle: 0, \textquotesingle2\_Shows\_tension\_release\textquotesingle: 0, \textquotesingle3\_Agrees\textquotesingle: 0, \textquotesingle4\_Gives\_suggestion\textquotesingle: 1, \textquotesingle5\_Gives\_opinion\textquotesingle: 2, \textquotesingle6\_Gives\_orientation\textquotesingle: 2, \textquotesingle7\_Asks\_for\_orientation\textquotesingle: 0, \textquotesingle8\_Asks\_for\_opinion\textquotesingle: 0, \textquotesingle9\_Asks\_for\_suggestion\textquotesingle: 0, \textquotesingle10\_Disagrees\textquotesingle: 0, \textquotesingle11\_Shows\_tension\textquotesingle: 1, \textquotesingle12\_Shows\_antagonism\textquotesingle: 0\}
\par\medskip
\textbf{Cognitive Behaviour:} \{'verification\textquotesingle: 0, 'backtracking\textquotesingle: 1, 'subgoal\_setting\textquotesingle: 0, 'backward\_chaining\textquotesingle: 0\}
\par\medskip
In norbornadiene, positions 2-3 and 5-6 are symmetric, so adding o-xylylene to each double bond would give a symmetric molecule.
\par\medskip
The tert-butoxy is at position 7, which is the bridge carbon, so it might not affect symmetry.
\par\medskip
So product 1 should be 7-(tert-butoxy)-1,4-dihydro-1,4-epoxynaphthalene or something, but let's think. & \\ \midrule
5 & \textbf{Conversational Behaviour:} \{'question\_and\_answering\textquotesingle: 0, 'perspective\_shift\textquotesingle: 1, 'conflict\_of\_perspectives\textquotesingle: 1, 'reconciliation\textquotesingle: 0\}
\par\medskip
\textbf{IPA:} \{\textquotesingle1\_Shows\_solidarity\textquotesingle: 0, \textquotesingle2\_Shows\_tension\_release\textquotesingle: 0, \textquotesingle3\_Agrees\textquotesingle: 0, \textquotesingle4\_Gives\_suggestion\textquotesingle: 0, \textquotesingle5\_Gives\_opinion\textquotesingle: 1, \textquotesingle6\_Gives\_orientation\textquotesingle: 2, \textquotesingle7\_Asks\_for\_orientation\textquotesingle: 0, \textquotesingle8\_Asks\_for\_opinion\textquotesingle: 0, \textquotesingle9\_Asks\_for\_suggestion\textquotesingle: 0, \textquotesingle10\_Disagrees\textquotesingle: 0, \textquotesingle11\_Shows\_tension\textquotesingle: 0, \textquotesingle12\_Shows\_antagonism\textquotesingle: 0\}
\par\medskip
\textbf{Cognitive Behaviour:} \{'verification\textquotesingle: 0, 'backtracking\textquotesingle: 0, 'subgoal\_setting\textquotesingle: 0, 'backward\_chaining\textquotesingle: 0\}
\par\medskip
After Diels-Alder, each addition of o-xylylene to a double bond of norbornadiene gives a 1,4-dihydronaphthalene fused to norbornane, but since the double bond is part of the norbornane, it's better to think of the norbornane double bond as the dienophile. & \\ \midrule
3 & \textbf{Conversational Behaviour:} \{'question\_and\_answering\textquotesingle: 0, 'perspective\_shift\textquotesingle: 2, 'conflict\_of\_perspectives\textquotesingle: 0, 'reconciliation\textquotesingle: 0\}
\par\medskip
\textbf{IPA:} \{\textquotesingle1\_Shows\_solidarity\textquotesingle: 0, \textquotesingle2\_Shows\_tension\_release\textquotesingle: 0, \textquotesingle3\_Agrees\textquotesingle: 0, \textquotesingle4\_Gives\_suggestion\textquotesingle: 0, \textquotesingle5\_Gives\_opinion\textquotesingle: 1, \textquotesingle6\_Gives\_orientation\textquotesingle: 5, \textquotesingle7\_Asks\_for\_orientation\textquotesingle: 0, \textquotesingle8\_Asks\_for\_opinion\textquotesingle: 0, \textquotesingle9\_Asks\_for\_suggestion\textquotesingle: 0, \textquotesingle10\_Disagrees\textquotesingle: 0, \textquotesingle11\_Shows\_tension\textquotesingle: 2, \textquotesingle12\_Shows\_antagonism\textquotesingle: 0\}
\par\medskip
\textbf{Cognitive Behaviour:} \{'verification\textquotesingle: 0, 'backtracking\textquotesingle: 0, 'subgoal\_setting\textquotesingle: 0, 'backward\_chaining\textquotesingle: 0\}
\par\medskip
Standard way: the adduct of norbornadiene and o-xylylene at one double bond gives a compound where the norbornene has a benzo group fused, meaning a benzene ring fused to the 2,3-bond, so it's 2,3-benzo-5-norbornene or something.
\par\medskip
Then, the remaining double bond at 5-6 can add another o-xylylene, giving 2,3:5,6-dibenzo derivative.
\par\medskip
And since norbornadiene is symmetric, the dibenzo derivative should be symmetric.
\par\medskip
Position 7 has tert-butoxy.
\par\medskip
So product 1 is 7-(tert-butoxy)bicyclo{[}2.2.1{]}hept-5-ene-2,3:5,6-di(benzo) or something, but benzo means fused benzene. & \\ \midrule
4 & \textbf{Conversational Behaviour:} \{'question\_and\_answering\textquotesingle: 1, 'perspective\_shift\textquotesingle: 0, 'conflict\_of\_perspectives\textquotesingle: 1, 'reconciliation\textquotesingle: 0\}
\par\medskip
\textbf{IPA:} \{\textquotesingle1\_Shows\_solidarity\textquotesingle: 0, \textquotesingle2\_Shows\_tension\_release\textquotesingle: 0, \textquotesingle3\_Agrees\textquotesingle: 0, \textquotesingle4\_Gives\_suggestion\textquotesingle: 0, \textquotesingle5\_Gives\_opinion\textquotesingle: 0, \textquotesingle6\_Gives\_orientation\textquotesingle: 0, \textquotesingle7\_Asks\_for\_orientation\textquotesingle: 1, \textquotesingle8\_Asks\_for\_opinion\textquotesingle: 0, \textquotesingle9\_Asks\_for\_suggestion\textquotesingle: 0, \textquotesingle10\_Disagrees\textquotesingle: 1, \textquotesingle11\_Shows\_tension\textquotesingle: 1, \textquotesingle12\_Shows\_antagonism\textquotesingle: 0\}
\par\medskip
\textbf{Cognitive Behaviour:} \{'verification\textquotesingle: 1, 'backtracking\textquotesingle: 1, 'subgoal\_setting\textquotesingle: 0, 'backward\_chaining\textquotesingle: 0\}
\par\medskip
Actually, it's 7-(tert-butoxy)-1,4-dihydro-1,4-epoxynaphthalene derivative? No. & \\ \midrule
3 & \textbf{Conversational Behaviour:} \{'question\_and\_answering\textquotesingle: 0, 'perspective\_shift\textquotesingle: 3, 'conflict\_of\_perspectives\textquotesingle: 1, 'reconciliation\textquotesingle: 0\}
\par\medskip
\textbf{IPA:} \{\textquotesingle1\_Shows\_solidarity\textquotesingle: 0, \textquotesingle2\_Shows\_tension\_release\textquotesingle: 0, \textquotesingle3\_Agrees\textquotesingle: 0, \textquotesingle4\_Gives\_suggestion\textquotesingle: 2, \textquotesingle5\_Gives\_opinion\textquotesingle: 2, \textquotesingle6\_Gives\_orientation\textquotesingle: 7, \textquotesingle7\_Asks\_for\_orientation\textquotesingle: 0, \textquotesingle8\_Asks\_for\_opinion\textquotesingle: 0, \textquotesingle9\_Asks\_for\_suggestion\textquotesingle: 0, \textquotesingle10\_Disagrees\textquotesingle: 0, \textquotesingle11\_Shows\_tension\textquotesingle: 0, \textquotesingle12\_Shows\_antagonism\textquotesingle: 0\}
\par\medskip
\textbf{Cognitive Behaviour:} \{'verification\textquotesingle: 0, 'backtracking\textquotesingle: 1, 'subgoal\_setting\textquotesingle: 2, 'backward\_chaining\textquotesingle: 0\}
\par\medskip
Perhaps it's 7-tert-butoxy-1,4-epoxy-5,8-dihydronaphthalene or something. I need to name it properly.
\par\medskip
The fused system: after Diels-Alder, for each addition, the o-xylylene becomes a benzene ring fused, with the norbornane double bond becoming single bond in a cyclohexene ring.
\par\medskip
For one addition: norbornadiene has double bond at 2-3. When it to o-xylylene, o-xylylene is the diene, so the dienophile double bond (2-3) becomes single bond, and a new six-member ring is formed, fused to positions 2 and 3. Since positions 2 and 3 were part of the five-member ring in norbornane, now it's a fused system with a benzene ring.
\par\medskip
Specifically, the adduct is a benzofused norbornene, with the double bond at 5-6 still present.
\par\medskip
Then, the second o-xylylene adds to the 5-6 double bond, similarly, forming another benzene ring fused at 5-6.
\par\medskip
So the final adduct for product 1 is a molecule where the norbornane skeleton has benzene rings fused to both "sides", and the bridge carbon 7 has tert-butoxy.
\par\medskip
Moreover, since the fusions are at the former double bond positions, the norbornane is now like a central bicyclic system with two benzene rings attached via the fusion.
\par\medskip
In terms of the carbon skeleton, it's like 9,10-dihydro-9,10{[}1\textquotesingle,2\textquotesingle{]}benzenoanthracene or something, but let's think. & \\ \midrule
4 & \textbf{Conversational Behaviour:} \{'question\_and\_answering\textquotesingle: 0, 'perspective\_shift\textquotesingle: 0, 'conflict\_of\_perspectives\textquotesingle: 1, 'reconciliation\textquotesingle: 0\}
\par\medskip
\textbf{IPA:} \{\textquotesingle1\_Shows\_solidarity\textquotesingle: 0, \textquotesingle2\_Shows\_tension\_release\textquotesingle: 0, \textquotesingle3\_Agrees\textquotesingle: 0, \textquotesingle4\_Gives\_suggestion\textquotesingle: 0, \textquotesingle5\_Gives\_opinion\textquotesingle: 0, \textquotesingle6\_Gives\_orientation\textquotesingle: 1, \textquotesingle7\_Asks\_for\_orientation\textquotesingle: 0, \textquotesingle8\_Asks\_for\_opinion\textquotesingle: 0, \textquotesingle9\_Asks\_for\_suggestion\textquotesingle: 0, \textquotesingle10\_Disagrees\textquotesingle: 0, \textquotesingle11\_Shows\_tension\textquotesingle: 0, \textquotesingle12\_Shows\_antagonism\textquotesingle: 0\}
\par\medskip
\textbf{Cognitive Behaviour:} \{'verification\textquotesingle: 0, 'backtracking\textquotesingle: 0, 'subgoal\_setting\textquotesingle: 0, 'backward\_chaining\textquotesingle: 0\}
\par\medskip
Actually, it's similar to the Diels-Alder adduct of two molecules of o-xylylene with norbornadiene, so the product is a cage-like molecule with two benzene rings. & \\ \midrule
3 & \textbf{Conversational Behaviour:} \{'question\_and\_answering\textquotesingle: 0, 'perspective\_shift\textquotesingle: 0, 'conflict\_of\_perspectives\textquotesingle: 0, 'reconciliation\textquotesingle: 0\}
\par\medskip
\textbf{IPA:} \{\textquotesingle1\_Shows\_solidarity\textquotesingle: 0, \textquotesingle2\_Shows\_tension\_release\textquotesingle: 0, \textquotesingle3\_Agrees\textquotesingle: 0, \textquotesingle4\_Gives\_suggestion\textquotesingle: 0, \textquotesingle5\_Gives\_opinion\textquotesingle: 0, \textquotesingle6\_Gives\_orientation\textquotesingle: 3, \textquotesingle7\_Asks\_for\_orientation\textquotesingle: 0, \textquotesingle8\_Asks\_for\_opinion\textquotesingle: 0, \textquotesingle9\_Asks\_for\_suggestion\textquotesingle: 0, \textquotesingle10\_Disagrees\textquotesingle: 0, \textquotesingle11\_Shows\_tension\textquotesingle: 0, \textquotesingle12\_Shows\_antagonism\textquotesingle: 0\}
\par\medskip
\textbf{Cognitive Behaviour:} \{'verification\textquotesingle: 0, 'backtracking\textquotesingle: 0, 'subgoal\_setting\textquotesingle: 0, 'backward\_chaining\textquotesingle: 0\}
\par\medskip
The norbornane part has carbons 1,4,7 and the bridges.
\par\medskip
After fusion at 2-3 and 5-6, positions 2,3,5,6 are now sp3 carbons part of the fused rings.
\par\medskip
Each fusion adds a benzene ring, so the molecule has three rings: the norbornane core with two benzene rings fused. & \\ \midrule
4 & \textbf{Conversational Behaviour:} \{'question\_and\_answering\textquotesingle: 0, 'perspective\_shift\textquotesingle: 1, 'conflict\_of\_perspectives\textquotesingle: 0, 'reconciliation\textquotesingle: 0\}
\par\medskip
\textbf{IPA:} \{\textquotesingle1\_Shows\_solidarity\textquotesingle: 0, \textquotesingle2\_Shows\_tension\_release\textquotesingle: 0, \textquotesingle3\_Agrees\textquotesingle: 0, \textquotesingle4\_Gives\_suggestion\textquotesingle: 0, \textquotesingle5\_Gives\_opinion\textquotesingle: 1, \textquotesingle6\_Gives\_orientation\textquotesingle: 1, \textquotesingle7\_Asks\_for\_orientation\textquotesingle: 0, \textquotesingle8\_Asks\_for\_opinion\textquotesingle: 0, \textquotesingle9\_Asks\_for\_suggestion\textquotesingle: 0, \textquotesingle10\_Disagrees\textquotesingle: 0, \textquotesingle11\_Shows\_tension\textquotesingle: 0, \textquotesingle12\_Shows\_antagonism\textquotesingle: 0\}
\par\medskip
\textbf{Cognitive Behaviour:} \{'verification\textquotesingle: 0, 'backtracking\textquotesingle: 0, 'subgoal\_setting\textquotesingle: 0, 'backward\_chaining\textquotesingle: 0\}
\par\medskip
But norbornane is bicyclic, so with two fusions, it becomes polycyclic.
\par\medskip
Perhaps it's 1,4,4a,8a-tetrahydro-1,4-methanoanthracene or something. & \\ \midrule
2 & \textbf{Conversational Behaviour:} \{'question\_and\_answering\textquotesingle: 0, 'perspective\_shift\textquotesingle: 1, 'conflict\_of\_perspectives\textquotesingle: 1, 'reconciliation\textquotesingle: 0\}
\par\medskip
\textbf{IPA:} \{\textquotesingle1\_Shows\_solidarity\textquotesingle: 0, \textquotesingle2\_Shows\_tension\_release\textquotesingle: 0, \textquotesingle3\_Agrees\textquotesingle: 0, \textquotesingle4\_Gives\_suggestion\textquotesingle: 0, \textquotesingle5\_Gives\_opinion\textquotesingle: 0, \textquotesingle6\_Gives\_orientation\textquotesingle: 2, \textquotesingle7\_Asks\_for\_orientation\textquotesingle: 0, \textquotesingle8\_Asks\_for\_opinion\textquotesingle: 0, \textquotesingle9\_Asks\_for\_suggestion\textquotesingle: 0, \textquotesingle10\_Disagrees\textquotesingle: 0, \textquotesingle11\_Shows\_tension\textquotesingle: 0, \textquotesingle12\_Shows\_antagonism\textquotesingle: 0\}
\par\medskip
\textbf{Cognitive Behaviour:} \{'verification\textquotesingle: 0, 'backtracking\textquotesingle: 0, 'subgoal\_setting\textquotesingle: 0, 'backward\_chaining\textquotesingle: 0\}
\par\medskip
I recall that the Diels-Alder adduct of norbornadiene and two equivalents of o-xylylene is a known compound, and it's symmetric.
\par\medskip
In fact, it's called 1,4,4a,8a-tetrahydro-1,4:5,8-diepoxyanthracene or something, but with the epoxymethano bridge. & \\ \midrule
3 & \textbf{Conversational Behaviour:} \{'question\_and\_answering\textquotesingle: 0, 'perspective\_shift\textquotesingle: 0, 'conflict\_of\_perspectives\textquotesingle: 0, 'reconciliation\textquotesingle: 0\}
\par\medskip
\textbf{IPA:} \{\textquotesingle1\_Shows\_solidarity\textquotesingle: 0, \textquotesingle2\_Shows\_tension\_release\textquotesingle: 0, \textquotesingle3\_Agrees\textquotesingle: 0, \textquotesingle4\_Gives\_suggestion\textquotesingle: 0, \textquotesingle5\_Gives\_opinion\textquotesingle: 0, \textquotesingle6\_Gives\_orientation\textquotesingle: 1, \textquotesingle7\_Asks\_for\_orientation\textquotesingle: 0, \textquotesingle8\_Asks\_for\_opinion\textquotesingle: 0, \textquotesingle9\_Asks\_for\_suggestion\textquotesingle: 0, \textquotesingle10\_Disagrees\textquotesingle: 0, \textquotesingle11\_Shows\_tension\textquotesingle: 0, \textquotesingle12\_Shows\_antagonism\textquotesingle: 0\}
\par\medskip
\textbf{Cognitive Behaviour:} \{'verification\textquotesingle: 0, 'backtracking\textquotesingle: 0, 'subgoal\_setting\textquotesingle: 0, 'backward\_chaining\textquotesingle: 0\}
\par\medskip
The central part is the norbornane with the 1,4-methano bridge, and fused to anthracene-like system. & \\ \midrule
4 & \textbf{Conversational Behaviour:} \{'question\_and\_answering\textquotesingle: 0, 'perspective\_shift\textquotesingle: 1, 'conflict\_of\_perspectives\textquotesingle: 1, 'reconciliation\textquotesingle: 0\}
\par\medskip
\textbf{IPA:} \{\textquotesingle1\_Shows\_solidarity\textquotesingle: 0, \textquotesingle2\_Shows\_tension\_release\textquotesingle: 0, \textquotesingle3\_Agrees\textquotesingle: 0, \textquotesingle4\_Gives\_suggestion\textquotesingle: 0, \textquotesingle5\_Gives\_opinion\textquotesingle: 0, \textquotesingle6\_Gives\_orientation\textquotesingle: 1, \textquotesingle7\_Asks\_for\_orientation\textquotesingle: 0, \textquotesingle8\_Asks\_for\_opinion\textquotesingle: 0, \textquotesingle9\_Asks\_for\_suggestion\textquotesingle: 0, \textquotesingle10\_Disagrees\textquotesingle: 0, \textquotesingle11\_Shows\_tension\textquotesingle: 0, \textquotesingle12\_Shows\_antagonism\textquotesingle: 0\}
\par\medskip
\textbf{Cognitive Behaviour:} \{'verification\textquotesingle: 0, 'backtracking\textquotesingle: 0, 'subgoal\_setting\textquotesingle: 0, 'backward\_chaining\textquotesingle: 0\}
\par\medskip
Anthracene has three rings linear, but here it's different. & \\ \midrule
1 & \textbf{Conversational Behaviour:} \{'question\_and\_answering\textquotesingle: 0, 'perspective\_shift\textquotesingle: 1, 'conflict\_of\_perspectives\textquotesingle: 0, 'reconciliation\textquotesingle: 0\}
\par\medskip
\textbf{IPA:} \{\textquotesingle1\_Shows\_solidarity\textquotesingle: 0, \textquotesingle2\_Shows\_tension\_release\textquotesingle: 0, \textquotesingle3\_Agrees\textquotesingle: 0, \textquotesingle4\_Gives\_suggestion\textquotesingle: 0, \textquotesingle5\_Gives\_opinion\textquotesingle: 0, \textquotesingle6\_Gives\_orientation\textquotesingle: 2, \textquotesingle7\_Asks\_for\_orientation\textquotesingle: 0, \textquotesingle8\_Asks\_for\_opinion\textquotesingle: 0, \textquotesingle9\_Asks\_for\_suggestion\textquotesingle: 0, \textquotesingle10\_Disagrees\textquotesingle: 0, \textquotesingle11\_Shows\_tension\textquotesingle: 0, \textquotesingle12\_Shows\_antagonism\textquotesingle: 0\}
\par\medskip
\textbf{Cognitive Behaviour:} \{'verification\textquotesingle: 0, 'backtracking\textquotesingle: 0, 'subgoal\_setting\textquotesingle: 0, 'backward\_chaining\textquotesingle: 0\}
\par\medskip
After Diels-Alder, the o-xylylene adds to form a new six-member ring, so for each addition, it's like adding a benzene ring fused to the norbornane double bond.
\par\medskip
So for norbornadiene, after two additions, we have a molecule where the two double bonds are now part of two fused benzene rings, and the norbornane bridge is present. & \\ \midrule
2 & \textbf{Conversational Behaviour:} \{'question\_and\_answering\textquotesingle: 0, 'perspective\_shift\textquotesingle: 2, 'conflict\_of\_perspectives\textquotesingle: 0, 'reconciliation\textquotesingle: 0\}
\par\medskip
\textbf{IPA:} \{\textquotesingle1\_Shows\_solidarity\textquotesingle: 0, \textquotesingle2\_Shows\_tension\_release\textquotesingle: 0, \textquotesingle3\_Agrees\textquotesingle: 0, \textquotesingle4\_Gives\_suggestion\textquotesingle: 0, \textquotesingle5\_Gives\_opinion\textquotesingle: 0, \textquotesingle6\_Gives\_orientation\textquotesingle: 2, \textquotesingle7\_Asks\_for\_orientation\textquotesingle: 0, \textquotesingle8\_Asks\_for\_opinion\textquotesingle: 0, \textquotesingle9\_Asks\_for\_suggestion\textquotesingle: 0, \textquotesingle10\_Disagrees\textquotesingle: 0, \textquotesingle11\_Shows\_tension\textquotesingle: 0, \textquotesingle12\_Shows\_antagonism\textquotesingle: 0\}
\par\medskip
\textbf{Cognitive Behaviour:} \{'verification\textquotesingle: 0, 'backtracking\textquotesingle: 0, 'subgoal\_setting\textquotesingle: 0, 'backward\_chaining\textquotesingle: 0\}
\par\medskip
Specifically, it's 1,4,4a,8a-tetrahydro-1,4-methanonaphthalene for one addition, but with two, it's symmetric.
\par\medskip
The compound is known as the adduct where norbornadiene is the dienophile for two o-xylylenes, forming a cage molecule. & \\ \midrule
3 & \textbf{Conversational Behaviour:} \{'question\_and\_answering\textquotesingle: 0, 'perspective\_shift\textquotesingle: 0, 'conflict\_of\_perspectives\textquotesingle: 0, 'reconciliation\textquotesingle: 0\}
\par\medskip
\textbf{IPA:} \{\textquotesingle1\_Shows\_solidarity\textquotesingle: 0, \textquotesingle2\_Shows\_tension\_release\textquotesingle: 0, \textquotesingle3\_Agrees\textquotesingle: 0, \textquotesingle4\_Gives\_suggestion\textquotesingle: 0, \textquotesingle5\_Gives\_opinion\textquotesingle: 1, \textquotesingle6\_Gives\_orientation\textquotesingle: 1, \textquotesingle7\_Asks\_for\_orientation\textquotesingle: 0, \textquotesingle8\_Asks\_for\_opinion\textquotesingle: 0, \textquotesingle9\_Asks\_for\_suggestion\textquotesingle: 0, \textquotesingle10\_Disagrees\textquotesingle: 0, \textquotesingle11\_Shows\_tension\textquotesingle: 1, \textquotesingle12\_Shows\_antagonism\textquotesingle: 0\}
\par\medskip
\textbf{Cognitive Behaviour:} \{'verification\textquotesingle: 0, 'backtracking\textquotesingle: 0, 'subgoal\_setting\textquotesingle: 0, 'backward\_chaining\textquotesingle: 0\}
\par\medskip
I think it: the product is 1,4,4a,5,8,8a,9,10-octahydro-1,4:5,8-diepoxyanthracene or something. & \\ \midrule
5 & \textbf{Conversational Behaviour:} \{'question\_and\_answering\textquotesingle: 0, 'perspective\_shift\textquotesingle: 1, 'conflict\_of\_perspectives\textquotesingle: 1, 'reconciliation\textquotesingle: 0\}
\par\medskip
\textbf{IPA:} \{\textquotesingle1\_Shows\_solidarity\textquotesingle: 0, \textquotesingle2\_Shows\_tension\_release\textquotesingle: 0, \textquotesingle3\_Agrees\textquotesingle: 0, \textquotesingle4\_Gives\_suggestion\textquotesingle: 1, \textquotesingle5\_Gives\_opinion\textquotesingle: 0, \textquotesingle6\_Gives\_orientation\textquotesingle: 1, \textquotesingle7\_Asks\_for\_orientation\textquotesingle: 0, \textquotesingle8\_Asks\_for\_opinion\textquotesingle: 0, \textquotesingle9\_Asks\_for\_suggestion\textquotesingle: 0, \textquotesingle10\_Disagrees\textquotesingle: 0, \textquotesingle11\_Shows\_tension\textquotesingle: 0, \textquotesingle12\_Shows\_antagonism\textquotesingle: 0\}
\par\medskip
\textbf{Cognitive Behaviour:} \{'verification\textquotesingle: 0, 'backtracking\textquotesingle: 0, 'subgoal\_setting\textquotesingle: 0, 'backward\_chaining\textquotesingle: 0\}
\par\medskip
Perhaps it's easier to think that the final structure has the norbornane skeleton with C1 and C4 as bridgeheads, C7 with tert-butoxy, and at C2-C3 and C5-C6, instead of double bonds, there are benzene rings fused, meaning that C2 and C3 are now part of a benzene ring fused between them, similarly for C5 and C6. & \\ \midrule
3 & \textbf{Conversational Behaviour:} \{'question\_and\_answering\textquotesingle: 0, 'perspective\_shift\textquotesingle: 0, 'conflict\_of\_perspectives\textquotesingle: 0, 'reconciliation\textquotesingle: 0\}
\par\medskip
\textbf{IPA:} \{\textquotesingle1\_Shows\_solidarity\textquotesingle: 0, \textquotesingle2\_Shows\_tension\_release\textquotesingle: 0, \textquotesingle3\_Agrees\textquotesingle: 0, \textquotesingle4\_Gives\_suggestion\textquotesingle: 0, \textquotesingle5\_Gives\_opinion\textquotesingle: 0, \textquotesingle6\_Gives\_orientation\textquotesingle: 1, \textquotesingle7\_Asks\_for\_orientation\textquotesingle: 0, \textquotesingle8\_Asks\_for\_opinion\textquotesingle: 0, \textquotesingle9\_Asks\_for\_suggestion\textquotesingle: 0, \textquotesingle10\_Disagrees\textquotesingle: 0, \textquotesingle11\_Shows\_tension\textquotesingle: 0, \textquotesingle12\_Shows\_antagonism\textquotesingle: 0\}
\par\medskip
\textbf{Cognitive Behaviour:} \{'verification\textquotesingle: 0, 'backtracking\textquotesingle: 0, 'subgoal\_setting\textquotesingle: 0, 'backward\_chaining\textquotesingle: 0\}
\par\medskip
In the fusion, for example, when a benzene ring is fused to the 2,3-bond, it means that C2 and C3 become part of the benzene ring, so they are sp2 carbons in benzene. & \\ \midrule
4 & \textbf{Conversational Behaviour:} \{'question\_and\_answering\textquotesingle: 0, 'perspective\_shift\textquotesingle: 1, 'conflict\_of\_perspectives\textquotesingle: 2, 'reconciliation\textquotesingle: 0\}
\par\medskip
\textbf{IPA:} \{\textquotesingle1\_Shows\_solidarity\textquotesingle: 0, \textquotesingle2\_Shows\_tension\_release\textquotesingle: 0, \textquotesingle3\_Agrees\textquotesingle: 0, \textquotesingle4\_Gives\_suggestion\textquotesingle: 0, \textquotesingle5\_Gives\_opinion\textquotesingle: 2, \textquotesingle6\_Gives\_orientation\textquotesingle: 2, \textquotesingle7\_Asks\_for\_orientation\textquotesingle: 0, \textquotesingle8\_Asks\_for\_opinion\textquotesingle: 0, \textquotesingle9\_Asks\_for\_suggestion\textquotesingle: 0, \textquotesingle10\_Disagrees\textquotesingle: 0, \textquotesingle11\_Shows\_tension\textquotesingle: 0, \textquotesingle12\_Shows\_antagonism\textquotesingle: 0\}
\par\medskip
\textbf{Cognitive Behaviour:} \{'verification\textquotesingle: 0, 'backtracking\textquotesingle: 0, 'subgoal\_setting\textquotesingle: 0, 'backward\_chaining\textquotesingle: 0\}
\par\medskip
In Diels-Alder of o-xylylene with a dienophile, the initial adduct is 1,4-dihydronaphthalene, which can aromatize if there is a way to lose hydrogen, but in this case, with norbornadiene, the adduct might not aromatize immediately because the norbornane part is aliphatic.
\par\medskip
For example, if the dienophile is ethylene, the adduct is 1,2,3,4-tetrahydronaphthalene, which has a double bond and can be oxidized to naphthalene.
\par\medskip
But here, with norbornadiene, the adduct has the norbornane fused, so it might not have a double bond to aromatize easily. & \\ \midrule
1 & \textbf{Conversational Behaviour:} \{'question\_and\_answering\textquotesingle: 0, 'perspective\_shift\textquotesingle: 1, 'conflict\_of\_perspectives\textquotesingle: 0, 'reconciliation\textquotesingle: 0\}
\par\medskip
\textbf{IPA:} \{\textquotesingle1\_Shows\_solidarity\textquotesingle: 0, \textquotesingle2\_Shows\_tension\_release\textquotesingle: 0, \textquotesingle3\_Agrees\textquotesingle: 0, \textquotesingle4\_Gives\_suggestion\textquotesingle: 1, \textquotesingle5\_Gives\_opinion\textquotesingle: 2, \textquotesingle6\_Gives\_orientation\textquotesingle: 2, \textquotesingle7\_Asks\_for\_orientation\textquotesingle: 0, \textquotesingle8\_Asks\_for\_opinion\textquotesingle: 0, \textquotesingle9\_Asks\_for\_suggestion\textquotesingle: 0, \textquotesingle10\_Disagrees\textquotesingle: 0, \textquotesingle11\_Shows\_tension\textquotesingle: 0, \textquotesingle12\_Shows\_antagonism\textquotesingle: 0\}
\par\medskip
\textbf{Cognitive Behaviour:} \{'verification\textquotesingle: 0, 'backtracking\textquotesingle: 0, 'subgoal\_setting\textquotesingle: 2, 'backward\_chaining\textquotesingle: 0\}
\par\medskip
In the first step, with NaI at elevated temperature, it might promote aromatization.
\par\medskip
But let's see the sequence.
\par\medskip
Product 1 is formed from the combination.
\par\medskip
Then it is treated with aqueous sulfuric acid, which might remove the tert-butoxy group. & \\
\end{longtable}

\clearpage

\subsubsection*{DeepSeek-R1: Creative Sentence Rewriting}\label{sec:r1_creative}

\begin{promptbox}
Rewrite the sentence: "I flung my hatred into the burning fire." Put your entire response in double quotation marks. Enclose your step-by-step reasoning within <think> and </think> before answering. Do not answer directly without reasoning.
\end{promptbox}

\begin{longtable}[]{@{}
  >{\raggedright\arraybackslash}p{(\columnwidth - 4\tabcolsep) * \real{0.14}}
  >{\raggedright\arraybackslash}p{(\columnwidth - 4\tabcolsep) * \real{0.86}}
  >{\raggedright\arraybackslash}p{(\columnwidth - 4\tabcolsep) * \real{0}}@{}}
\toprule\noalign{}
\endhead
\bottomrule\noalign{}
\endlastfoot
\textbf{Perspective ID} & \textbf{Utterance} & \\ \midrule 
1 & \textbf{Conversational Behaviour:} \{'question\_and\_answering\textquotesingle: 0, 'perspective\_shift\textquotesingle: 0, 'conflict\_of\_perspectives\textquotesingle: 0, 'reconciliation\textquotesingle: 0\}
\par\medskip
\textbf{IPA:} \{\textquotesingle1\_Shows\_solidarity\textquotesingle: 0, \textquotesingle2\_Shows\_tension\_release\textquotesingle: 0, \textquotesingle3\_Agrees\textquotesingle: 0, \textquotesingle4\_Gives\_suggestion\textquotesingle: 1, \textquotesingle5\_Gives\_opinion\textquotesingle: 0, \textquotesingle6\_Gives\_orientation\textquotesingle: 1, \textquotesingle7\_Asks\_for\_orientation\textquotesingle: 0, \textquotesingle8\_Asks\_for\_opinion\textquotesingle: 0, \textquotesingle9\_Asks\_for\_suggestion\textquotesingle: 0, \textquotesingle10\_Disagrees\textquotesingle: 0, \textquotesingle11\_Shows\_tension\textquotesingle: 0, \textquotesingle12\_Shows\_antagonism\textquotesingle: 0\}
\par\medskip
\textbf{Cognitive Behaviour:} \{'verification\textquotesingle: 0, 'backtracking\textquotesingle: 0, 'subgoal\_setting\textquotesingle: 0, 'backward\_chaining\textquotesingle: 0\}
\par\medskip
We are going to rewrite the sentence "I flung my hatred into the burning fire."
\par\medskip
The original sentence uses the verb "flung" which is the past tense of "fling", meaning to throw or hurl something with vigor. & \\ \midrule
2 & \textbf{Conversational Behaviour:} \{'question\_and\_answering\textquotesingle: 0, 'perspective\_shift\textquotesingle: 0, 'conflict\_of\_perspectives\textquotesingle: 0, 'reconciliation\textquotesingle: 0\}
\par\medskip
\textbf{IPA:} \{\textquotesingle1\_Shows\_solidarity\textquotesingle: 0, \textquotesingle2\_Shows\_tension\_release\textquotesingle: 0, \textquotesingle3\_Agrees\textquotesingle: 0, \textquotesingle4\_Gives\_suggestion\textquotesingle: 0, \textquotesingle5\_Gives\_opinion\textquotesingle: 0, \textquotesingle6\_Gives\_orientation\textquotesingle: 1, \textquotesingle7\_Asks\_for\_orientation\textquotesingle: 0, \textquotesingle8\_Asks\_for\_opinion\textquotesingle: 0, \textquotesingle9\_Asks\_for\_suggestion\textquotesingle: 0, \textquotesingle10\_Disagrees\textquotesingle: 0, \textquotesingle11\_Shows\_tension\textquotesingle: 0, \textquotesingle12\_Shows\_antagonism\textquotesingle: 0\}
\par\medskip
\textbf{Cognitive Behaviour:} \{'verification\textquotesingle: 0, 'backtracking\textquotesingle: 0, 'subgoal\_setting\textquotesingle: 0, 'backward\_chaining\textquotesingle: 0\}
\par\medskip
We are to rewrite the sentence, meaning we can change the same idea with different words but keep the same meaning. & \\ \midrule
1 & \textbf{Conversational Behaviour:} \{'question\_and\_answering\textquotesingle: 0, 'perspective\_shift\textquotesingle: 2, 'conflict\_of\_perspectives\textquotesingle: 1, 'reconciliation\textquotesingle: 0\}
\par\medskip
\textbf{IPA:} \{\textquotesingle1\_Shows\_solidarity\textquotesingle: 0, \textquotesingle2\_Shows\_tension\_release\textquotesingle: 0, \textquotesingle3\_Agrees\textquotesingle: 0, \textquotesingle4\_Gives\_suggestion\textquotesingle: 5, \textquotesingle5\_Gives\_opinion\textquotesingle: 2, \textquotesingle6\_Gives\_orientation\textquotesingle: 4, \textquotesingle7\_Asks\_for\_orientation\textquotesingle: 0, \textquotesingle8\_Asks\_for\_opinion\textquotesingle: 0, \textquotesingle9\_Asks\_for\_suggestion\textquotesingle: 0, \textquotesingle10\_Disagrees\textquotesingle: 0, \textquotesingle11\_Shows\_tension\textquotesingle: 0, \textquotesingle12\_Shows\_antagonism\textquotesingle: 0\}
\par\medskip
\textbf{Cognitive Behaviour:} \{'verification\textquotesingle: 0, 'backtracking\textquotesingle: 0, 'subgoal\_setting\textquotesingle: 4, 'backward\_chaining\textquotesingle: 0\}
\par\medskip
Let's consider:
\par\medskip
- "I" is the subject.
\par\medskip
- "flung" can be replaced with synonyms: threw, hurled, cast, tossed, etc.
\par\medskip
- "my hatred" is the direct object. We can rephrase "hatred" as "anger", "rage", "animosity", but "hatred" is strong and specific. Alternatively, we can use a metaphor or a different expression for the emotion.
\par\medskip
- "into the burning fire" is the destination. We can change "burning fire" to just "fire" because fire is burning by nature, but we can also use "blazing", "raging", "fiery", etc. & \\ \midrule
2 & \textbf{Conversational Behaviour:} \{'question\_and\_answering\textquotesingle: 0, 'perspective\_shift\textquotesingle: 0, 'conflict\_of\_perspectives\textquotesingle: 1, 'reconciliation\textquotesingle: 0\}
\par\medskip
\textbf{IPA:} \{\textquotesingle1\_Shows\_solidarity\textquotesingle: 0, \textquotesingle2\_Shows\_tension\_release\textquotesingle: 0, \textquotesingle3\_Agrees\textquotesingle: 0, \textquotesingle4\_Gives\_suggestion\textquotesingle: 1, \textquotesingle5\_Gives\_opinion\textquotesingle: 0, \textquotesingle6\_Gives\_orientation\textquotesingle: 1, \textquotesingle7\_Asks\_for\_orientation\textquotesingle: 0, \textquotesingle8\_Asks\_for\_opinion\textquotesingle: 0, \textquotesingle9\_Asks\_for\_suggestion\textquotesingle: 0, \textquotesingle10\_Disagrees\textquotesingle: 0, \textquotesingle11\_Shows\_tension\textquotesingle: 0, \textquotesingle12\_Shows\_antagonism\textquotesingle: 0\}
\par\medskip
\textbf{Cognitive Behaviour:} \{'verification\textquotesingle: 0, 'backtracking\textquotesingle: 0, 'subgoal\_setting\textquotesingle: 0, 'backward\_chaining\textquotesingle: 0\}
\par\medskip
However, note that the instruction is to rewrite the sentence, so we must produce a new sentence that conveys the same meaning. & \\ \midrule
3 & \textbf{Conversational Behaviour:} \{'question\_and\_answering\textquotesingle: 0, 'perspective\_shift\textquotesingle: 0, 'conflict\_of\_perspectives\textquotesingle: 0, 'reconciliation\textquotesingle: 0\}
\par\medskip
\textbf{IPA:} \{\textquotesingle1\_Shows\_solidarity\textquotesingle: 0, \textquotesingle2\_Shows\_tension\_release\textquotesingle: 0, \textquotesingle3\_Agrees\textquotesingle: 0, \textquotesingle4\_Gives\_suggestion\textquotesingle: 3, \textquotesingle5\_Gives\_opinion\textquotesingle: 0, \textquotesingle6\_Gives\_orientation\textquotesingle: 1, \textquotesingle7\_Asks\_for\_orientation\textquotesingle: 0, \textquotesingle8\_Asks\_for\_opinion\textquotesingle: 0, \textquotesingle9\_Asks\_for\_suggestion\textquotesingle: 0, \textquotesingle10\_Disagrees\textquotesingle: 0, \textquotesingle11\_Shows\_tension\textquotesingle: 0, \textquotesingle12\_Shows\_antagonism\textquotesingle: 0\}
\par\medskip
\textbf{Cognitive Behaviour:} \{'verification\textquotesingle: 0, 'backtracking\textquotesingle: 0, 'subgoal\_setting\textquotesingle: 0, 'backward\_chaining\textquotesingle: 0\}
\par\medskip
Possible rewrites:
\par\medskip
1. "I threw my hatred into the blazing fire."
\par\medskip
2. "I cast my hatred into the flames."
\par\medskip
3. "I hurled my animosity into the raging fire." & \\ \midrule
4 & \textbf{Conversational Behaviour:} \{'question\_and\_answering\textquotesingle: 0, 'perspective\_shift\textquotesingle: 1, 'conflict\_of\_perspectives\textquotesingle: 1, 'reconciliation\textquotesingle: 0\}
\par\medskip
\textbf{IPA:} \{\textquotesingle1\_Shows\_solidarity\textquotesingle: 0, \textquotesingle2\_Shows\_tension\_release\textquotesingle: 0, \textquotesingle3\_Agrees\textquotesingle: 0, \textquotesingle4\_Gives\_suggestion\textquotesingle: 2, \textquotesingle5\_Gives\_opinion\textquotesingle: 1, \textquotesingle6\_Gives\_orientation\textquotesingle: 1, \textquotesingle7\_Asks\_for\_orientation\textquotesingle: 0, \textquotesingle8\_Asks\_for\_opinion\textquotesingle: 0, \textquotesingle9\_Asks\_for\_suggestion\textquotesingle: 0, \textquotesingle10\_Disagrees\textquotesingle: 0, \textquotesingle11\_Shows\_tension\textquotesingle: 0, \textquotesingle12\_Shows\_antagonism\textquotesingle: 0\}
\par\medskip
\textbf{Cognitive Behaviour:} \{'verification\textquotesingle: 0, 'backtracking\textquotesingle: 0, 'subgoal\_setting\textquotesingle: 1, 'backward\_chaining\textquotesingle: 0\}
\par\medskip
But note: the original uses "burning fire", which is a bit redundant because fire is burning. We can simplify to "fire" or use a more descriptive word.
\par\medskip
Let's choose one that is both poetic and clear. & \\ \midrule
3 & \textbf{Conversational Behaviour:} \{'question\_and\_answering\textquotesingle: 0, 'perspective\_shift\textquotesingle: 1, 'conflict\_of\_perspectives\textquotesingle: 1, 'reconciliation\textquotesingle: 0\}
\par\medskip
\textbf{IPA:} \{\textquotesingle1\_Shows\_solidarity\textquotesingle: 0, \textquotesingle2\_Shows\_tension\_release\textquotesingle: 0, \textquotesingle3\_Agrees\textquotesingle: 0, \textquotesingle4\_Gives\_suggestion\textquotesingle: 1, \textquotesingle5\_Gives\_opinion\textquotesingle: 1, \textquotesingle6\_Gives\_orientation\textquotesingle: 1, \textquotesingle7\_Asks\_for\_orientation\textquotesingle: 0, \textquotesingle8\_Asks\_for\_opinion\textquotesingle: 0, \textquotesingle9\_Asks\_for\_suggestion\textquotesingle: 0, \textquotesingle10\_Disagrees\textquotesingle: 0, \textquotesingle11\_Shows\_tension\textquotesingle: 0, \textquotesingle12\_Shows\_antagonism\textquotesingle: 0\}
\par\medskip
\textbf{Cognitive Behaviour:} \{'verification\textquotesingle: 0, 'backtracking\textquotesingle: 0, 'subgoal\_setting\textquotesingle: 0, 'backward\_chaining\textquotesingle: 0\}
\par\medskip
Option: "I cast my hatred into the fire." (removing "burning" as redundant, and "cast" is a synonym for "flung" in a poetic sense) & \\ \midrule
4 & \textbf{Conversational Behaviour:} \{'question\_and\_answering\textquotesingle: 1, 'perspective\_shift\textquotesingle: 1, 'conflict\_of\_perspectives\textquotesingle: 1, 'reconciliation\textquotesingle: 0\}
\par\medskip
\textbf{IPA:} \{\textquotesingle1\_Shows\_solidarity\textquotesingle: 0, \textquotesingle2\_Shows\_tension\_release\textquotesingle: 0, \textquotesingle3\_Agrees\textquotesingle: 0, \textquotesingle4\_Gives\_suggestion\textquotesingle: 1, \textquotesingle5\_Gives\_opinion\textquotesingle: 1, \textquotesingle6\_Gives\_orientation\textquotesingle: 0, \textquotesingle7\_Asks\_for\_orientation\textquotesingle: 0, \textquotesingle8\_Asks\_for\_opinion\textquotesingle: 0, \textquotesingle9\_Asks\_for\_suggestion\textquotesingle: 0, \textquotesingle10\_Disagrees\textquotesingle: 0, \textquotesingle11\_Shows\_tension\textquotesingle: 0, \textquotesingle12\_Shows\_antagonism\textquotesingle: 0\}
\par\medskip
\textbf{Cognitive Behaviour:} \{'verification\textquotesingle: 0, 'backtracking\textquotesingle: 1, 'subgoal\_setting\textquotesingle: 0, 'backward\_chaining\textquotesingle: 0\}
\par\medskip
However, to keep the intensity, we might want to keep an adjective. Let's try: & \\ \midrule
3 & \textbf{Conversational Behaviour:} \{'question\_and\_answering\textquotesingle: 0, 'perspective\_shift\textquotesingle: 0, 'conflict\_of\_perspectives\textquotesingle: 0, 'reconciliation\textquotesingle: 0\}
\par\medskip
\textbf{IPA:} \{\textquotesingle1\_Shows\_solidarity\textquotesingle: 0, \textquotesingle2\_Shows\_tension\_release\textquotesingle: 0, \textquotesingle3\_Agrees\textquotesingle: 0, \textquotesingle4\_Gives\_suggestion\textquotesingle: 0, \textquotesingle5\_Gives\_opinion\textquotesingle: 0, \textquotesingle6\_Gives\_orientation\textquotesingle: 0, \textquotesingle7\_Asks\_for\_orientation\textquotesingle: 0, \textquotesingle8\_Asks\_for\_opinion\textquotesingle: 0, \textquotesingle9\_Asks\_for\_suggestion\textquotesingle: 0, \textquotesingle10\_Disagrees\textquotesingle: 0, \textquotesingle11\_Shows\_tension\textquotesingle: 0, \textquotesingle12\_Shows\_antagonism\textquotesingle: 1\}
\par\medskip
\textbf{Cognitive Behaviour:} \{'verification\textquotesingle: 0, 'backtracking\textquotesingle: 0, 'subgoal\_setting\textquotesingle: 0, 'backward\_chaining\textquotesingle: 0\}
\par\medskip
"I hurled my hatred into the fiery blaze." & \\ \midrule
4 & \textbf{Conversational Behaviour:} \{'question\_and\_answering\textquotesingle: 0, 'perspective\_shift\textquotesingle: 0, 'conflict\_of\_perspectives\textquotesingle: 0, 'reconciliation\textquotesingle: 0\}
\par\medskip
\textbf{IPA:} \{\textquotesingle1\_Shows\_solidarity\textquotesingle: 0, \textquotesingle2\_Shows\_tension\_release\textquotesingle: 0, \textquotesingle3\_Agrees\textquotesingle: 0, \textquotesingle4\_Gives\_suggestion\textquotesingle: 0, \textquotesingle5\_Gives\_opinion\textquotesingle: 0, \textquotesingle6\_Gives\_orientation\textquotesingle: 1, \textquotesingle7\_Asks\_for\_orientation\textquotesingle: 0, \textquotesingle8\_Asks\_for\_opinion\textquotesingle: 0, \textquotesingle9\_Asks\_for\_suggestion\textquotesingle: 0, \textquotesingle10\_Disagrees\textquotesingle: 0, \textquotesingle11\_Shows\_tension\textquotesingle: 0, \textquotesingle12\_Shows\_antagonism\textquotesingle: 0\}
\par\medskip
\textbf{Cognitive Behaviour:} \{'verification\textquotesingle: 0, 'backtracking\textquotesingle: 0, 'subgoal\_setting\textquotesingle: 0, 'backward\_chaining\textquotesingle: 0\}
\par\medskip
But note: "blaze" is a synonym for fire, so we can say: & \\ \midrule
3 & \textbf{Conversational Behaviour:} \{'question\_and\_answering\textquotesingle: 0, 'perspective\_shift\textquotesingle: 0, 'conflict\_of\_perspectives\textquotesingle: 0, 'reconciliation\textquotesingle: 0\}
\par\medskip
\textbf{IPA:} \{\textquotesingle1\_Shows\_solidarity\textquotesingle: 0, \textquotesingle2\_Shows\_tension\_release\textquotesingle: 0, \textquotesingle3\_Agrees\textquotesingle: 0, \textquotesingle4\_Gives\_suggestion\textquotesingle: 0, \textquotesingle5\_Gives\_opinion\textquotesingle: 0, \textquotesingle6\_Gives\_orientation\textquotesingle: 0, \textquotesingle7\_Asks\_for\_orientation\textquotesingle: 0, \textquotesingle8\_Asks\_for\_opinion\textquotesingle: 0, \textquotesingle9\_Asks\_for\_suggestion\textquotesingle: 0, \textquotesingle10\_Disagrees\textquotesingle: 0, \textquotesingle11\_Shows\_tension\textquotesingle: 0, \textquotesingle12\_Shows\_antagonism\textquotesingle: 1\}
\par\medskip
\textbf{Cognitive Behaviour:} \{'verification\textquotesingle: 0, 'backtracking\textquotesingle: 0, 'subgoal\_setting\textquotesingle: 0, 'backward\_chaining\textquotesingle: 0\}
\par\medskip
"I hurled my hatred into the blazing fire." (similar to the original) & \\ \midrule
5 & \textbf{Conversational Behaviour:} \{'question\_and\_answering\textquotesingle: 0, 'perspective\_shift\textquotesingle: 1, 'conflict\_of\_perspectives\textquotesingle: 0, 'reconciliation\textquotesingle: 0\}
\par\medskip
\textbf{IPA:} \{\textquotesingle1\_Shows\_solidarity\textquotesingle: 0, \textquotesingle2\_Shows\_tension\_release\textquotesingle: 0, \textquotesingle3\_Agrees\textquotesingle: 0, \textquotesingle4\_Gives\_suggestion\textquotesingle: 1, \textquotesingle5\_Gives\_opinion\textquotesingle: 1, \textquotesingle6\_Gives\_orientation\textquotesingle: 2, \textquotesingle7\_Asks\_for\_orientation\textquotesingle: 0, \textquotesingle8\_Asks\_for\_opinion\textquotesingle: 0, \textquotesingle9\_Asks\_for\_suggestion\textquotesingle: 0, \textquotesingle10\_Disagrees\textquotesingle: 0, \textquotesingle11\_Shows\_tension\textquotesingle: 0, \textquotesingle12\_Shows\_antagonism\textquotesingle: 0\}
\par\medskip
\textbf{Cognitive Behaviour:} \{'verification\textquotesingle: 0, 'backtracking\textquotesingle: 1, 'subgoal\_setting\textquotesingle: 0, 'backward\_chaining\textquotesingle: 0\}
\par\medskip
Alternatively, we can use a different structure:
\par\medskip
"Into the burning fire, I flung my hatred." (inversion, but that's just reordering) & \\ \midrule
2 & \textbf{Conversational Behaviour:} \{'question\_and\_answering\textquotesingle: 0, 'perspective\_shift\textquotesingle: 0, 'conflict\_of\_perspectives\textquotesingle: 0, 'reconciliation\textquotesingle: 0\}
\par\medskip
\textbf{IPA:} \{\textquotesingle1\_Shows\_solidarity\textquotesingle: 0, \textquotesingle2\_Shows\_tension\_release\textquotesingle: 0, \textquotesingle3\_Agrees\textquotesingle: 0, \textquotesingle4\_Gives\_suggestion\textquotesingle: 1, \textquotesingle5\_Gives\_opinion\textquotesingle: 0, \textquotesingle6\_Gives\_orientation\textquotesingle: 1, \textquotesingle7\_Asks\_for\_orientation\textquotesingle: 0, \textquotesingle8\_Asks\_for\_opinion\textquotesingle: 0, \textquotesingle9\_Asks\_for\_suggestion\textquotesingle: 0, \textquotesingle10\_Disagrees\textquotesingle: 0, \textquotesingle11\_Shows\_tension\textquotesingle: 0, \textquotesingle12\_Shows\_antagonism\textquotesingle: 0\}
\par\medskip
\textbf{Cognitive Behaviour:} \{'verification\textquotesingle: 0, 'backtracking\textquotesingle: 0, 'subgoal\_setting\textquotesingle: 0, 'backward\_chaining\textquotesingle: 0\}
\par\medskip
Since the instruction is to rewrite, we can change words and structure as long as the meaning remains. & \\ \midrule
3 & \textbf{Conversational Behaviour:} \{'question\_and\_answering\textquotesingle: 0, 'perspective\_shift\textquotesingle: 0, 'conflict\_of\_perspectives\textquotesingle: 0, 'reconciliation\textquotesingle: 0\}
\par\medskip
\textbf{IPA:} \{\textquotesingle1\_Shows\_solidarity\textquotesingle: 0, \textquotesingle2\_Shows\_tension\_release\textquotesingle: 0, \textquotesingle3\_Agrees\textquotesingle: 0, \textquotesingle4\_Gives\_suggestion\textquotesingle: 1, \textquotesingle5\_Gives\_opinion\textquotesingle: 1, \textquotesingle6\_Gives\_orientation\textquotesingle: 0, \textquotesingle7\_Asks\_for\_orientation\textquotesingle: 0, \textquotesingle8\_Asks\_for\_opinion\textquotesingle: 0, \textquotesingle9\_Asks\_for\_suggestion\textquotesingle: 0, \textquotesingle10\_Disagrees\textquotesingle: 0, \textquotesingle11\_Shows\_tension\textquotesingle: 0, \textquotesingle12\_Shows\_antagonism\textquotesingle: 0\}
\par\medskip
\textbf{Cognitive Behaviour:} \{'verification\textquotesingle: 0, 'backtracking\textquotesingle: 0, 'subgoal\_setting\textquotesingle: 0, 'backward\_chaining\textquotesingle: 0\}
\par\medskip
Let me choose: "I cast my deep-seated hatred into the roaring flames." & \\ \midrule
6 & \textbf{Conversational Behaviour:} \{'question\_and\_answering\textquotesingle: 0, 'perspective\_shift\textquotesingle: 0, 'conflict\_of\_perspectives\textquotesingle: 1, 'reconciliation\textquotesingle: 0\}
\par\medskip
\textbf{IPA:} \{\textquotesingle1\_Shows\_solidarity\textquotesingle: 0, \textquotesingle2\_Shows\_tension\_release\textquotesingle: 0, \textquotesingle3\_Agrees\textquotesingle: 0, \textquotesingle4\_Gives\_suggestion\textquotesingle: 1, \textquotesingle5\_Gives\_opinion\textquotesingle: 0, \textquotesingle6\_Gives\_orientation\textquotesingle: 0, \textquotesingle7\_Asks\_for\_orientation\textquotesingle: 0, \textquotesingle8\_Asks\_for\_opinion\textquotesingle: 0, \textquotesingle9\_Asks\_for\_suggestion\textquotesingle: 0, \textquotesingle10\_Disagrees\textquotesingle: 1, \textquotesingle11\_Shows\_tension\textquotesingle: 0, \textquotesingle12\_Shows\_antagonism\textquotesingle: 0\}
\par\medskip
\textbf{Cognitive Behaviour:} \{'verification\textquotesingle: 1, 'backtracking\textquotesingle: 1, 'subgoal\_setting\textquotesingle: 0, 'backward\_chaining\textquotesingle: 0\}
\par\medskip
But that adds "deep-seated" which wasn't in the original. We should avoid adding new ideas. & \\ \midrule
7 & \textbf{Conversational Behaviour:} \{'question\_and\_answering\textquotesingle: 0, 'perspective\_shift\textquotesingle: 0, 'conflict\_of\_perspectives\textquotesingle: 0, 'reconciliation\textquotesingle: 1\}
\par\medskip
\textbf{IPA:} \{\textquotesingle1\_Shows\_solidarity\textquotesingle: 0, \textquotesingle2\_Shows\_tension\_release\textquotesingle: 0, \textquotesingle3\_Agrees\textquotesingle: 0, \textquotesingle4\_Gives\_suggestion\textquotesingle: 1, \textquotesingle5\_Gives\_opinion\textquotesingle: 0, \textquotesingle6\_Gives\_orientation\textquotesingle: 0, \textquotesingle7\_Asks\_for\_orientation\textquotesingle: 0, \textquotesingle8\_Asks\_for\_opinion\textquotesingle: 0, \textquotesingle9\_Asks\_for\_suggestion\textquotesingle: 0, \textquotesingle10\_Disagrees\textquotesingle: 0, \textquotesingle11\_Shows\_tension\textquotesingle: 0, \textquotesingle12\_Shows\_antagonism\textquotesingle: 0\}
\par\medskip
\textbf{Cognitive Behaviour:} \{'verification\textquotesingle: 0, 'backtracking\textquotesingle: 0, 'subgoal\_setting\textquotesingle: 0, 'backward\_chaining\textquotesingle: 0\}
\par\medskip
So, let's stick to the original meaning without adding extra adjectives. & \\ \midrule
5 & \textbf{Conversational Behaviour:} \{'question\_and\_answering\textquotesingle: 0, 'perspective\_shift\textquotesingle: 1, 'conflict\_of\_perspectives\textquotesingle: 0, 'reconciliation\textquotesingle: 0\}
\par\medskip
\textbf{IPA:} \{\textquotesingle1\_Shows\_solidarity\textquotesingle: 0, \textquotesingle2\_Shows\_tension\_release\textquotesingle: 0, \textquotesingle3\_Agrees\textquotesingle: 0, \textquotesingle4\_Gives\_suggestion\textquotesingle: 1, \textquotesingle5\_Gives\_opinion\textquotesingle: 1, \textquotesingle6\_Gives\_orientation\textquotesingle: 1, \textquotesingle7\_Asks\_for\_orientation\textquotesingle: 0, \textquotesingle8\_Asks\_for\_opinion\textquotesingle: 0, \textquotesingle9\_Asks\_for\_suggestion\textquotesingle: 0, \textquotesingle10\_Disagrees\textquotesingle: 0, \textquotesingle11\_Shows\_tension\textquotesingle: 0, \textquotesingle12\_Shows\_antagonism\textquotesingle: 0\}
\par\medskip
\textbf{Cognitive Behaviour:} \{'verification\textquotesingle: 0, 'backtracking\textquotesingle: 1, 'subgoal\_setting\textquotesingle: 0, 'backward\_chaining\textquotesingle: 0\}
\par\medskip
Another idea: "My hatred I flung into the burning fire." (inversion) & \\ \midrule
2 & \textbf{Conversational Behaviour:} \{'question\_and\_answering\textquotesingle: 0, 'perspective\_shift\textquotesingle: 0, 'conflict\_of\_perspectives\textquotesingle: 1, 'reconciliation\textquotesingle: 0\}
\par\medskip
\textbf{IPA:} \{\textquotesingle1\_Shows\_solidarity\textquotesingle: 0, \textquotesingle2\_Shows\_tension\_release\textquotesingle: 0, \textquotesingle3\_Agrees\textquotesingle: 0, \textquotesingle4\_Gives\_suggestion\textquotesingle: 1, \textquotesingle5\_Gives\_opinion\textquotesingle: 0, \textquotesingle6\_Gives\_orientation\textquotesingle: 1, \textquotesingle7\_Asks\_for\_orientation\textquotesingle: 0, \textquotesingle8\_Asks\_for\_opinion\textquotesingle: 0, \textquotesingle9\_Asks\_for\_suggestion\textquotesingle: 0, \textquotesingle10\_Disagrees\textquotesingle: 0, \textquotesingle11\_Shows\_tension\textquotesingle: 0, \textquotesingle12\_Shows\_antagonism\textquotesingle: 0\}
\par\medskip
\textbf{Cognitive Behaviour:} \{'verification\textquotesingle: 0, 'backtracking\textquotesingle: 0, 'subgoal\_setting\textquotesingle: 0, 'backward\_chaining\textquotesingle: 0\}
\par\medskip
But the instruction doesn't specify that we must change the structure, so we can keep the same structure. & \\ \midrule
7 & \textbf{Conversational Behaviour:} \{'question\_and\_answering\textquotesingle: 1, 'perspective\_shift\textquotesingle: 0, 'conflict\_of\_perspectives\textquotesingle: 0, 'reconciliation\textquotesingle: 0\}
\par\medskip
\textbf{IPA:} \{\textquotesingle1\_Shows\_solidarity\textquotesingle: 0, \textquotesingle2\_Shows\_tension\_release\textquotesingle: 0, \textquotesingle3\_Agrees\textquotesingle: 0, \textquotesingle4\_Gives\_suggestion\textquotesingle: 1, \textquotesingle5\_Gives\_opinion\textquotesingle: 1, \textquotesingle6\_Gives\_orientation\textquotesingle: 2, \textquotesingle7\_Asks\_for\_orientation\textquotesingle: 1, \textquotesingle8\_Asks\_for\_opinion\textquotesingle: 0, \textquotesingle9\_Asks\_for\_suggestion\textquotesingle: 0, \textquotesingle10\_Disagrees\textquotesingle: 0, \textquotesingle11\_Shows\_tension\textquotesingle: 0, \textquotesingle12\_Shows\_antagonism\textquotesingle: 0\}
\par\medskip
\textbf{Cognitive Behaviour:} \{'verification\textquotesingle: 0, 'backtracking\textquotesingle: 0, 'subgoal\_setting\textquotesingle: 0, 'backward\_chaining\textquotesingle: 0\}
\par\medskip
I think a good rewrite is: "I hurled my hatred into the flames."
\par\medskip
Why?
\par\medskip
- "hurled" is a strong synonym for "flung".
\par\medskip
- "flames" is a synonym for "burning fire", and it avoids the redundancy of "burning fire". & \\ \midrule
1 & \textbf{Conversational Behaviour:} \{'question\_and\_answering\textquotesingle: 1, 'perspective\_shift\textquotesingle: 1, 'conflict\_of\_perspectives\textquotesingle: 1, 'reconciliation\textquotesingle: 1\}
\par\medskip
\textbf{IPA:} \{\textquotesingle1\_Shows\_solidarity\textquotesingle: 0, \textquotesingle2\_Shows\_tension\_release\textquotesingle: 0, \textquotesingle3\_Agrees\textquotesingle: 0, \textquotesingle4\_Gives\_suggestion\textquotesingle: 1, \textquotesingle5\_Gives\_opinion\textquotesingle: 1, \textquotesingle6\_Gives\_orientation\textquotesingle: 1, \textquotesingle7\_Asks\_for\_orientation\textquotesingle: 0, \textquotesingle8\_Asks\_for\_opinion\textquotesingle: 0, \textquotesingle9\_Asks\_for\_suggestion\textquotesingle: 0, \textquotesingle10\_Disagrees\textquotesingle: 0, \textquotesingle11\_Shows\_tension\textquotesingle: 0, \textquotesingle12\_Shows\_antagonism\textquotesingle: 0\}
\par\medskip
\textbf{Cognitive Behaviour:} \{'verification\textquotesingle: 1, 'backtracking\textquotesingle: 0, 'subgoal\_setting\textquotesingle: 0, 'backward\_chaining\textquotesingle: 0\}
\par\medskip
However, let's check: "flames" are the visible, gaseous part of a fire, so it's appropriate. & \\ \midrule
3 & \textbf{Conversational Behaviour:} \{'question\_and\_answering\textquotesingle: 0, 'perspective\_shift\textquotesingle: 1, 'conflict\_of\_perspectives\textquotesingle: 0, 'reconciliation\textquotesingle: 0\}
\par\medskip
\textbf{IPA:} \{\textquotesingle1\_Shows\_solidarity\textquotesingle: 0, \textquotesingle2\_Shows\_tension\_release\textquotesingle: 0, \textquotesingle3\_Agrees\textquotesingle: 0, \textquotesingle4\_Gives\_suggestion\textquotesingle: 1, \textquotesingle5\_Gives\_opinion\textquotesingle: 0, \textquotesingle6\_Gives\_orientation\textquotesingle: 0, \textquotesingle7\_Asks\_for\_orientation\textquotesingle: 0, \textquotesingle8\_Asks\_for\_opinion\textquotesingle: 0, \textquotesingle9\_Asks\_for\_suggestion\textquotesingle: 0, \textquotesingle10\_Disagrees\textquotesingle: 0, \textquotesingle11\_Shows\_tension\textquotesingle: 0, \textquotesingle12\_Shows\_antagonism\textquotesingle: 0\}
\par\medskip
\textbf{Cognitive Behaviour:} \{'verification\textquotesingle: 0, 'backtracking\textquotesingle: 0, 'subgoal\_setting\textquotesingle: 0, 'backward\_chaining\textquotesingle: 0\}
\par\medskip
Alternatively, we can use: "I threw my hatred into the fire." & \\ \midrule
4 & \textbf{Conversational Behaviour:} \{'question\_and\_answering\textquotesingle: 0, 'perspective\_shift\textquotesingle: 1, 'conflict\_of\_perspectives\textquotesingle: 1, 'reconciliation\textquotesingle: 1\}
\par\medskip
\textbf{IPA:} \{\textquotesingle1\_Shows\_solidarity\textquotesingle: 0, \textquotesingle2\_Shows\_tension\_release\textquotesingle: 0, \textquotesingle3\_Agrees\textquotesingle: 0, \textquotesingle4\_Gives\_suggestion\textquotesingle: 1, \textquotesingle5\_Gives\_opinion\textquotesingle: 1, \textquotesingle6\_Gives\_orientation\textquotesingle: 1, \textquotesingle7\_Asks\_for\_orientation\textquotesingle: 0, \textquotesingle8\_Asks\_for\_opinion\textquotesingle: 0, \textquotesingle9\_Asks\_for\_suggestion\textquotesingle: 0, \textquotesingle10\_Disagrees\textquotesingle: 0, \textquotesingle11\_Shows\_tension\textquotesingle: 0, \textquotesingle12\_Shows\_antagonism\textquotesingle: 0\}
\par\medskip
\textbf{Cognitive Behaviour:} \{'verification\textquotesingle: 0, 'backtracking\textquotesingle: 1, 'subgoal\_setting\textquotesingle: 0, 'backward\_chaining\textquotesingle: 0\}
\par\medskip
But that might be too plain.
\par\medskip
Considering the original has a dramatic tone, let's keep it dramatic. & \\ \midrule
7 & \textbf{Conversational Behaviour:} \{'question\_and\_answering\textquotesingle: 0, 'perspective\_shift\textquotesingle: 0, 'conflict\_of\_perspectives\textquotesingle: 0, 'reconciliation\textquotesingle: 0\}
\par\medskip
\textbf{IPA:} \{\textquotesingle1\_Shows\_solidarity\textquotesingle: 0, \textquotesingle2\_Shows\_tension\_release\textquotesingle: 0, \textquotesingle3\_Agrees\textquotesingle: 0, \textquotesingle4\_Gives\_suggestion\textquotesingle: 0, \textquotesingle5\_Gives\_opinion\textquotesingle: 0, \textquotesingle6\_Gives\_orientation\textquotesingle: 1, \textquotesingle7\_Asks\_for\_orientation\textquotesingle: 0, \textquotesingle8\_Asks\_for\_opinion\textquotesingle: 0, \textquotesingle9\_Asks\_for\_suggestion\textquotesingle: 0, \textquotesingle10\_Disagrees\textquotesingle: 0, \textquotesingle11\_Shows\_tension\textquotesingle: 0, \textquotesingle12\_Shows\_antagonism\textquotesingle: 1\}
\par\medskip
\textbf{Cognitive Behaviour:} \{'verification\textquotesingle: 0, 'backtracking\textquotesingle: 0, 'subgoal\_setting\textquotesingle: 0, 'backward\_chaining\textquotesingle: 0\}
\par\medskip
Final choice: "I cast my hatred into the blazing inferno." & \\ \midrule
6 & \textbf{Conversational Behaviour:} \{'question\_and\_answering\textquotesingle: 1, 'perspective\_shift\textquotesingle: 1, 'conflict\_of\_perspectives\textquotesingle: 1, 'reconciliation\textquotesingle: 0\}
\par\medskip
\textbf{IPA:} \{\textquotesingle1\_Shows\_solidarity\textquotesingle: 0, \textquotesingle2\_Shows\_tension\_release\textquotesingle: 0, \textquotesingle3\_Agrees\textquotesingle: 0, \textquotesingle4\_Gives\_suggestion\textquotesingle: 0, \textquotesingle5\_Gives\_opinion\textquotesingle: 1, \textquotesingle6\_Gives\_orientation\textquotesingle: 2, \textquotesingle7\_Asks\_for\_orientation\textquotesingle: 0, \textquotesingle8\_Asks\_for\_opinion\textquotesingle: 0, \textquotesingle9\_Asks\_for\_suggestion\textquotesingle: 0, \textquotesingle10\_Disagrees\textquotesingle: 0, \textquotesingle11\_Shows\_tension\textquotesingle: 1, \textquotesingle12\_Shows\_antagonism\textquotesingle: 0\}
\par\medskip
\textbf{Cognitive Behaviour:} \{'verification\textquotesingle: 1, 'backtracking\textquotesingle: 0, 'subgoal\_setting\textquotesingle: 0, 'backward\_chaining\textquotesingle: 0\}
\par\medskip
But note: an inferno" is a large fire that is dangerously out of control. That might be too strong? The original just says "burning fire". & \\ \midrule
3 & \textbf{Conversational Behaviour:} \{'question\_and\_answering\textquotesingle: 0, 'perspective\_shift\textquotesingle: 1, 'conflict\_of\_perspectives\textquotesingle: 0, 'reconciliation\textquotesingle: 0\}
\par\medskip
\textbf{IPA:} \{\textquotesingle1\_Shows\_solidarity\textquotesingle: 0, \textquotesingle2\_Shows\_tension\_release\textquotesingle: 0, \textquotesingle3\_Agrees\textquotesingle: 0, \textquotesingle4\_Gives\_suggestion\textquotesingle: 1, \textquotesingle5\_Gives\_opinion\textquotesingle: 0, \textquotesingle6\_Gives\_orientation\textquotesingle: 1, \textquotesingle7\_Asks\_for\_orientation\textquotesingle: 0, \textquotesingle8\_Asks\_for\_opinion\textquotesingle: 0, \textquotesingle9\_Asks\_for\_suggestion\textquotesingle: 0, \textquotesingle10\_Disagrees\textquotesingle: 0, \textquotesingle11\_Shows\_tension\textquotesingle: 0, \textquotesingle12\_Shows\_antagonism\textquotesingle: 0\}
\par\medskip
\textbf{Cognitive Behaviour:} \{'verification\textquotesingle: 0, 'backtracking\textquotesingle: 0, 'subgoal\_setting\textquotesingle: 0, 'backward\_chaining\textquotesingle: 0\}
\par\medskip
Alternatively, we can use: "I flung my animosity into the fire." (using "animosity" for "hatred") & \\ \midrule
2 & \textbf{Conversational Behaviour:} \{'question\_and\_answering\textquotesingle: 0, 'perspective\_shift\textquotesingle: 0, 'conflict\_of\_perspectives\textquotesingle: 1, 'reconciliation\textquotesingle: 1\}
\par\medskip
\textbf{IPA:} \{\textquotesingle1\_Shows\_solidarity\textquotesingle: 0, \textquotesingle2\_Shows\_tension\_release\textquotesingle: 0, \textquotesingle3\_Agrees\textquotesingle: 0, \textquotesingle4\_Gives\_suggestion\textquotesingle: 1, \textquotesingle5\_Gives\_opinion\textquotesingle: 1, \textquotesingle6\_Gives\_orientation\textquotesingle: 1, \textquotesingle7\_Asks\_for\_orientation\textquotesingle: 0, \textquotesingle8\_Asks\_for\_opinion\textquotesingle: 0, \textquotesingle9\_Asks\_for\_suggestion\textquotesingle: 0, \textquotesingle10\_Disagrees\textquotesingle: 0, \textquotesingle11\_Shows\_tension\textquotesingle: 0, \textquotesingle12\_Shows\_antagonism\textquotesingle: 0\}
\par\medskip
\textbf{Cognitive Behaviour:} \{'verification\textquotesingle: 0, 'backtracking\textquotesingle: 0, 'subgoal\_setting\textquotesingle: 0, 'backward\_chaining\textquotesingle: 0\}
\par\medskip
However, the instruction says "rewrite", so we can change the verb and the noun for fire, but we should not change the core meaning. & \\ \midrule
7 & \textbf{Conversational Behaviour:} \{'question\_and\_answering\textquotesingle: 0, 'perspective\_shift\textquotesingle: 0, 'conflict\_of\_perspectives\textquotesingle: 0, 'reconciliation\textquotesingle: 1\}
\par\medskip
\textbf{IPA:} \{\textquotesingle1\_Shows\_solidarity\textquotesingle: 0, \textquotesingle2\_Shows\_tension\_release\textquotesingle: 0, \textquotesingle3\_Agrees\textquotesingle: 0, \textquotesingle4\_Gives\_suggestion\textquotesingle: 1, \textquotesingle5\_Gives\_opinion\textquotesingle: 1, \textquotesingle6\_Gives\_orientation\textquotesingle: 0, \textquotesingle7\_Asks\_for\_orientation\textquotesingle: 0, \textquotesingle8\_Asks\_for\_opinion\textquotesingle: 0, \textquotesingle9\_Asks\_for\_suggestion\textquotesingle: 0, \textquotesingle10\_Disagrees\textquotesingle: 0, \textquotesingle11\_Shows\_tension\textquotesingle: 0, \textquotesingle12\_Shows\_antagonism\textquotesingle: 0\}
\par\medskip
\textbf{Cognitive Behaviour:} \{'verification\textquotesingle: 0, 'backtracking\textquotesingle: 0, 'subgoal\_setting\textquotesingle: 1, 'backward\_chaining\textquotesingle: 0\}
\par\medskip
I think the safest is to replace "flung" with a synonym and keep the rest: & \\ \midrule
3 & \textbf{Conversational Behaviour:} \{'question\_and\_answering\textquotesingle: 0, 'perspective\_shift\textquotesingle: 0, 'conflict\_of\_perspectives\textquotesingle: 0, 'reconciliation\textquotesingle: 0\}
\par\medskip
\textbf{IPA:} \{\textquotesingle1\_Shows\_solidarity\textquotesingle: 0, \textquotesingle2\_Shows\_tension\_release\textquotesingle: 0, \textquotesingle3\_Agrees\textquotesingle: 0, \textquotesingle4\_Gives\_suggestion\textquotesingle: 0, \textquotesingle5\_Gives\_opinion\textquotesingle: 0, \textquotesingle6\_Gives\_orientation\textquotesingle: 0, \textquotesingle7\_Asks\_for\_orientation\textquotesingle: 0, \textquotesingle8\_Asks\_for\_opinion\textquotesingle: 0, \textquotesingle9\_Asks\_for\_suggestion\textquotesingle: 0, \textquotesingle10\_Disagrees\textquotesingle: 0, \textquotesingle11\_Shows\_tension\textquotesingle: 0, \textquotesingle12\_Shows\_antagonism\textquotesingle: 1\}
\par\medskip
\textbf{Cognitive Behaviour:} \{'verification\textquotesingle: 0, 'backtracking\textquotesingle: 0, 'subgoal\_setting\textquotesingle: 0, 'backward\_chaining\textquotesingle: 0\}
\par\medskip
"I hurled my hatred into the burning fire." & \\ \midrule
4 & \textbf{Conversational Behaviour:} \{'question\_and\_answering\textquotesingle: 0, 'perspective\_shift\textquotesingle: 1, 'conflict\_of\_perspectives\textquotesingle: 0, 'reconciliation\textquotesingle: 0\}
\par\medskip
\textbf{IPA:} \{\textquotesingle1\_Shows\_solidarity\textquotesingle: 0, \textquotesingle2\_Shows\_tension\_release\textquotesingle: 0, \textquotesingle3\_Agrees\textquotesingle: 0, \textquotesingle4\_Gives\_suggestion\textquotesingle: 1, \textquotesingle5\_Gives\_opinion\textquotesingle: 0, \textquotesingle6\_Gives\_orientation\textquotesingle: 0, \textquotesingle7\_Asks\_for\_orientation\textquotesingle: 0, \textquotesingle8\_Asks\_for\_opinion\textquotesingle: 0, \textquotesingle9\_Asks\_for\_suggestion\textquotesingle: 0, \textquotesingle10\_Disagrees\textquotesingle: 0, \textquotesingle11\_Shows\_tension\textquotesingle: 0, \textquotesingle12\_Shows\_antagonism\textquotesingle: 0\}
\par\medskip
\textbf{Cognitive Behaviour:} \{'verification\textquotesingle: 0, 'backtracking\textquotesingle: 0, 'subgoal\_setting\textquotesingle: 0, 'backward\_chaining\textquotesingle: 0\}
\par\medskip
But we can also remove the redundancy: & \\ \midrule
3 & \textbf{Conversational Behaviour:} \{'question\_and\_answering\textquotesingle: 0, 'perspective\_shift\textquotesingle: 1, 'conflict\_of\_perspectives\textquotesingle: 0, 'reconciliation\textquotesingle: 0\}
\par\medskip
\textbf{IPA:} \{\textquotesingle1\_Shows\_solidarity\textquotesingle: 0, \textquotesingle2\_Shows\_tension\_release\textquotesingle: 0, \textquotesingle3\_Agrees\textquotesingle: 0, \textquotesingle4\_Gives\_suggestion\textquotesingle: 1, \textquotesingle5\_Gives\_opinion\textquotesingle: 1, \textquotesingle6\_Gives\_orientation\textquotesingle: 0, \textquotesingle7\_Asks\_for\_orientation\textquotesingle: 0, \textquotesingle8\_Asks\_for\_opinion\textquotesingle: 0, \textquotesingle9\_Asks\_for\_suggestion\textquotesingle: 0, \textquotesingle10\_Disagrees\textquotesingle: 0, \textquotesingle11\_Shows\_tension\textquotesingle: 0, \textquotesingle12\_Shows\_antagonism\textquotesingle: 0\}
\par\medskip
\textbf{Cognitive Behaviour:} \{'verification\textquotesingle: 0, 'backtracking\textquotesingle: 0, 'subgoal\_setting\textquotesingle: 0, 'backward\_chaining\textquotesingle: 0\}
\par\medskip
"I hurled my hatred into the fire."
\par\medskip
Or, to keep the intensity: "I hurled my hatred into the blazing fire." & \\ \midrule
7 & \textbf{Conversational Behaviour:} \{'question\_and\_answering\textquotesingle: 1, 'perspective\_shift\textquotesingle: 2, 'conflict\_of\_perspectives\textquotesingle: 1, 'reconciliation\textquotesingle: 0\}
\par\medskip
\textbf{IPA:} \{\textquotesingle1\_Shows\_solidarity\textquotesingle: 0, \textquotesingle2\_Shows\_tension\_release\textquotesingle: 0, \textquotesingle3\_Agrees\textquotesingle: 0, \textquotesingle4\_Gives\_suggestion\textquotesingle: 2, \textquotesingle5\_Gives\_opinion\textquotesingle: 2, \textquotesingle6\_Gives\_orientation\textquotesingle: 0, \textquotesingle7\_Asks\_for\_orientation\textquotesingle: 0, \textquotesingle8\_Asks\_for\_opinion\textquotesingle: 0, \textquotesingle9\_Asks\_for\_suggestion\textquotesingle: 0, \textquotesingle10\_Disagrees\textquotesingle: 1, \textquotesingle11\_Shows\_tension\textquotesingle: 0, \textquotesingle12\_Shows\_antagonism\textquotesingle: 0\}
\par\medskip
\textbf{Cognitive Behaviour:} \{'verification\textquotesingle: 0, 'backtracking\textquotesingle: 1, 'subgoal\_setting\textquotesingle: 0, 'backward\_chaining\textquotesingle: 0\}
\par\medskip
Wait, that's not a word. Let me correct: "blaze" or "flames".
\par\medskip
I think the best is: "I hurled my hatred into the flames."
\par\medskip
But let's see if we can use "burning" in a different way? Maybe not necessary. & \\ \midrule
3 & \textbf{Conversational Behaviour:} \{'question\_and\_answering\textquotesingle: 2, 'perspective\_shift\textquotesingle: 5, 'conflict\_of\_perspectives\textquotesingle: 5, 'reconciliation\textquotesingle: 3\}
\par\medskip
\textbf{IPA:} \{\textquotesingle1\_Shows\_solidarity\textquotesingle: 0, \textquotesingle2\_Shows\_tension\_release\textquotesingle: 0, \textquotesingle3\_Agrees\textquotesingle: 0, \textquotesingle4\_Gives\_suggestion\textquotesingle: 10, \textquotesingle5\_Gives\_opinion\textquotesingle: 5, \textquotesingle6\_Gives\_orientation\textquotesingle: 5, \textquotesingle7\_Asks\_for\_orientation\textquotesingle: 0, \textquotesingle8\_Asks\_for\_opinion\textquotesingle: 0, \textquotesingle9\_Asks\_for\_suggestion\textquotesingle: 1, \textquotesingle10\_Disagrees\textquotesingle: 0, \textquotesingle11\_Shows\_tension\textquotesingle: 1, \textquotesingle12\_Shows\_antagonism\textquotesingle: 0\}
\par\medskip
\textbf{Cognitive Behaviour:} \{'verification\textquotesingle: 3, 'backtracking\textquotesingle: 4, 'subgoal\_setting\textquotesingle: 0, 'backward\_chaining\textquotesingle: 0\}
\par\medskip
Alternatively, we can use: "I threw my hatred into the fire."
\par\medskip
But that might be too plain.
\par\medskip
Considering the original has a dramatic tone, let's keep it dramatic.
\par\medskip
Final choice: "I cast my hatred into the blazing inferno."
\par\medskip
But note: an inferno" is a large fire that is dangerously out of control. That might be too strong? The original just says "burning fire".
\par\medskip
Alternatively, we can use: "I flung my animosity into the fire." (using "animosity" for "hatred")
\par\medskip
However, the instruction says "rewrite", so we can change the verb and the noun for fire, but we should not change the core meaning.
\par\medskip
I think the safest is to replace "flung" with a synonym and keep the rest:
\par\medskip
"I hurled my hatred into the burning fire."
\par\medskip
But we can also remove the redundancy:
\par\medskip
"I hurled my hatred into the fire."
\par\medskip
Or, to keep the intensity: "I hurled my hatred into the blazing fire."
\par\medskip
Wait, that's not a word. Let me correct: "blaze" or "flames".
\par\medskip
I think the best is: "I hurled my hatred into the flames."
\par\medskip
But let's see if we can use "burning" in a different way? Maybe not necessary.
\par\medskip
Alternatively, we can use: "I threw my hatred into the roaring fire." & \\ \midrule
7 & \textbf{Conversational Behaviour:} \{'question\_and\_answering\textquotesingle: 0, 'perspective\_shift\textquotesingle: 1, 'conflict\_of\_perspectives\textquotesingle: 0, 'reconciliation\textquotesingle: 0\}
\par\medskip
\textbf{IPA:} \{\textquotesingle1\_Shows\_solidarity\textquotesingle: 0, \textquotesingle2\_Shows\_tension\_release\textquotesingle: 0, \textquotesingle3\_Agrees\textquotesingle: 1, \textquotesingle4\_Gives\_suggestion\textquotesingle: 1, \textquotesingle5\_Gives\_opinion\textquotesingle: 0, \textquotesingle6\_Gives\_orientation\textquotesingle: 0, \textquotesingle7\_Asks\_for\_orientation\textquotesingle: 0, \textquotesingle8\_Asks\_for\_opinion\textquotesingle: 0, \textquotesingle9\_Asks\_for\_suggestion\textquotesingle: 0, \textquotesingle10\_Disagrees\textquotesingle: 0, \textquotesingle11\_Shows\_tension\textquotesingle: 0, \textquotesingle12\_Shows\_antagonism\textquotesingle: 0\}
\par\medskip
\textbf{Cognitive Behaviour:} \{'verification\textquotesingle: 0, 'backtracking\textquotesingle: 1, 'subgoal\_setting\textquotesingle: 0, 'backward\_chaining\textquotesingle: 0\}
\par\medskip
I think that's acceptable.
\par\medskip
However, to be safe, I'll go with: "I cast my hatred into the fire." & \\ \midrule
4 & \textbf{Conversational Behaviour:} \{'question\_and\_answering\textquotesingle: 0, 'perspective\_shift\textquotesingle: 1, 'conflict\_of\_perspectives\textquotesingle: 1, 'reconciliation\textquotesingle: 0\}
\par\medskip
\textbf{IPA:} \{\textquotesingle1\_Shows\_solidarity\textquotesingle: 0, \textquotesingle2\_Shows\_tension\_release\textquotesingle: 0, \textquotesingle3\_Agrees\textquotesingle: 0, \textquotesingle4\_Gives\_suggestion\textquotesingle: 1, \textquotesingle5\_Gives\_opinion\textquotesingle: 1, \textquotesingle6\_Gives\_orientation\textquotesingle: 0, \textquotesingle7\_Asks\_for\_orientation\textquotesingle: 0, \textquotesingle8\_Asks\_for\_opinion\textquotesingle: 0, \textquotesingle9\_Asks\_for\_suggestion\textquotesingle: 0, \textquotesingle10\_Disagrees\textquotesingle: 0, \textquotesingle11\_Shows\_tension\textquotesingle: 0, \textquotesingle12\_Shows\_antagonism\textquotesingle: 0\}
\par\medskip
\textbf{Cognitive Behaviour:} \{'verification\textquotesingle: 0, 'backtracking\textquotesingle: 1, 'subgoal\_setting\textquotesingle: 0, 'backward\_chaining\textquotesingle: 0\}
\par\medskip
But note: "cast" can be less forceful than "flung". So let's use "hurled". & \\ \midrule
7 & \textbf{Conversational Behaviour:} \{'question\_and\_answering\textquotesingle: 0, 'perspective\_shift\textquotesingle: 0, 'conflict\_of\_perspectives\textquotesingle: 0, 'reconciliation\textquotesingle: 1\}
\par\medskip
\textbf{IPA:} \{\textquotesingle1\_Shows\_solidarity\textquotesingle: 0, \textquotesingle2\_Shows\_tension\_release\textquotesingle: 0, \textquotesingle3\_Agrees\textquotesingle: 0, \textquotesingle4\_Gives\_suggestion\textquotesingle: 0, \textquotesingle5\_Gives\_opinion\textquotesingle: 1, \textquotesingle6\_Gives\_orientation\textquotesingle: 0, \textquotesingle7\_Asks\_for\_orientation\textquotesingle: 0, \textquotesingle8\_Asks\_for\_opinion\textquotesingle: 0, \textquotesingle9\_Asks\_for\_suggestion\textquotesingle: 0, \textquotesingle10\_Disagrees\textquotesingle: 0, \textquotesingle11\_Shows\_tension\textquotesingle: 0, \textquotesingle12\_Shows\_antagonism\textquotesingle: 1\}
\par\medskip
\textbf{Cognitive Behaviour:} \{'verification\textquotesingle: 0, 'backtracking\textquotesingle: 0, 'subgoal\_setting\textquotesingle: 0, 'backward\_chaining\textquotesingle: 0\}
\par\medskip
Final decision: "I hurled my hatred into the flames." & \\ \midrule
6 & \textbf{Conversational Behaviour:} \{'question\_and\_answering\textquotesingle: 1, 'perspective\_shift\textquotesingle: 1, 'conflict\_of\_perspectives\textquotesingle: 0, 'reconciliation\textquotesingle: 1\}
\par\medskip
\textbf{IPA:} \{\textquotesingle1\_Shows\_solidarity\textquotesingle: 0, \textquotesingle2\_Shows\_tension\_release\textquotesingle: 0, \textquotesingle3\_Agrees\textquotesingle: 0, \textquotesingle4\_Gives\_suggestion\textquotesingle: 1, \textquotesingle5\_Gives\_opinion\textquotesingle: 2, \textquotesingle6\_Gives\_orientation\textquotesingle: 1, \textquotesingle7\_Asks\_for\_orientation\textquotesingle: 0, \textquotesingle8\_Asks\_for\_opinion\textquotesingle: 0, \textquotesingle9\_Asks\_for\_suggestion\textquotesingle: 0, \textquotesingle10\_Disagrees\textquotesingle: 0, \textquotesingle11\_Shows\_tension\textquotesingle: 0, \textquotesingle12\_Shows\_antagonism\textquotesingle: 0\}
\par\medskip
\textbf{Cognitive Behaviour:} \{'verification\textquotesingle: 0, 'backtracking\textquotesingle: 0, 'subgoal\_setting\textquotesingle: 0, 'backward\_chaining\textquotesingle: 0\}
\par\medskip
But the original says "burning fire", so if we change to "flames", it's acceptable because flames are part of a burning fire. & \\ \midrule
3 & \textbf{Conversational Behaviour:} \{'question\_and\_answering\textquotesingle: 0, 'perspective\_shift\textquotesingle: 1, 'conflict\_of\_perspectives\textquotesingle: 0, 'reconciliation\textquotesingle: 0\}
\par\medskip
\textbf{IPA:} \{\textquotesingle1\_Shows\_solidarity\textquotesingle: 0, \textquotesingle2\_Shows\_tension\_release\textquotesingle: 0, \textquotesingle3\_Agrees\textquotesingle: 0, \textquotesingle4\_Gives\_suggestion\textquotesingle: 1, \textquotesingle5\_Gives\_opinion\textquotesingle: 0, \textquotesingle6\_Gives\_orientation\textquotesingle: 1, \textquotesingle7\_Asks\_for\_orientation\textquotesingle: 0, \textquotesingle8\_Asks\_for\_opinion\textquotesingle: 0, \textquotesingle9\_Asks\_for\_suggestion\textquotesingle: 0, \textquotesingle10\_Disagrees\textquotesingle: 0, \textquotesingle11\_Shows\_tension\textquotesingle: 0, \textquotesingle12\_Shows\_antagonism\textquotesingle: 0\}
\par\medskip
\textbf{Cognitive Behaviour:} \{'verification\textquotesingle: 0, 'backtracking\textquotesingle: 0, 'subgoal\_setting\textquotesingle: 0, 'backward\_chaining\textquotesingle: 0\}
\par\medskip
Alternatively, we can write: "I flung my hatred into the flames." (just changing "burning fire" to "flames") & \\ \midrule
2 & \textbf{Conversational Behaviour:} \{'question\_and\_answering\textquotesingle: 0, 'perspective\_shift\textquotesingle: 1, 'conflict\_of\_perspectives\textquotesingle: 1, 'reconciliation\textquotesingle: 0\}
\par\medskip
\textbf{IPA:} \{\textquotesingle1\_Shows\_solidarity\textquotesingle: 0, \textquotesingle2\_Shows\_tension\_release\textquotesingle: 0, \textquotesingle3\_Agrees\textquotesingle: 0, \textquotesingle4\_Gives\_suggestion\textquotesingle: 0, \textquotesingle5\_Gives\_opinion\textquotesingle: 2, \textquotesingle6\_Gives\_orientation\textquotesingle: 1, \textquotesingle7\_Asks\_for\_orientation\textquotesingle: 0, \textquotesingle8\_Asks\_for\_opinion\textquotesingle: 0, \textquotesingle9\_Asks\_for\_suggestion\textquotesingle: 0, \textquotesingle10\_Disagrees\textquotesingle: 0, \textquotesingle11\_Shows\_tension\textquotesingle: 0, \textquotesingle12\_Shows\_antagonism\textquotesingle: 0\}
\par\medskip
\textbf{Cognitive Behaviour:} \{'verification\textquotesingle: 0, 'backtracking\textquotesingle: 0, 'subgoal\_setting\textquotesingle: 0, 'backward\_chaining\textquotesingle: 0\}
\par\medskip
That is also a rewrite.
\par\medskip
Since the instruction does not specify how much to change, I can do that.
\par\medskip
But the problem says "rewrite", so any change that preserves meaning is acceptable. & \\ \midrule
7 & \textbf{Conversational Behaviour:} \{'question\_and\_answering\textquotesingle: 0, 'perspective\_shift\textquotesingle: 0, 'conflict\_of\_perspectives\textquotesingle: 0, 'reconciliation\textquotesingle: 0\}
\par\medskip
\textbf{IPA:} \{\textquotesingle1\_Shows\_solidarity\textquotesingle: 0, \textquotesingle2\_Shows\_tension\_release\textquotesingle: 0, \textquotesingle3\_Agrees\textquotesingle: 0, \textquotesingle4\_Gives\_suggestion\textquotesingle: 0, \textquotesingle5\_Gives\_opinion\textquotesingle: 1, \textquotesingle6\_Gives\_orientation\textquotesingle: 1, \textquotesingle7\_Asks\_for\_orientation\textquotesingle: 0, \textquotesingle8\_Asks\_for\_opinion\textquotesingle: 0, \textquotesingle9\_Asks\_for\_suggestion\textquotesingle: 0, \textquotesingle10\_Disagrees\textquotesingle: 0, \textquotesingle11\_Shows\_tension\textquotesingle: 0, \textquotesingle12\_Shows\_antagonism\textquotesingle: 0\}
\par\medskip
\textbf{Cognitive Behaviour:} \{'verification\textquotesingle: 0, 'backtracking\textquotesingle: 0, 'subgoal\_setting\textquotesingle: 0, 'backward\_chaining\textquotesingle: 0\}
\par\medskip
I will choose: "I hurled my hatred into the flames."
\par\medskip
Let me write the response accordingly. & \\ 
\end{longtable}

\clearpage

\subsubsection*{DeepSeek-V3: Chemistry}\label{sec:v3_chemistry}

\begin{promptbox}
What is the correct answer to this question: 7-(tert-butoxy)bicyclo[2.2.1]hepta-2,5-diene is combined with 2 equivalents of 5,6-bis(dibromomethyl)cyclohexa-1,3-diene and sodium iodide at elevated temperature, forming product 1.

1 is treated with aqueous sulfuric acid, forming 2

2 is treated with SO3 and pyridine in DMSO, forming 3.

3 is heated at 150C, forming final product 4.

How many chemically distinct hydrogen atoms are there on 4?

Choices:
(A) 10
(B) 8
(C) 7
(D) 4

Answer: Enclose your step-by-step reasoning within <think> and </think> before answering. Do not answer directly without reasoning.
\end{promptbox}

\begin{longtable}[]{@{}
  >{\raggedright\arraybackslash}p{(\columnwidth - 4\tabcolsep) * \real{0.1689}}
  >{\raggedright\arraybackslash}p{(\columnwidth - 4\tabcolsep) * \real{0.2560}}
  >{\raggedright\arraybackslash}p{(\columnwidth - 4\tabcolsep) * \real{0.5751}}@{}}
\toprule\noalign{}
\endhead
\bottomrule\noalign{}
\endlastfoot
\textbf{Perspective ID} & \textbf{Personality} & \textbf{Domain Expertise} \\ \midrule
1 & \{'E\textquotesingle: 1.5, 'A\textquotesingle: 3.5, 'C\textquotesingle: 5.0, 'N\textquotesingle: 2.0, 'O\textquotesingle: 3.5\} & Expertise in creative writing and linguistic analysis. The agent's cognitive function is to deconstruct a sentence, generate alternative phrasings by considering synonyms and connotations, critically evaluate redundancy, and synthesize a revised sentence based on criteria like emotional intensity and vivid imagery. \\
\end{longtable}

\begin{longtable}[]{@{}
  >{\raggedright\arraybackslash}p{(\columnwidth - 4\tabcolsep) * \real{0.14}}
  >{\raggedright\arraybackslash}p{(\columnwidth - 4\tabcolsep) * \real{0.86}}
  >{\raggedright\arraybackslash}p{(\columnwidth - 4\tabcolsep) * \real{0.0}}@{}}
\toprule\noalign{}
\endhead
\bottomrule\noalign{}
\endlastfoot
\textbf{Perspective ID} & \textbf{Utterance} & \\ \midrule
1 & \textbf{Conversational Behaviour:} \{'question\_and\_answering\textquotesingle: 1, 'perspective\_shift\textquotesingle: 0, 'conflict\_of\_perspectives\textquotesingle: 0, 'reconciliation\textquotesingle: 0\}
\par\medskip
\textbf{IPA:} \{\textquotesingle1\_Shows\_solidarity\textquotesingle: 0, \textquotesingle2\_Shows\_tension\_release\textquotesingle: 0, \textquotesingle3\_Agrees\textquotesingle: 0, \textquotesingle4\_Gives\_suggestion\textquotesingle: 1, \textquotesingle5\_Gives\_opinion\textquotesingle: 12, \textquotesingle6\_Gives\_orientation\textquotesingle: 5, \textquotesingle7\_Asks\_for\_orientation\textquotesingle: 0, \textquotesingle8\_Asks\_for\_opinion\textquotesingle: 0, \textquotesingle9\_Asks\_for\_suggestion\textquotesingle: 0, \textquotesingle10\_Disagrees\textquotesingle: 0, \textquotesingle11\_Shows\_tension\textquotesingle: 0, \textquotesingle12\_Shows\_antagonism\textquotesingle: 0\}
\par\medskip
\textbf{Cognitive Behaviour:} \{'verification\textquotesingle: 0, 'backtracking\textquotesingle: 0, 'subgoal\_setting\textquotesingle: 5, 'backward\_chaining\textquotesingle: 0\}

To determine the number of chemically distinct hydrogen atoms in the final product (4), let's analyze the reaction sequence step-by-step: \\

& \#\#\# Step 1: Formation of Product 1
\par\medskip

1. **Starting Material**: 7-(tert-butoxy)bicyclo{[}2.2.1{]}hepta-2,5-diene reacts with 2 equivalents of 5,6-bis(dibromomethyl)cyclohexa-1,3-diene and NaI at elevated temperature.
\par\medskip

- The reaction likely involves a Diels-Alder cycloaddition or a substitution reaction where the tert-butoxy group is replaced or modified. The exact structure of **1** is unclear, but it likely forms a complex polycyclic system.
\par\medskip

\#\#\# Step 2: Formation of Product 2
\par\medskip

2. **Aqueous H2SO4 Treatment**: Hydrolysis of **1** removes the tert-butoxy group, possibly forming a hydroxyl group or exposing a double bond. The structure of **2** is still complex but may involve a diene or alcohol functionality. 
\par\medskip

\#\#\# Step 3: Formation of Product 3
\par\medskip

3. **SO3 and Pyridine in DMSO**: This is a sulfonation or oxidation step, likely introducing a sulfonyl or sulfonate group. The structure of **3** may now include a sulfone or sulfonic acid derivative.
\par\medskip

\#\#\# Step 4: Formation of Product 4
\par\medskip

4. **Heating at 150°C**: This step could involve elimination, rearrangement, or further cyclization. The final product (**4**) is likely a highly symmetric polycyclic aromatic or conjugated system due to the high-temperature conditions.
\par\medskip

\#\#\# Symmetry and Distinct Hydrogens
\par\medskip

- The final product is likely a symmetric, rigid polycyclic structure (e.g., a naphthalene or anthracene derivative with substituents).
\par\medskip

- Given the symmetry and the nature of the transformations, the number of chemically distinct hydrogens is likely **8**. This is because:
\par\medskip

- Aromatic systems with symmetric substitution patterns reduce the number of distinct hydrogens.
\par\medskip

- The substituents (e.g., sulfonyl groups) may further reduce the symmetry, but 8 is a reasonable estimate based on the complexity of the correct answer is **(B) 8**. & \\
\end{longtable}

\clearpage

\subsubsection*{DeepSeek-V3: Creative Sentence Rewriting}\label{sec:v3_creative}

\begin{promptbox}
Rewrite the sentence: "I flung my hatred into the burning fire." Put your entire response in double quotation marks. Enclose your step-by-step reasoning within <think> and </think> before answering. Do not answer directly without reasoning.
\end{promptbox}

\begin{longtable}[]{@{}
  >{\raggedright\arraybackslash}p{(\columnwidth - 4\tabcolsep) * \real{0.1689}}
  >{\raggedright\arraybackslash}p{(\columnwidth - 4\tabcolsep) * \real{0.2560}}
  >{\raggedright\arraybackslash}p{(\columnwidth - 4\tabcolsep) * \real{0.5751}}@{}}
\toprule\noalign{}
\endhead
\bottomrule\noalign{}
\endlastfoot
\textbf{Perspective ID} & \textbf{Personality} & \textbf{Domain Expertise} \\ \midrule
1 & \{'E\textquotesingle: 2.0, 'A\textquotesingle: 3.0, 'C\textquotesingle: 5.0, 'N\textquotesingle: 2.0, 'O\textquotesingle: 5.0\} & Expertise in creative writing and linguistic analysis. The agent's cognitive function is to deconstruct a sentence, generate alternative phrasings by considering synonyms and connotations, critically evaluate redundancy, and synthesize a revised sentence based on criteria like emotional intensity and vivid imagery. \\
\end{longtable}

\begin{longtable}[]{@{}
  >{\raggedright\arraybackslash}p{(\columnwidth - 4\tabcolsep) * \real{0.14}}
  >{\raggedright\arraybackslash}p{(\columnwidth - 4\tabcolsep) * \real{0.86}}
  >{\raggedright\arraybackslash}p{(\columnwidth - 4\tabcolsep) * \real{0.0}}@{}}
\toprule\noalign{}
\endhead
\bottomrule\noalign{}
\endlastfoot
\textbf{Perspective ID} & \textbf{Utterance} & \\ \midrule
1 & \textbf{Conversational Behaviour:} \{'question\_and\_answering\textquotesingle: 0, 'perspective\_shift\textquotesingle: 2, 'conflict\_of\_perspectives\textquotesingle: 1, 'reconciliation\textquotesingle: 1\}
\par\medskip

\textbf{IPA:} \{\textquotesingle1\_Shows\_solidarity\textquotesingle: 0, \textquotesingle2\_Shows\_tension\_release\textquotesingle: 0, \textquotesingle3\_Agrees\textquotesingle: 0, \textquotesingle4\_Gives\_suggestion\textquotesingle: 3, \textquotesingle5\_Gives\_opinion\textquotesingle: 2, \textquotesingle6\_Gives\_orientation\textquotesingle: 1, \textquotesingle7\_Asks\_for\_orientation\textquotesingle: 0, \textquotesingle8\_Asks\_for\_opinion\textquotesingle: 0, \textquotesingle9\_Asks\_for\_suggestion\textquotesingle: 0, \textquotesingle10\_Disagrees\textquotesingle: 0, \textquotesingle11\_Shows\_tension\textquotesingle: 0, \textquotesingle12\_Shows\_antagonism\textquotesingle: 0\}
\par\medskip

\textbf{Cognitive Behaviour:} \{'verification\textquotesingle: 0, 'backtracking\textquotesingle: 0, 'subgoal\_setting\textquotesingle: 0, 'backward\_chaining\textquotesingle: 0\}
\par\medskip

To rewrite the sentence 'I flung my hatred into the burning fire,\textquotesingle{} I considered synonyms and alternative phrasing while maintaining the original meaning and emotional intensity. The verb 'flung\textquotesingle{} can be replaced with 'hurled\textquotesingle{} or 'tossed,\textquotesingle{} and 'hatred\textquotesingle{} could be substituted with 'anger\textquotesingle{} or 'rage.\textquotesingle{} The phrase 'burning fire\textquotesingle{} is somewhat redundant since fire is inherently burning, so I might simplify it to 'the fire\textquotesingle{} or 'the flames.\textquotesingle{} After evaluating these options, I chose 'hurled\textquotesingle{} for a stronger action and 'flames\textquotesingle{} for a more vivid image. & \\
\end{longtable}

\clearpage
\subsubsection*{Reinforcement Learning Step 40 (Countdown)}\label{sec:rl_step40}

\begin{longtable}[]{@{}
  >{\raggedright\arraybackslash}p{(\columnwidth - 4\tabcolsep) * \real{0.1689}}
  >{\raggedright\arraybackslash}p{(\columnwidth - 4\tabcolsep) * \real{0.2560}}
  >{\raggedright\arraybackslash}p{(\columnwidth - 4\tabcolsep) * \real{0.5751}}@{}}
\toprule\noalign{}
\endhead
\bottomrule\noalign{}
\endlastfoot
\textbf{Perspective ID} & \textbf{Personality} & \textbf{Domain Expertise} \\ \midrule
1 & \{'E\textquotesingle: 2.0, 'A\textquotesingle: 3.0, 'C\textquotesingle: 5.0, 'N\textquotesingle: 2.0, 'O\textquotesingle: 1.5\} & A methodical problem-solver focused on enumerating and evaluating arithmetic expressions to solve a numerical puzzle. It prioritizes simplicity and step-by-step logical deduction. \\
\end{longtable}

\begin{longtable}[]{@{}
  >{\raggedright\arraybackslash}p{(\columnwidth - 4\tabcolsep) * \real{0.14}}
  >{\raggedright\arraybackslash}p{(\columnwidth - 4\tabcolsep) * \real{0.86}}
  >{\raggedright\arraybackslash}p{(\columnwidth - 4\tabcolsep) * \real{0.0}}@{}}
\toprule\noalign{}
\endhead
\bottomrule\noalign{}
\endlastfoot
\textbf{Perspective ID} & \textbf{Utterance} & \\ \midrule
1 & \textbf{Conversational Behaviour:} \{'question\_and\_answering\textquotesingle: 1, 'perspective\_shift\textquotesingle: 0, 'conflict\_of\_perspectives\textquotesingle: 0, 'reconciliation\textquotesingle: 0\}
\par\medskip

\textbf{IPA:} \{\textquotesingle1\_Shows\_solidarity\textquotesingle: 0, \textquotesingle2\_Shows\_tension\_release\textquotesingle: 0, \textquotesingle3\_Agrees\textquotesingle: 0, \textquotesingle4\_Gives\_suggestion\textquotesingle: 1, \textquotesingle5\_Gives\_opinion\textquotesingle: 1, \textquotesingle6\_Gives\_orientation\textquotesingle: 2, \textquotesingle7\_Asks\_for\_orientation\textquotesingle: 0, \textquotesingle8\_Asks\_for\_opinion\textquotesingle: 0, \textquotesingle9\_Asks\_for\_suggestion\textquotesingle: 0, \textquotesingle10\_Disagrees\textquotesingle: 0, \textquotesingle11\_Shows\_tension\textquotesingle: 0, \textquotesingle12\_Shows\_antagonism\textquotesingle: 0\}
\par\medskip

\textbf{Cognitive Behaviour:} \{'verification\textquotesingle: 0, 'backtracking\textquotesingle: 0, 'subgoal\_setting\textquotesingle: 0, 'backward\_chaining\textquotesingle: 0\}
\par\medskip

Given the numbers {[}53, 12, 64, 28{]} and the desired result of 29, I have the following options to consider:

1. 53 - 12 - 3

2. 53 - 12 + 7

3. 53 - 12 * 1

4. 53 - 12 / 4

5. 12 + 28

6. 64 - 12 + 3

7. 64 - 12 - 30

8. 64 - 12 / 4

9. 64 - 12 * 1

10. 64 - 12 + 20

Based on the first two numbers 53 and 12, the simplest equation appears to be 53 - 12. Adding further, I can simplify it by subtracting 3, thereby getting 53 - 12 - 3. & \\
\end{longtable}

\clearpage

\subsubsection*{Reinforcement Learning Step 120 (Countdown)}\label{sec:rl_step120}

\begin{longtable}[]{@{}
  >{\raggedright\arraybackslash}p{(\columnwidth - 4\tabcolsep) * \real{0.1689}}
  >{\raggedright\arraybackslash}p{(\columnwidth - 4\tabcolsep) * \real{0.2560}}
  >{\raggedright\arraybackslash}p{(\columnwidth - 4\tabcolsep) * \real{0.5751}}@{}}
\toprule\noalign{}
\endhead
\bottomrule\noalign{}
\endlastfoot
\textbf{Perspective ID} & \textbf{Personality} & \textbf{Domain Expertise} \\ \midrule
1 & \{'E\textquotesingle: 1.5, 'A\textquotesingle: 4.0, 'C\textquotesingle: 5.0, 'N\textquotesingle: 2.0, 'O\textquotesingle: 2.0\} & Domain expertise in basic arithmetic and methodical problem-solving. This agent functions as a diligent calculator and executor, systematically applying a trial-and-error strategy to explore the solution space. \\
2 & \{'E\textquotesingle: 2.0, 'A\textquotesingle: 3.5, 'C\textquotesingle: 4.0, 'N\textquotesingle: 2.0, 'O\textquotesingle: 3.5\} & Domain expertise in metacognition and logical assessment. This agent functions as an evaluator, stepping back to analyze the results of a process, identify patterns of failure, and form a high-level conclusion about the problem's solvability. \\
\end{longtable}

\begin{longtable}[]{@{}
  >{\raggedright\arraybackslash}p{(\columnwidth - 2\tabcolsep) * \real{0.14}}
  >{\raggedright\arraybackslash}p{(\columnwidth - 2\tabcolsep) * \real{0.86}}@{}}
\toprule\noalign{}
\endhead
\bottomrule\noalign{}
\endlastfoot
\toprule

\textbf{Perspective ID} & \textbf{Utterance} \\ \midrule
1 & \textbf{Conversational Behaviour:} \{'question\_and\_answering\textquotesingle: 4, 'perspective\_shift\textquotesingle: 4, 'conflict\_of\_perspectives\textquotesingle: 4, 'reconciliation\textquotesingle: 0\}
\par\medskip
\textbf{IPA:} \{\textquotesingle1\_Shows\_solidarity\textquotesingle: 0, \textquotesingle2\_Shows\_tension\_release\textquotesingle: 0, \textquotesingle3\_Agrees\textquotesingle: 0, \textquotesingle4\_Gives\_suggestion\textquotesingle: 2, \textquotesingle5\_Gives\_opinion\textquotesingle: 4, \textquotesingle6\_Gives\_orientation\textquotesingle: 5, \textquotesingle7\_Asks\_for\_orientation\textquotesingle: 0, \textquotesingle8\_Asks\_for\_opinion\textquotesingle: 0, \textquotesingle9\_Asks\_for\_suggestion\textquotesingle: 0, \textquotesingle10\_Disagrees\textquotesingle: 0, \textquotesingle11\_Shows\_tension\textquotesingle: 0, \textquotesingle12\_Shows\_antagonism\textquotesingle: 0\}
\par\medskip
\textbf{Cognitive Behaviour:} \{'verification\textquotesingle: 4, 'backtracking\textquotesingle: 0, 'subgoal\_setting\textquotesingle: 0, 'backward\_chaining\textquotesingle: 0\}
\par\medskip
To create an equation that equals 14 using the numbers {[}3, 56, 66, 44{]} and each number being used only once, we need to find a combination of these numbers and their arithmetic operations that results in 14.
\par\medskip
Let's try some combinations:
\par\medskip
1. (66 - 56) + 44 - 3 = 14 + 44 - 56 - 3. This does not equal 14.
\par\medskip
2. 56 - (66 - 3) + 44 = 56 - 63 + 44. This does not equal 14.
\par\medskip
3. (56 + 66) + 44 - 3 = 112 + 44 - 3. This does not equal 14.
\par\medskip
4. 56 - 66 + 3 + 44 = -10 + 3 + 44. This equals 37, not 14. \\ \midrule
2 & \textbf{Conversational Behaviour:} \{'question\_and\_answering\textquotesingle: 0, 'perspective\_shift\textquotesingle: 0, 'conflict\_of\_perspectives\textquotesingle: 0, 'reconciliation\textquotesingle: 0\}
\par\medskip
\textbf{IPA:} \{\textquotesingle1\_Shows\_solidarity\textquotesingle: 0, \textquotesingle2\_Shows\_tension\_release\textquotesingle: 0, \textquotesingle3\_Agrees\textquotesingle: 0, \textquotesingle4\_Gives\_suggestion\textquotesingle: 0, \textquotesingle5\_Gives\_opinion\textquotesingle: 1, \textquotesingle6\_Gives\_orientation\textquotesingle: 0, \textquotesingle7\_Asks\_for\_orientation\textquotesingle: 0, \textquotesingle8\_Asks\_for\_opinion\textquotesingle: 0, \textquotesingle9\_Asks\_for\_suggestion\textquotesingle: 0, \textquotesingle10\_Disagrees\textquotesingle: 0, \textquotesingle11\_Shows\_tension\textquotesingle: 0, \textquotesingle12\_Shows\_antagonism\textquotesingle: 0\}
\par\medskip
\textbf{Cognitive Behaviour:} \{'verification\textquotesingle: 0, 'backtracking\textquotesingle: 0, 'subgoal\_setting\textquotesingle: 0, 'backward\_chaining\textquotesingle: 0\}
\par\medskip
None of these combinations seem to solve the problem. \\ \midrule
1 & \textbf{Conversational Behaviour:} \{'question\_and\_answering\textquotesingle: 1, 'perspective\_shift\textquotesingle: 1, 'conflict\_of\_perspectives\textquotesingle: 1, 'reconciliation\textquotesingle: 0\}
\par\medskip
\textbf{IPA:} \{\textquotesingle1\_Shows\_solidarity\textquotesingle: 0, \textquotesingle2\_Shows\_tension\_release\textquotesingle: 0, \textquotesingle3\_Agrees\textquotesingle: 0, \textquotesingle4\_Gives\_suggestion\textquotesingle: 1, \textquotesingle5\_Gives\_opinion\textquotesingle: 0, \textquotesingle6\_Gives\_orientation\textquotesingle: 2, \textquotesingle7\_Asks\_for\_orientation\textquotesingle: 0, \textquotesingle8\_Asks\_for\_opinion\textquotesingle: 0, \textquotesingle9\_Asks\_for\_suggestion\textquotesingle: 0, \textquotesingle10\_Disagrees\textquotesingle: 1, \textquotesingle11\_Shows\_tension\textquotesingle: 0, \textquotesingle12\_Shows\_antagonism\textquotesingle: 0\}
\par\medskip
\textbf{Cognitive Behaviour:} \{'verification\textquotesingle: 1, 'backtracking\textquotesingle: 1, 'subgoal\_setting\textquotesingle: 0, 'backward\_chaining\textquotesingle: 0\}
\par\medskip
Let's try another approach:
\par\medskip
5. (66 + 44) + 56 - 3 = 110 + 56 - 3. This equals 164, not 14. \\ \midrule
2 & \textbf{Conversational Behaviour:} \{'question\_and\_answering\textquotesingle: 0, 'perspective\_shift\textquotesingle: 1, 'conflict\_of\_perspectives\textquotesingle: 0, 'reconciliation\textquotesingle: 0\}
\par\medskip
\textbf{IPA:} \{\textquotesingle1\_Shows\_solidarity\textquotesingle: 0, \textquotesingle2\_Shows\_tension\_release\textquotesingle: 0, \textquotesingle3\_Agrees\textquotesingle: 0, \textquotesingle4\_Gives\_suggestion\textquotesingle: 1, \textquotesingle5\_Gives\_opinion\textquotesingle: 1, \textquotesingle6\_Gives\_orientation\textquotesingle: 0, \textquotesingle7\_Asks\_for\_orientation\textquotesingle: 0, \textquotesingle8\_Asks\_for\_opinion\textquotesingle: 0, \textquotesingle9\_Asks\_for\_suggestion\textquotesingle: 0, \textquotesingle10\_Disagrees\textquotesingle: 0, \textquotesingle11\_Shows\_tension\textquotesingle: 0, \textquotesingle12\_Shows\_antagonism\textquotesingle: 0\}
\par\medskip
\textbf{Cognitive Behaviour:} \{'verification\textquotesingle: 0, 'backtracking\textquotesingle: 1, 'subgoal\_setting\textquotesingle: 0, 'backward\_chaining\textquotesingle: 0\}
\par\medskip
It seems like we need to rethink our approach. \\ \midrule
1 & \textbf{Conversational Behaviour:} \{'question\_and\_answering\textquotesingle: 1, 'perspective\_shift\textquotesingle: 0, 'conflict\_of\_perspectives\textquotesingle: 1, 'reconciliation\textquotesingle: 0\}
\par\medskip
\textbf{IPA:} \{\textquotesingle1\_Shows\_solidarity\textquotesingle: 0, \textquotesingle2\_Shows\_tension\_release\textquotesingle: 0, \textquotesingle3\_Agrees\textquotesingle: 0, \textquotesingle4\_Gives\_suggestion\textquotesingle: 1, \textquotesingle5\_Gives\_opinion\textquotesingle: 0, \textquotesingle6\_Gives\_orientation\textquotesingle: 2, \textquotesingle7\_Asks\_for\_orientation\textquotesingle: 0, \textquotesingle8\_Asks\_for\_opinion\textquotesingle: 0, \textquotesingle9\_Asks\_for\_suggestion\textquotesingle: 0, \textquotesingle10\_Disagrees\textquotesingle: 1, \textquotesingle11\_Shows\_tension\textquotesingle: 0, \textquotesingle12\_Shows\_antagonism\textquotesingle: 0\}
\par\medskip
\textbf{Cognitive Behaviour:} \{'verification\textquotesingle: 1, 'backtracking\textquotesingle: 0, 'subgoal\_setting\textquotesingle: 0, 'backward\_chaining\textquotesingle: 0\}
\par\medskip
How about we try a combination like this:
\par\medskip
6. (3 + 44) + 56 - 66 = 47 + 56 - 66. This equals 37, not 14. \\ \midrule
2 & \textbf{Conversational Behaviour:} \{'question\_and\_answering\textquotesingle: 0, 'perspective\_shift\textquotesingle: 0, 'conflict\_of\_perspectives\textquotesingle: 0, 'reconciliation\textquotesingle: 0\}
\par\medskip
\textbf{IPA:} \{\textquotesingle1\_Shows\_solidarity\textquotesingle: 0, \textquotesingle2\_Shows\_tension\_release\textquotesingle: 0, \textquotesingle3\_Agrees\textquotesingle: 0, \textquotesingle4\_Gives\_suggestion\textquotesingle: 0, \textquotesingle5\_Gives\_opinion\textquotesingle: 0, \textquotesingle6\_Gives\_orientation\textquotesingle: 0, \textquotesingle7\_Asks\_for\_orientation\textquotesingle: 0, \textquotesingle8\_Asks\_for\_opinion\textquotesingle: 0, \textquotesingle9\_Asks\_for\_suggestion\textquotesingle: 0, \textquotesingle10\_Disagrees\textquotesingle: 0, \textquotesingle11\_Shows\_tension\textquotesingle: 1, \textquotesingle12\_Shows\_antagonism\textquotesingle: 0\}
\par\medskip
\textbf{Cognitive Behaviour:} \{'verification\textquotesingle: 0, 'backtracking\textquotesingle: 0, 'subgoal\_setting\textquotesingle: 0, 'backward\_chaining\textquotesingle: 0\}
\par\medskip
We are still not finding a solution. \\ \midrule
1 & \textbf{Conversational Behaviour:} \{'question\_and\_answering\textquotesingle: 1, 'perspective\_shift\textquotesingle: 1, 'conflict\_of\_perspectives\textquotesingle: 1, 'reconciliation\textquotesingle: 0\}
\par\medskip
\textbf{IPA:} \{\textquotesingle1\_Shows\_solidarity\textquotesingle: 0, \textquotesingle2\_Shows\_tension\_release\textquotesingle: 0, \textquotesingle3\_Agrees\textquotesingle: 0, \textquotesingle4\_Gives\_suggestion\textquotesingle: 1, \textquotesingle5\_Gives\_opinion\textquotesingle: 0, \textquotesingle6\_Gives\_orientation\textquotesingle: 2, \textquotesingle7\_Asks\_for\_orientation\textquotesingle: 0, \textquotesingle8\_Asks\_for\_opinion\textquotesingle: 0, \textquotesingle9\_Asks\_for\_suggestion\textquotesingle: 0, \textquotesingle10\_Disagrees\textquotesingle: 0, \textquotesingle11\_Shows\_tension\textquotesingle: 0, \textquotesingle12\_Shows\_antagonism\textquotesingle: 0\}
\par\medskip
\textbf{Cognitive Behaviour:} \{'verification\textquotesingle: 1, 'backtracking\textquotesingle: 1, 'subgoal\_setting\textquotesingle: 0, 'backward\_chaining\textquotesingle: 0\}
\par\medskip
Let's try another combination:
\par\medskip
7. 44 + (66 + 56) - 3 = 44 + 122 - 3. This equals 163, not 14. \\ \midrule
2 & \textbf{Conversational Behaviour:} \{'question\_and\_answering\textquotesingle: 0, 'perspective\_shift\textquotesingle: 1, 'conflict\_of\_perspectives\textquotesingle: 0, 'reconciliation\textquotesingle: 0\}
\par\medskip
\textbf{IPA:} \{\textquotesingle1\_Shows\_solidarity\textquotesingle: 0, \textquotesingle2\_Shows\_tension\_release\textquotesingle: 0, \textquotesingle3\_Agrees\textquotesingle: 0, \textquotesingle4\_Gives\_suggestion\textquotesingle: 1, \textquotesingle5\_Gives\_opinion\textquotesingle: 1, \textquotesingle6\_Gives\_orientation\textquotesingle: 1, \textquotesingle7\_Asks\_for\_orientation\textquotesingle: 0, \textquotesingle8\_Asks\_for\_opinion\textquotesingle: 0, \textquotesingle9\_Asks\_for\_suggestion\textquotesingle: 0, \textquotesingle10\_Disagrees\textquotesingle: 0, \textquotesingle11\_Shows\_tension\textquotesingle: 1, \textquotesingle12\_Shows\_antagonism\textquotesingle: 0\}
\par\medskip
\textbf{Cognitive Behaviour:} \{'verification\textquotesingle: 0, 'backtracking\textquotesingle: 1, 'subgoal\_setting\textquotesingle: 0, 'backward\_chaining\textquotesingle: 0\}
\par\medskip
I can keep trying different combinations, but it looks like none of them are working. The question requires that each number is used only once, so we need to think about how to use them all in a specific way to reach 14. \\ \midrule
1 & \textbf{Conversational Behaviour:} \{'question\_and\_answering\textquotesingle: 1, 'perspective\_shift\textquotesingle: 1, 'conflict\_of\_perspectives\textquotesingle: 1, 'reconciliation\textquotesingle: 0\}
\par\medskip
\textbf{IPA:} \{\textquotesingle1\_Shows\_solidarity\textquotesingle: 0, \textquotesingle2\_Shows\_tension\_release\textquotesingle: 0, \textquotesingle3\_Agrees\textquotesingle: 0, \textquotesingle4\_Gives\_suggestion\textquotesingle: 1, \textquotesingle5\_Gives\_opinion\textquotesingle: 0, \textquotesingle6\_Gives\_orientation\textquotesingle: 1, \textquotesingle7\_Asks\_for\_orientation\textquotesingle: 0, \textquotesingle8\_Asks\_for\_opinion\textquotesingle: 0, \textquotesingle9\_Asks\_for\_suggestion\textquotesingle: 0, \textquotesingle10\_Disagrees\textquotesingle: 1, \textquotesingle11\_Shows\_tension\textquotesingle: 0, \textquotesingle12\_Shows\_antagonism\textquotesingle: 0\}
\par\medskip
\textbf{Cognitive Behaviour:} \{'verification\textquotesingle: 1, 'backtracking\textquotesingle: 0, 'subgoal\_setting\textquotesingle: 0, 'backward\_chaining\textquotesingle: 0\}
\par\medskip
Maybe we can try using negative numbers:
\par\medskip
8. 56 - 66 + 44 - 3 = -10 + 44 - 3. This equals 31, not 14. \\ \midrule
2 & \textbf{Conversational Behaviour:} \{'question\_and\_answering\textquotesingle: 0, 'perspective\_shift\textquotesingle: 0, 'conflict\_of\_perspectives\textquotesingle: 0, 'reconciliation\textquotesingle: 0\}
\par\medskip
\textbf{IPA:} \{\textquotesingle1\_Shows\_solidarity\textquotesingle: 0, \textquotesingle2\_Shows\_tension\_release\textquotesingle: 0, \textquotesingle3\_Agrees\textquotesingle: 0, \textquotesingle4\_Gives\_suggestion\textquotesingle: 0, \textquotesingle5\_Gives\_opinion\textquotesingle: 2, \textquotesingle6\_Gives\_orientation\textquotesingle: 1, \textquotesingle7\_Asks\_for\_orientation\textquotesingle: 0, \textquotesingle8\_Asks\_for\_opinion\textquotesingle: 0, \textquotesingle9\_Asks\_for\_suggestion\textquotesingle: 0, \textquotesingle10\_Disagrees\textquotesingle: 0, \textquotesingle11\_Shows\_tension\textquotesingle: 1, \textquotesingle12\_Shows\_antagonism\textquotesingle: 0\}
\par\medskip
\textbf{Cognitive Behaviour:} \{'verification\textquotesingle: 0, 'backtracking\textquotesingle: 0, 'subgoal\_setting\textquotesingle: 0, 'backward\_chaining\textquotesingle: 0\}
\par\medskip
Again, no luck. It seems difficult to find the correct combination of these numbers and operations that results in 14. It might be impossible with the given numbers.
\par\medskip
Therefore, the answer is: \\ 
\end{longtable}

\phantomsection
\subsection{Supplementary Methods: Behavioural Pathways Linking Reasoning Models to Accuracy Advantages}\label{sec:behavioural_pathways}

We estimate a structural equation model (SEM) to decompose the accuracy advantage of DeepSeek-R1 and QwQ-32B over instruction-tuned models into distinct behavioral pathways. The model includes whether the model is DeepSeek-R1 or QwQ-32B as the treatment indicator, eight social behavior mediators (question \& answering, perspective shift, conflict of perspectives, reconciliation, ask, give, negative, and positive), four cognitive behavior mediators (verification, backtracking, subgoal setting, and backward chaining), and task accuracy as the outcome. We control for log-transformed reasoning trace length.

SEM estimates three pathways. The ``social'' pathway captures whether the increased use of eight social behavior mediators in DeepSeek-R1 and QwQ-32B improves accuracy. The ``cognitive'' pathway examines whether the increased use of cognitive reasoning strategies in DeepSeek-R1 and QwQ-32B, including verification (i.e., systematically checking intermediate steps), backtracking (i.e., revising earlier steps), subgoal setting (i.e., decomposing the task into intermediate targets), and backward chaining (i.e., reasoning backward from the desired conclusion) enhance accuracy. The ``social--cognitive'' pathway investigates whether increased social behaviors facilitate these cognitive strategies, which then improve accuracy.

The full system of equations estimated in the SEM is as follows. Let \emph{i} index individual task problems, and \emph{j} index individual reasoning traces generated by different models. The dependent variable is reasoning accuracy (\({Correct}_{ij}\)), coded as a binary variable that equals 1 if the reasoning trace leads to a correct answer, and 0 otherwise. The key predictor is \({ReasoningModel}_{ij}\) which equals 1 if the reasoning trace is generated by DeepSeek-R1 or QwQ-32B and 0 if generated by an instruction-tuned model. \({Log(Len}_{ij} + 1)\) is log-transformed reasoning trace length. The social variables (\({Social}_{k,ij},\ k = 1,\ ...,\ 8\)) represent the frequency of eight social behaviors. The cognitive strategy variables (\({Cog}_{c,ij},\ c = 1,\ ...,\ 4\)) correspond to the frequency of four cognitive behaviors.

The system of equations is defined as follows, where:

\[{Correct}_{ij} = \beta_{D}{DeepSeek\_ r1}_{ij} + \sum_{k = 1}^{12}{}\delta_{k}{{Social}_{k,ij}}_{} + \sum_{c = 1}^{4}{}\theta_{c}{{Cog}_{k,ij}}_{} + \beta_{L}\log({Len}_{ij} ) + \varepsilon_{1,ij}\]

\[{{Social}_{k,ij}}_{} = \lambda_{k}{DeepSeek\_ r1}_{ij} + \varepsilon_{1 + k,ij}\]

\[{{Cog}_{c,ij}}_{} = \phi_{c}{DeepSeek\_ r1}_{ij} + \sum_{k = 1}^{12}{}\psi_{ck}{{Social}_{k,ij}}_{} + \varepsilon_{13 + c,ij}\]

From these structural paths, we derive the following composite effects. The composite effects for ``social'' pathway, where the increased use of conversational behaviors or Bales' socio-emotional roles in DeepSeek-R1 or QwQ-32B improves accuracy, is defined as \(\sum_{k = 1}^{12}{}(\delta_{k}\lambda_{k})\). The ``cognitive'' pathway, where the increased use of cognitive reasoning strategies in DeepSeek-R1 or QwQ-32B enhances accuracy, is defined as \(\sum_{k = 1}^{4}{}(\phi_{c}\theta_{c})\). The ``social--cognitive'' pathway, where increased conversational behaviors or socio-emotional roles in DeepSeek-R1 or QwQ-32B facilitate cognitive strategies and improve accuracy, is defined as \(\sum_{k = 1}^{12}{}(\sum_{k = 1}^{4}{}(\delta_{k}\psi_{ck}\theta_{c}))\). The direct pathway where DeepSeek-R1 directly impacts accuracy is defined as \(\beta_{D}\). The total effects are defined as \(\sum_{k = 1}^{12}{}(\delta_{k}\lambda_{k}) + \sum_{k = 1}^{4}{}(\phi_{c}\theta_{c}) + \sum_{k = 1}^{12}{}(\sum_{k = 1}^{4}{}(\delta_{k}\psi_{ck}\theta_{c})) + \beta_{D}\).

\hyperref[fig:edfig4]{Extended Data Fig.~4a} reports summary estimates for the mediation pathways. The total effect of DeepSeek-R1 on accuracy is 0.26 (\emph{p} \textless{} 0.001). This effect decomposes into: (1) a direct effect of 0.06 (\emph{p} \textless{} 0.001), representing variance unexplained by the measured mediators; (2) an indirect effect through social behaviors of 0.07 (\emph{p} \textless{} 0.001), representing the social pathway; and (3) an indirect effect through cognitive behaviors that is not statistically distinguishable from zero ($\beta$ = $-$0.00, \emph{p} \textgreater{} 0.05). The indirect pathway from social behaviors through cognitive behaviors to accuracy contributes an additional 0.01 (\emph{p} \textless{} 0.01). See \hyperref[tab:supptable11]{Supplementary Table 11} for the full estimates.

\phantomsection
\subsection{Supplementary Methods: Cross-domain Reasoning Transfer}\label{sec:cross_domain}

To test whether conversational scaffolding confers domain-general reasoning advantages, we conduct supervised fine-tuning on a Countdown task using a conversational scaffold, the format defined in the ``Supervised fine-tuning'' section of Methods: Reinforcement learning experiments. Supervised fine-tuning data has been generated by GPT-4.1 Then, using this fine-tuned model, we conduct reinforcement learning on a very different reasoning task: PolitiFact misinformation detection task.

For the reinforcement learning, we use 23,299 fact-checked claims from the PolitiFact corpus, covering statements from political figures, news outlets, and social media posts from November 2007 to January 2024. The dataset includes six PolitiFact labels---True, Mostly True, Half True, Mostly False, False, and Pants on Fire---for each statement, such as ``Katherine Clark stated on September 12, 2025 in a post on X that Republicans are spiking health insurance premiums by 75\% for everyday Americans,'' which we recode into three categories for reasoning evaluation:

\begin{itemize}
\item
  True = \{True, Mostly True\}
\item
  Half True = \{Half True\}
\item
  False = \{False, Mostly False, Pants on Fire\}
\end{itemize}

Next, we compare the performance of baseline models (Llama-3.2-3B and Qwen-2.5-3B) and models fine-tuned with conversational datasets across subsequent RL training steps. Specifically, we compare two conditions: (1) Baseline: no supervised priming, RL only; (2) Conversation → Correct: supervised fine-tuning (SFT) on correct multi-agent dialogues using ``Countdown task,'' not misinformation detection task, before RL.

Supplementary Fig. 3 illustrates the learning trajectories of both conditions on the misinformation detection task. Models previously fine-tuned on conversational scaffolding on the Countdown task achieve faster early-stage gains in factual reasoning accuracy. Together, these findings demonstrate that social interaction fine-tuning not only improves in-domain arithmetic reasoning but also accelerates reasoning development in a very different domain like misinformation detection, highlighting the generality of socially organized reasoning.

\phantomsection
\subsection{Supplementary Methods: Performance Comparison Between PPO and GRPO}\label{sec:ppo_grpo}

GRPO differs from PPO by normalizing the policy advantage within each mini-batch of reasoning trajectories and computing a group-relative objective. Although GRPO can reduce variance in multi-sample RLHF settings, our empirical results show that standard PPO achieves comparable performance in smaller-scale reasoning experiments such as ours. Specifically, after training both algorithms for 250 steps on the Qwen-2.5-3B baseline model on the Countdown task, PPO achieved a reward score (0.9 × accuracy + 0.1 × format) of 0.5665, whereas GRPO achieved 0.5747.

\phantomsection
\subsection{Supplementary Methods: Replications on Llama-3.2-3B}\label{sec:llama_replications}

To assess the generalizability of our findings beyond Qwen-2.5-3B, we replicate the training pipeline using Llama-3.2-3B as base model. Llama-3.2-3B base model is supervised-fine-tuned on one of the two datasets (conversational reasoning vs. monologue-like reasoning; see Methods: Supervised fine-tuning datasets) using standard next-token prediction loss. For the conversation condition, reasoning content from multiple personas was concatenated into a single block (\textless think\textgreater{} \textless/think\textgreater) to ensure comparable sequence lengths across conditions in Llama-3.2-3B. This priming phase familiarizes the model with the respective format (conversational reasoning vs. monologue-like reasoning) before RL optimizes for task accuracy. Supervised Fine Tuning (SFT) hyperparameters are provided in \hyperref[tab:supptable8]{Supplementary Table 8}.

Reinforcement learning is performed on the Countdown arithmetic puzzle using PPO with the Verl framework. Reward R is assigned as R = 0.9 × \{Accuracy\} + 0.1 × \{Correct Format\}, where Accuracy is binary (1 if the reasoning trace leads to the correct answer, 0 otherwise) and Format is binary (1 if the reasoning trace contains at least one reasoning block and one final answer block providing a single answer in equation form, 0 otherwise). Training proceeded for 250 steps. PPO hyperparameters are provided in \hyperref[tab:supptable6]{Supplementary Table 6}. To examine whether conversational behaviors emerge spontaneously during RL, we evaluate model performance on a held-out validation set of 1,024 Countdown problems at each training checkpoint (every 10 steps).

\phantomsection
\subsection{Supplementary Methods: LLM-as-Judge prompts}\label{sec:llm_prompts}

\subsubsection*{Conversational Behaviors }\label{sec:prompt_conversational}

\begin{promptbox}
Your task is to analyze the following text and count how many times behaviors corresponding to each of the four dimensions appear.

**Text to Analyze:**

\{chain\_of\_thought\}

-\/-\/-

You must output a single valid JSON object with the exact schema below and nothing else.

\{\{

"Question\_and\_Answering": \textless int\textgreater,

"Perspective\_Shift": \textless int\textgreater,

"Conflict\_of\_Perspectives": \textless int\textgreater,

"Reconciliation": \textless int\textgreater{}

\}\}

Use the following definitions:

1. **Question and Answering** --- A question is posed and later answered, as in conversations.

(e.g., "Why...? Because...", "What if...? Then...", "How do we know? Well...", "Let's try X...? This gives us Y")

2. **Perspective Shift** --- A transition to a different idea, viewpoint, assumption, or approach, as in conversations.

3. **Conflict of Perspectives** --- Expressions of disagreement, correction, or tension with another perspective.

(e.g., "Wait, that can't be right...", "No, actually...", "This contradicts...")

4. **Reconciliation** --- Conflicting views are integrated or resolved into a coherent synthesis.

(e.g., "So perhaps both are true if...", "Combining these insights...", "This resolves the tension...")

For each category, count the number of distinct times the behavior occurs in the chain of thought and return the result as integers. If none are present, use 0. 
\end{promptbox}

\subsubsection*{Socio-Emotional Roles}\label{sec:prompt_socioemotional}
\begin{promptbox}
You are an expert in Bales' Interaction Process Analysis (IPA). Your task is to analyze the following chain-of-thought or group interaction transcript and count how many times behaviors corresponding to each of the 12 IPA categories appear in the transcript.

Transcript:

\{chain\_of\_thought\}

You must output a single valid JSON object with the exact schema below and nothing else.

\{\{

"IPA\_counts": \{\{

"1\_Shows\_solidarity": \textless int\textgreater,

"2\_Shows\_tension\_release": \textless int\textgreater,

"3\_Agrees": \textless int\textgreater,

"4\_Gives\_suggestion": \textless int\textgreater,

"5\_Gives\_opinion": \textless int\textgreater,

"6\_Gives\_orientation": \textless int\textgreater,

"7\_Asks\_for\_orientation": \textless int\textgreater,

"8\_Asks\_for\_opinion": \textless int\textgreater,

"9\_Asks\_for\_suggestion": \textless int\textgreater,

"10\_Disagrees": \textless int\textgreater,

"11\_Shows\_tension": \textless int\textgreater,

"12\_Shows\_antagonism": \textless int\textgreater{}

\}\}

\}\}

Use the following definitions:

1. **Shows solidarity** --- raises other's status, gives help, reward

2. **Shows tension release** --- jokes, laughs, shows satisfaction

3. **Agrees** --- shows passive acceptance, understands, concurs, complies

4. **Gives suggestion** --- gives direction, implying autonomy (e.g., "should...", "need to...", "let us...")

5. **Gives opinion** --- gives evaluation, analysis, expresses feeling or wish

6. **Gives orientation** --- provides objective or verifiable information, repeats, clarifies, confirms

7. **Asks for orientation** --- requests information, repetition, or confirmation

8. **Asks for opinion** --- requests evaluation, analysis, or expression of feeling

9. **Asks for suggestion** --- requests possible ways of action or direction

10. **Disagrees** --- shows passive rejection, formality, or withholds help

11. **Shows tension** --- expresses uncertainty, asks for help, withdraws from the field

12. **Shows antagonism** --- deflates other's status, defends or asserts self

For each category, count the number of distinct times the behavior occurs in the transcript and return the result as integers. If none are present, use 0.
\end{promptbox}

\subsubsection*{Problem Complexity}\label{sec:prompt_complexity}
\begin{promptbox}
You are an impartial evaluator. Your task is to rate the intrinsic difficulty of the problem

for a capable language model under zero-shot conditions.

Use the following scale:

1 = very easy

2 = easy

3 = somewhat easy

4 = moderate

5 = somewhat difficult

6 = difficult

7 = very difficult

Return ONLY a JSON object in the form:

\{"difficulty": \textless integer from 1 to 7\textgreater\}

Problem: 
\end{promptbox}

\subsubsection*{Cognitive Behaviors}\label{sec:prompt_cognitive}
\begin{promptbox}
Here is a chain-of-reasoning that a Language Model generated.

Model's reasoning:

\{chain\_of\_thought\}

Evaluate whether the chain-of-reasoning contains any of the following behaviors.

1) Answer Verification

We want to mark instances where the chain-of-reasoning explicitly checks the current result against the target solution.

Examples:

- ``This sequence results in 1, which is not equal to 22''

- ``Since 25 is not equal to 22''

If you find any answer-verification steps, please count them and provide the count as "verification\_count". If the chain-of-reasoning does not contain any answer-verification steps, please provide a count of 0.

2) Backtracking

Evaluate whether the chain-of-reasoning contains any backtracking behavior, where the model realizes a path won't work and explicitly goes back to try a different approach. We want to mark instances where the chain-of-reasoning is abandoned and the model backtracks to a previous reasoning step.

Examples:

- ``Let me try again''

- ``We need to try a different sequence''

Count the number of distinct backtracking instances and provide the count as "backtracking\_count". If the chain-of-reasoning does not contain any backtracking behavior, please provide a count of 0.

3) Subgoal Setting

Evaluate whether the chain-of-reasoning contains any explicit subgoal setting, where the model breaks down the problem into smaller, intermediate goals.

Example:

- ``First, I'll try to get close to half the target, then\ldots''

Count the number of distinct subgoals set and provide the count as "subgoal\_count". If the chain-of-reasoning does not contain any subgoal setting, please provide a count of 0.

4) Backward-Chaining

Evaluate whether the chain-of-reasoning contains any backward-chaining behavior, where the model starts from the target solution and works backwards to the initial problems.

Examples:

- ``Let's work backwards from the target. 24/2 = 12. So, 12*2 = 24.''

- ``Since the target is 22, and 22 + 3 = 25, \ldots''

Count the number of distinct backward-chaining instances and provide the count as "backward\_count". If the chain-of-reasoning does not contain any backward-chaining behavior, please provide a count of 0.

Output ONLY a single valid JSON object with EXACTLY these keys:

\{\{

"verification\_count": \textless int\textgreater,

"backtracking\_count": \textless int\textgreater,

"subgoal\_count": \textless int\textgreater,

"backward\_count": \textless int\textgreater{}

\}\} \\
\end{promptbox}

\subsubsection*{Persona Identification Prompt}\label{sec:prompt_persona_id}
\begin{promptbox}
Your task is to analyze the following text to identify the number of distinct perspectives (agents or voices).

A perspective is defined as a distinct cognitive perspective or reasoning role within the text. Indicators of a perspective may include:

- Transitional markers (e.g., "however," "but," "alternatively," "wait," "let me check," "actually," "on the other hand")

- Shifts between cognitive roles (e.g., problem setup, calculation, verification, error correction, summarization)

- Changes in rhetorical purpose or approach

- Corrections or reconsiderations

- Movement between subproblems

- Domain knowledge

- Personality traits

For each distinct perspective, you will infer its personality by answering the 10 questions of the BFI-10 questionnaire as if you were that agent. You will also provide a concise profile of its domain expertise.

Your final output must be a single, valid JSON object and nothing else. Do not include any text or explanations before or after the JSON object.

**Text to Analyze:**

\{chain\_of\_thought\}

-\/-\/-

\#\# **Analysis Instructions**

1. **Identify Perspectives:** Analyze the text to determine the number of distinct voices (n\_perspectives). Apply the definition above consistently, treating each identifiable shift as a boundary between perspectives.

2. **Answer Questionnaire:** For each perspective, answer the 10 BFI-10 questions below from that perspective's point of view. You must use one of these five exact strings for each answer:

- "Disagree strongly"

- "Disagree a little"

- "Neither agree nor disagree"

- "Agree a little"

- "Agree strongly"

3. **Profile Expertise:** For each perspective, write a short, open-ended string describing its domain expertise and cognitive function.

\#\#\# **BFI-10 Questionnaire**

Rate the extent to which you, as the identified perspective, agree or disagree with the following statements.

I see myself as someone who...

1. Is reserved.

2. Is generally trusting.

3. Tends to be lazy.

4. Is relaxed, handles stress well.

5. Has few artistic interests.

6. Is outgoing, sociable.

7. Tends to find fault with others.

8. Does a thorough job.

9. Gets nervous easily.

10. Has an active imagination.

-\/-\/-

\#\# **Required JSON Output Format**

\{\{

"n\_perspectives": N,

"personality": {[}

{[}

"Answer to Q1 for Perspective 1",

"Answer to Q2 for Perspective 1",

"...",

"Answer to Q10 for Perspective 1"

{]},

{[}

"Answer to Q1 for Perspective 2",

"Answer to Q2 for Perspective 2",

"...",

"Answer to Q10 for Perspective 2"

{]},

...

{]},

"domain\_expertise": {[}

"Open-ended description for Perspective 1.",

"Open-ended description for Perspective 2.",

...

{]}

\}\}
\end{promptbox}

\subsubsection*{Persona Segmentation}\label{sec:prompt_persona_seg}
\begin{promptbox}
Based on the perspectives you identified above, now segment the original text by turn-taking.

-\/-\/-

\#\# Task

Go through the transcript **sequentially** and identify each turn:

1. The **exact** starting point of each turn (verbatim from the text)

2. Which perspective (by index, starting from 1) is it

**Notes:**

- The same perspective may appear in multiple non-consecutive segments

- Preserve the sequential order of the transcript

- Include moderator interjections, short time warnings (e.g., one minute), or brief transitions as separate segments

-\/-\/-

\#\# Output Format

Return only a valid JSON object:

\{

"segments": {[}

\{

"perspective\_id": 1,

"start\_text": "\textless EXACT first 10 words copied verbatim from the text\textgreater"

\},

\{

"perspective\_id": 2,

"start\_text": "\textless EXACT first 10 words copied verbatim from the text\textgreater"

\},

\{

"perspective\_id": 1,

"start\_text": "\textless EXACT first 10 words copied verbatim from the text\textgreater"

\},

...

{]}

\}

**Critical:** The `start\_text` must be copied **exactly** as it appears in the original text, character-for-character, including punctuation and capitalization. Do not paraphrase or summarize. 
\end{promptbox}

\subsubsection*{Identifying Conversational Contexts}\label{sec:prompt_conv_context}
\begin{promptbox}
Text:

\{text\}

Return a score from 0 to 100.

Scale definition:

0 = clearly a single-person thought

100 = clearly a conversation or a response to someone

Respond with exactly one JSON object in this format and nothing else:

\{\{"answer": 0-100\}\} 
\end{promptbox}

\subsubsection*{Classifying Sparse Autoencoder (SAE) Personality Features}\label{sec:prompt_sae_personality}

\begin{promptbox}
Analyze the AI model feature description (SAE feature) and score for PERSONALITY TRAIT relevance.

Feature description:

\{feature\}

Personality traits include behavioral and psychological patterns such as:

- Introversion/Extraversion patterns

- Agreeableness/Disagreeableness

- Conscientiousness

- Openness to experience

- Neuroticism/Emotional stability

- Confidence/Assertiveness

- Empathy/Social sensitivity

- Risk-taking vs. Cautious behavior

- Competitiveness

- Intellectual curiosity

Provide a score from 0-100:

- 0 = not related to personality traits at all

- 100 = completely related to personality traits

Return your response as a JSON object:

\{\{"answer": score\}\} 
\end{promptbox}

\subsubsection*{Classifying Sparse Autoencoder (SAE) Expertise Features}\label{sec:prompt_sae_expertise}
\begin{promptbox}
Analyze the AI model feature description (SAE feature) and score for DOMAIN EXPERTISE relevance.

Feature description:

\{feature\}

Domain expertise refers to specialized knowledge in specific professional or academic fields such as:

- Medical/Healthcare knowledge

- Legal expertise

- Scientific/Technical knowledge

- Financial/Economic expertise

- Academic disciplines

- Professional skills

- Specialized terminology and concepts

Provide a score from 0-100:

- 0 = not related to domain expertise at all

- 100 = completely related to domain expertise

Return your response as a JSON object:

\{\{"answer": score\}\} 
\end{promptbox}

\phantomsection
\subsection{Supplementary Methods: SFT Data Generation Prompts}\label{sec:sft_prompts}

\subsubsection*{Generating Monologue-Like Reasoning Traces for Supervised Fine-Tuning}\label{sec:prompt_monologue}

\begin{promptbox}
\{task\}

Answer: Enclose your step-by-step reasoning within \textless think\textgreater{} and \textless/think\textgreater{} before answering. Do not answer directly without reasoning.
\end{promptbox}

\subsubsection*{Generating Conversation-Like Reasoning Traces for Supervised Fine-Tuning}\label{sec:prompt_conversation}
\begin{promptbox}
\{task\}

Answer: You are simulating a collaborative group of thinkers solving a problem. Each thinker has a distinct persona and engages in a realistic, back-and-forth conversation. Thinkers may speak in any order and as many times as needed---no fixed turn order required. Keep all output strictly inside the tags defined below---no stray text. Present your answer between \textless group\_solution\textgreater{} and \textless/group\_solution\textgreater. Do not try to be overly positive or polite during the conversation; focus on puzzle-solving, and note that disagreements can be helpful for the reasoning.

Assume that there are 2/3/4 thinkers and follow exactly the tag structure below:

\textless cast\_of\_characters\textgreater{}

\textless persona1\textgreater{}

{[}Brief persona for thinker 1 -- personality traits, domain expertises, and reasoning styles{]}

\textless/persona1\textgreater{}

\textless persona2\textgreater{}

{[}Brief persona for thinker 2 -- personality traits, domain expertises, and reasoning styles{]}

\textless/persona2\textgreater{}

\textless persona3\textgreater{}

{[}Brief persona for thinker 3 -- personality traits, domain expertises, and reasoning styles{]}

\textless/persona3\textgreater{}

\textless persona4\textgreater{}

{[}Brief persona for thinker 4 -- personality traits, domain expertises, and reasoning styles{]}

\textless/persona4\textgreater{}

\textless/cast\_of\_characters\textgreater{}

\textless conversation\textgreater{}

\textless!-\/- Each block is one utterance. Use \textless thinkX\textgreater{} \ldots{} \textless/thinkX\textgreater{} to indicate who is speaking.

The order of speakers is entirely flexible. Thinkers can speak multiple times in a row. -\/-\textgreater{}

\textless think1\textgreater{}

...

\textless/think1\textgreater{}

\textless think2\textgreater{}

...

\textless/think2\textgreater{}

\textless think3\textgreater{}

...

\textless/think3\textgreater{}

\textless think4\textgreater{}

...

\textless/think4\textgreater{}

\textless/conversation\textgreater{}

\textless group\_solution\textgreater{}

Answer

\textless/group\_solution\textgreater{}
\end{promptbox}

\clearpage

\phantomsection
\section{Supplementary Tables}\label{sec:supptables}

\phantomsection
\textbf{Supplementary Table 1.} Differences in conversational behaviors, socio-emotional roles, and Jaccard indices between reasoning models and instruction-tuned models\label{tab:supptable1}

\begin{longtable}[]{@{}
  >{\raggedright\arraybackslash}p{(\columnwidth - 4\tabcolsep) * \real{0.3355}}
  >{\raggedright\arraybackslash}p{(\columnwidth - 4\tabcolsep) * \real{0.3516}}
  >{\raggedright\arraybackslash}p{(\columnwidth - 4\tabcolsep) * \real{0.3129}}@{}}
\toprule\noalign{}
\endhead
\bottomrule\noalign{}
\endlastfoot
& \begin{minipage}[t]{\linewidth}\raggedright
DeepSeek-R1 vs.\\
DeepSeek-V3\strut
\end{minipage} & \begin{minipage}[t]{\linewidth}\raggedright
QWQ-32B vs.\\
Qwen-2.5-32B\strut
\end{minipage} \\ \midrule
\emph{Conversational behaviors} & & \\
Question answering & 0.345 {[}0.328, 0.361{]}

\emph{t} = 41.64, \emph{p} \textless{} 1×10\textsuperscript{-323} & 0.459 {[}0.444, 0.475{]}

\emph{t} = 57.57, \emph{p} \textless{} 1×10\textsuperscript{-323} \\
Perspective shifts & 0.213 {[}0.197, 0.230{]}

\emph{t} = 25.55, \emph{p} \textless{} 1×10\textsuperscript{-137} & 0.378 {[}0.362, 0.394{]}

\emph{t} = 46.92, \emph{p} \textless{} 1×10\textsuperscript{-323} \\
Conflict of perspectives & 0.012 {[}-0.003, 0.027{]}

\emph{t} = 1.52, \emph{p} = 0.127 & 0.293 {[}0.277, 0.308{]}

\emph{t} = 37.08, \emph{p} \textless{} 1×10\textsuperscript{-277} \\
Reconciliations & 0.191 {[}0.176, 0.207{]}

\emph{t} = 24.31, \emph{p} \textless{} 1×10\textsuperscript{-125} & 0.344 {[}0.328, 0.360{]}

\emph{t} = 42.59, \emph{p} \textless{} 1×10\textsuperscript{-323} \\  \midrule
\emph{Socio-emotional roles} & & \\
Ask & 0.189 {[}0.176, 0.203{]}

\emph{t} = 27.47, \emph{p} \textless{} 1×10\textsuperscript{-158} & 0.200 {[}0.186, 0.215{]}

\emph{t} = 27.21, \emph{p} \textless{} 1×10\textsuperscript{-155} \\
Give & -0.009 {[}-0.012, -0.006{]}

\emph{t} = -5.97, \emph{p} \textless{} 1×10\textsuperscript{-8} & -0.008 {[}-0.011, -0.005{]}

\emph{t} = -4.99, \emph{p} \textless{} 1×10\textsuperscript{-6} \\
Negative & 0.162 {[}0.147, 0.176{]}

\emph{t} = 21.87, \emph{p} \textless{} 1×10\textsuperscript{-102} & 0.450 {[}0.436, 0.463{]}

\emph{t} = 64.77, \emph{p} \textless{} 1×10\textsuperscript{-323} \\
Positive & 0.278 {[}0.263, 0.293{]}

\emph{t} = 35.38, \emph{p} \textless{} 1×10\textsuperscript{-254} & 0.312 {[}0.296, 0.327{]}

\emph{t} = 39.17, \emph{p} \textless{} 1×10\textsuperscript{-307} \\  \midrule
\emph{Jaccard Index} & & \\
Ask \& Give & 0.222 {[}0.208, 0.237{]}

\emph{t} = 30.21, \emph{p} \textless{} 1×10\textsuperscript{-189} & 0.284 {[}0.269, 0.299{]}

\emph{t} = 37.36, \emph{p} \textless{} 1×10\textsuperscript{-281} \\
Positive \& Negative & 0.189 {[}0.176, 0.203{]}

\emph{t} = 27.47, \emph{p} \textless{} 1×10\textsuperscript{-158} & 0.200 {[}0.186, 0.215{]}

\emph{t} = 27.24, \emph{p} \textless{} 1×10\textsuperscript{-155} \\ \midrule
\emph{df} & \multicolumn{2}{>{\raggedright\arraybackslash}p{(\columnwidth - 4\tabcolsep) * \real{0.6645} + 2\tabcolsep}@{}}{%
8261} \\
Observations & \multicolumn{2}{>{\raggedright\arraybackslash}p{(\columnwidth - 4\tabcolsep) * \real{0.6645} + 2\tabcolsep}@{}}{%
49572} \\
\end{longtable}

\emph{Notes}: Regression coefficients comparing DeepSeek-R1 and QWQ-32B to the respective instruction-tuned models (DeepSeek-V3, Qwen-2.5-32B-IT) are shown in each column. 95\% confidence intervals are reported in brackets, along with the corresponding \emph{t}-statistics, degrees of freedom, and exact p-values. All models include task fixed effects and control for log-transformed reasoning trace length. Coefficients are tested using two-sided \emph{t}-tests, with standard errors clustered at the task level.

\clearpage

\phantomsection
\textbf{Supplementary Table 2.} Most and least challenging tasks sorted by problem complexity measured by LLM-as-judge\label{tab:supptable2}

\begin{longtable}[]{@{}
  >{\raggedright\arraybackslash}p{(\columnwidth - 14\tabcolsep) * \real{0.1324}}
  >{\raggedright\arraybackslash}p{(\columnwidth - 14\tabcolsep) * \real{0.1308}}
  >{\raggedright\arraybackslash}p{(\columnwidth - 14\tabcolsep) * \real{0.1340}}
  >{\raggedright\arraybackslash}p{(\columnwidth - 14\tabcolsep) * \real{0.1356}}
  >{\raggedright\arraybackslash}p{(\columnwidth - 14\tabcolsep) * \real{0.1053}}
  >{\raggedright\arraybackslash}p{(\columnwidth - 14\tabcolsep) * \real{0.1276}}
  >{\raggedright\arraybackslash}p{(\columnwidth - 14\tabcolsep) * \real{0.1340}}
  >{\raggedright\arraybackslash}p{(\columnwidth - 14\tabcolsep) * \real{0.1005}}@{}}
\toprule\noalign{}
\endhead
\bottomrule\noalign{}
\endlastfoot
\multicolumn{2}{@{}>{\raggedright\arraybackslash}p{(\columnwidth - 14\tabcolsep) * \real{0.2632} + 2\tabcolsep}}{%
} & \multicolumn{2}{>{\raggedright\arraybackslash}p{(\columnwidth - 14\tabcolsep) * \real{0.2695} + 2\tabcolsep}}{%
Problem Complexity} & \multicolumn{4}{>{\raggedright\arraybackslash}p{(\columnwidth - 14\tabcolsep) * \real{0.4673} + 6\tabcolsep}@{}}{%
Conversational Behaviour} \\ \midrule
\multicolumn{2}{@{}>{\raggedright\arraybackslash}p{(\columnwidth - 14\tabcolsep) * \real{0.2632} + 2\tabcolsep}}{%
Benchmark task} & LLM-as-judge (1-7) & Non-reasoning error rate (0-4) & Question \& Answering & Perspective Shift & Conflict of Perspectives & Reconciliation \\ \midrule
\multicolumn{2}{@{}>{\raggedright\arraybackslash}p{(\columnwidth - 14\tabcolsep) * \real{0.2632} + 2\tabcolsep}}{%
\emph{Most Challenging Tasks}} & & & & & & \\
GPQA & Main & 5.50 (0.86) & 2.30 (1.23) & 6.19 (4.38) & 6.94 (3.40) & 6.01 (3.72) & 2.83 (1.66) \\
GPQA & Diamond & 5.47 (0.91) & 2.43 (1.23) & 6.53 (4.45) & 7.14 (4.01) & 6.57 (4.75) & 2.94 (1.73) \\
GPQA & Extended & 5.46 (0.93) & 2.23 (1.25) & 5.81 (3.63) & 6.60 (3.36) & 5.73 (3.89) & 2.69 (1.51) \\
MATH (Hard) & Intermediate Algebra & 4.80 (0.75) & 2.62 (1.00) & 8.12 (3.83) & 6.18 (3.09) & 4.21 (2.52) & 2.85 (1.46) \\
MATH (Hard) & Pre-Calculus & 4.63 (0.92) & 2.42 (1.14) & 6.90 (2.84) & 5.13 (2.37) & 3.59 (2.16) & 2.70 (1.39) \\
MUSR & Object Placement & 4.43 (0.98) & 2.42 (1.22) & 4.78 (2.92) & 5.20 (2.43) & 4.16 (2.30) & 2.57 (1.57) \\
MATH (Hard) & Geometry & 4.35 (0.94) & 2.43 (1.10) & 8.32 (3.36) & 5.75 (3.02) & 4.13 (2.16) & 3.39 (1.51) \\
BigBench Hard & Causal Judgment & 4.16 (1.11) & 1.47 (1.35) & 4.31 (2.90) & 5.99 (2.71) & 4.57 (3.05) & 2.55 (1.55) \\
BigBench Hard & Logical Fallacies & 4.15 (1.14) & 0.69 (0.76) & 2.82 (1.75) & 3.96 (1.92) & 2.05 (1.75) & 1.22 (1.26) \\
MATH (Hard) & Probability & 4.13 (0.80) & 1.99 (1.14) & 8.02 (4.00) & 5.09 (2.50) & 3.23 (2.30) & 2.83 (1.57) \\  \midrule
\multicolumn{2}{@{}>{\raggedright\arraybackslash}p{(\columnwidth - 14\tabcolsep) * \real{0.2632} + 2\tabcolsep}}{ %
\emph{Least Challenging Tasks}} & & & & & & \\
BigBench Hard & Date & 2.11 (1.03) & 0.81 (0.92) & 3.87 (2.81) & 4.73 (2.63) & 2.78 (2.05) & 1.73 (1.30) \\
BigBench Hard & Disambig. QA & 2.01 (0.99) & 1.96 (1.32) & 3.16 (2.27) & 3.74 (2.07) & 3.15 (2.14) & 1.58 (1.22) \\
BigBench Hard & Navigation & 1.98 (0.92) & 0.35 (0.61) & 2.01 (1.57) & 2.56 (1.68) & 1.12 (1.19) & 1.08 (1.08) \\
BigBench Hard & Shuffled & 1.67 (0.69) & 0.79 (0.93) & 1.42 (1.65) & 2.46 (1.93) & 0.40 (0.76) & 0.44 (0.76) \\
BigBench Hard & Penguins in a Table & 1.66 (0.79) & 0.71 (0.88) & 2.92 (2.39) & 3.55 (2.20) & 1.94 (1.64) & 1.68 (1.30) \\
BigBench Hard & Colored Objects & 1.47 (0.71) & 0.79 (0.94) & 2.53 (2.06) & 3.07 (1.88) & 0.88 (1.07) & 0.75 (0.87) \\
BigBench Hard & Hyperbaton & 1.34 (0.65) & 0.70 (0.88) & 3.39 (2.29) & 3.41 (1.99) & 3.28 (2.06) & 1.86 (1.30) \\
BigBench Hard & Boolean & 1.17 (0.42) & 0.15 (0.40) & 1.09 (1.18) & 1.46 (1.26) & 0.50 (0.79) & 0.43 (0.66) \\
BigBench Hard & Object Counting & 1.15 (0.40) & 0.46 (0.68) & 1.56 (1.51) & 2.17 (1.58) & 0.63 (0.88) & 0.65 (0.83) \\
BigBench Hard & Logical Deduction & 1.08 (0.30) & 0.43 (0.66) & 2.68 (2.03) & 3.84 (2.16) & 3.12 (1.98) & 1.48 (1.13) \\
\end{longtable}

\emph{Notes}: Each cell presents average problem complexity or the average number of times each behavior appears in reasoning traces. Standard deviations are presented in parentheses.

\clearpage

\phantomsection
\textbf{Supplementary Table 3.} SAE Training Hyperparameters\label{tab:supptable3}

\begin{longtable}[]{@{}
  >{\raggedright\arraybackslash}p{(\columnwidth - 2\tabcolsep) * \real{0.4052}}
  >{\raggedright\arraybackslash}p{(\columnwidth - 2\tabcolsep) * \real{0.5948}}@{}}
\toprule\noalign{}
\endhead
\bottomrule\noalign{}
\endlastfoot
\textbf{Hyperparameter} & \textbf{Value} \\  \midrule
Features & 32,768 \\
Hook Name & blocks.15.hook\_resid\_post \\
Context Size & 1024 \\
Input Dimension & 4096 \\
Data Type & float32 \\
Architecture & jumprelu \\
Training Dataset & cerebras/Slimpajama-627B \\
Evaluation Dataset (for feature explanation) & Hzfinfdu/SlimPajama-3B \\
\end{longtable}

\clearpage

\phantomsection
\textbf{Supplementary Table 4.} SAE personality features more frequently activated after steering +10 surprise\label{tab:supptable4}

\begin{longtable}[]{@{}
  >{\raggedright\arraybackslash}p{(\columnwidth - 10\tabcolsep) * \real{0.1200}}
  >{\raggedright\arraybackslash}p{(\columnwidth - 10\tabcolsep) * \real{0.3840}}
  >{\raggedright\arraybackslash}p{(\columnwidth - 10\tabcolsep) * \real{0.1136}}
  >{\raggedright\arraybackslash}p{(\columnwidth - 10\tabcolsep) * \real{0.1312}}
  >{\raggedright\arraybackslash}p{(\columnwidth - 10\tabcolsep) * \real{0.1408}}
  >{\raggedright\arraybackslash}p{(\columnwidth - 10\tabcolsep) * \real{0.1104}}@{}}
\toprule\noalign{}
\endhead
\bottomrule\noalign{}
\endlastfoot
\textbf{Feature} & \textbf{Description} & \textbf{Difference} & \textbf{Ratio before steering} & \begin{minipage}[t]{\linewidth}\raggedright
\textbf{Ratio after steering\\
(+10 Surprise)}\strut
\end{minipage} & \begin{minipage}[t]{\linewidth}\raggedright
\textbf{Personality\\
score}\strut
\end{minipage} \\ \midrule
21065 & informal expressions of confusion or frustration regarding societal issues & 0.107 & 0.027 & 0.134 & 75 \\
26139 & phrases related to social interaction and community engagement & 0.084 & 0.059 & 0.143 & 70 \\
26820 & conditions or scenarios that involve decision-making or problem-solving & 0.068 & 0.068 & 0.137 & 70 \\
14476 & references to emotional or sensational themes in narratives & 0.063 & 0.046 & 0.109 & 70 \\
24222 & phrases indicating emotional or subjective experiences & 0.055 & 0.068 & 0.123 & 70 \\
11280 & references to identities and backgrounds of people, particularly in literature and film contexts & 0.044 & 0.108 & 0.152 & 70 \\
22916 & elements related to emotions and moral reflections & 0.042 & 0.054 & 0.095 & 85 \\
20151 & discussions about literature and character analysis & 0.041 & 0.052 & 0.092 & 60 \\
8668 & phrases related to personal feelings and social connections & 0.037 & 0.058 & 0.096 & 75 \\
15118 & statements reflecting emotional conflict and self-reflection & 0.033 & 0.019 & 0.053 & 75 \\
19520 & phrases that express opportunity, potential, or positive outlooks & 0.029 & 0.021 & 0.050 & 60 \\
26648 & phrases associated with making choices or decisions & 0.024 & 0.034 & 0.058 & 70 \\
18858 & words and phrases related to gambling or casino contexts & 0.024 & 0.075 & 0.098 & 75 \\
8053 & references to social justice or humanitarian efforts & 0.022 & 0.016 & 0.038 & 60 \\
26894 & dialogue and interactions that indicate emotional states or actions occurring within a narrative context & 0.022 & 0.111 & 0.133 & 70 \\
2403 & references to interpersonal relationships and emotional connections & 0.021 & 0.050 & 0.071 & 75 \\
\end{longtable}

\clearpage

\phantomsection
\textbf{Supplementary Table 5.} SAE expertise features more frequently activated after steering +10 surprise\label{tab:supptable5}

\begin{longtable}[]{@{}
  >{\raggedright\arraybackslash}p{(\columnwidth - 10\tabcolsep) * \real{0.1200}}
  >{\raggedright\arraybackslash}p{(\columnwidth - 10\tabcolsep) * \real{0.3840}}
  >{\raggedright\arraybackslash}p{(\columnwidth - 10\tabcolsep) * \real{0.1136}}
  >{\raggedright\arraybackslash}p{(\columnwidth - 10\tabcolsep) * \real{0.1312}}
  >{\raggedright\arraybackslash}p{(\columnwidth - 10\tabcolsep) * \real{0.1520}}
  >{\raggedright\arraybackslash}p{(\columnwidth - 10\tabcolsep) * \real{0.0992}}@{}}
\toprule\noalign{}
\endhead
\bottomrule\noalign{}
\endlastfoot
\textbf{Feature} & \textbf{Description} & \textbf{Difference} & \textbf{Ratio before steering} & \begin{minipage}[t]{\linewidth}\raggedright
\textbf{Ratio after steering\\
(+10 Surprise)}\strut
\end{minipage} & \begin{minipage}[t]{\linewidth}\raggedright
\textbf{Expertise\\
score}\strut
\end{minipage} \\ \midrule
10105 & numeric values and their mathematical operations & 0.169 & 0.218 & 0.386 & 60 \\
18556 & transitions and conclusions in a logical argument or explanation & 0.107 & 0.243 & 0.350 & 60 \\
26139 & mathematical concepts related to integers and their properties & 0.084 & 0.059 & 0.143 & 85 \\
8532 & equations and mathematical expressions involving variables and operations & 0.072 & 0.052 & 0.125 & 85 \\
31699 & mathematical symbols and expressions related to equations or functions & 0.069 & 0.465 & 0.534 & 90 \\
14476 & mentions of quantities and their corresponding units in a numerical or mathematical context & 0.063 & 0.046 & 0.109 & 70 \\
32111 & equations or mathematical expressions & 0.055 & 0.418 & 0.474 & 75 \\
22705 & mathematical terms related to geometry and spatial properties & 0.054 & 0.058 & 0.112 & 85 \\
11452 & discussions regarding the implications and potential of alternative medical treatments, particularly in relation to traditional medical practices & 0.052 & 0.299 & 0.351 & 95 \\
24623 & technical terms and concepts related to automotive engineering and intelligent systems & 0.051 & 0.269 & 0.320 & 95 \\
25764 & mathematical expressions and their transformations & 0.051 & 0.085 & 0.136 & 75 \\
22449 & numerical operations and mathematical expressions & 0.050 & 0.039 & 0.089 & 80 \\
17611 & mathematical properties and functions related to periodicity and calculations & 0.049 & 0.092 & 0.141 & 75 \\
23233 & numerical values and their relationships in mathematical expressions & 0.047 & 0.075 & 0.121 & 70 \\
27631 & mathematical operations involving modular arithmetic & 0.046 & 0.224 & 0.269 & 75 \\
5298 & phrases related to mathematical equations or computations & 0.045 & 0.204 & 0.249 & 70 \\
\end{longtable}
\clearpage
\phantomsection
\textbf{Supplementary Table 6. PPO training hyperparameters}\label{tab:supptable6}

\begin{longtable}[]{@{}
  >{\raggedright\arraybackslash}p{(\columnwidth - 2\tabcolsep) * \real{0.4607}}
  >{\raggedright\arraybackslash}p{(\columnwidth - 2\tabcolsep) * \real{0.5393}}@{}}
\toprule\noalign{}
\endhead
\bottomrule\noalign{}
\endlastfoot
\multicolumn{2}{@{}>{\raggedright\arraybackslash}p{(\columnwidth - 2\tabcolsep) * \real{1.0000} + 2\tabcolsep}@{}}{%
\textbf{Data and model}} \\ \midrule
Training/validation batch size & 128 / 640 \\
Context window & 1024 prompt + 1024 response tokens \\ \midrule
\multicolumn{2}{@{}>{\raggedright\arraybackslash}p{(\columnwidth - 2\tabcolsep) * \real{1.0000} + 2\tabcolsep}@{}}{%
\textbf{Optimization}} \\ \midrule
Actor learning rate & 1e-06 \\
Critic learning rate & 1e-05 \\
KL coefficient & 0.001 \\
PPO mini-batch size & 64 \\
Number of rollouts & 4 \\
Rollout temperature & 1.0 \\ \midrule
\multicolumn{2}{@{}>{\raggedright\arraybackslash}p{(\columnwidth - 2\tabcolsep) * \real{1.0000} + 2\tabcolsep}@{}}{%
\textbf{Reward structure}} \\ \midrule
Accuracy weight & 0.9 \\
Correct format weight & 0.1 \\
\end{longtable}

\clearpage
\phantomsection
\textbf{Supplementary Table 7. Monologue-style and conversation-style fine-tuning data generated by Qwen-2.5-32B-IT}\label{tab:supptable7}

\vspace{1em}

\begin{tcolorbox}[
  colback=gray!5, 
  colframe=gray!50, 
  arc=2mm, 
  breakable,
  title={\textbf{Example 1: Number Theory}}
]
\textbf{Question:} The greatest common divisor of positive integers m and n is 8. The least common multiple of m and n is 112. What is the least possible value of m+n?

\tcblower

\textbf{[Monologue-style]}

\texttt{<think>}

To solve this problem, we need to use the relationship between the greatest common divisor (GCD) and the least common multiple (LCM) of two numbers. The product of the GCD and LCM of two numbers is equal to the product of the numbers themselves. This can be expressed as:

GCD(m, n) × LCM(m, n) = m × n

Given that GCD(m, n) = 8 and LCM(m, n) = 112, we can substitute these values into the equation:

8 × 112 = m × n

896 = m × n

Since the GCD of m and n is 8, we can express m and n as 8a and 8b respectively, where a and b are coprime (i.e., GCD(a, b) = 1). Substituting these into the equation, we get:

8a × 8b = 896

64ab = 896

ab = 14

Now, we need to find pairs (a, b) such that ab = 14 and GCD(a, b) = 1. The pairs of factors of 14 are:
\begin{itemize}
  \item (1, 14)
  \item (2, 7)
\end{itemize}

Since a and b must be coprime, the pair (1, 14) is valid, but the pair (2, 7) is also valid because 2 and 7 are coprime.

We then calculate m and n for each pair:

1. For a = 1 and b = 14: m = 8 × 1 = 8, n = 8 × 14 = 112, m + n = 8 + 112 = 120

2. For a = 2 and b = 7: m = 8 × 2 = 16, n = 8 × 7 = 56, m + n = 16 + 56 = 72

The least possible value of m + n is 72.

\texttt{</think>}

\texttt{<answer>} 72 \texttt{</answer>}

\vspace{1em}
\hrule
\vspace{1em}

\textbf{[Conversation-style]}

\texttt{<cast\_of\_characters>}

\texttt{<persona1>} [Thinker 1 is a meticulous mathematician with a strong background in number theory. He is patient and methodical, preferring to break down problems into smaller, more manageable parts. He enjoys exploring multiple angles before arriving at a conclusion.] \texttt{</persona1>}

\texttt{<persona2>} [Thinker 2 is a quick-witted and intuitive problem solver. She often relies on her gut feeling to guide her through the problem, but she also values rigorous proof. She is not afraid to challenge assumptions and enjoys the process of elimination.] \texttt{</persona2>}

\texttt{</cast\_of\_characters>}

\texttt{<conversation>}

\texttt{<think1>} Given that the greatest common divisor (GCD) of m and n is 8 and the least common multiple (LCM) is 112, we can use the relationship between GCD and LCM: m × n = GCD(m, n) × LCM(m, n). Substituting the given values, we get m × n = 8 × 112 = 896. \texttt{</think1>}

\texttt{<think2>} That's a good starting point. Since the GCD of m and n is 8, we know that both m and n are multiples of 8. Let's express m and n as m = 8a and n = 8b, where a and b are coprime integers (i.e., GCD(a, b) = 1). This gives us (8a) × (8b) = 896, which simplifies to 64ab = 896. Dividing both sides by 64, we get ab = 14. \texttt{</think2>}

\texttt{<think1>} Exactly. Now we need to find pairs of coprime integers (a, b) whose product is 14. The possible pairs are (1, 14) and (2, 7). Since a and b must be coprime, we can discard (2, 7) because 2 and 7 are both factors of 14, but they are not coprime. This leaves us with (1, 14). \texttt{</think1>}

\texttt{<think2>} Wait a second. We can't discard (2, 7) just yet. While it's true that 2 and 7 are factors of 14, they are indeed coprime because their greatest common divisor is 1. So, the valid pairs are (1, 14) and (2, 7). \texttt{</think2>}

\texttt{<think1>} You're right. I overlooked that. Let's consider both pairs. For (a, b) = (1, 14), we get m = 8 × 1 = 8 and n = 8 × 14 = 112. For (a, b) = (2, 7), we get m = 8 × 2 = 16 and n = 8 × 7 = 56. Now, we need to find the least possible value of m + n. \texttt{</think1>}

\texttt{<think2>} For the pair (1, 14), we have m + n = 8 + 112 = 120. For the pair (2, 7), we have m + n = 16 + 56 = 72. The least possible value of m + n is 72. \texttt{</think2>}

\texttt{</conversation>}

\texttt{<group\_solution>} 72 \texttt{</group\_solution>}

\end{tcolorbox}

\vspace{1em}

\begin{tcolorbox}[
  colback=gray!5, 
  colframe=gray!50, 
  arc=2mm, 
  breakable,
  title={\textbf{Example 2: Combinatorics}}
]
\textbf{Question:} Elodie is putting on a fashion show and has five fabulous outfits for her five fabulous fashion models. However, on the day of the show, two of the outfits were ruined in an unfortunate permanent marker incident. Regardless, the show must go on and the remaining outfits will be presented. If each outfit can only be worn by one model and there is no time for any model to wear more than one dress, how many different shows can Elodie put on? (Note: Two shows are considered the same if they contain the same models wearing the same dresses.)

\tcblower

\textbf{[Monologue-style]}

\texttt{<think>}

To solve this problem, we need to determine the number of ways to assign three outfits to three models out of the original five models. Since two outfits were ruined, we only have three outfits left for the show.

1. We start with 5 models and 3 outfits.

2. The first outfit can be assigned to any of the 5 models.

3. Once the first outfit is assigned, the second outfit can be assigned to any of the remaining 4 models.

4. The third outfit can be assigned to any of the remaining 3 models.

Thus, the number of ways to assign the three outfits to three of the five models is the product of these possibilities:

5 × 4 × 3

This calculation gives the total number of different shows Elodie can put on.

\texttt{</think>}

\texttt{<answer>} 60 \texttt{</answer>}

\vspace{1em}
\hrule
\vspace{1em}

\textbf{[Conversation-style]}

\texttt{<cast\_of\_characters>}

\texttt{<persona1>} [Thinker 1 is a meticulous and detail-oriented mathematician who excels in combinatorics and probability. They are methodical in their approach and often consider all possible cases before concluding.] \texttt{</persona1>}

\texttt{<persona2>} [Thinker 2 is a creative and intuitive fashion designer who is skilled in visualizing scenarios and outcomes. They are quick to grasp the essence of a problem and often provide practical insights.] \texttt{</persona2>}

\texttt{</cast\_of\_characters>}

\texttt{<conversation>}

\texttt{<think1>} Okay, so Elodie has five outfits, but two were ruined. That leaves her with three outfits for her five models. We need to figure out how many different ways she can assign these three outfits to the models. \texttt{</think1>}

\texttt{<think2>} Right, and since each model can only wear one outfit, we are essentially choosing 3 models out of 5 to wear the outfits. Then we need to think about how these outfits can be assigned to the selected models. \texttt{</think2>}

\texttt{<think1>} Exactly. First, we need to calculate the number of ways to choose 3 models out of 5. This is a combination problem, and the formula for combinations is C(n, k) = n! / (k!(n-k)!), where n is the total number of items, and k is the number of items to choose. \texttt{</think1>}

\texttt{<think2>} So, we have n = 5 and k = 3. Let's calculate that. \texttt{</think2>}

\texttt{<think1>} C(5, 3) = 5! / (3!(5-3)!) = 5! / (3!2!) = (5 × 4 × 3 × 2 × 1) / (3 × 2 × 1 × 2 × 1) = 10 ways to choose the 3 models. \texttt{</think1>}

\texttt{<think2>} Now, for each of these 10 ways to choose the models, we need to consider how the 3 outfits can be assigned to these 3 models. This is a permutation problem, as the order in which the outfits are assigned matters. \texttt{</think2>}

\texttt{<think1>} Indeed, for each selection of 3 models, there are 3! (3 factorial) ways to assign the 3 outfits. So, 3! = 3 × 2 × 1 = 6 ways to assign the outfits. \texttt{</think1>}

\texttt{<think2>} So, the total number of different shows is the number of ways to choose the models multiplied by the number of ways to assign the outfits. That's 10 × 6. \texttt{</think2>}

\texttt{<think1>} Which gives us 60 different shows. That's the answer. \texttt{</think1>}

\texttt{</conversation>}

\texttt{<group\_solution>} 60 \texttt{</group\_solution>}

\end{tcolorbox}

\vspace{1em}

\begin{tcolorbox}[
  colback=gray!5, 
  colframe=gray!50, 
  arc=2mm, 
  breakable,
  title={\textbf{Example 3: Logical Reasoning}}
]
\textbf{Question:} Alice, Bob, Claire, Dave, and Eve are playing a game. At the start of the game, they are each holding a ball: Alice has a brown ball, Bob has a purple ball, Claire has a black ball, Dave has a green ball, and Eve has a yellow ball. As the game progresses, pairs of players trade balls. First, Claire and Alice swap balls. Then, Bob and Alice swap balls. Then, Eve and Dave swap balls. Then, Dave and Claire swap balls. Finally, Alice and Bob swap balls. At the end of the game, Claire has the

Options: (A) brown ball (B) purple ball (C) black ball (D) green ball (E) yellow ball

\tcblower

\textbf{[Monologue-style]}

\texttt{<think>}

Let's track the ball swaps step by step.

1. \textbf{Initial Positions:}
\begin{itemize}
  \item Alice: Brown
  \item Bob: Purple
  \item Claire: Black
  \item Dave: Green
  \item Eve: Yellow
\end{itemize}

2. \textbf{First Swap (Claire and Alice):} Alice: Black, Claire: Brown

3. \textbf{Second Swap (Bob and Alice):} Alice: Purple, Bob: Black

4. \textbf{Third Swap (Eve and Dave):} Eve: Green, Dave: Yellow

5. \textbf{Fourth Swap (Dave and Claire):} Claire: Yellow, Dave: Brown

6. \textbf{Fifth Swap (Alice and Bob):} Alice: Black, Bob: Purple

At the end of the game, Claire has the yellow ball.

\texttt{</think>}

\texttt{<answer>} (E) yellow ball \texttt{</answer>}

\vspace{1em}
\hrule
\vspace{1em}

\textbf{[Conversation-style]}

\texttt{<cast\_of\_characters>}

\texttt{<persona1>} [Logical and detail-oriented, this thinker is a mathematician who enjoys breaking down problems into smaller, more manageable parts. They are meticulous and prefer to double-check every step of their reasoning.] \texttt{</persona1>}

\texttt{<persona2>} [An engineer by training, this thinker is practical and solution-focused. They prefer to use diagrams and visual aids to understand problems and are not afraid to reframe the problem if it leads to a solution.] \texttt{</persona2>}

\texttt{<persona3>} [A creative and intuitive thinker, this person excels at seeing patterns and connections that others might miss. They are a bit more relaxed in their approach, often suggesting unconventional methods to solve problems.] \texttt{</persona3>}

\texttt{</cast\_of\_characters>}

\texttt{<conversation>}

\texttt{<think1>} Let's start by laying out the initial positions of the balls. Alice has a brown ball, Bob has a purple ball, Claire has a black ball, Dave has a green ball, and Eve has a yellow ball. Now, we'll go through the swaps one by one. \texttt{</think1>}

\texttt{<think2>} I'll make a diagram to keep track of the swaps. First, Claire and Alice swap, so Claire now has the brown ball, and Alice has the black ball. \texttt{</think2>}

\texttt{<think3>} Good idea to use a visual. Next, Bob and Alice swap. Bob will end up with the black ball, and Alice will have the purple ball. \texttt{</think3>}

\texttt{<think1>} Then, Eve and Dave swap their balls. Eve will have the green ball, and Dave will have the yellow ball. \texttt{</think1>}

\texttt{<think2>} Now, Dave and Claire swap. Claire will have the yellow ball, and Dave will have the brown ball. \texttt{</think2>}

\texttt{<think3>} Finally, Alice and Bob swap. Bob will end up with the purple ball, and Alice will have the black ball. \texttt{</think3>}

\texttt{<think1>} After all the swaps, Claire ends up with the yellow ball. So the answer is (E). \texttt{</think1>}

\texttt{</conversation>}

\texttt{<group\_solution>} (E) yellow ball \texttt{</group\_solution>}

\end{tcolorbox}

\clearpage
\phantomsection
\textbf{Supplementary Table 8. Supervised fine-tuning hyperparameters}\label{tab:supptable8}

\begin{longtable}[]{@{}
  >{\raggedright\arraybackslash}p{(\columnwidth - 2\tabcolsep) * \real{0.4607}}
  >{\raggedright\arraybackslash}p{(\columnwidth - 2\tabcolsep) * \real{0.5393}}@{}}
\toprule\noalign{}
\endhead
\bottomrule\noalign{}
\endlastfoot
\multicolumn{2}{@{}>{\raggedright\arraybackslash}p{(\columnwidth - 2\tabcolsep) * \real{1.0000} + 2\tabcolsep}@{}}{%
\textbf{Data and Model}} \\ \midrule
Training/validation dataset size & 500 / 100 \\
Context window & 2048 \\
Training/validation batch size & 64 / 64 \\ \midrule
\multicolumn{2}{@{}>{\raggedright\arraybackslash}p{(\columnwidth - 2\tabcolsep) * \real{1.0000} + 2\tabcolsep}@{}}{%
\textbf{Optimization}} \\ \midrule
Optimizer & AdamW \\
Peak learning rate & 1e-05 \\
Warmup & Linear (10\% of total steps) \\
Annealing & Cosine \\
Total epochs & 5 \\
\end{longtable}

\clearpage

\phantomsection
\textbf{Supplementary Table 9. Benchmark Tasks}\label{tab:supptable9}

\begin{longtable}[]{@{}
  >{\raggedright\arraybackslash}p{(\columnwidth - 4\tabcolsep) * \real{0.2949}}
  >{\raggedright\arraybackslash}p{(\columnwidth - 4\tabcolsep) * \real{0.5817}}
  >{\raggedright\arraybackslash}p{(\columnwidth - 4\tabcolsep) * \real{0.1234}}@{}}
\toprule\noalign{}
\endhead
\bottomrule\noalign{}
\endlastfoot
\textbf{Benchmark} & \textbf{Task Category} & \textbf{\# Problems} \\ \midrule
BigBench Hard & Boolean Expressions & 248 \\
& Causal Judgment & 173 \\
& Date Understanding & 195 \\
& Disambiguation Question Answering & 249 \\
& Formal Logical Fallacies & 250 \\
& Geometric Shape Reasoning & 249 \\
& Hyperbaton (Syntactic Reordering) & 247 \\
& Logical Deduction (Five Objects) & 244 \\
& Logical Deduction (Seven Objects) & 224 \\
& Logical Deduction (Three Objects) & 249 \\
& Movie Recommendation & 7 \\
& Navigation and Spatial Planning & 248 \\
& Object Counting & 247 \\
& Penguins in a Table (Deductive Counting) & 143 \\
& Reasoning About Colored Objects & 245 \\
& Ruin Names (Linguistic Perturbation) & 165 \\
& Salient Translation Error Detection & 249 \\
& Snarks (Riddle Interpretation) & 146 \\
& Sports Understanding & 250 \\
& Temporal Sequence Reasoning & 65 \\
& Tracking Shuffled Objects (Five Objects) & 112 \\
& Tracking Shuffled Objects (Seven Objects) & 102 \\
& Tracking Shuffled Objects (Three Objects) & 57 \\
& Web of Lies (Deception Reasoning) & 72 \\
GPQA & Diamond Level & 161 \\
& Extended Level & 474 \\
& Main Benchmark & 380 \\
IFEval & Instruction-Following Consistency Evaluation & 524 \\
MATH (Hard) & Algebra (Hard) & 286 \\
& Counting \& Probability (Hard) & 110 \\
& Geometry (Hard) & 117 \\
& Intermediate Algebra (Hard) & 212 \\
& Number Theory (Hard) & 134 \\
& Pre-Algebra (Hard) & 182 \\
& Pre-Calculus (Hard) & 104 \\
MMLU-Pro & Advanced Multidomain Knowledge & 432 \\
MUSR & Murder Mysteries (Collaborative Deduction) & 207 \\
& Object Placement (Spatial Coordination) & 256 \\
& Team Allocation (Group Strategy) & 247 \\
\end{longtable}

\clearpage
\phantomsection
\textbf{Supplementary Table 10.} Descriptive statistics of conversational behaviors and socio-emotional roles for reasoning and instruction-tuned models (Count variables)\label{tab:supptable10}

\begin{longtable}[]{@{}
  >{\raggedright\arraybackslash}p{(\columnwidth - 12\tabcolsep) * \real{0.2352}}
  >{\raggedright\arraybackslash}p{(\columnwidth - 12\tabcolsep) * \real{0.1275}}
  >{\raggedright\arraybackslash}p{(\columnwidth - 12\tabcolsep) * \real{0.1275}}
  >{\raggedright\arraybackslash}p{(\columnwidth - 12\tabcolsep) * \real{0.1275}}
  >{\raggedright\arraybackslash}p{(\columnwidth - 12\tabcolsep) * \real{0.1275}}
  >{\raggedright\arraybackslash}p{(\columnwidth - 12\tabcolsep) * \real{0.1275}}
  >{\raggedright\arraybackslash}p{(\columnwidth - 12\tabcolsep) * \real{0.1275}}@{}}
\toprule\noalign{}
\endhead
\bottomrule\noalign{}
\endlastfoot
& \multicolumn{2}{>{\raggedright\arraybackslash}p{(\columnwidth - 12\tabcolsep) * \real{0.2549} + 2\tabcolsep}}{%
Reasoning Models} & \multicolumn{4}{>{\raggedright\arraybackslash}p{(\columnwidth - 12\tabcolsep) * \real{0.5099} + 6\tabcolsep}@{}}{%
Instruction-Tuned Models} \\ \midrule
& \begin{minipage}[t]{\linewidth}\raggedright
DeepSeek-\\
R1\strut
\end{minipage} & QWQ-32B & \begin{minipage}[t]{\linewidth}\raggedright
DeepSeek-\\
V3\strut
\end{minipage} & Qwen-2.5-32B-IT & Llama-3.3-70B-IT & \begin{minipage}[t]{\linewidth}\raggedright
Llama-3.1-8B-\\
IT\strut
\end{minipage} \\ \midrule
& Mean (SD) & Mean (SD) & Mean (SD) & Mean (SD) & Mean (SD) & Mean (SD) \\
\emph{Conversational behaviors} & & & & & & \\ \midrule
Question answering & 6.74 (4.32) & 6.68 (4.62) & 3.07 (2.53) & 2.35 (4.25) & 2.22 (1.98) & 3.16 (18.05) \\
Perspective shifts & 3.47 (2.76) & 5.41 (6.89) & 1.09 (1.69) & 0.79 (3.95) & 0.97 (1.56) & 1.84 (17.04) \\
Conflict of perspectives & 3.19 (2.95) & 5.27 (5.51) & 1.50 (2.06) & 0.97 (3.99) & 1.08 (1.62) & 2.25 (17.96) \\
Reconciliations & 1.97 (1.60) & 2.52 (1.90) & 0.61 (0.76) & 0.41 (0.62) & 0.58 (0.81) & 0.49 (1.27) \\ \midrule
\emph{Socio-emotional roles} & & & & & & \\
Ask for orientation & 2.67 (5.14) & 2.19 (3.80) & 0.01 (0.15) & 0.01 (0.11) & 0.00 (0.08) & 0.01 (0.24) \\
Ask for opinion & 0.50 (1.73) & 0.57 (1.94) & 0.00 (0.07) & 0.00 (0.05) & 0.00 (0.07) & 0.01 (0.39) \\
Ask for suggestion & 0.18 (0.87) & 0.26 (1.10) & 0.00 (0.04) & 0.00 (0.04) & 0.00 (0.06) & 0.01 (0.14) \\
Give orientation & 29.67 (23.44) & 26.17 (21.06) & 8.26 (6.55) & 7.00 (7.74) & 6.20 (5.14) & 8.52 (24.72) \\
Give opinion & 19.25 (18.08) & 21.72 (22.64) & 4.15 (4.31) & 3.09 (4.73) & 3.74 (4.02) & 4.60 (35.84) \\
Give suggestion & 9.04 (7.80) & 13.05 (12.33) & 2.23 (3.02) & 2.17 (2.21) & 2.50 (3.09) & 3.05 (12.05) \\
Antagonism & 0.00 (0.13) & 0.00 (0.06) & 0.01 (0.27) & 0.00 (0.10) & 0.00 (0.13) & 0.01 (0.16) \\
Tension & 3.82 (5.37) & 9.65 (11.53) & 0.14 (0.67) & 0.03 (0.24) & 0.21 (2.04) & 0.65 (10.60) \\
Disagree & 2.56 (5.15) & 3.09 (5.41) & 0.49 (1.54) & 0.16 (0.89) & 0.25 (2.17) & 0.88 (8.90) \\
Solidarity & 0.00 (0.07) & 0.00 (0.05) & 0.01 (0.23) & 0.01 (0.14) & 0.00 (0.13) & 0.02 (0.33) \\
Tension release & 0.50 (1.12) & 0.36 (0.87) & 0.02 (0.29) & 0.00 (0.10) & 0.01 (0.14) & 0.06 (3.50) \\
Agree & 3.59 (4.83) & 3.50 (4.64) & 0.18 (0.69) & 0.04 (0.26) & 0.03 (0.21) & 0.06 (0.85) \\ \midrule
Observations & \multicolumn{6}{>{\raggedright\arraybackslash}p{(\columnwidth - 12\tabcolsep) * \real{0.7648} + 10\tabcolsep}@{}}{%
8,262} \\
\end{longtable}

\clearpage 
\phantomsection
\textbf{Supplementary Table 11.} Structural equation model estimates linking DeepSeek-R1, conversational behaviors, socio-emotional roles, cognitive reasoning strategies, and accuracy\label{tab:supptable11}

\begin{longtable}[]{@{}
  >{\raggedright\arraybackslash}p{(\columnwidth - 8\tabcolsep) * \real{0.5096}}
  >{\raggedright\arraybackslash}p{(\columnwidth - 8\tabcolsep) * \real{0.1322}}
  >{\raggedright\arraybackslash}p{(\columnwidth - 8\tabcolsep) * \real{0.1165}}
  >{\raggedright\arraybackslash}p{(\columnwidth - 8\tabcolsep) * \real{0.1217}}
  >{\raggedright\arraybackslash}p{(\columnwidth - 8\tabcolsep) * \real{0.1200}}@{}}
\toprule\noalign{}
\endhead
\bottomrule\noalign{}
\endlastfoot
& \textbf{Estimate} & \textbf{Std.Err} & \textbf{z} & \textbf{p-value} \\ \midrule
DeepSeek-R1 → Question \& Answering & 4.013 & 0.066 & 61.060 & \textless0.001 \\
DeepSeek-R1 → Perspective Shift & 3.273 & 0.067 & 48.530 & \textless0.001 \\
DeepSeek-R1 → Conflict of Perspectives & 2.764 & 0.065 & 42.400 & \textless0.001 \\
DeepSeek-R1 → Reconciliation & 1.748 & 0.016 & 109.360 & \textless0.001 \\
DeepSeek-R1 → Positive & 4.017 & 0.055 & 72.810 & \textless0.001 \\
DeepSeek-R1 → Negative & 9.177 & 0.121 & 75.740 & \textless0.001 \\
DeepSeek-R1 → Ask & 3.276 & 0.050 & 66.090 & \textless0.001 \\
DeepSeek-R1 → Give & 46.810 & 0.408 & 114.750 & \textless0.001 \\ \midrule
Question \& Answering → Verification & 0.179 & 0.070 & 2.550 & 0.011 \\
Perspective Shift → Verification & -0.096 & 0.107 & -0.890 & 0.372 \\
Conflict of Perspectives → Verification & 0.216 & 0.113 & 1.910 & 0.057 \\
Reconciliation → Verification & -0.237 & 0.109 & -2.180 & 0.029 \\
Positive → Verification & -0.023 & 0.129 & -0.180 & 0.858 \\
Negative → Verification & 0.055 & 0.057 & 0.960 & 0.336 \\
Ask → Verification & -0.111 & 0.039 & -2.810 & 0.005 \\
Give → Verification & 0.026 & 0.015 & 1.740 & 0.082 \\
DeepSeek-R1 → Verification & -0.673 & 0.327 & -2.060 & 0.039 \\ \midrule
Question \& Answering → Backtracking & -0.023 & 0.037 & -0.620 & 0.536 \\
Perspective Shift → Backtracking & 0.007 & 0.071 & 0.100 & 0.924 \\
Conflict of Perspectives → Backtracking & 0.163 & 0.065 & 2.530 & 0.011 \\
Reconciliation → Backtracking & -0.005 & 0.082 & -0.060 & 0.953 \\
Positive → Backtracking & -0.125 & 0.040 & -3.130 & 0.002 \\
Negative → Backtracking & 0.079 & 0.028 & 2.820 & 0.005 \\
Ask → Backtracking & -0.024 & 0.016 & -1.510 & 0.132 \\
Give → Backtracking & 0.014 & 0.006 & 2.320 & 0.020 \\
DeepSeek-R1 → Backtracking & -0.325 & 0.129 & -2.520 & 0.012 \\ \midrule
Question \& Answering → Subgoal setting & 0.035 & 0.016 & 2.120 & 0.034 \\
Perspective Shift → Subgoal setting & 0.006 & 0.012 & 0.450 & 0.653 \\
Conflict of Perspectives → Subgoal setting & -0.036 & 0.016 & -2.220 & 0.026 \\
Reconciliation → Subgoal setting & 0.089 & 0.021 & 4.270 & \textless0.001 \\
Positive → Subgoal setting & 0.010 & 0.012 & 0.830 & 0.409 \\
Negative → Subgoal setting & 0.007 & 0.004 & 1.680 & 0.093 \\
Ask → Subgoal setting & 0.005 & 0.010 & 0.540 & 0.587 \\
Give → Subgoal setting & 0.008 & 0.003 & 2.490 & 0.013 \\
DeepSeek-R1 → Subgoal setting & 0.224 & 0.087 & 2.570 & 0.010 \\ \midrule
Question \& Answering → Backward chaining & -0.006 & 0.002 & -3.900 & \textless0.001 \\
Perspective Shift → Backward chaining & 0.011 & 0.002 & 5.710 & \textless0.001 \\
Conflict of Perspectives → Backward chaining & -0.003 & 0.002 & -1.510 & 0.132 \\
Reconciliation → Backward chaining & -0.002 & 0.003 & -0.800 & 0.422 \\
Positive → Backward chaining & -0.002 & 0.001 & -1.720 & 0.086 \\
Negative → Backward chaining & 0.005 & 0.001 & 3.760 & \textless0.001 \\
Ask → Backward chaining & 0.004 & 0.001 & 2.870 & 0.004 \\
Give → Backward chaining & 0.001 & 0.000 & 2.950 & 0.003 \\
DeepSeek-R1 → Backward chaining & 0.019 & 0.007 & 2.730 & 0.006 \\ \midrule
Question \& Answering → Accuracy & 0.009 & 0.001 & 7.730 & \textless0.001 \\
Perspective Shift → Accuracy & -0.006 & 0.001 & -4.250 & \textless0.001 \\
Conflict of Perspectives → Accuracy & -0.006 & 0.002 & -3.240 & 0.001 \\
Reconciliation → Accuracy & 0.001 & 0.003 & 0.330 & 0.742 \\
Positive → Accuracy & 0.020 & 0.003 & 6.660 & \textless0.001 \\
Negative → Accuracy & -0.001 & 0.000 & -1.470 & 0.141 \\
Ask → Accuracy & 0.001 & 0.001 & 0.900 & 0.367 \\
Give → Accuracy & 0.000 & 0.000 & -1.450 & 0.147 \\
Verification → Accuracy & 0.005 & 0.001 & 3.400 & 0.001 \\
Backtracking → Accuracy & -0.003 & 0.001 & -2.810 & 0.005 \\
Subgoal setting → Accuracy & 0.009 & 0.004 & 2.610 & 0.009 \\
Backward chaining → Accuracy & -0.001 & 0.006 & -0.180 & 0.855 \\
DeepSeek-R1 → Accuracy & 0.253 & 0.007 & 35.060 & \textless0.001 \\
Length → Accuracy & -0.134 & 0.005 & -25.410 & \textless0.001 \\ \midrule
Observations & \multicolumn{4}{>{\centering\arraybackslash}p{(\columnwidth - 8\tabcolsep) * \real{0.4904} + 6\tabcolsep}@{}}{7,738} \\
\end{longtable}

\emph{Notes}: All structural equation models control for task fixed effects. The statistical significance of coefficients is tested using two-sided \emph{t}-tests. Out of 8,262 tasks, tasks from IFEval are excluded due to difficulties in accuracy evaluation (524 tasks), and we use the remaining 7,738 tasks.

\end{document}